\documentclass{article}

\usepackage{arxiv}

\usepackage[utf8]{inputenc} %
\usepackage[T1]{fontenc}    %
\usepackage{hyperref}       %
\usepackage{url}            %
\usepackage{booktabs}       %
\usepackage{amsfonts}       %
\usepackage{nicefrac}       %
\usepackage{microtype}      %
\usepackage{lipsum}		%
\usepackage{graphicx}
\usepackage{natbib}
\usepackage{doi}

\title{Semantic and Geometric Unfolding of StyleGAN Latent Space}

\author{Mustafa Shukor \\
	InterDigital, Inc \thanks{Interdigital, Research and Innovation, Rennes, France. The work done during Mustafa's internship.} 

	\And
	Xu Yao \\
	InterDigital, Inc \\
	T\'el\'ecom Paris

	\And 
	Bharath Bhushan Damodaran \\
	InterDigital, Inc 
	\And 
	Pierre Hellier \\
	InterDigital, Inc 

}

\date{\texttt{\{firstname.lastname\}@interdigital.com}}

\def\eg{\emph{e.g.} }

\def\etal{\emph{et al.} }
\def\ie{\emph{i.e.} }

\def\wplus{$\mathcal{W}^+$ }
\def\wstar{$\mathcal{W}^{\star}$ }
\def\wplussp{$\mathcal{W}^+$}
\def\wstarsp{$\mathcal{W}^{\star}$}

\def\wstara{$\mathcal{W}^{\star}_a$ }
\def\wstarasp{$\mathcal{W}^{\star}_a$}

\usepackage{svg}
\usepackage{amsmath}
\usepackage{caption}
\usepackage{subcaption}
\usepackage{mwe}
\usepackage{graphicx}
\usepackage{rotating}
\usepackage{multirow}
\usepackage{array}
\usepackage{booktabs}
\usepackage{pifont}%
\newcommand{\cmark}{\color{green} \ding{51}}%

\usepackage{xcolor}
\usepackage{xparse}

\usepackage[normalem]{ulem}
\usepackage{ifthen}
\newboolean{final}
\setboolean{final}{false}

\newcommand{\FontFormatter}[2]{%
    \textcolor[rgb]{#1}{\textbf{#2}}%
}

\NewDocumentCommand{\TextFormatter}{ommm}{%
    \FontFormatter{#4}{%
        \IfNoValueTF{#1}{%
            $\scriptstyle \overset{#3}{++}$ #2%
        }{%
            \sout{#1} $\scriptstyle \xrightarrow{#3}$ #2%
        }%
    }%
}

\NewDocumentCommand{\TextFormatterFinal}{ommm}{#2}

\NewDocumentCommand{\pierre}{om}{%
    \TextFormatter[#1]{#2}{Pierre}{0.5, 0.0, 0.0}%
}

\NewDocumentCommand{\ms}{om}{%
    \TextFormatter[#1]{#2}{Mustafa}{0.0, 0.5, 0.0}%
}

\NewDocumentCommand{\bharath}{om}{%
\TextFormatter[#1]{#2}{Bharath}{0.0, 0.0, 1.0}%
}

\NewDocumentCommand{\xu}{om}{%
\TextFormatter[#1]{#2}{}{0.5, 0.0, 1.0}%
}

\hypersetup{
    colorlinks=true,%
}

\begin{document}
\maketitle

\begin{abstract}
Generative adversarial networks (GANs) have proven to be surprisingly efficient for image editing by inverting and manipulating the latent code corresponding to a natural image. This property emerges from the disentangled nature of the latent space. In this paper, we identify two geometric limitations of such latent space: (a) euclidean distances differ from image perceptual distance, and (b) disentanglement is not optimal and facial attribute separation using linear model is a limiting hypothesis. We thus propose a new method to learn a proxy latent representation using normalizing flows to remedy these limitations, and show that this leads to a more efficient space for face image editing.

\end{abstract}

\section{Introduction}
\label{sec:intro}

GANs \cite{gan} have shown tremendous success in generating high quality realistic images that are indistinguishable from real ones. Yet, several open problems regarding these models still exist, such as image generation control, latent space understanding and attributes disentanglement, which are important for both the generation and editing of high quality images.
Recently, many improvements have been proposed to the original GAN architecture \cite{karras2017progressive, brock2018large, miyato2018spectral}, which led to unprecedented image quality \cite{stylegan, styleganimporveing}. In particular, the state-of-the-art method StyleGAN \cite{stylegan, styleganimporveing} has been improved, leading to image editing methods using the latent representation \cite{collins2020editing, shen2020interpreting, abdal2020image2stylegan++, shen2020interfacegan}. Specifically, recent approaches \cite{shen2020interpreting, shen2020interfacegan} assume that the attributes are disentangled and can be separated by hyperplanes, which enables interpolation and one attribute manipulations. InterFaceGAN \cite{shen2020interfacegan} computes an editing direction for each facial attribute, orthogonal to the linear classification boundary for this attribute. We claim that the hyperplane classification boundary assumption is not perfectly true, and the attributes are not perfectly disentangled, which can further explain why these approaches do not lead to perfect attribute manipulation.

We argue that another current limitation of GANs latent space is that the Euclidean distance between two latent codes do not reflect the perceptual distance between the corresponding images. However, distance computation is critical for many applications including facial similarity, image comparison, and data clustering. Designing consistent distances between objects is crucial as shown in \cite{zhang2018unreasonable}.

\begin{figure}[h]
     \centering
     \includegraphics[width=\textwidth]{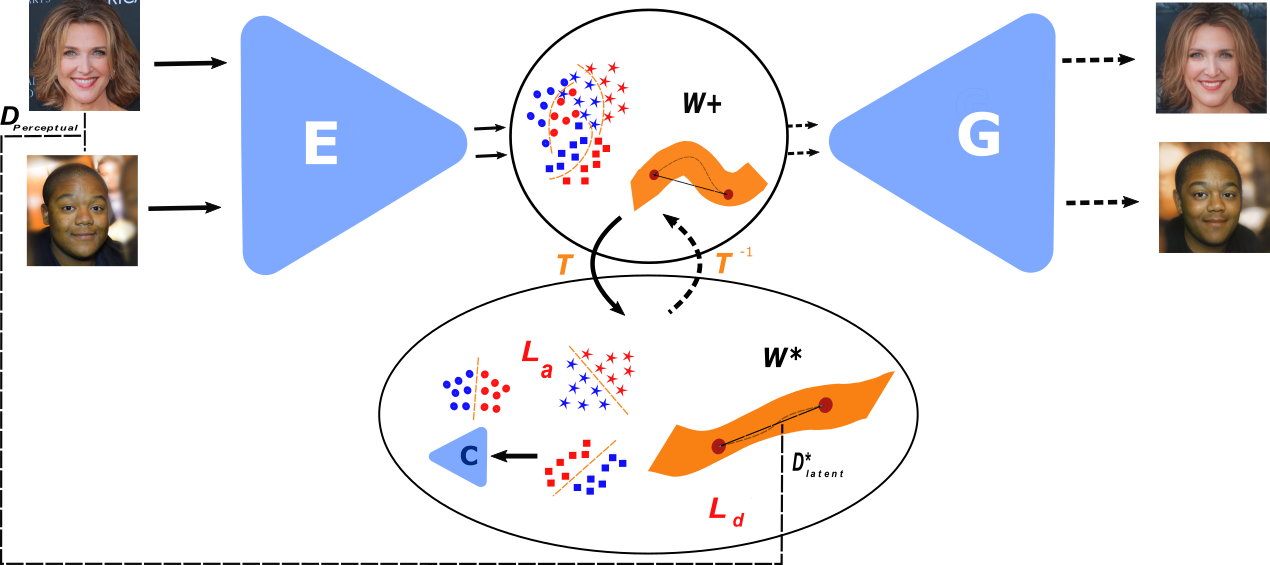}
     \vspace*{+2mm}
    \caption{%
    An illustration of our proposed approach. $E$, $G$, and $C$ are the StyleGAN2 encoder, generator and attributes classifier respectively. Only the NF mapping $T$ is trained. The existing StyleGAN2 latent space \wplus does not satisfy the linear separability assumption and the consistency between the latent euclidean distance (straight line) and the perceptual distance (dashed geodesic line). In our learned proxy latent space \wstar the attributes are disentangled, can be separated by hyperplanes (between the positive and negative regions) and the Euclidean latent distance mimics the perceptual distance.%
    }
    \hfill
    \label{fig:mainfigure}
\end{figure}

The first solution to these limitations could be the retraining of GAN with explicit constraints. However, it is known that the training of GANs is hard and computationally expensive. We propose in this paper an alternative approach without retraining the GAN. Specifically, we will focus on the aforementioned properties and learn a bijective transformation (\ie, Normalizing Flows) from the original latent space (\ie \wplussp) to a new proxy latent space (\wstarsp). In \wstar,  the facial attributes are linearly separable, disentangled and the latent Euclidean distance mimics the perceptual image distance. The choice of a bijective transformation allows to benefit from the pretrained StyleGAN2 generative capabilities. Figure \ref{fig:mainfigure} illustrates the proposed approach. Our contributions are the following: 
\begin{itemize}
    \item We propose to learn a derived latent representation where supervision is used to explicitly disentangle  the facial attributes. We also propose to enforce that Euclidean distance in the new latent space mimics the perceptual one in the image space.
    \item We propose to learn this proxy latent representation using normalizing flows, which can be applied to any pretrained GAN while preserving the generative capability of the original GAN.
    \item We show experimentally that the desired properties are indeed enforced in this new latent representation, leading to a more efficient image manipulation.
\end{itemize}
The rest of the paper is organised as follows: section \ref{sec:related works} presents related works on GANs, Normalizing Flows, perceptual distances, and attributes disentanglement. Our proposed method is detailed in section \ref{sec:method} and section \ref{sec:experiments} presents the experimental results and ablation study. Finally, section \ref{sec:discussion} discusses the approach and conclusions are drawn in section \ref{sec:conclusion}. The code will be released once the paper is accepted.

\section{Related work}
\label{sec:related works}
\paragraph{GANs} \quad
GANs \cite{gan} are one type of generative models that are trained adversarially to generate complex data distributions starting from a simple one. There are other types such as VAEs \cite{vae} and Normalizing Flows \cite{nf}, though the image quality generated by GANs is higher with reasonable model size. Several improvements have been proposed to improve GANs architecture \cite{radford2015unsupervised}, loss function \cite{mao2017least, wgan} and its training \cite{gulrajani2017improved, miyato2018spectral, karras2017progressive}. Recently, StyleGAN \cite{stylegan} was introduced as the state of the art in high-resolution image generation, especially for human faces. This architecture allows better control of the generation process, although it produces some artifacts which were solved in a new version (StyleGAN2 \cite{styleganimporveing}). The latent space of StyleGAN2 ($\mathcal{W}$) is better disentangled than the original $\mathcal{Z}$ space. Furthermore, the \wplus is better for image editing \cite{abdal2019image2stylegan, ghfeatxu2020generative, wei2021simplebase} and the focus in our work is to obtain a new space with better properties.
\paragraph{Disentangled Representations} \quad
There have been many efforts to obtain disentangled representation for generative models (\eg, VAEs and GANs). In $\beta$-VAE \cite{betavae} they put a strong constraint to obtain latent posterior distribution with uncorrelated dimensions. Similarly, FactorVAE \cite{factorvae} minimizes the total correlation (TC) of the latent space distribution to encourage independence. InfoGAN \cite{infogan} maximizes the mutual information between a small subset of the latent dimensions and the generator distribution. This is improved in InfoGAN-CR \cite{Lin2019InfoGANCRDG}, where each dimension in the latent code is encouraged to make a significant change to the image. Recently, Zheng \etal \cite{zheng2021unsupervised} leverage an auxiliary task (\ie face reconstruction) and a decorrelation loss to obtain disentanglement. All these approaches require retraining the generator. In addition, they are unsupervised which limits the range of attributes that they can manipulate, for instance, we can change the age, gender and adding makeup while in all these mentioned methods it is not clear how to achieve such manipulation.

\paragraph{Normalizing Flows (NFs)} \quad
NFs \cite{nf} are another type of generative models that consists of diffeomorphic transformations between a simple known distribution and any arbitrarily complex one. Due to the constraints that should be satisfied (\eg bijectivity, tractable inverse and jacobian determinant) the expressivity of such models is limited compared to others (\eg GANs). In recent years, there have been many attempts to improve them. Real NVP \cite{realnvp} is one of the Discrete NF models that was able to generate good quality images. MAF \cite{maf} was also proposed but it is not efficient as Real NVP as it is not parallelizable. Recently, Glow \cite{kingma2018glow} was proposed as the state of the art in high quality image generation . Unfortunately, its success depends on a very large number of parameters, making this model difficult to train in practice. Continuous NF \cite{grathwohl2018ffjord} has been proposed, which usually gives comparable expressivity to DNF with smaller models. However, since it is based on ODE solver, the inference is computationally demanding for real time applications such as video editing.

\paragraph{Perceptual distances} \quad
Computing image distance that is consistent with the human perception is of high interest for many computer vision applications. As the traditional L1/L2 distances do not have this property, many works were devoted to handcraft perceptual distances \cite{MSSSIM, sims2020frequency}. Recently, VGG16 \cite{johnson2016perceptual} was one of the first feature based distances that outperforms all the the handcrafted ones. It consists of computing the L2 distance between the features extracted from a pretrained VGG16 network. This was followed by LPIPS \cite{zhang2018unreasonable} which uses a pre-trained network fine-tuned on perceptual judgement prediction task. Some perceptual distances were designed for specific tasks \cite{loss_training_resto}.

\section{Method}
\label{sec:method}
In this section, we explain how to learn a proxy latent space (dubbed \wstarsp) that  satisfies two properties: (a) a latent Euclidean distance that mimics the perceptual one in the image space (Sec \ref{subsec:LD}) and (b) disentanglement and linear separation of the attributes (Sec \ref{subsec:AD}). In addition, we will explain how other properties that are useful for image editing can be satisfied. 

We assume that we have a pretrained StyleGAN2 generator $G$ that considers a latent code $w \in \mathcal{W}^{+}$ and generates a high resolution image $I$ (\ie, 1024 x 1024). A bijective transformation $T: \mathcal{W}^{+} \rightarrow \mathcal{W}^{\star}$ is trained to map a latent code $w \in \mathcal{W}^{+}$ to $w^{\star} \in \mathcal{W}^{\star}$. $T$ is a Normalizing Flows (NFs) model and can be inverted explicitly. The focus of our work is on real natural images, thus we assume that there exists a pretrained encoder $E$ that embeds any image in \wplus such that $G(E(I)) \simeq I$. 
Although, the transformation $T$ is modelled as a NF, it is noted that our work only requires the bijectivity, as such, we did not impose the prior distribution in the proxy latent space as we are not interested in the density estimation.

\subsection{Latent Distance Unfolding}\label{subsec:LD}
The objective here is to learn the mapping $T$ that maps the latent codes to the new latent space such that the latent distance in this space is similar to the perceptual one in the image space. This property is obtained by minimizing the distance between the proxy latent distance and perceptual distance as below:
\begin{equation}
\label{eq:latent-dist}
    \mathcal{L}_d = \frac{1}{N-1} \sum_{i, j \in S} (\Vert T(E(I_i)) - T(E(I_j))\Vert_2^2 - D_{perceptual}(I_i, I_j))^2
\end{equation}
where $S$ is a set of image samples indices of size $N$. The first term is the latent squared Euclidean distance  in \wstar ($D^{*}_{latent}(\cdot, \cdot))$  and $D_{perceptual}(\cdot, \cdot)$ is the perceptual distance.
$D_{perceptual}$ could be any perceptual distance. We refer to the latent distance in \wplus as $D_{latent}$.

\subsection{Attributes Disentanglement}
\label{subsec:AD}

The main objective here is to learn $T$ that maps the latent codes to the new space where the linear classification is optimal for each attribute.
This is done by minimizing the classification loss of a linear attribute classifier $C:$ \wstar $\rightarrow \{0, 1\}^K$, where $K$ is the number of attributes labeled in the image dataset. Choosing a linear model is mainly to enforce the linear separation between the positive and negative regions of each attribute while reducing the loss in general leads to better attributes disentanglement.
Instead of using one classification model for all the attributes, we found that it is better to use one binary classification model for each attribute and train these models jointly. For each sample $w$, the objective is to minimize:
\begin{equation}
\label{eq:att}
    \mathcal{L}_a = - \sum_{i=0}^{K} y_{i}\log(C_i(T(w)) + (1-y_{i})\log(1-C_i(T(w)))
\end{equation}
Where $C_i:$ \wstar $\rightarrow \{0, 1\}$ is the classifier for the ith attribute, $y_{i} \in \{0, 1\}$ is the label of the sample $w$ corresponding to the ith attribute. In \eqref{eq:att} the classifiers are fixed and only $T$ is optimized as we are interested in obtaining a linear separation for each attribute. Theoretically, any linear classifier could be used. However, since a form of linear separation already exists in \wplussp, we choose to pretrain the linear classifiers first in \wplus and fix it while optimizing for $T$. This provides some regularization so that $T$ only focuses on improving the pre-trained classifier. In addition, it helps to converge faster.

Our proposed proxy latent space \wstar with both properties; the attributes disentanglement/separation and latent distance unfolding are learned by minimizing the joint loss as:
\begin{equation}
\label{eq:wstar}
    \mathcal{L}_{W^{\star}} = \mathcal{L}_a + \lambda_d \mathcal{L}_d, 
\end{equation}
where $\lambda_d$ trade-off between the two losses.
\subsection{Regularization for Image Editing}
Specifically for image editing, we introduce additional regularizations in  \eqref{eq:wstar} to better condition the properties of \wstarsp.
For instance, it is important to preserve the person identity after editing. 
Identity preservation is enforced by minimizing the loss between the features extracted from a pretrained face recognition model $F$ before and after editing. Thus for a given image sample $I$, the identity preservation loss can be written as:
\begin{equation}
    \mathcal{L}_{ID} = \|F(G(E(I))) - F(G(T^{-1}(T(E(I)) + \epsilon)))\|_2^2
\end{equation}
where $\epsilon \sim \mathcal{N}(0, I)$  accounts for noise or image manipulation.
Thus the total loss with identity regularization is written as follows:
\begin{equation}
\label{eq:att_id}
    \mathcal{L}_{W^{\star}_{ID}} = \mathcal{L}_a + \lambda_d \mathcal{L}_d + \lambda_{ID} \mathcal{L}_{ID}
\end{equation}
where $\lambda_d$ and $\lambda_{ID}$ are the weights of the respective loss terms.

\section{Experiments}
\label{sec:experiments}
In this section we evaluate the properties of our proposed proxy latent space \wstarsp. First, we detail the implementation details, next we describe the quantitative metrics, and finally we present the experimental results.

\subsection{Implementation Details}
\label{sec:implementation}

We use a pretrained StyleGAN2 ($G$) on FFHQ dataset \cite{stylegan}. The images are encoded in \wplus using a pretrained StyleGAN2 encoder ($E$) \cite{encodinginstyle} (the parameters of the generator and the encoder remain fixed in all the experiments). The latent vector dimension in \wplus and \wstar is $18\times 512$.
Celeba-HQ \cite{karras2017progressive} is the image dataset that is used and consists of 30000 high quality images (\ie 1024x1024) of faces where each image has annotation for $K=40$ attributes. A single layer MLP model for each attribute ($C_i$) is used as linear classifier which is pretrained in \wplussp. 
For the NF model, Real NVP \cite{realnvp} is used without batch normalization. Each coupling layer consists of 3 fully connected (FC) layers for the translation function and 3 FC for the scale one with LeakyReLU as hidden activation and Tanh as output one. VGG16 \cite{johnson2016perceptual} features of blocks 2, 3 and 4,  pretrained on Imagenet is used to compute perceptual distance. For the face recognition model $F$ we use VGG16 pretrained on a face recognition dataset \cite{deepfacerecognition}. For all the experiments, Adam optimizer is used with $\beta_1=0.9$ and $\beta_2=0.999$, learning rate=$1e-4$ and the batch size=$8$.
\subsection{Quantitative metrics}
To assess the linear separation and disentanglement for attributes we use classification accuracy and DCI \cite{Eastwood2018AFF},  and for latent distance unfolding we use 2AFC score and some statistics.
\paragraph{Classification Accuracy:}
An SVM was trained from scratch for each attribute on 15000 latent codes in the corresponding space.
In \wplus the latent codes were obtained after encoding the images of Celeba-HQ using the pretrained encoder, and in \wstar were obtained after mapping the encoded codes using the trained NF model $T$. Among the available latent codes, we split 80\% for training and the rest of them for validation. The accuracy measures are reported using minimum classification accuracy (Min Class.Acc), maximum classification accuracy (Min Class.Acc) among the attributes and average accuracy (Avg Class.Acc).
\paragraph{DCI \cite{Eastwood2018AFF}:} DCI is used to assess disentanglement; Disentanglement (D) quantifies how much each dimension captures at most one attribute, Completeness (C) quantifies how much each attribute is captured by a single dimension and Informativeness (I) quantifies how much informative the latent code is for the attributes which is simply the classification error. 
We have used 40 Lasso regressors from scikit-learn library with $\alpha =0.02$ that is multiplied by the L1 regularizer. The dataset size is 2000 which is composed of the validation set of Celeba-HQ encoded using the pretrained encoder. The train and validation sets are splitted as 80\% and 20\% respectively. The regressors are trained using the RMSE loss.
\paragraph{Mean, STD:} To measure how much the latent distance deviates from perceptual distance, we report the mean and standard deviation of the difference between the latent and perceptual distances (Mean and STD). In order to compute this, we randomly sampled 600 pairs of latent codes (images are from the Celeba-HQ dataset and were encoded using the pretrained encoder) and computed the latent distance in both spaces (\wplus and \wstarsp) and the perceptual distance of the corresponding image pairs. 
\paragraph{2AFC score:} 2AFC test \cite{zhang2018unreasonable} consists of choosing which of the two distorted images (\ie I0, I1) is more similar to the reference one. In our case, the 3 images were selected at random from Celeba-HQ dataset, encoded using the pretrained encoder, and the latent distance was computed between the 2 pairs.The ground truth was obtained based on the judgment of the VGG16 perceptual distance pretrained on Imagenet.
The 2AFC score is computed as the number of right decisions divided by the total number of triplets.
\subsection{Results}
\subsubsection{Attributes Disentanglement and Latent Distance Unfolding}
\label{sec:results}
\setlength\tabcolsep{2 pt}
\begin{table}[t]
\begin{center}
\begin{tabular}{lccccccccccc}
\toprule
& 
\multicolumn{3}{c}{Linear separation} &
&
\multicolumn{3}{c}{Disentanglement} &
&
\multicolumn{3}{c}{Dist. Unfolding} \\ \cline{2-4} \cline{6-8} \cline{10-12}
Space & 
\begin{tabular}{@{}c@{}}Min Class.\\ Acc $\uparrow$ \end{tabular} & 
\begin{tabular}{@{}c@{}}Max Class.\\ Acc $\uparrow$\end{tabular} & 
\begin{tabular}{@{}c@{}}Avg. Class.\\ Acc $\uparrow$\end{tabular} & 
& 
D $\uparrow$ & 
C $\uparrow$ & 
I $\downarrow$ &
& 
Mean $\downarrow$ & 
STD $\downarrow$& 
\begin{tabular}{@{}c@{}} 2AFC $\uparrow$ score  \end{tabular} \\
\hline
\wplus & 0.635 & 0.979 & 0.834 & & 0.59 & 0.43 & 0.30 & & -1.185 & 0.970& 0.58 \\
\wstar & \textbf{0.822} & \textbf{0.989} & \textbf{0.913} & & \textbf{0.75} & \textbf{0.52} & \textbf{0.27} & & \textbf{0.035} & \textbf{0.384}& \textbf{0.83}\\
\bottomrule
\end{tabular}
\end{center}
\caption{The quantitative assessment of the attribute's linear separation, disentanglement and latent distance unfolding in the \wplus and \wstarsp. %
The $\uparrow$ ($\downarrow$) indicates the higher (lower) values are better and the best results in \textbf{bold}. We can notice that in \wstarsp; the latent distance is closer to the perceptual distance and the attributes are more disentangled and linearly separable. }
\label{tab:res_dist_att}
\end{table}
In this experiment, Real NVP consists of 13 coupling layers and we set $\lambda_d=10$ in eq. \eqref{eq:wstar}. The classifier $C$ is pretrained in \wplus and fixed to benefit from the properties of \wplussp. For this reason, we do not use batch normalization in the Real NVP, otherwise batch normalization normalizes the data and the hyperplanes of the pretrained classifiers will not hold. Table \ref{tab:res_dist_att} reports the quantitative metrics in \wstar and \wplussp. When the attributes disentanglement is considered, we can notice a significant improvement in the classification accuracy and DCI  metrics in \wstar over \wplussp. For example, the latent space \wstar is improved by 20\% with respect to the minimum classification accuracy. This demonstrates that in our proposed proxy latent space the linear separability of attributes is better and similar observation also holds for attributes disentanglement. On the other hand, when the latent distance is considered, the mean and std reveals that the Euclidean distance in \wplus is inconsistent with perceptual distance in the image space, whereas in \wstar it mimics better the perceptual distance (also check supplementary material). Further, the 2AFC score demonstrates that the latent distance $D^{*}_{\text{latent}}$ in \wstar is significantly higher with large margin than in \wplus and also Figure ~\ref{fig:2afc_att_inter_results} shows $D^{*}_{\text{latent}}$ mimics the human perception.

\begin{figure}[t]
\setlength\tabcolsep{2pt}%
\centering
\begin{tabular}{p{1.5cm}ccc|ccc|ccc}
\centering
  &
 \begin{tabular}{@{}c@{}} I 0 \\ \includegraphics[width=0.08\textwidth]{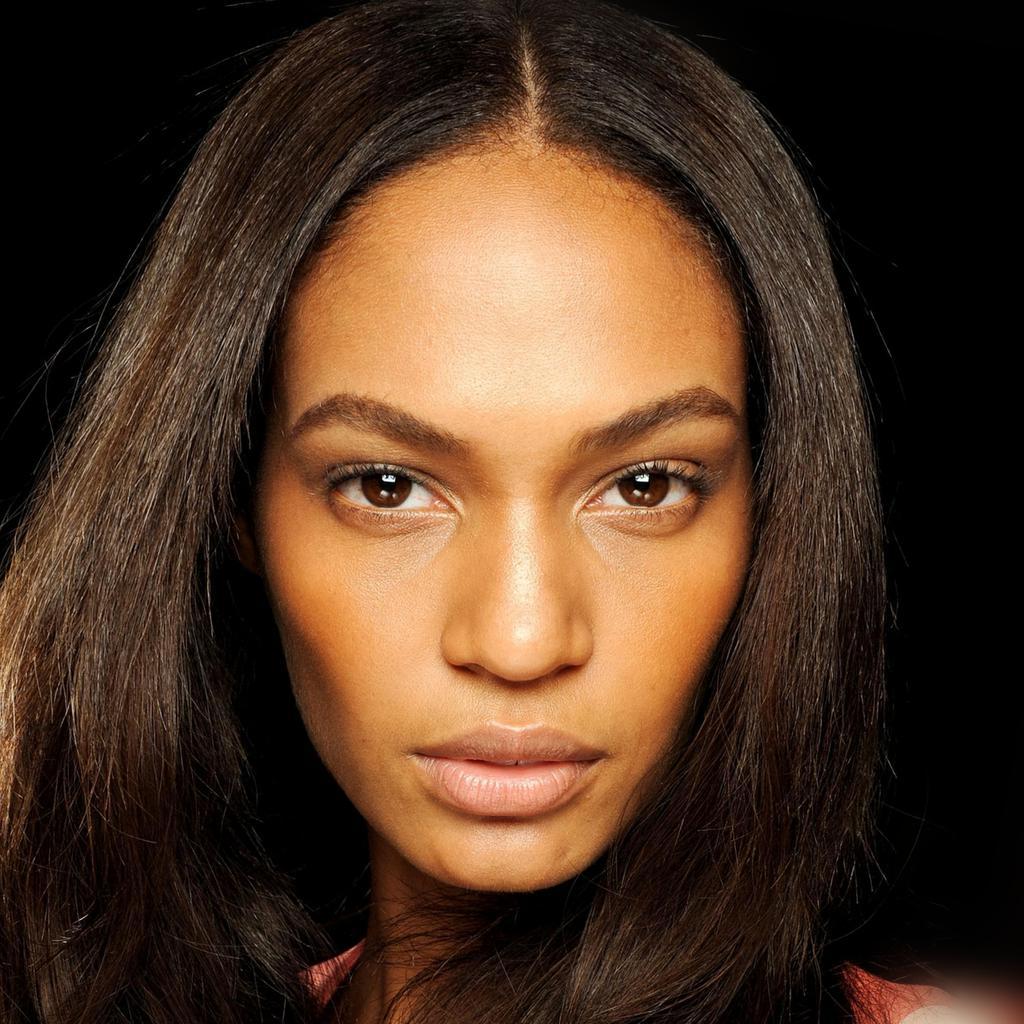} \end{tabular} & 
 \begin{tabular}{@{}c@{}} Ref \\ \includegraphics[width=0.08\textwidth]{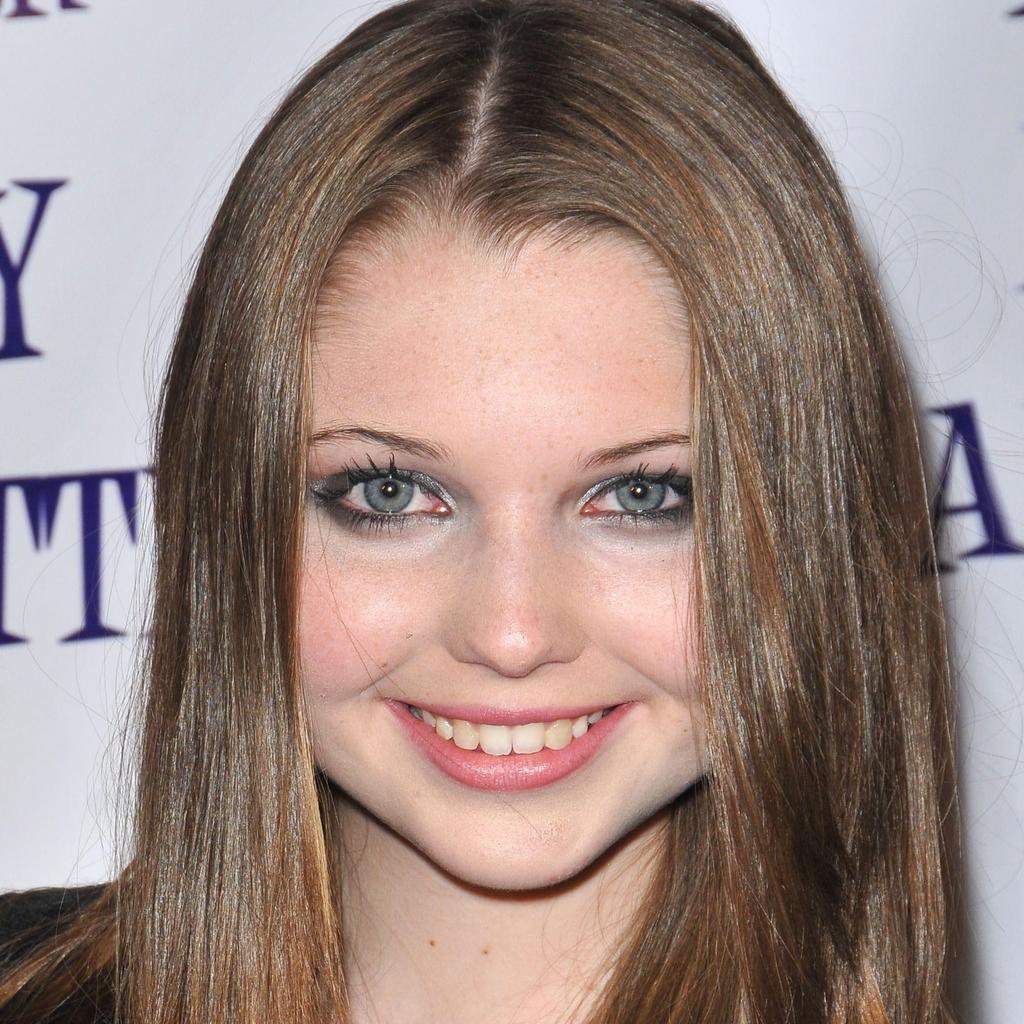} \end{tabular} & 
 \begin{tabular}{@{}c@{}} I 1  \\ \includegraphics[width=0.08\textwidth]{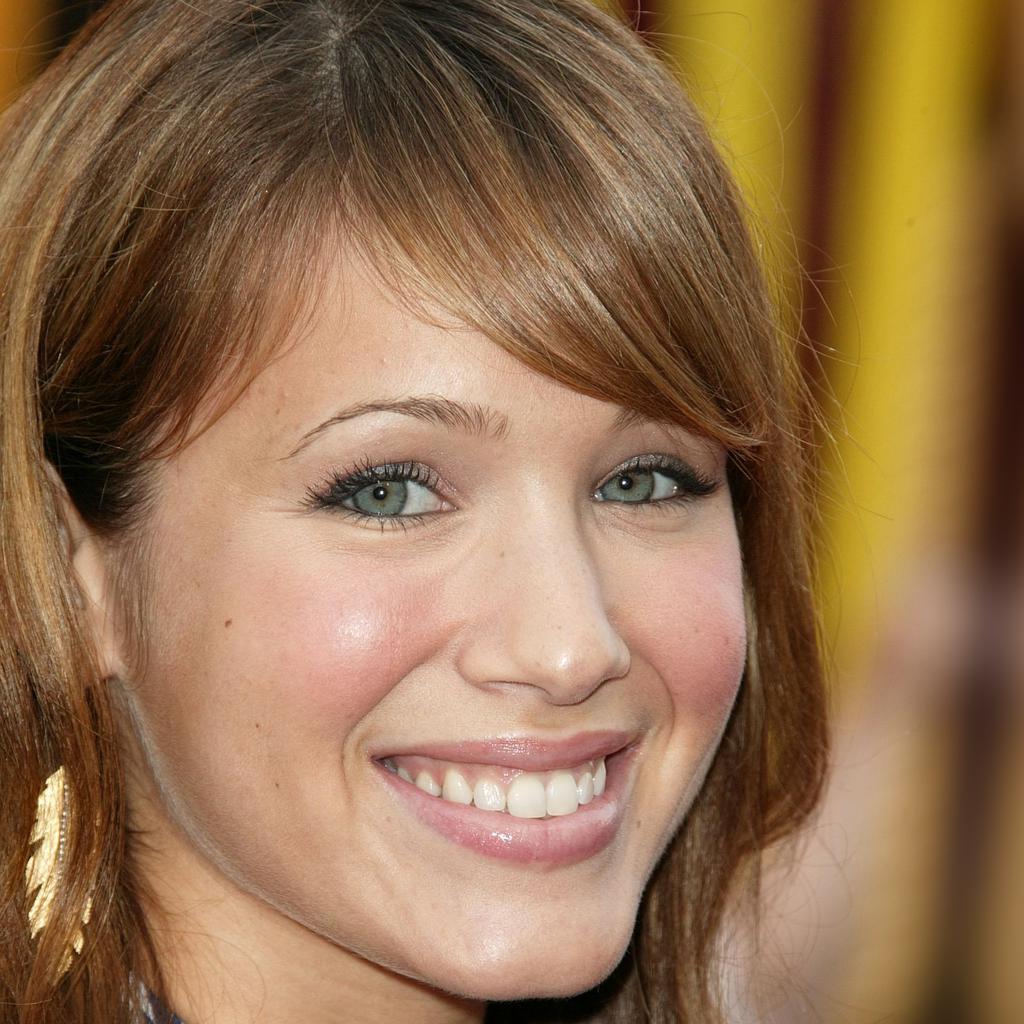} \end{tabular} & 
 \begin{tabular}{@{}c@{}} I 0 \\ \includegraphics[width=0.08\textwidth]{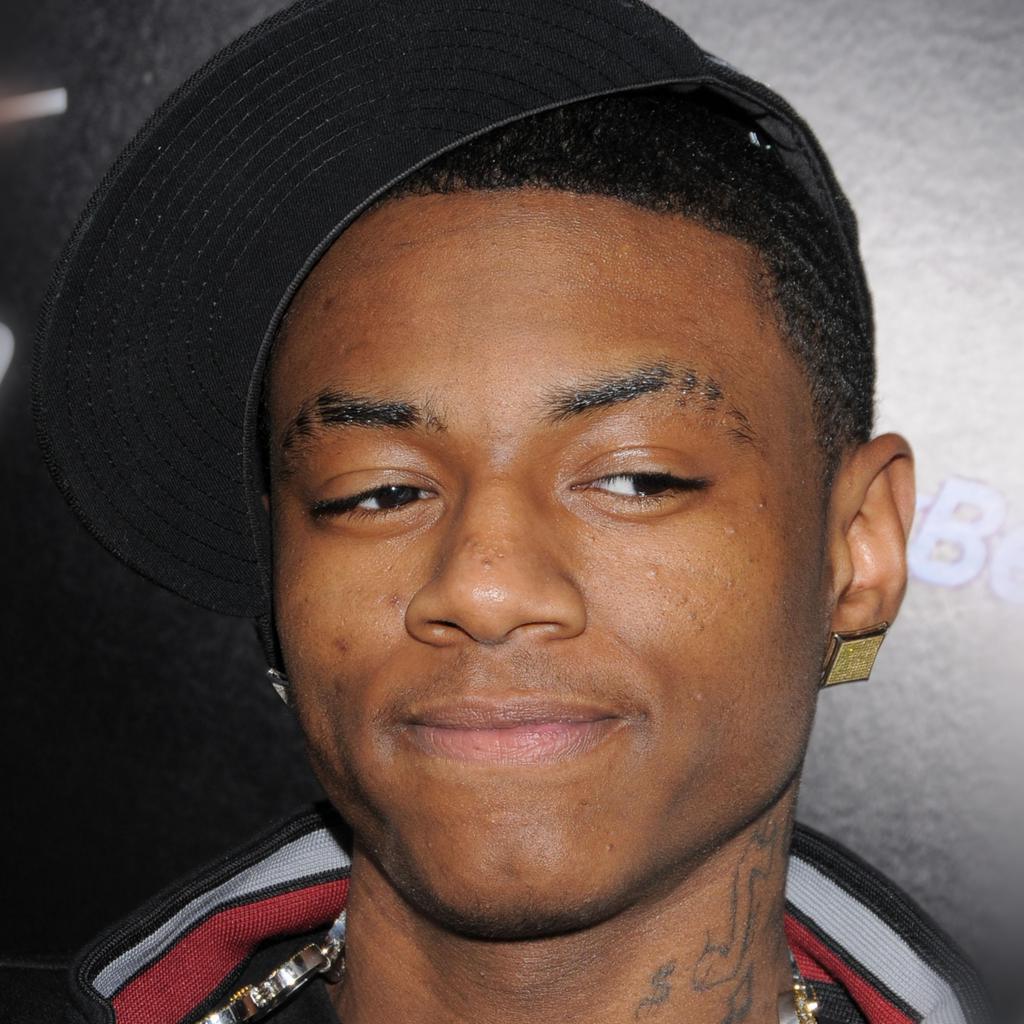} \end{tabular} & 
 \begin{tabular}{@{}c@{}} Ref  \\ \includegraphics[width=0.08\textwidth]{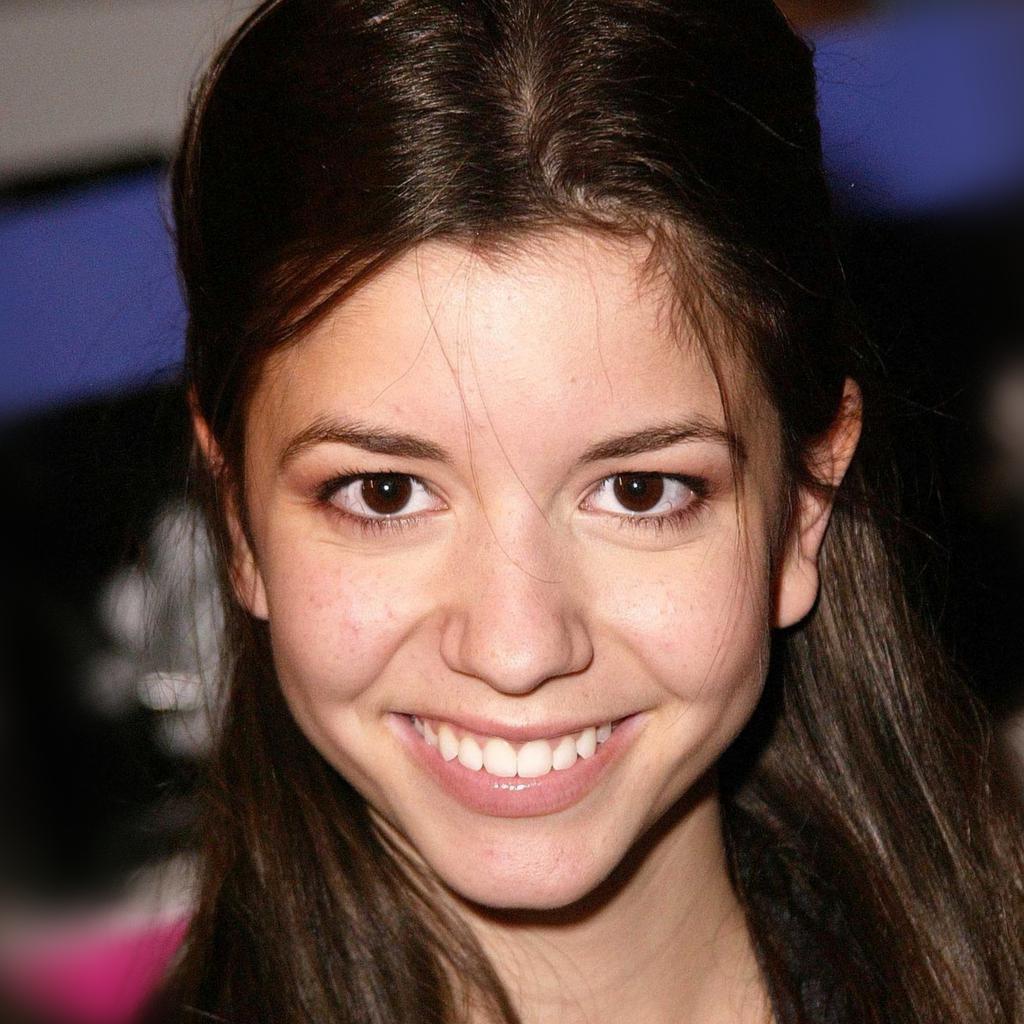} \end{tabular} & 
 \begin{tabular}{@{}c@{}} I 1 \\ \includegraphics[width=0.08\textwidth]{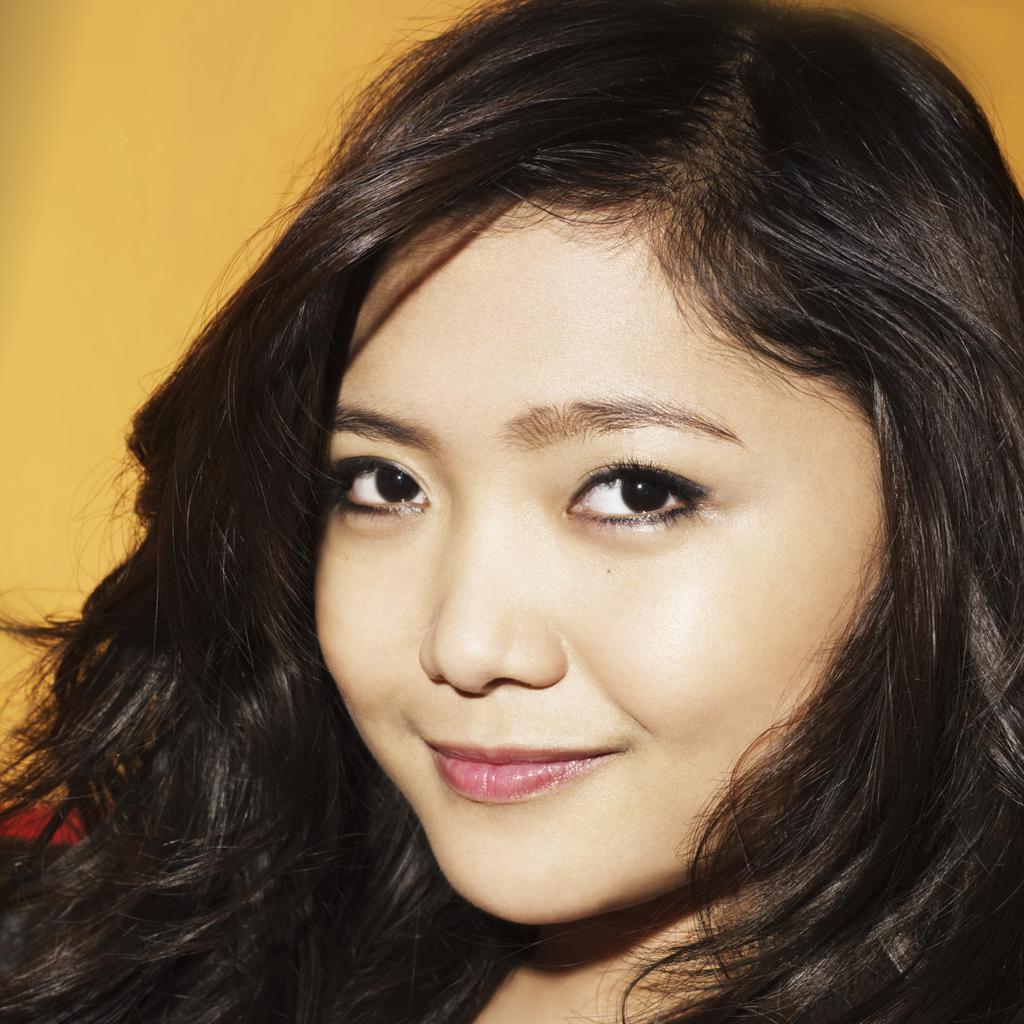} \end{tabular} & 
 \begin{tabular}{@{}c@{}} I 0 \\ \includegraphics[width=0.08\textwidth]{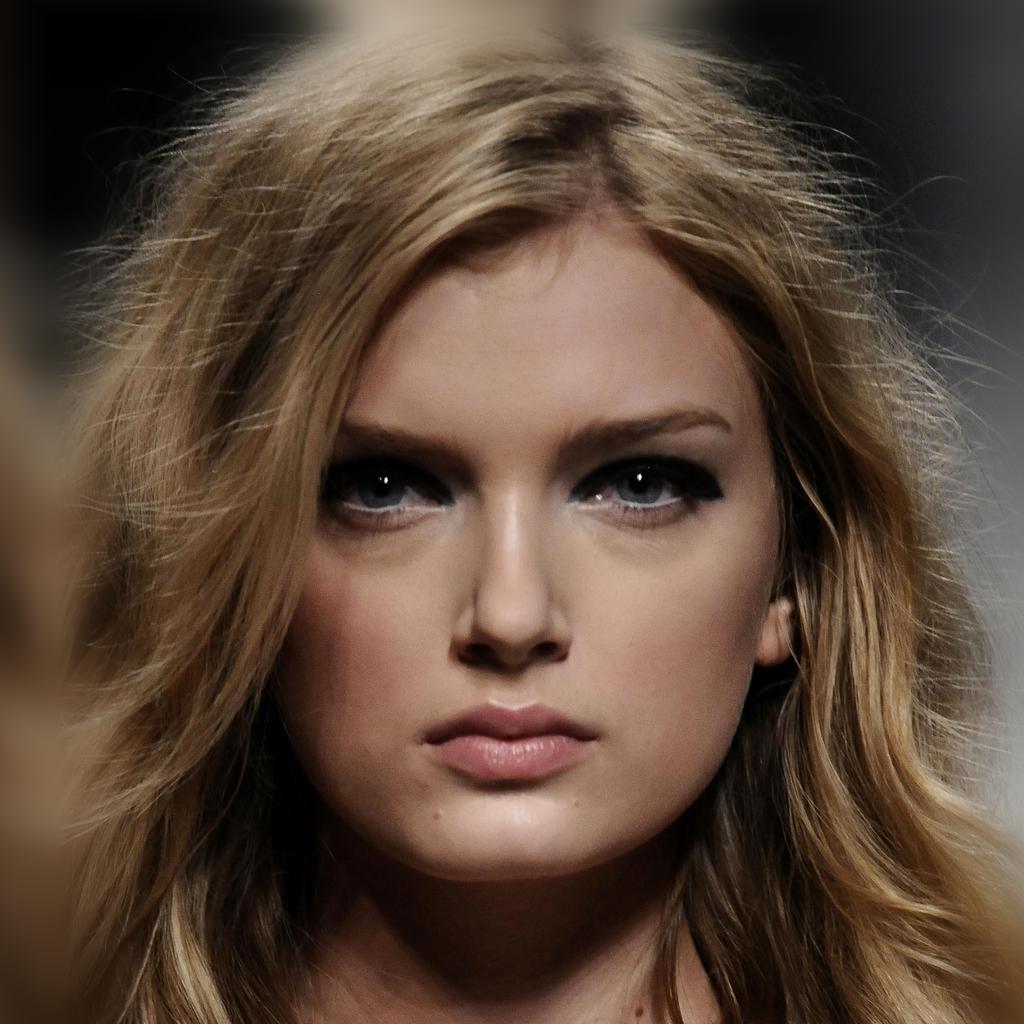} \end{tabular} &
 \begin{tabular}{@{}c@{}} Ref  \\ \includegraphics[width=0.08\textwidth]{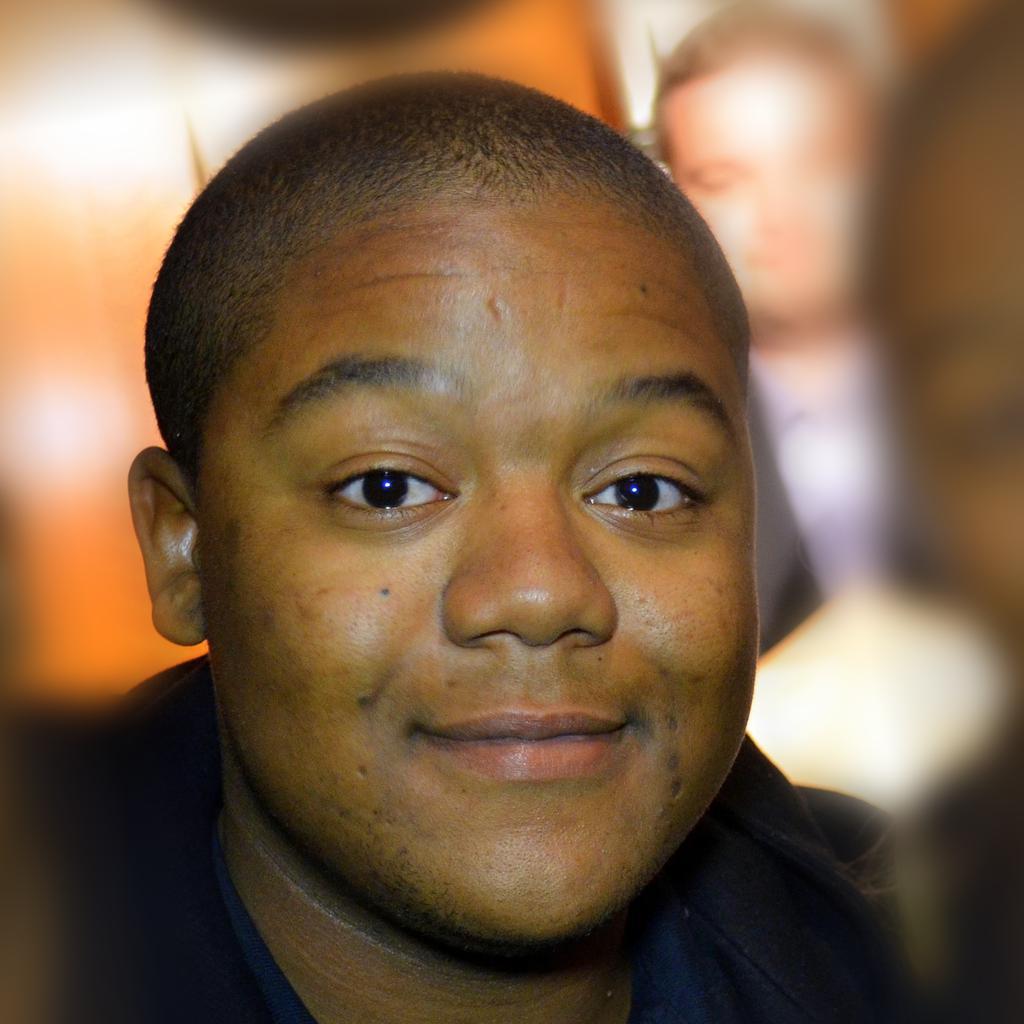} \end{tabular} &
 \begin{tabular}{@{}c@{}} I 1 \\ \includegraphics[width=0.08\textwidth]{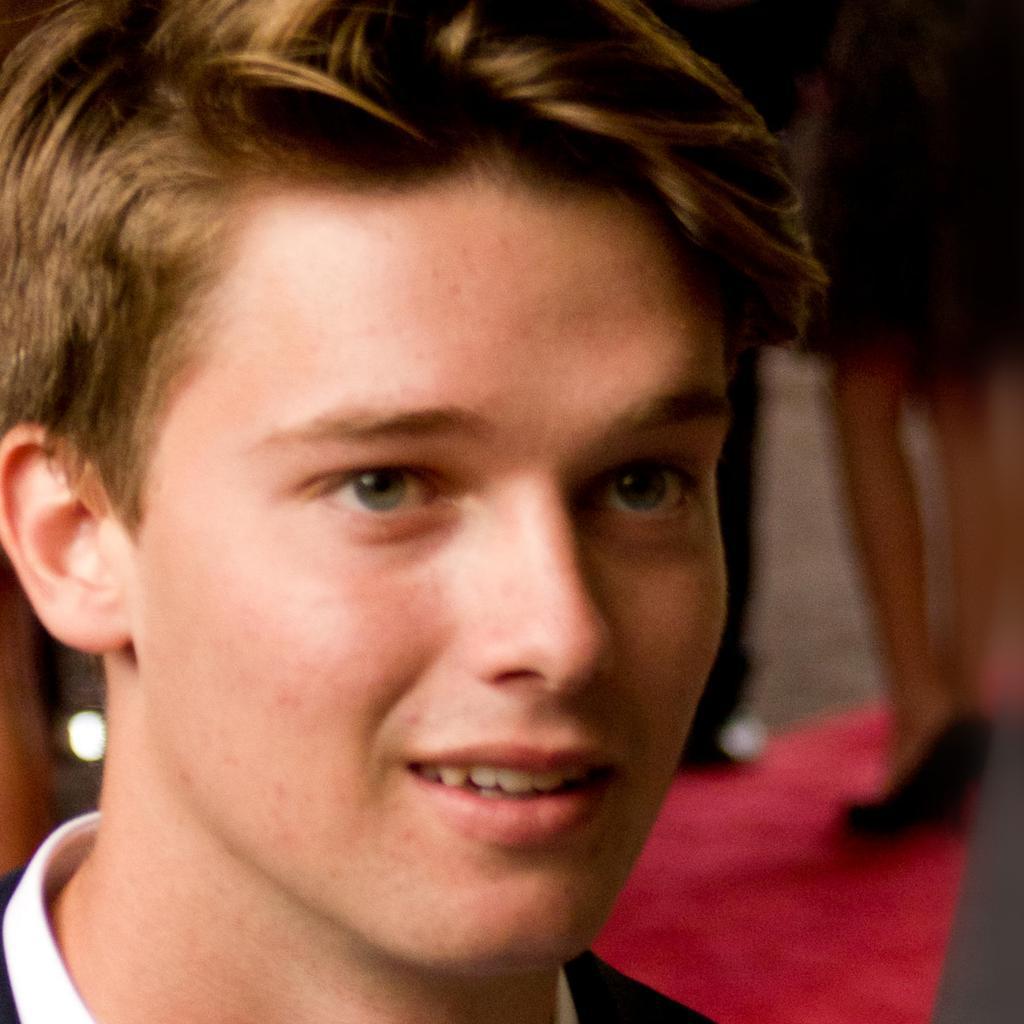} \end{tabular} \\

$D_{perceptual}$ &
   & 
   &
   \cmark &
   & 
   &
   \cmark &
   \cmark & 
   &\\
   
\hline
$D_{\text{latent}}$ &
  \cmark & 
   &
   &
  \cmark & 
   &
   &
   & 
   &
  \cmark \\
  \hline
$D^{*}_{\text{latent}}$(ours) &
   & 
   &
   \cmark &
   & 
   &
   \cmark &
   \cmark & 
   &

\end{tabular}
\caption{2AFC results on the triplets (middle reference image, on the left and right are the images to be compared with the reference one). $D_{perceptual}$, $D_{\text{latent}}$, and $D^{*}_{\text{latent}}$ are the VGG16 perceptual distance, Euclidean latent distance in \wplus and \wstar respectively. Compared to the distance computed in \wplussp, the distance in \wstar agrees with the perceptual one.}
\label{fig:2afc_att_inter_results}

\end{figure}

\subsubsection{Image Editing}
\label{sec:editing}
Here we qualitatively demonstrate the benefits of the new proxy space for the image editing task. InterFaceGAN \cite{shen2020interfacegan} was retrained to manipulate the attributes of a given real image in both \wplus and $\mathcal{W}^{\star}_{ID}$ (eq. \eqref{eq:att_id} was optimised). InterFaceGAN assumes that the positive and negative examples of each attribute are linearly separable, and the editing direction is simply the normal to the hyperplane that separates the positive and negative regions. These normal directions were obtained after training an SVM for each attribute in both spaces. The edited images were generated after editing the latent codes in \wplus or $\mathcal{W}^{\star}_{ID}$ before feeding them to the StyleGAN2 generator. For $\mathcal{W}^{\star}_{ID}$, the edited latent codes were mapped back to \wplus before feeding them to the generator. 
\begin{figure}[t]
\setlength\tabcolsep{2pt}%
\centering
\begin{tabular}{p{0.25cm}ccccccc}
\centering
 &
 \textbf{Original} &
 \textbf{Inverted} &
 \textbf{Makeup} &
 \textbf{Male} &
 \textbf{Mustache} &
 \textbf{Chubby} &
 \textbf{Lipstick} \\
 \begin{turn}{90} \hspace{0.5cm} \wplus\end{turn} &
 \includegraphics[width=0.125\textwidth]{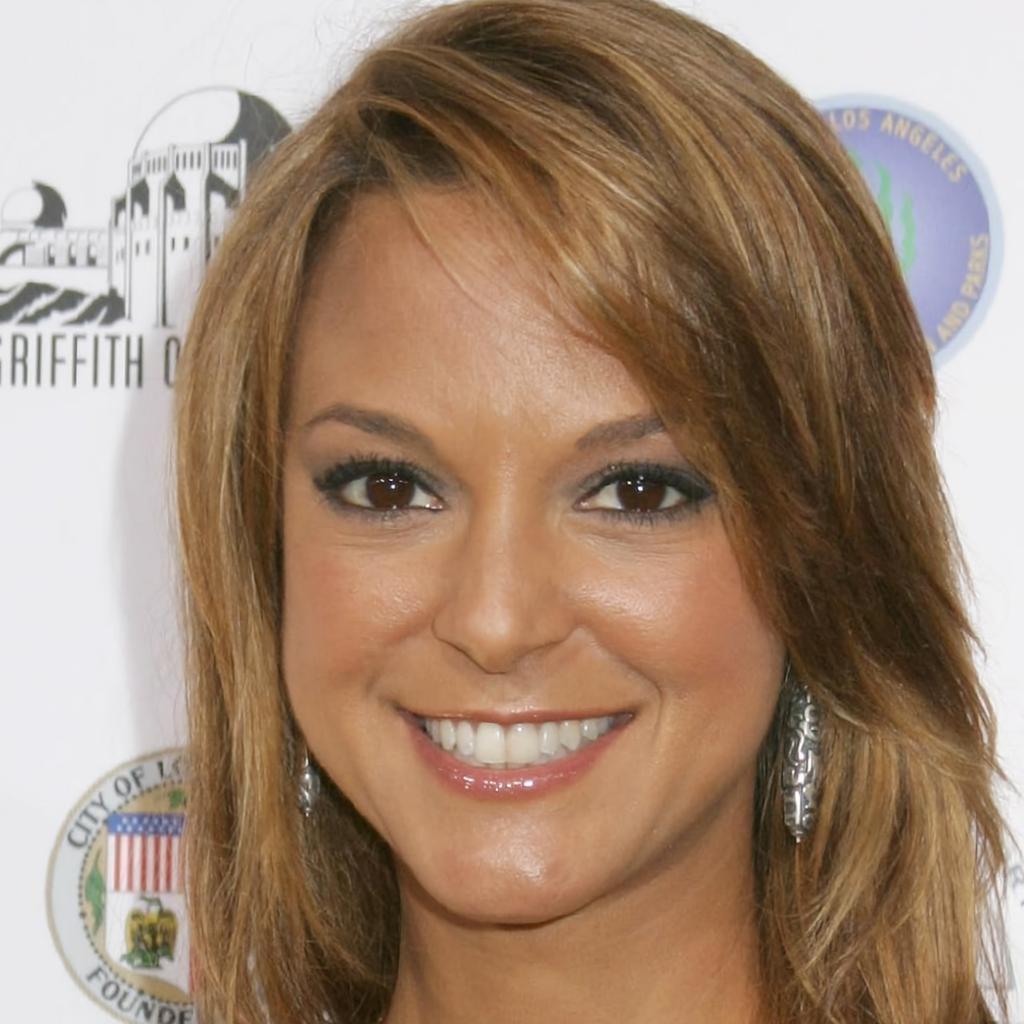} & 
 \includegraphics[width=0.125\textwidth]{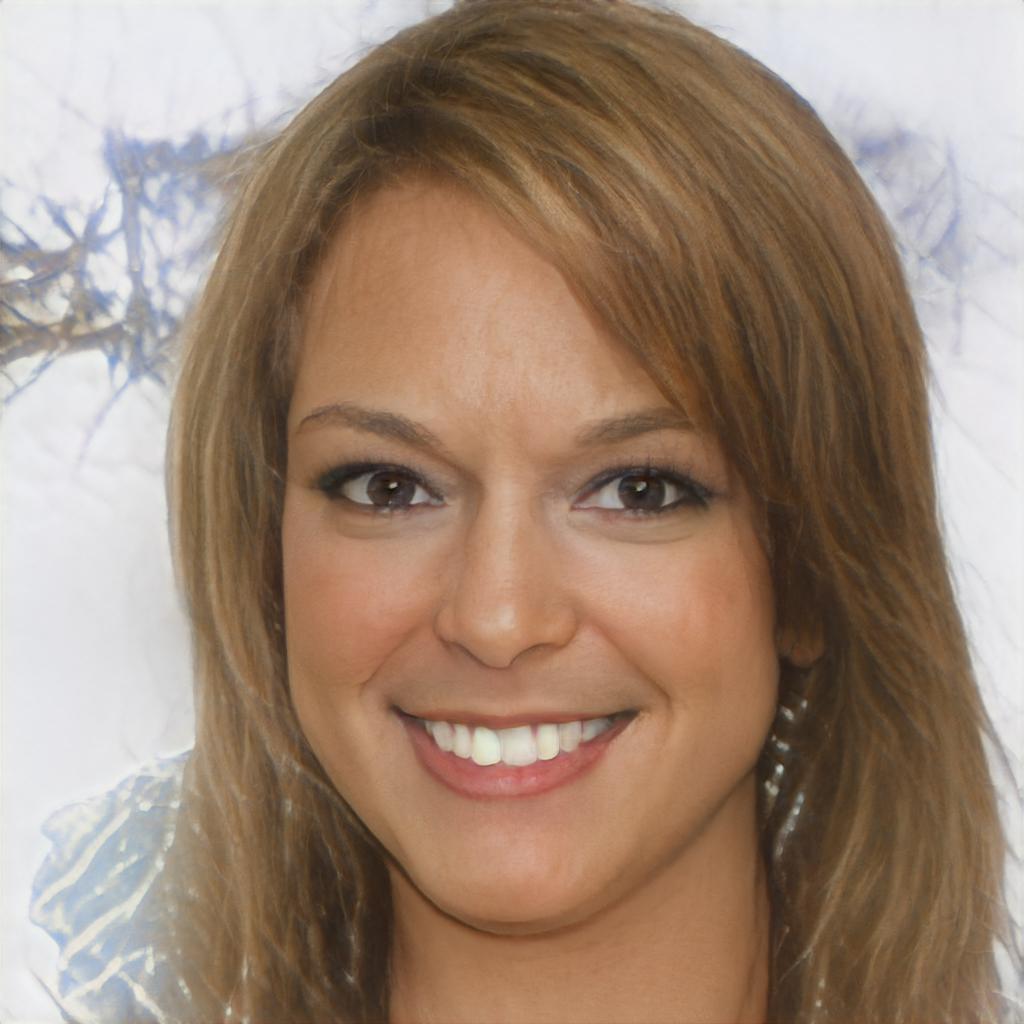} &
 \includegraphics[width=0.125\textwidth]{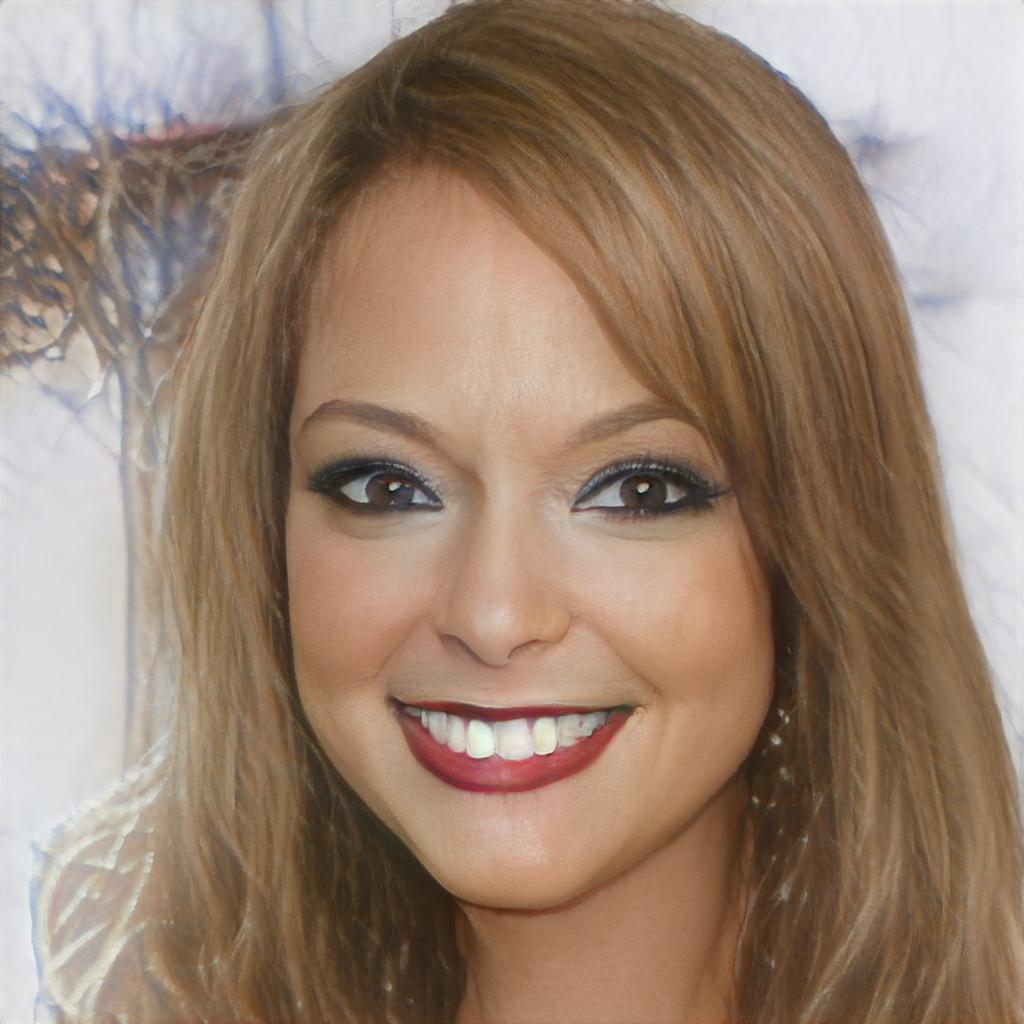} &
 \includegraphics[width=0.125\textwidth]{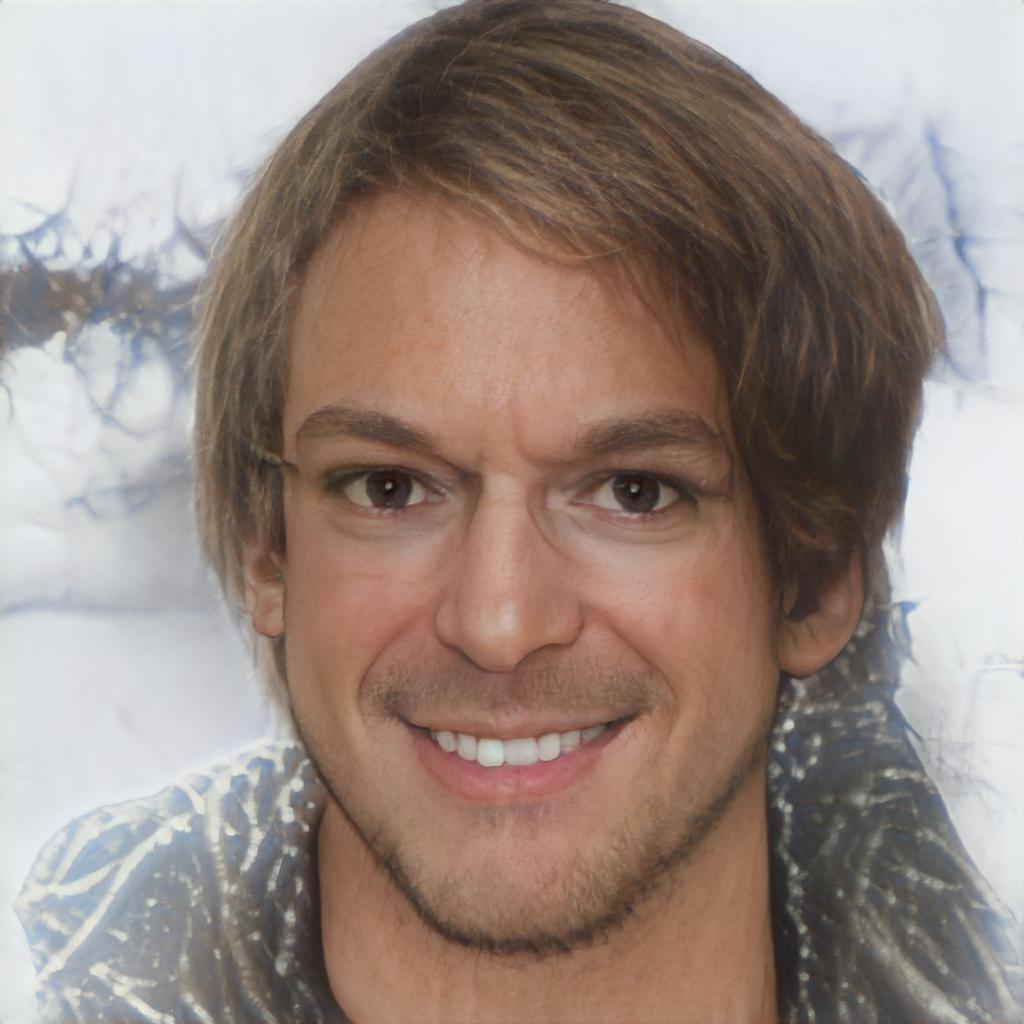} &
 \includegraphics[width=0.125\textwidth]{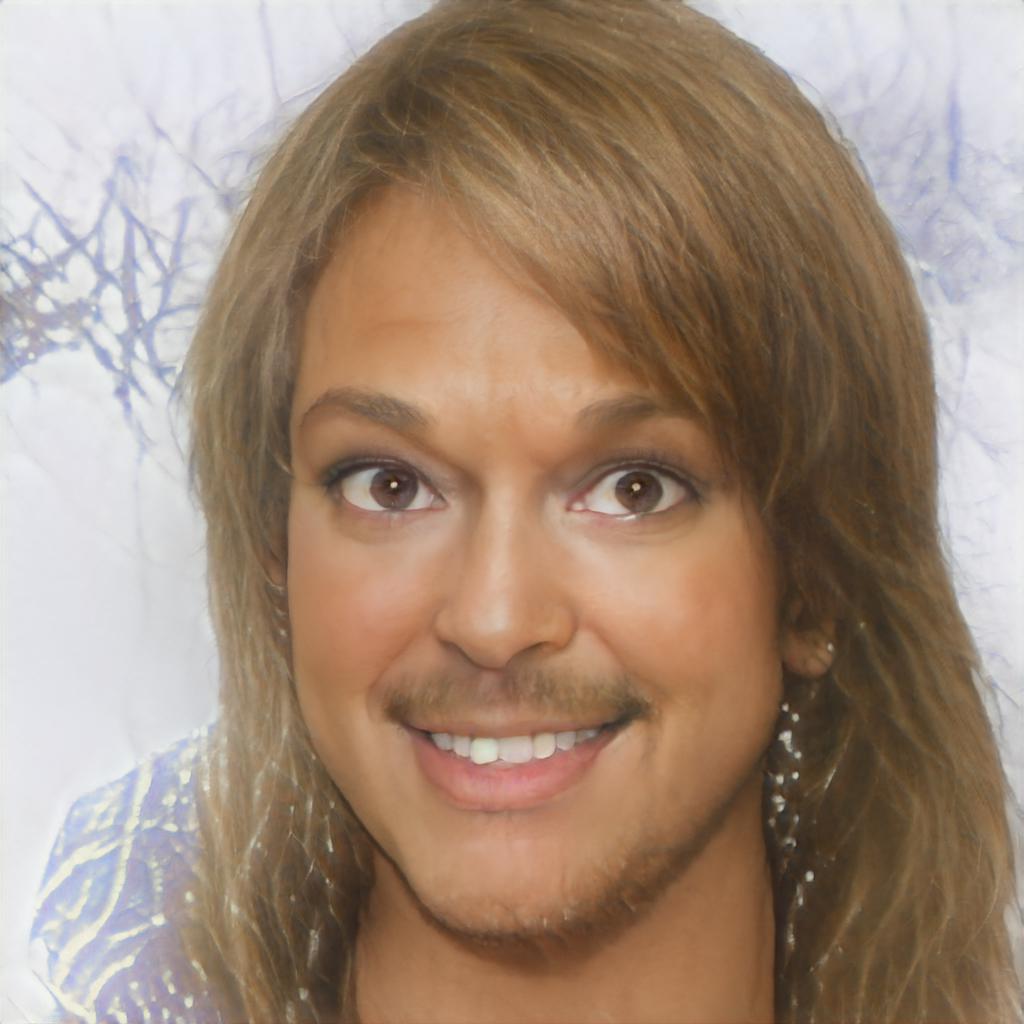} &
 \includegraphics[width=0.125\textwidth]{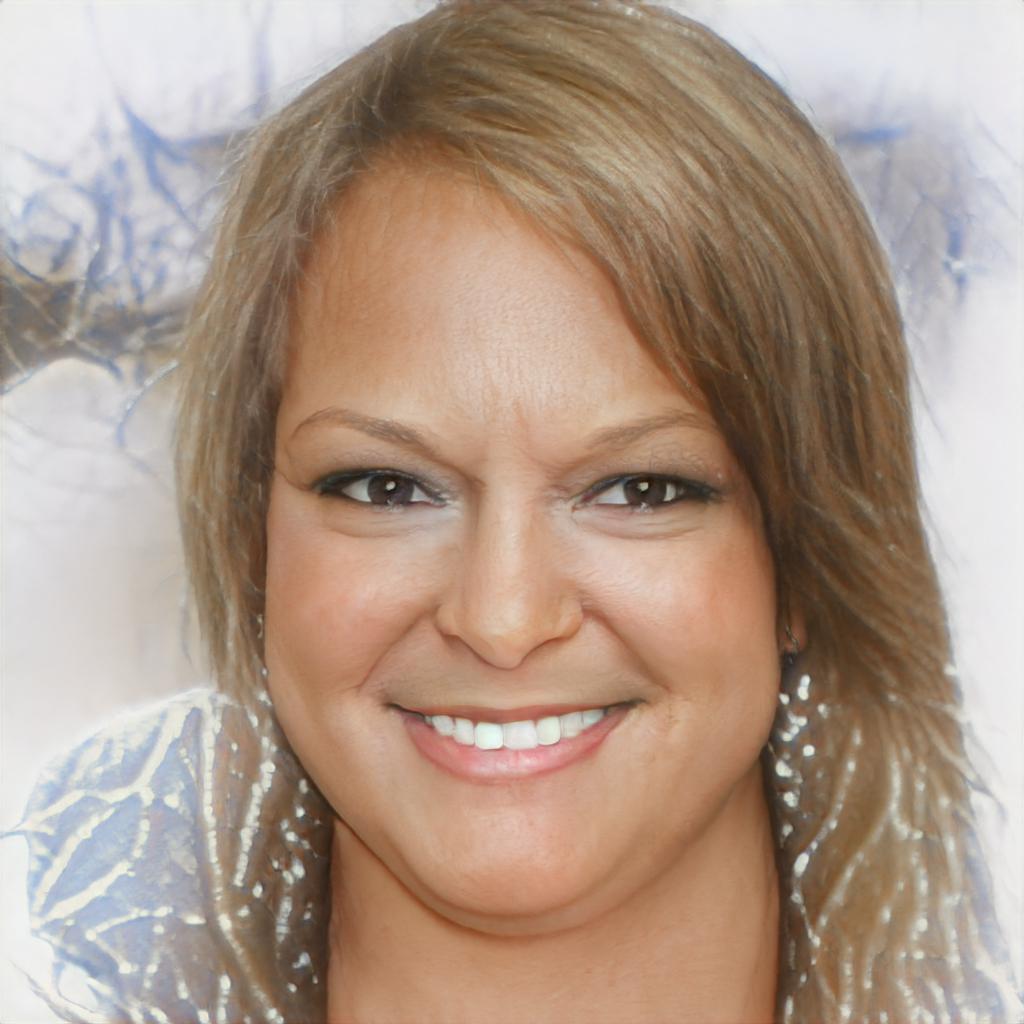} &
 \includegraphics[width=0.125\textwidth]{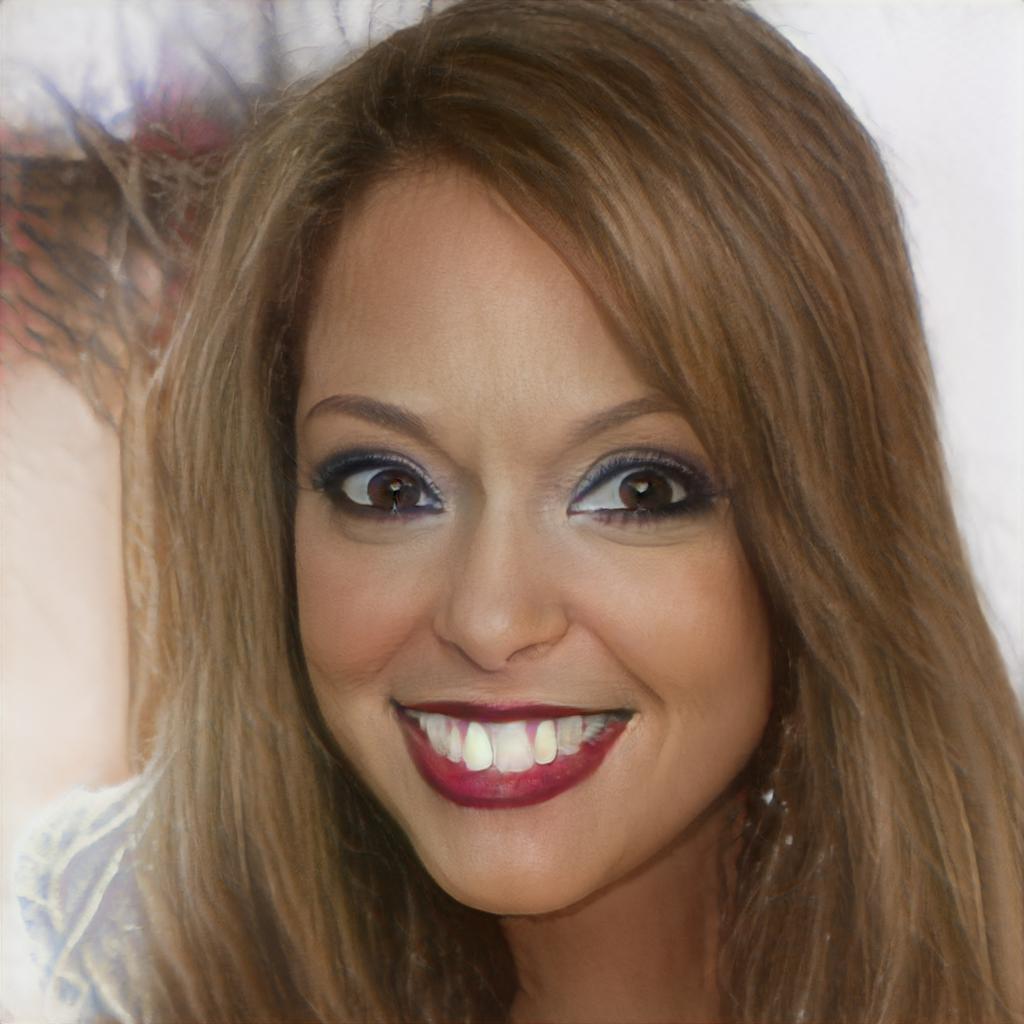} \\
 \begin{turn}{90} \hspace{0.5cm} $\mathcal{W}^{\star}_{ID}$ \end{turn} &
 \includegraphics[width=0.125\textwidth]{images/original/06002.jpg} & 
 \includegraphics[width=0.125\textwidth]{images/inversion/18_orig_img_1.jpg} &
 \includegraphics[width=0.125\textwidth, ]{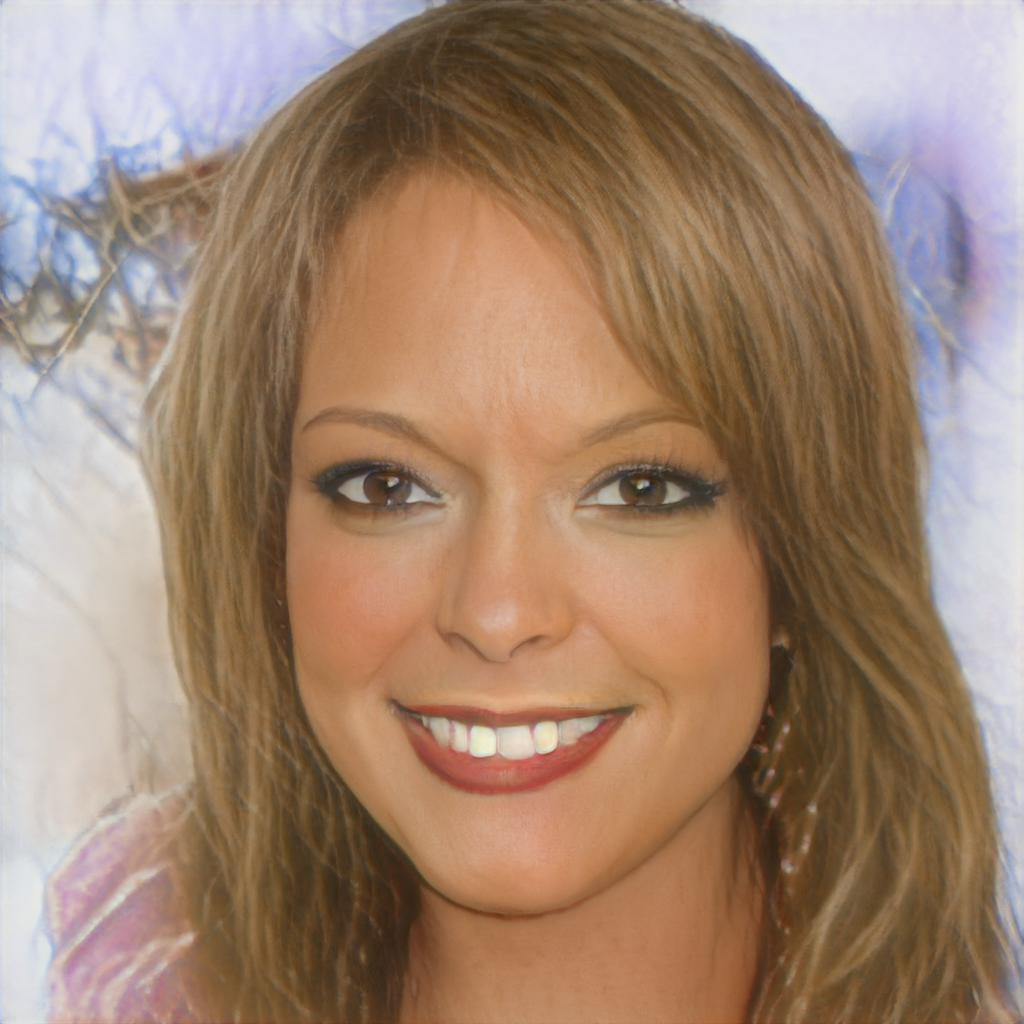} &
 \includegraphics[width=0.125\textwidth, ]{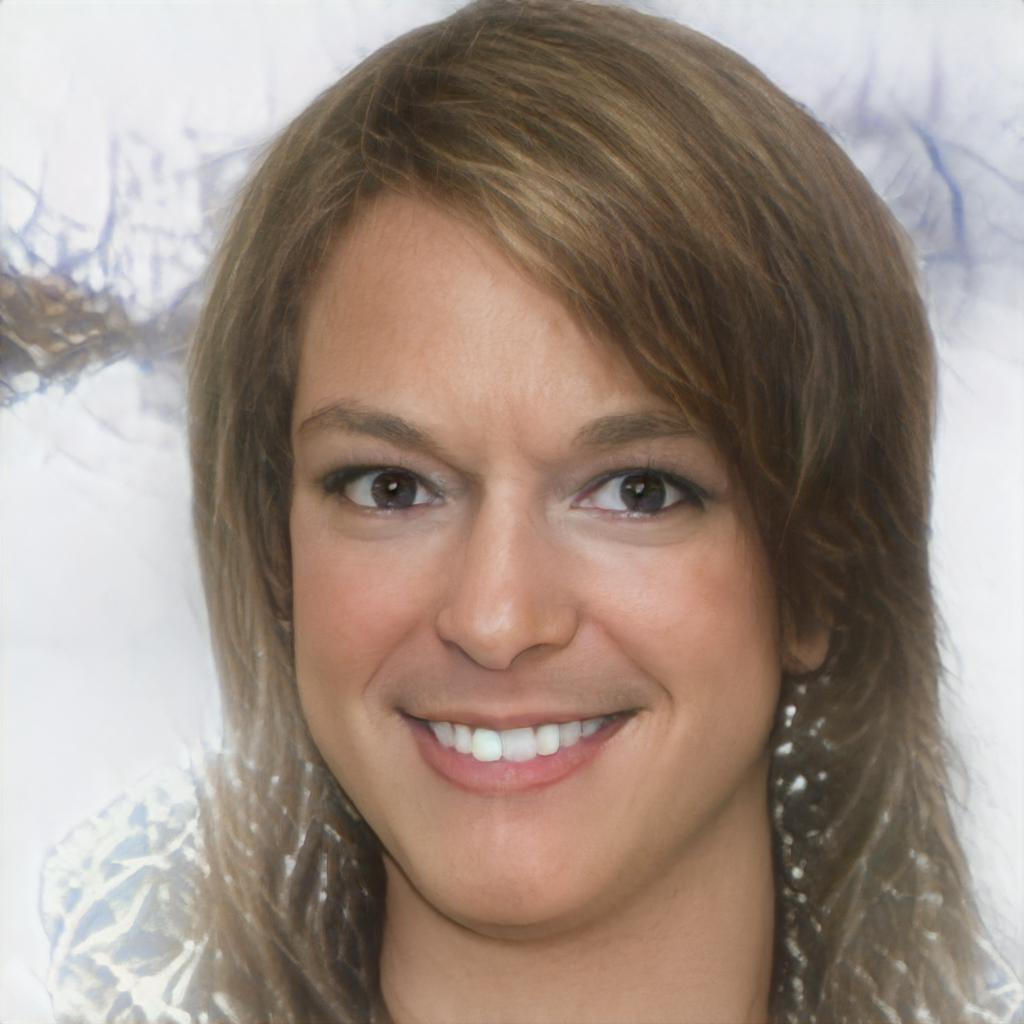} &
 \includegraphics[width=0.125\textwidth, ]{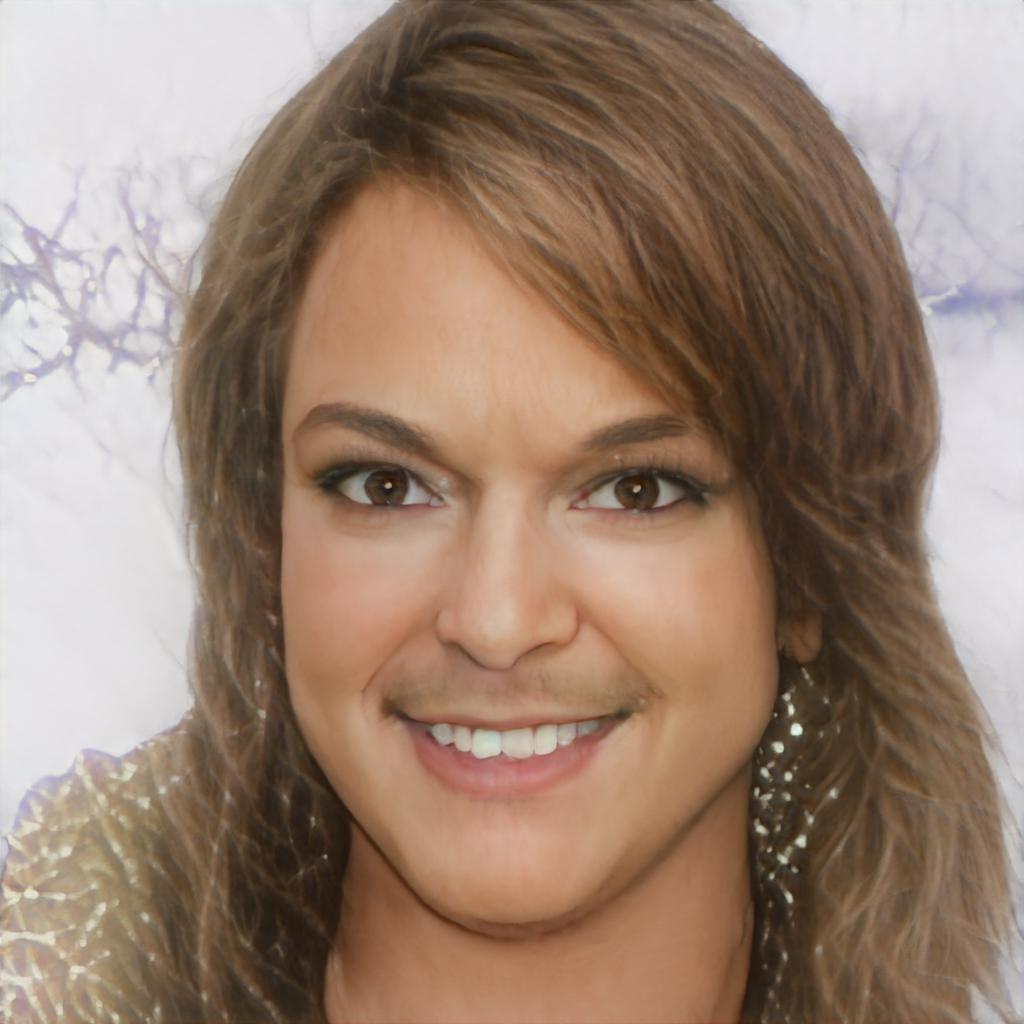} &
 \includegraphics[width=0.125\textwidth, ]{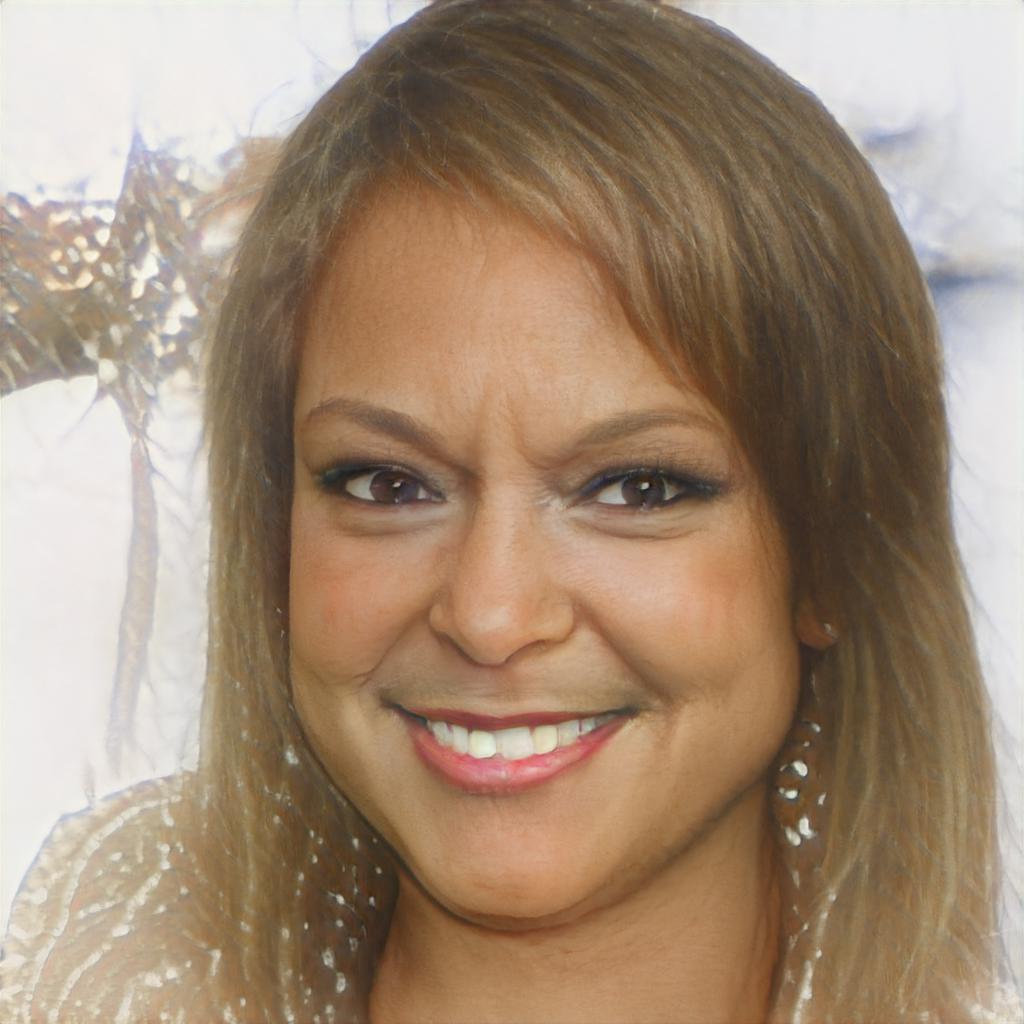} &
 \includegraphics[width=0.125\textwidth, ]{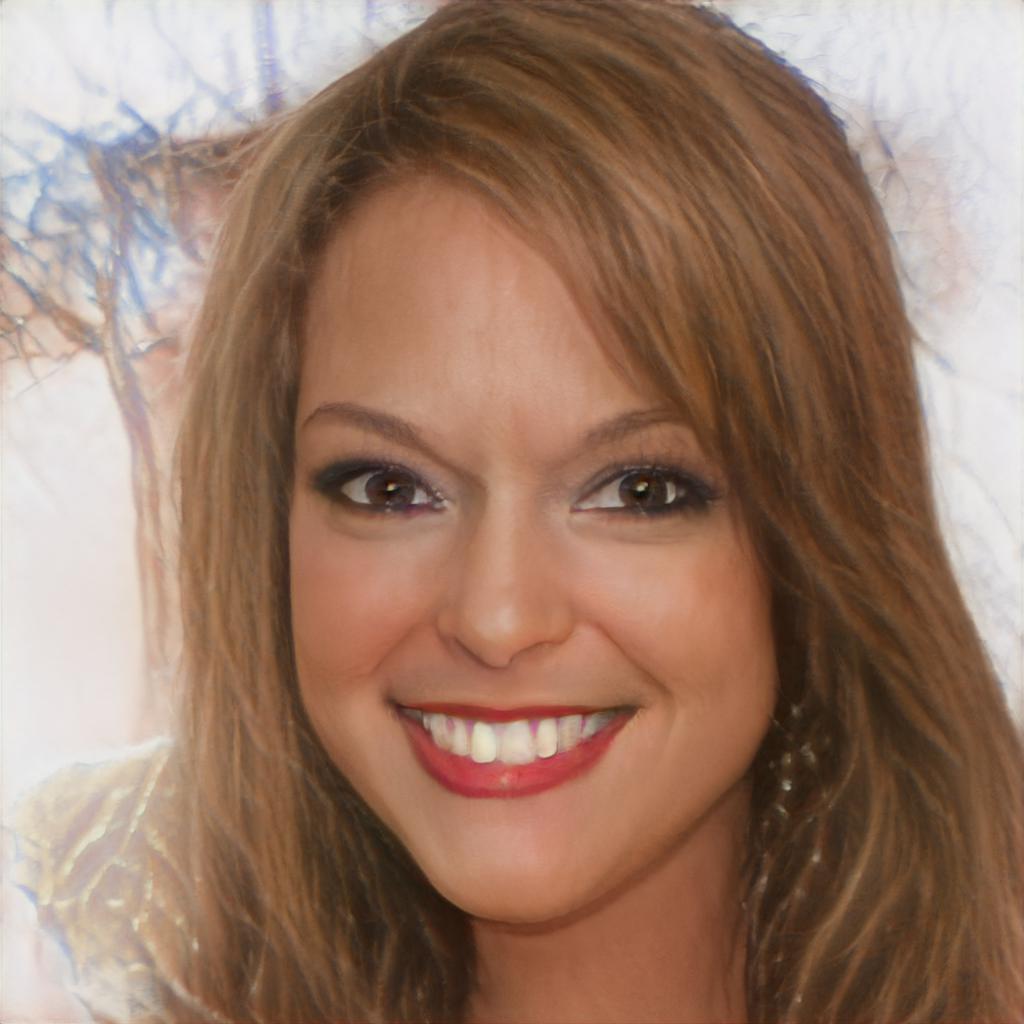} \\
 \midrule
 \begin{turn}{90} \hspace{0.5cm} \wplus\end{turn} &
 \includegraphics[width=0.125\textwidth]{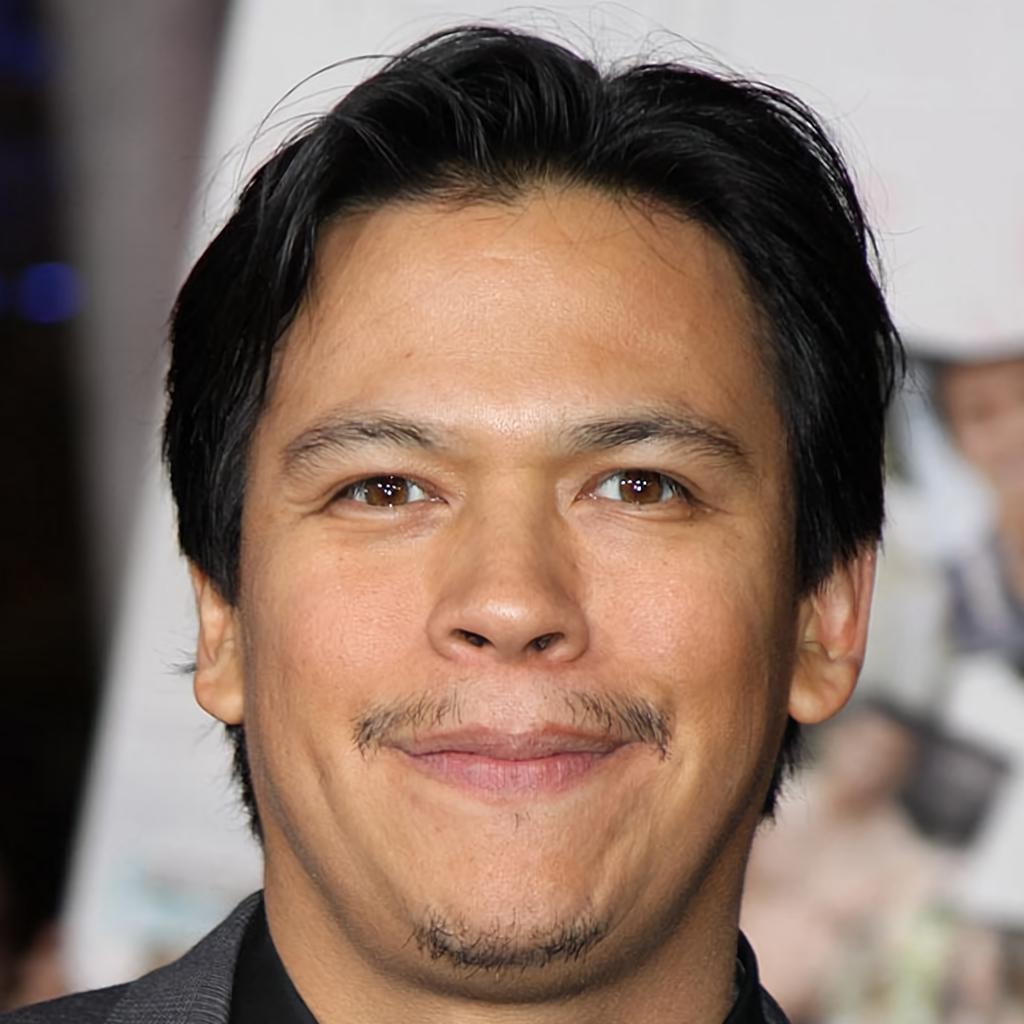} & 
 \includegraphics[width=0.125\textwidth, ]{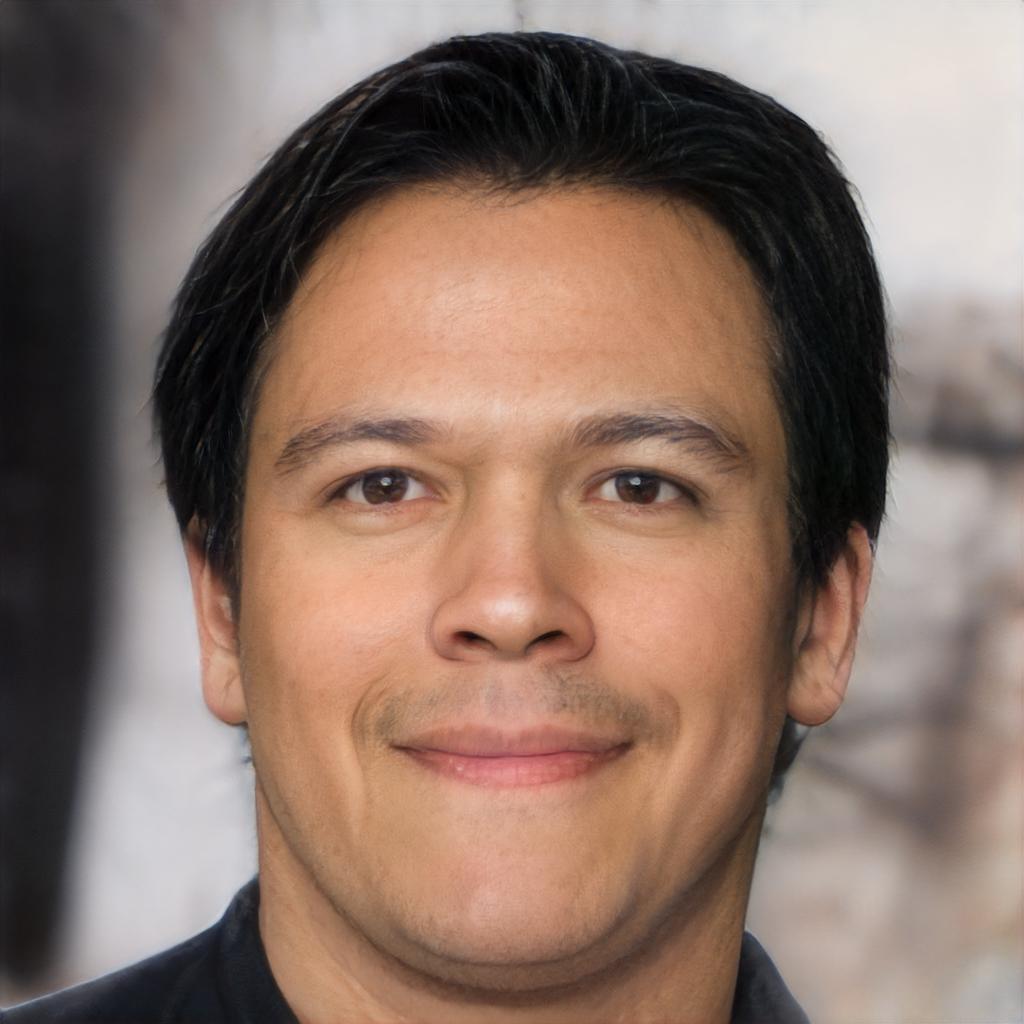} &
 \includegraphics[width=0.125\textwidth, ]{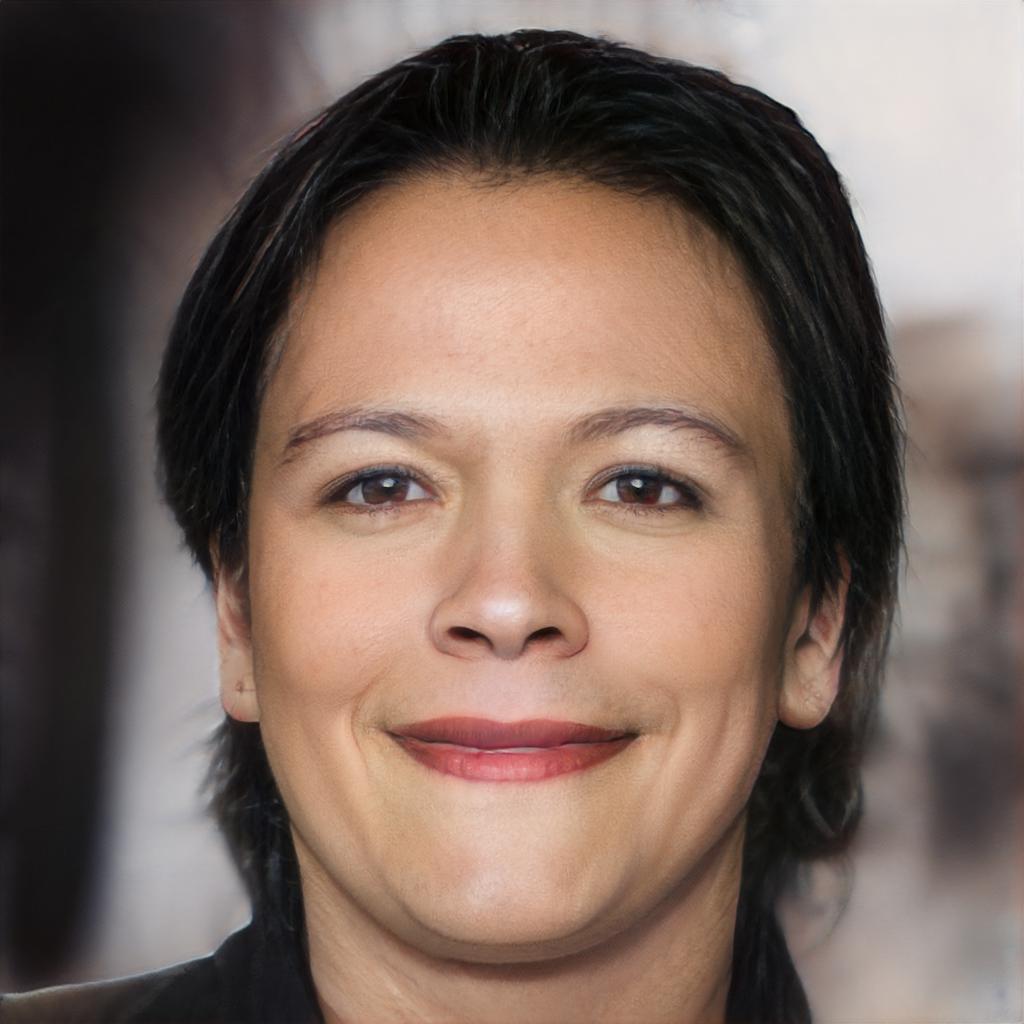} &
 \includegraphics[width=0.125\textwidth, ]{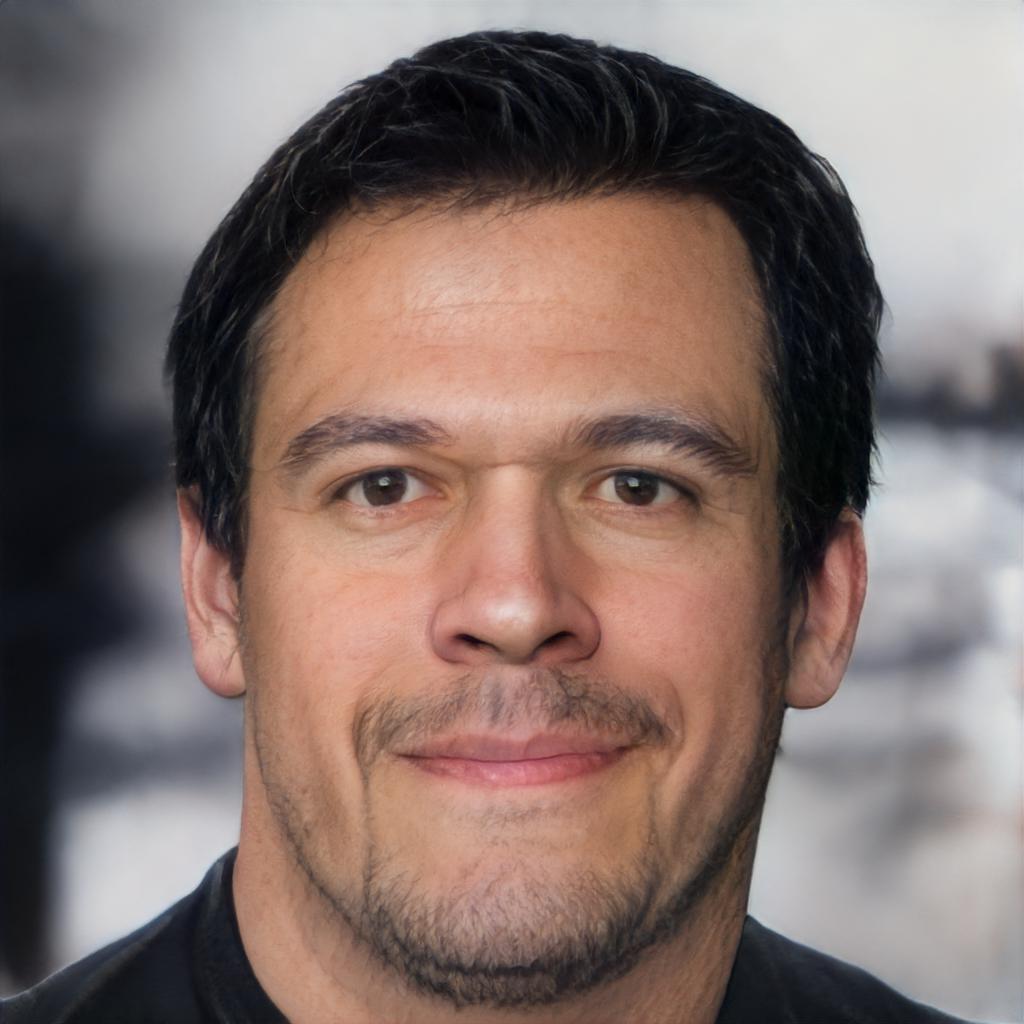} &
 \includegraphics[width=0.125\textwidth, ]{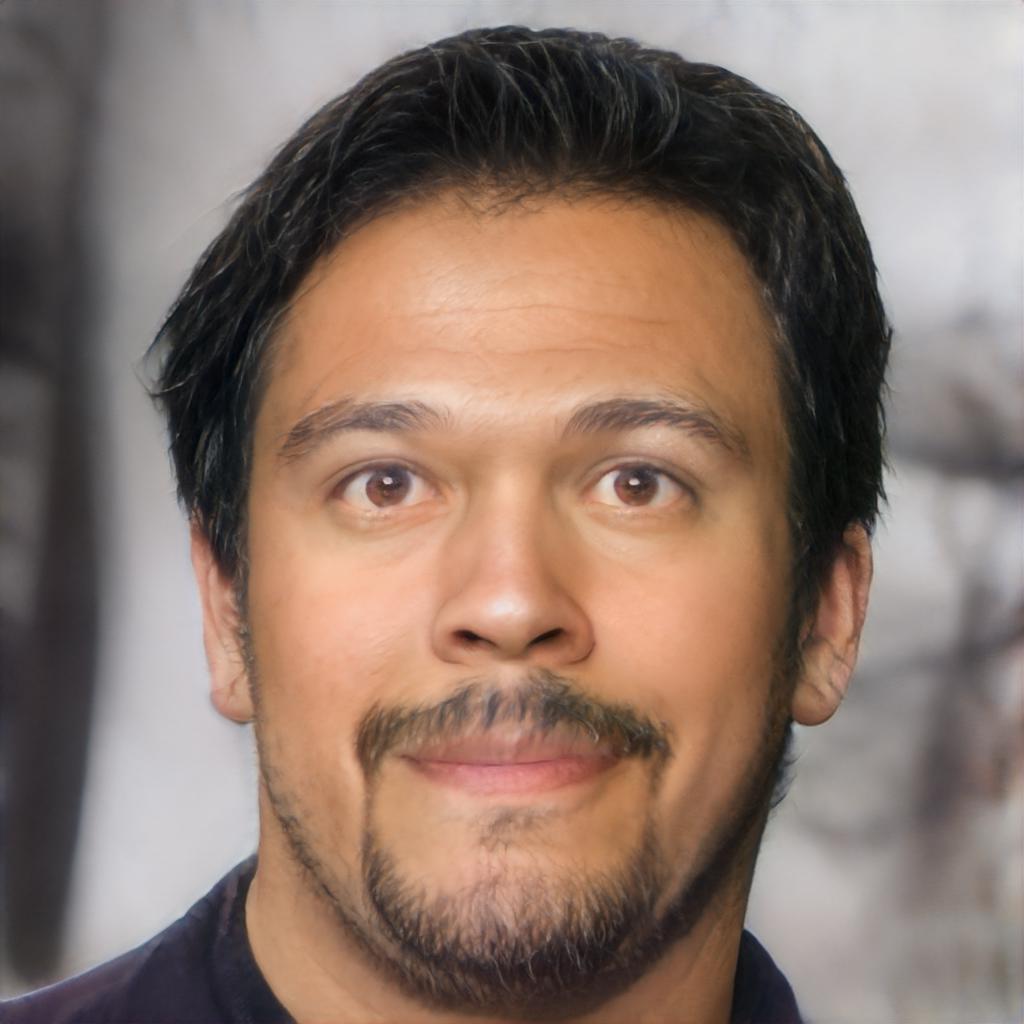} &
 \includegraphics[width=0.125\textwidth, ]{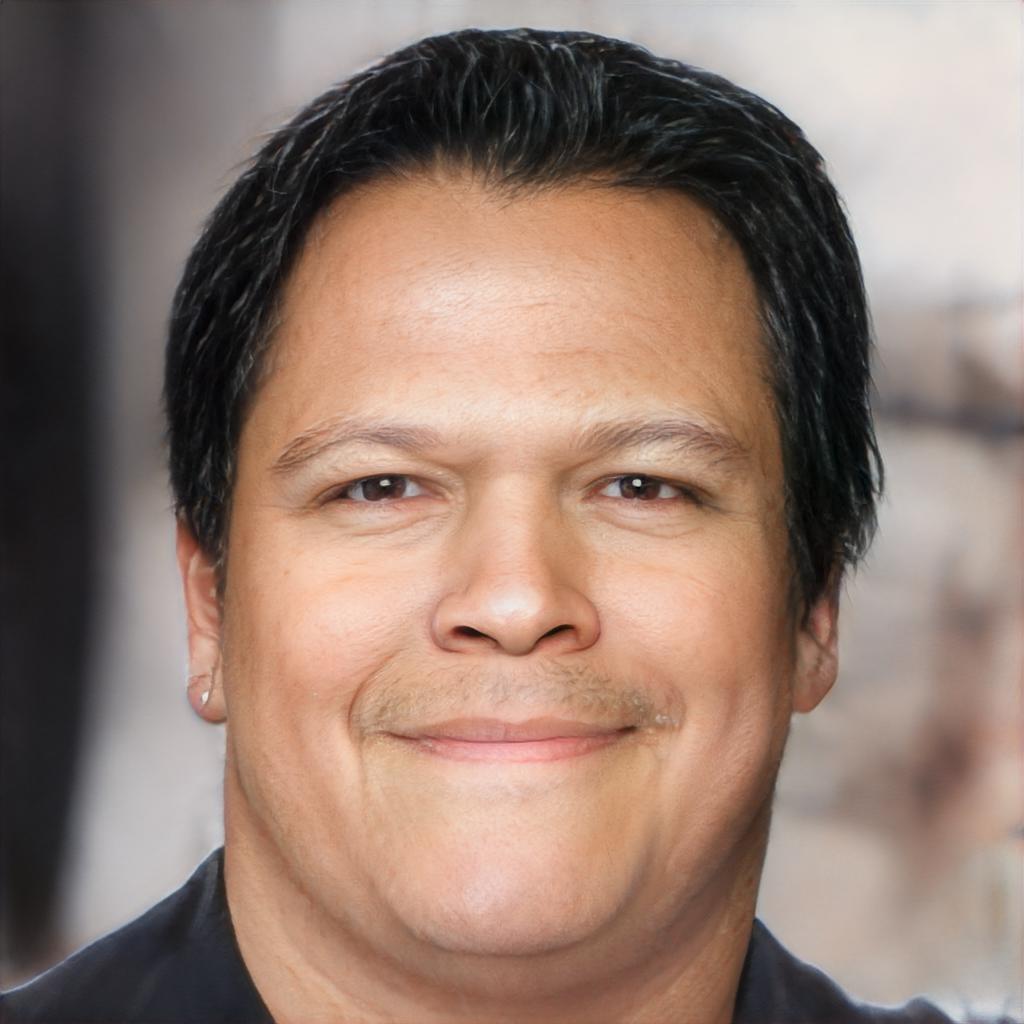} &
 \includegraphics[width=0.125\textwidth]{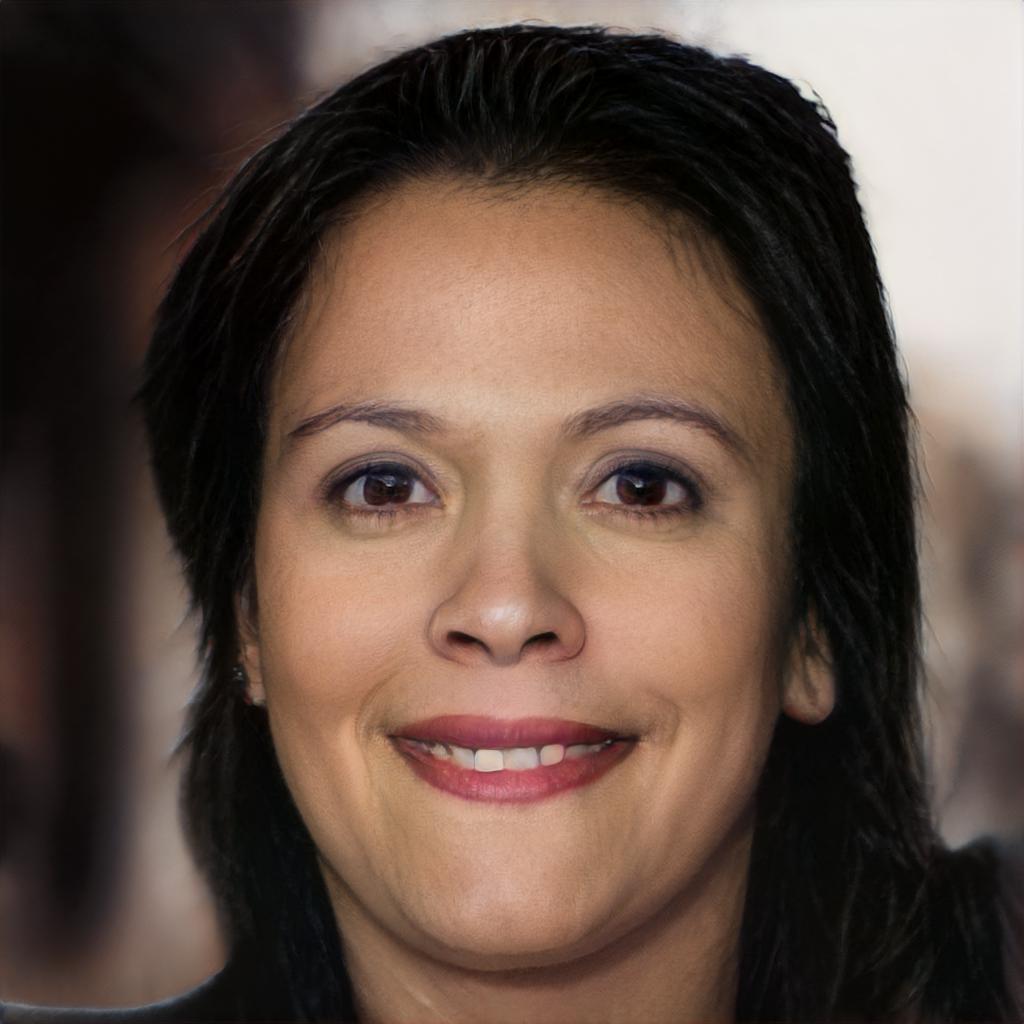} \\
 \begin{turn}{90} \hspace{0.5cm} $\mathcal{W}^{\star}_{ID}$ \end{turn} &
 \includegraphics[width=0.125\textwidth]{images/original/06004.jpg} & 
 \includegraphics[width=0.125\textwidth, ]{images/inversion/18_orig_img_3.jpg} &
 \includegraphics[width=0.125\textwidth, ]{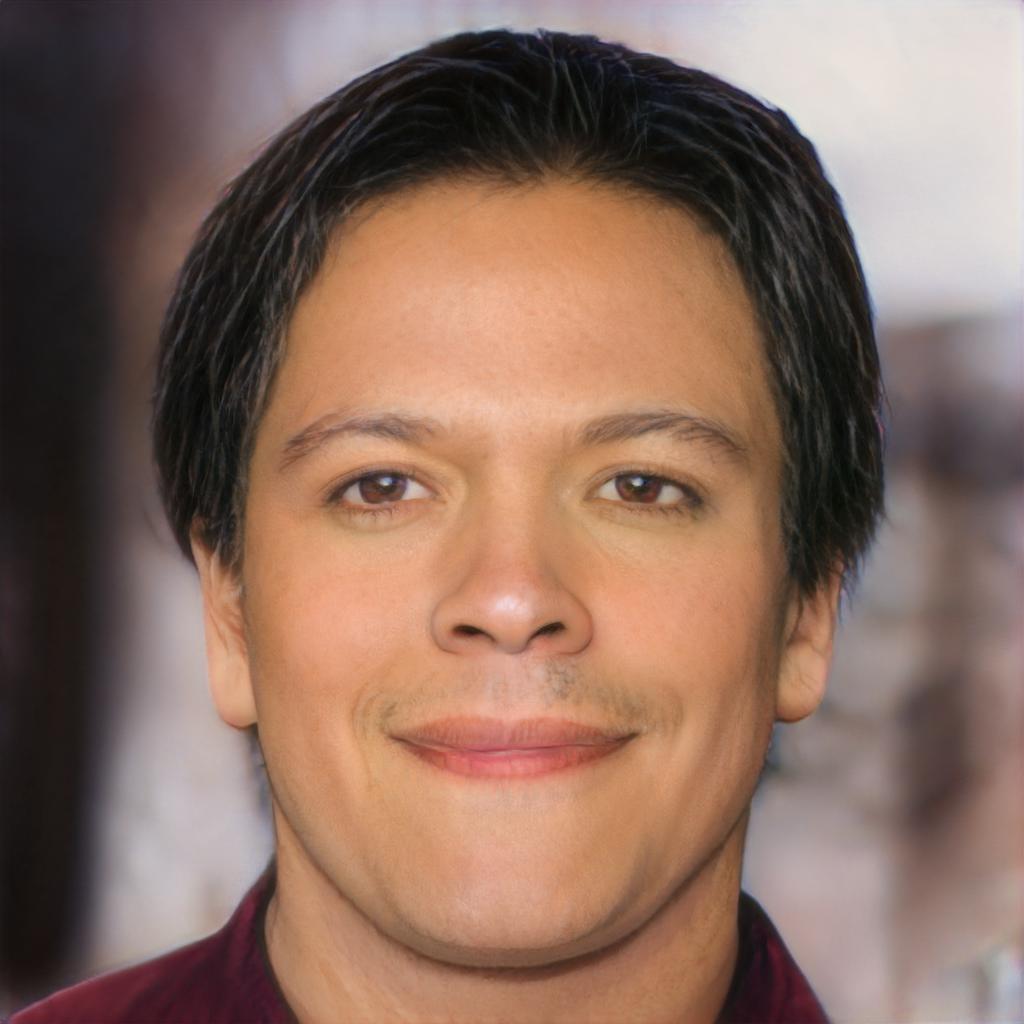} &
 \includegraphics[width=0.125\textwidth, ]{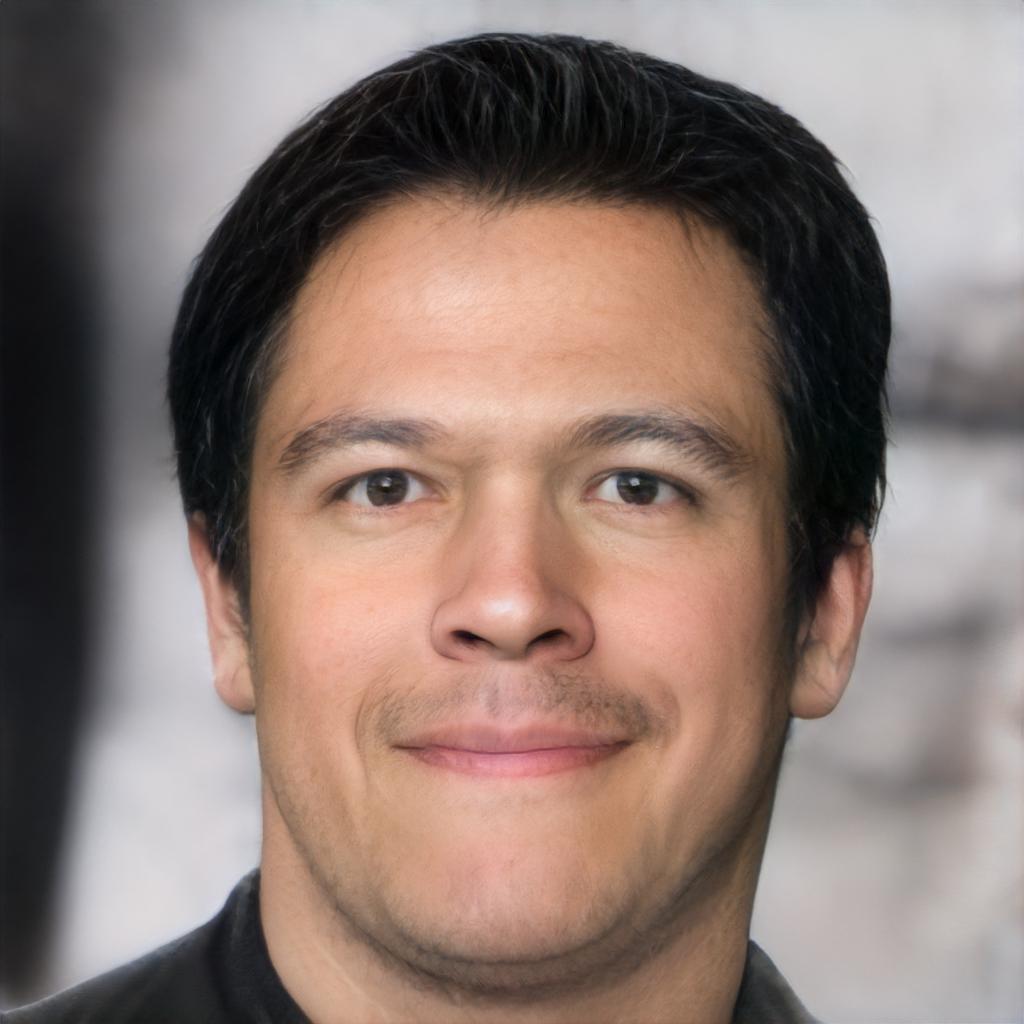} &
 \includegraphics[width=0.125\textwidth, ]{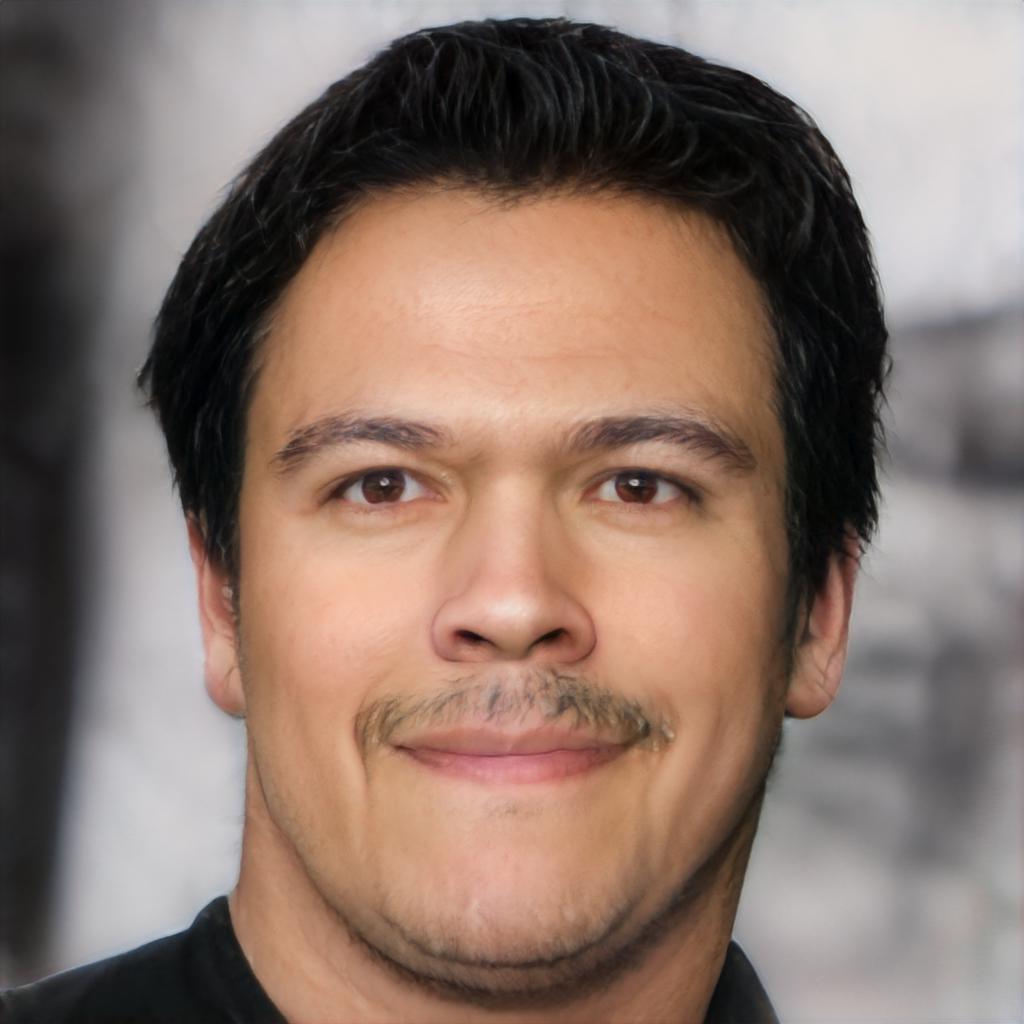} &
 \includegraphics[width=0.125\textwidth, ]{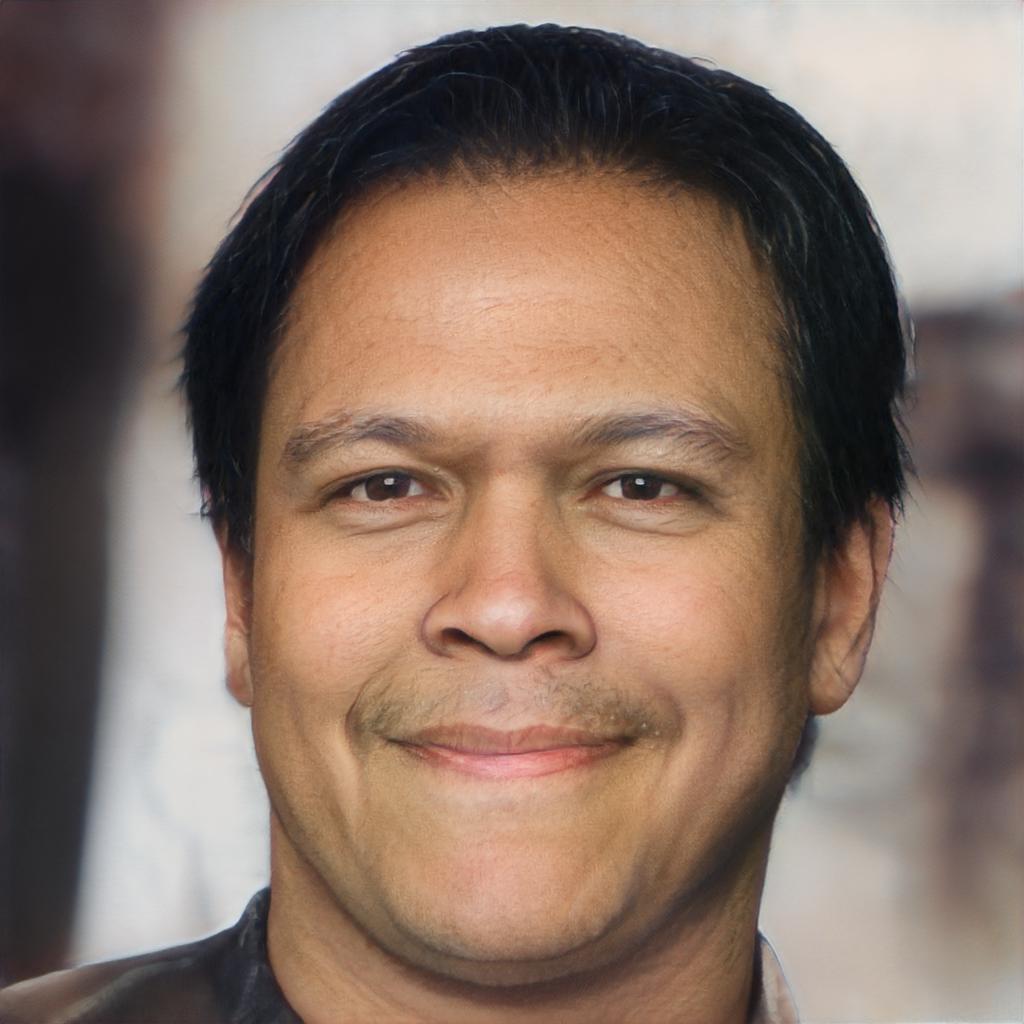} &
 \includegraphics[width=0.125\textwidth, ]{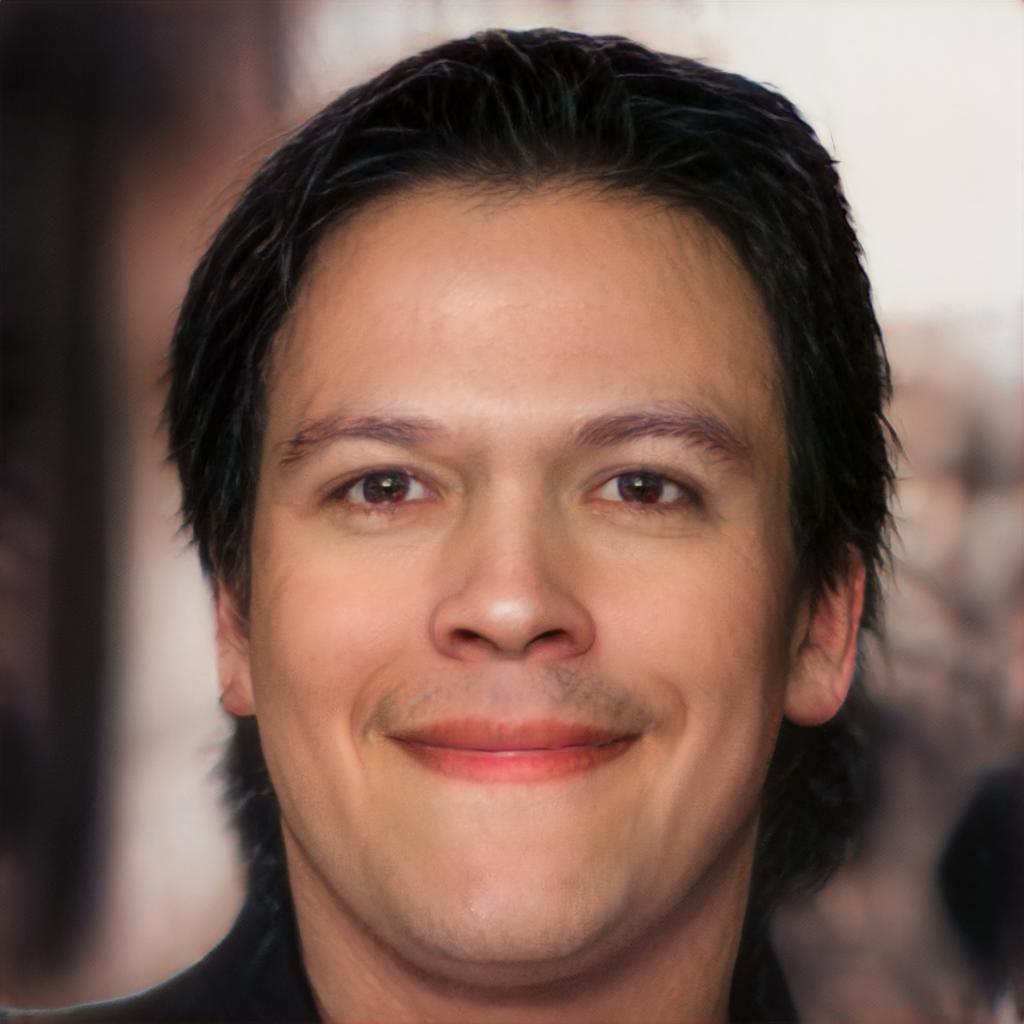} \\
 \midrule
 \begin{turn}{90} \hspace{0.5cm} \wplus \end{turn} &
 \includegraphics[width=0.125\textwidth, ]{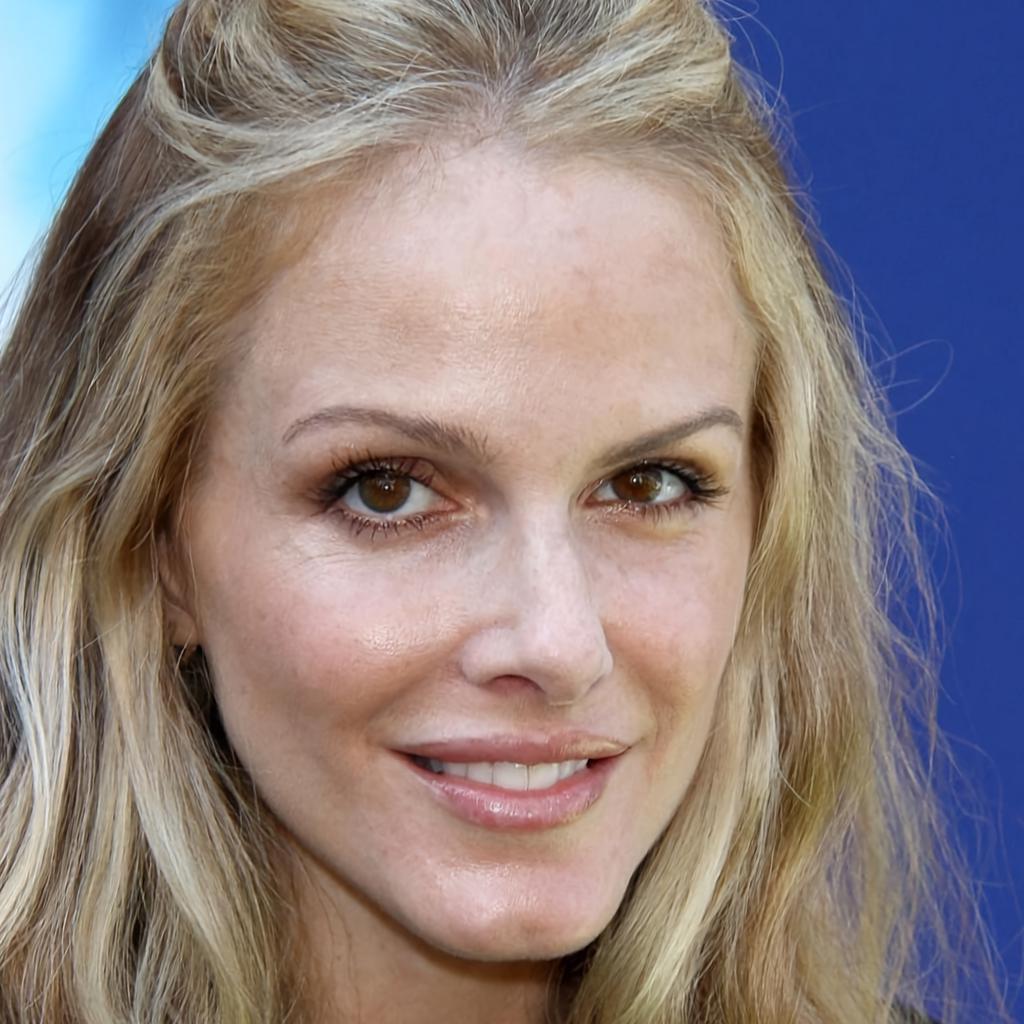} & 
 \includegraphics[width=0.125\textwidth, ]{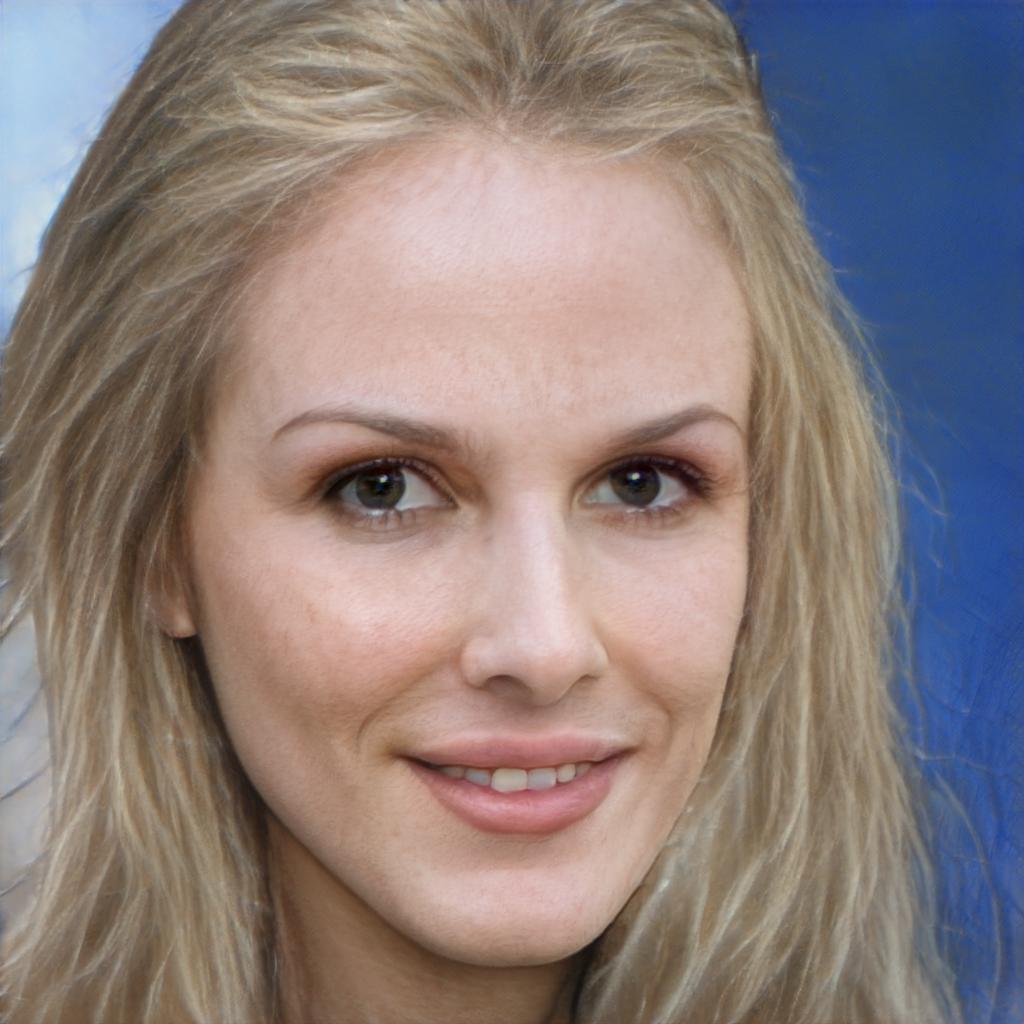} &
 \includegraphics[width=0.125\textwidth, ]{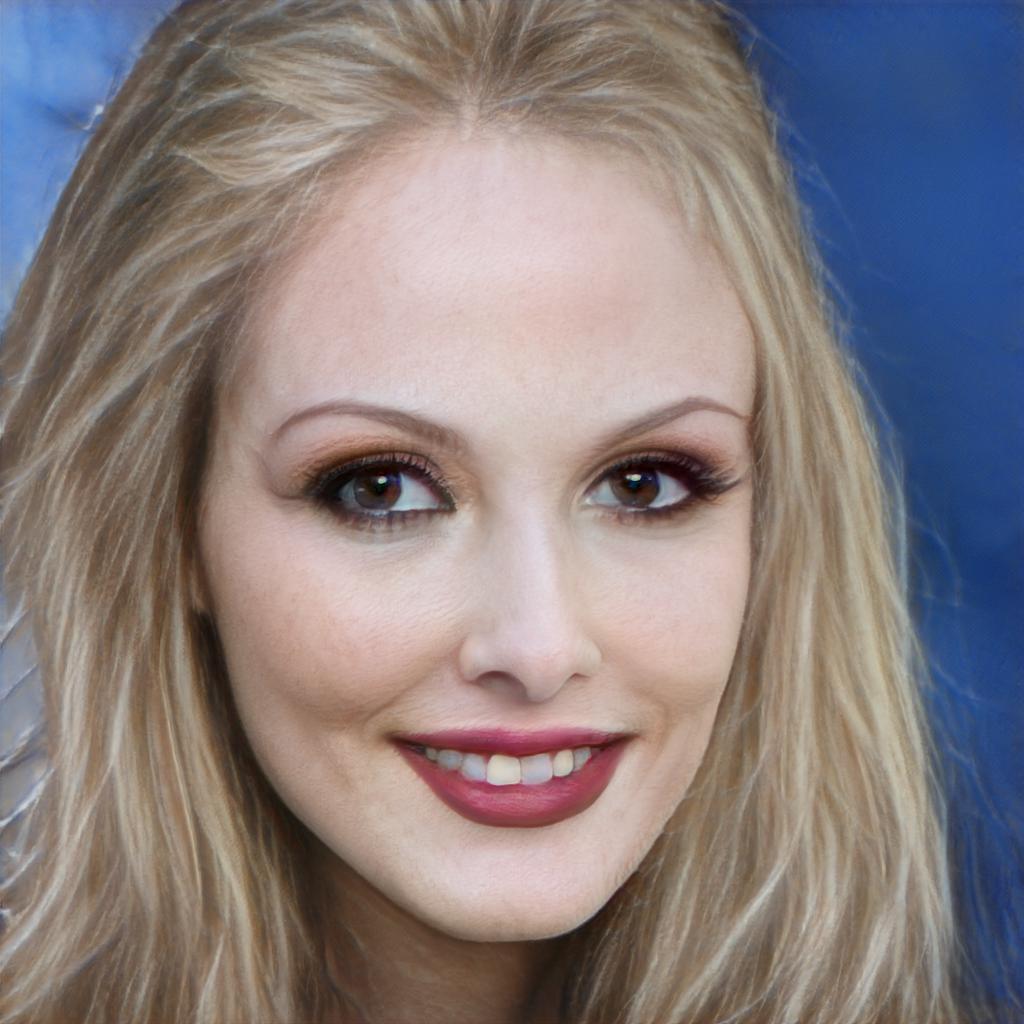} &
 \includegraphics[width=0.125\textwidth, ]{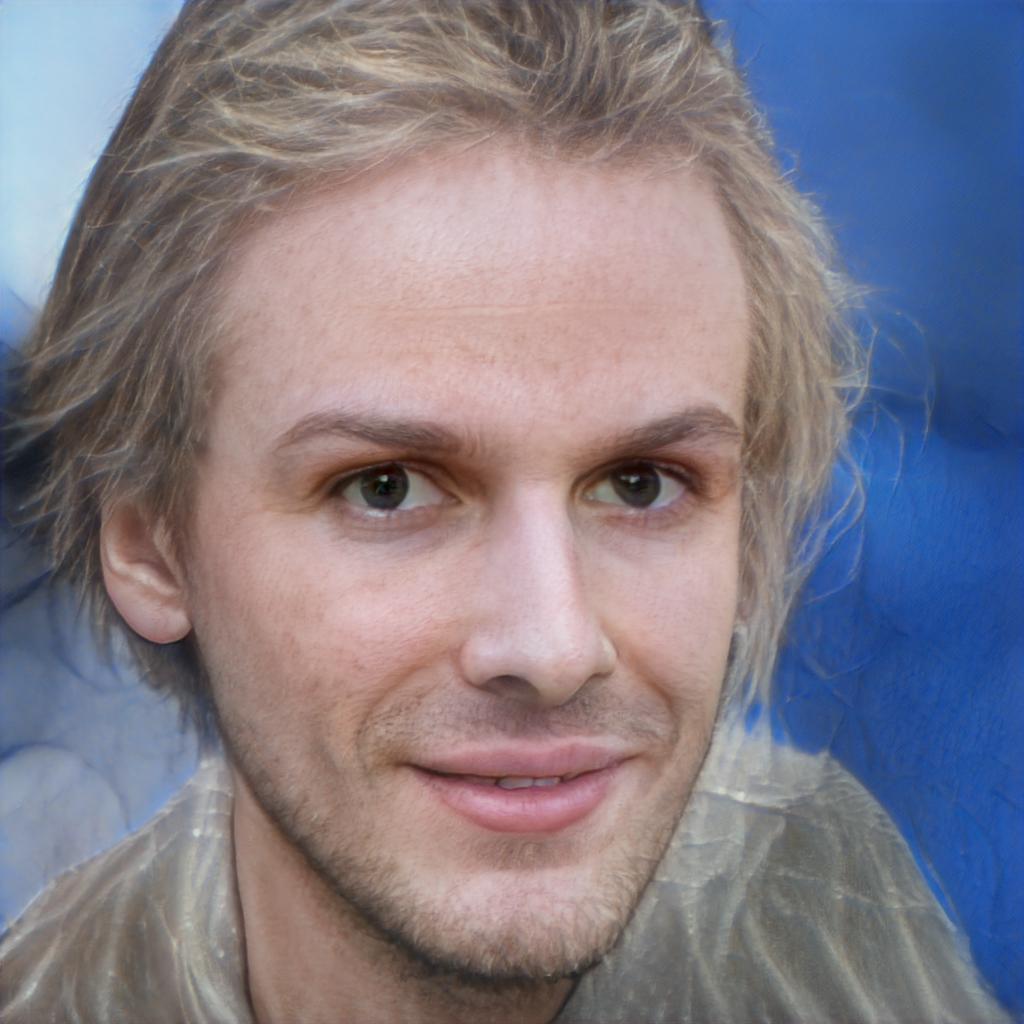} &
 \includegraphics[width=0.125\textwidth, ]{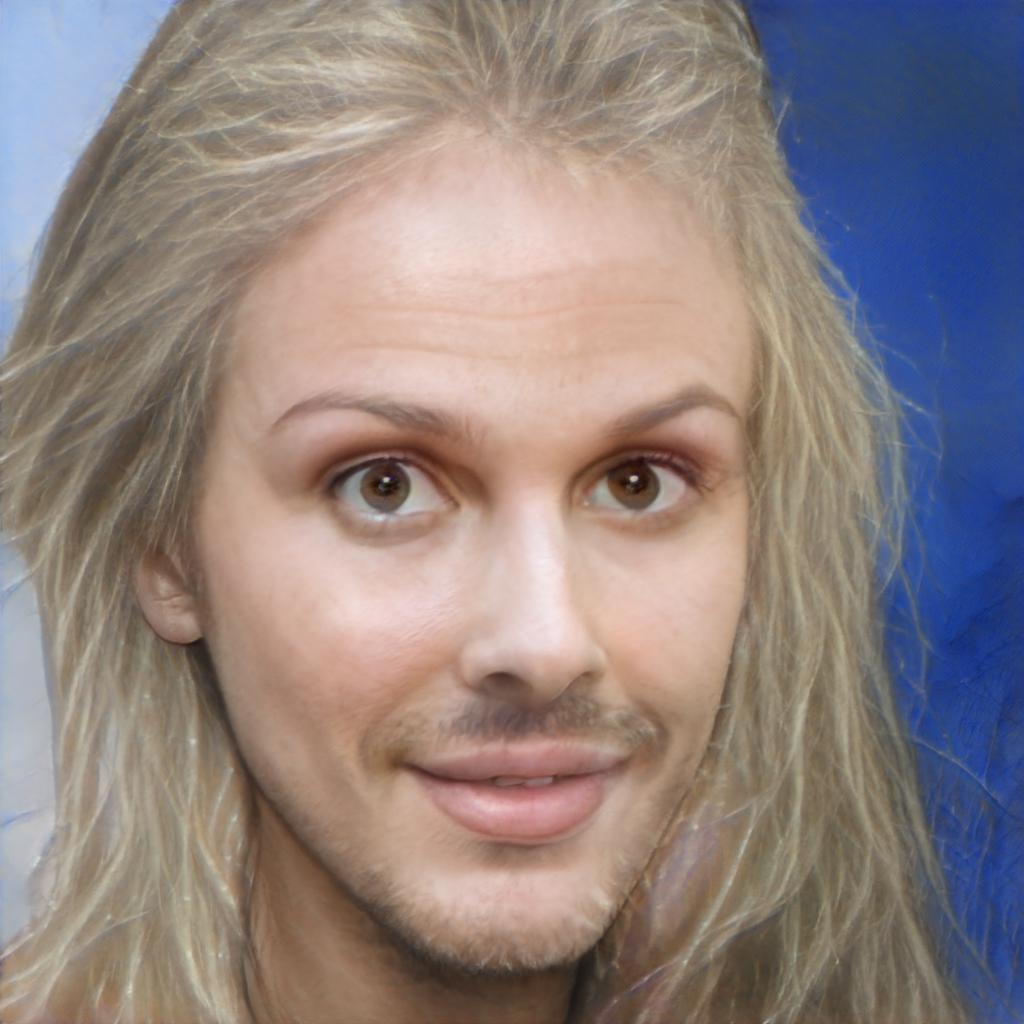} &
 \includegraphics[width=0.125\textwidth, ]{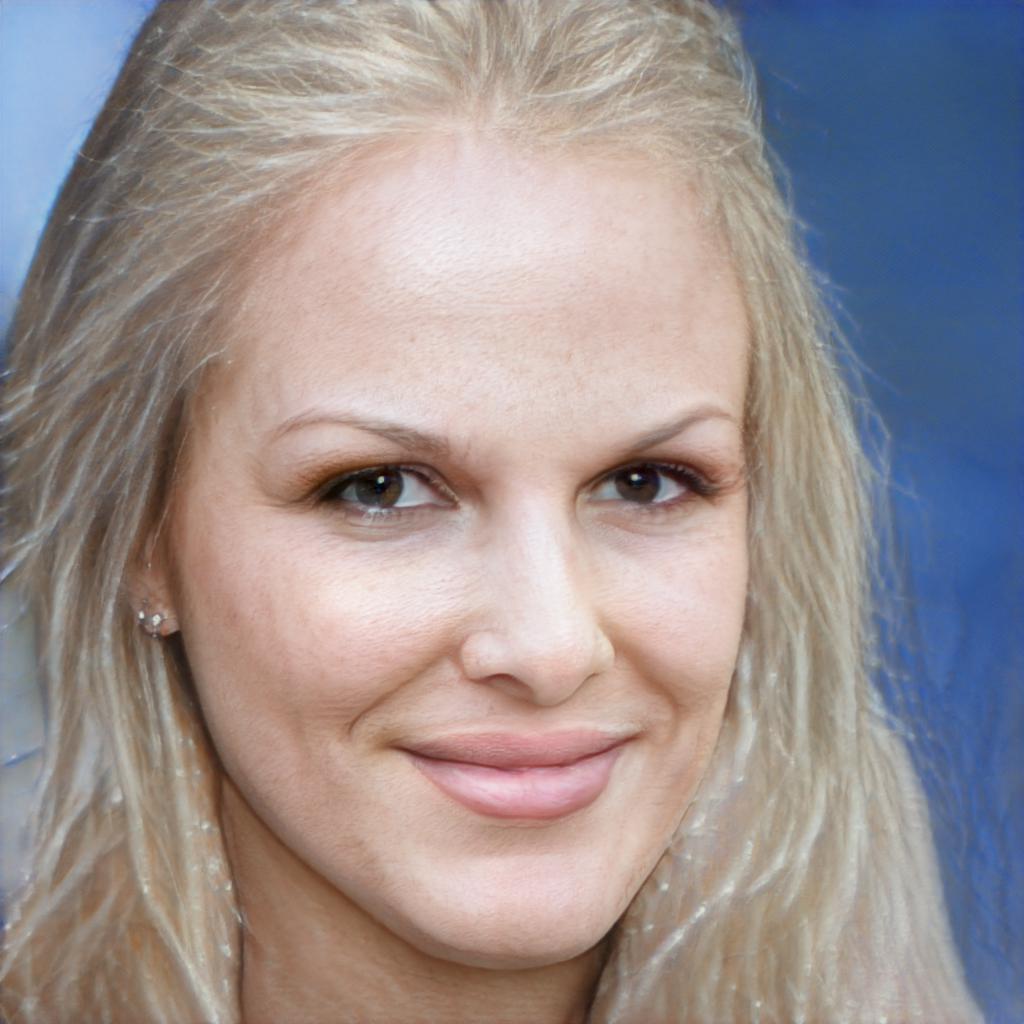} &
 \includegraphics[width=0.125\textwidth]{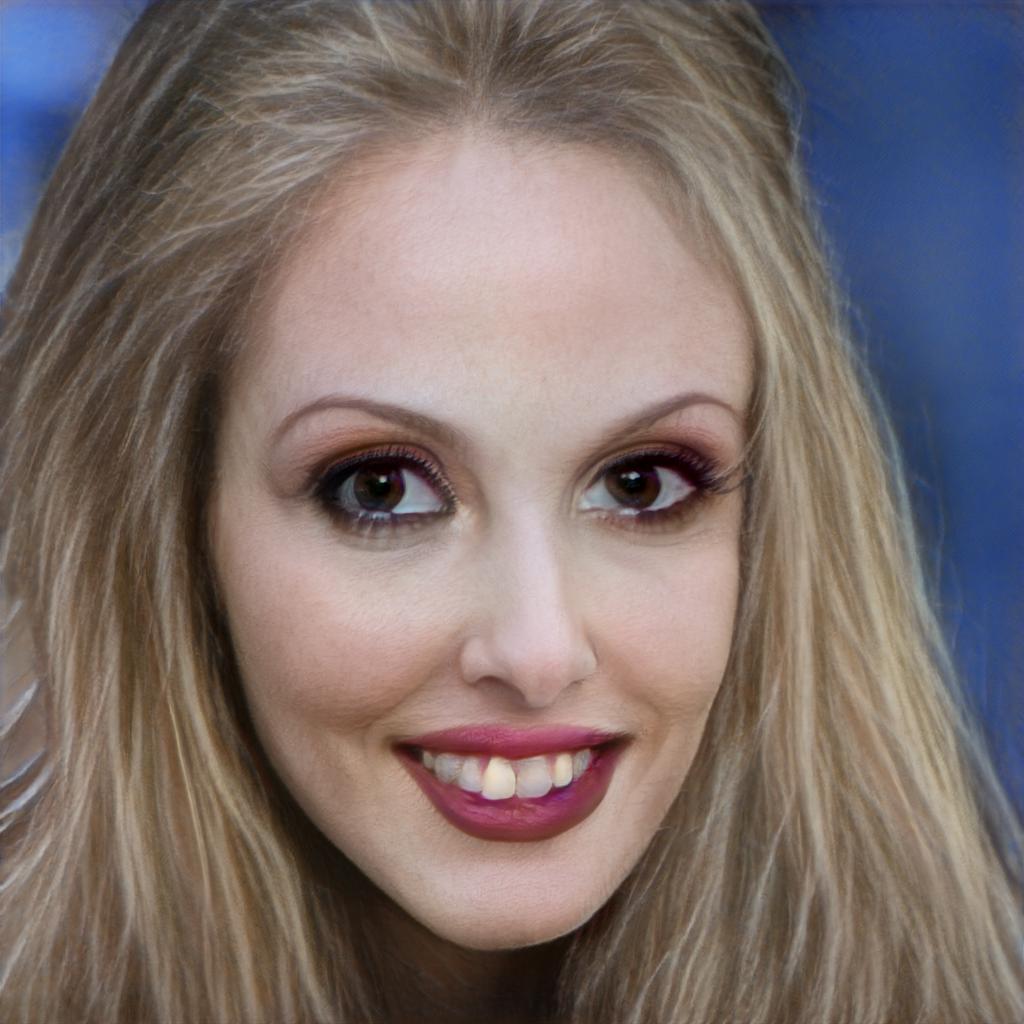} \\
 \begin{turn}{90} \hspace{0.5cm} $\mathcal{W}^{\star}_{ID}$ \end{turn} &
 \includegraphics[width=0.125\textwidth, ]{images/original/06011.jpg} & 
 \includegraphics[width=0.125\textwidth, ]{images/inversion/18_orig_img_10.jpg} &
 \includegraphics[width=0.125\textwidth,  ]{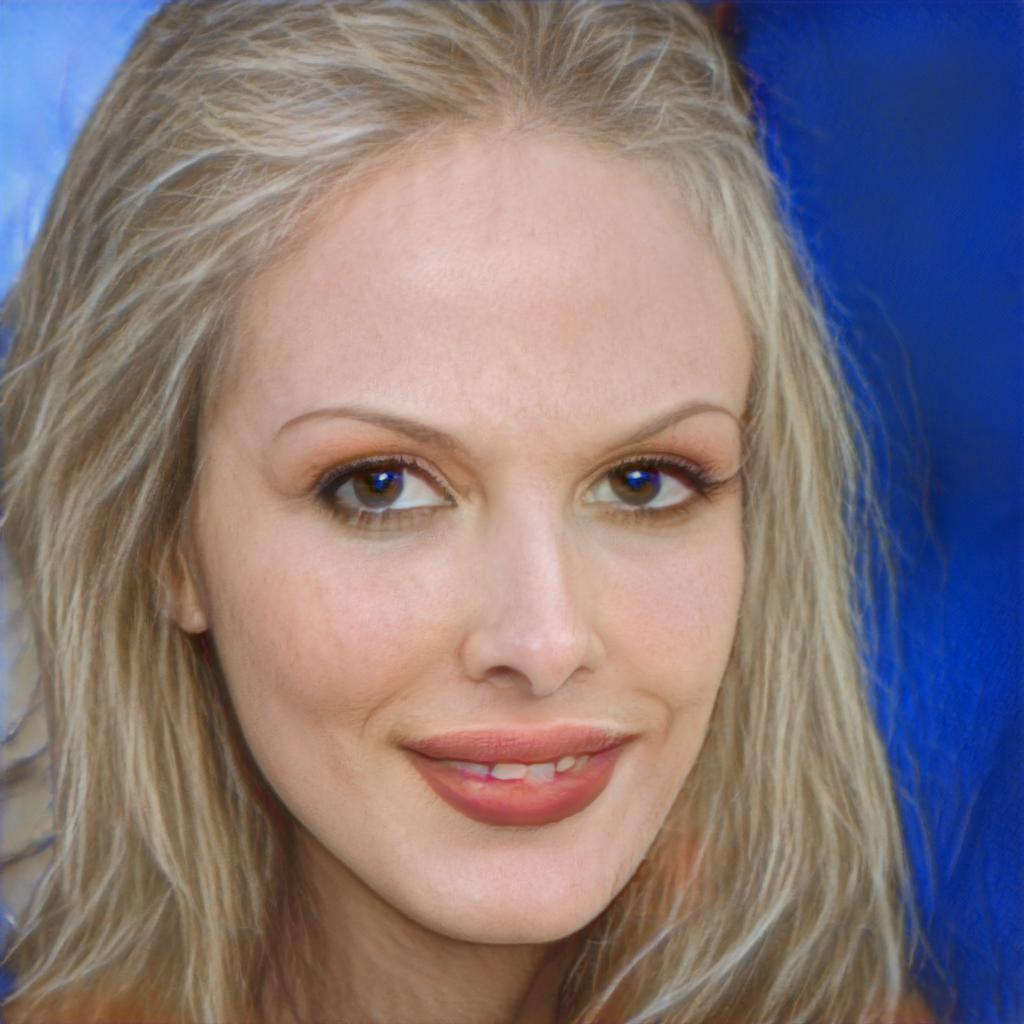} &
 \includegraphics[width=0.125\textwidth, ]{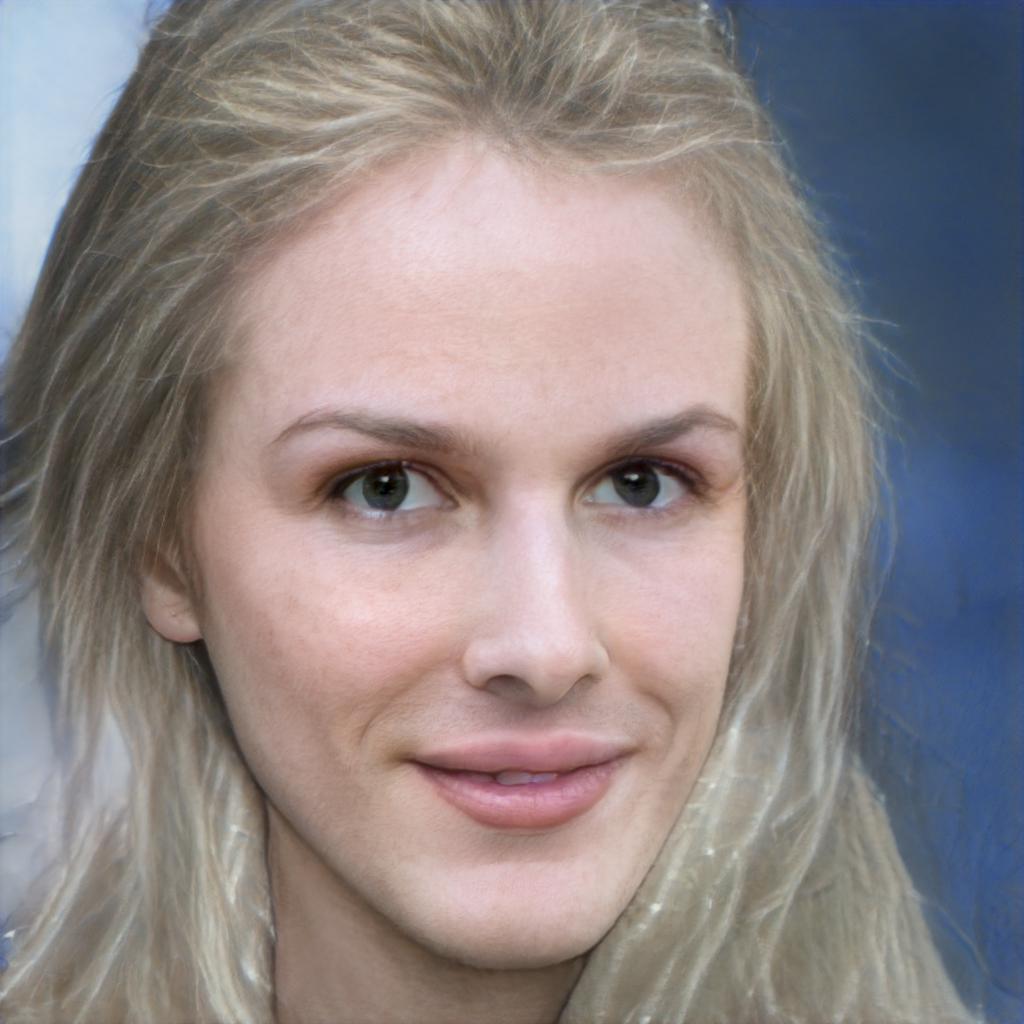} &
 \includegraphics[width=0.125\textwidth, ]{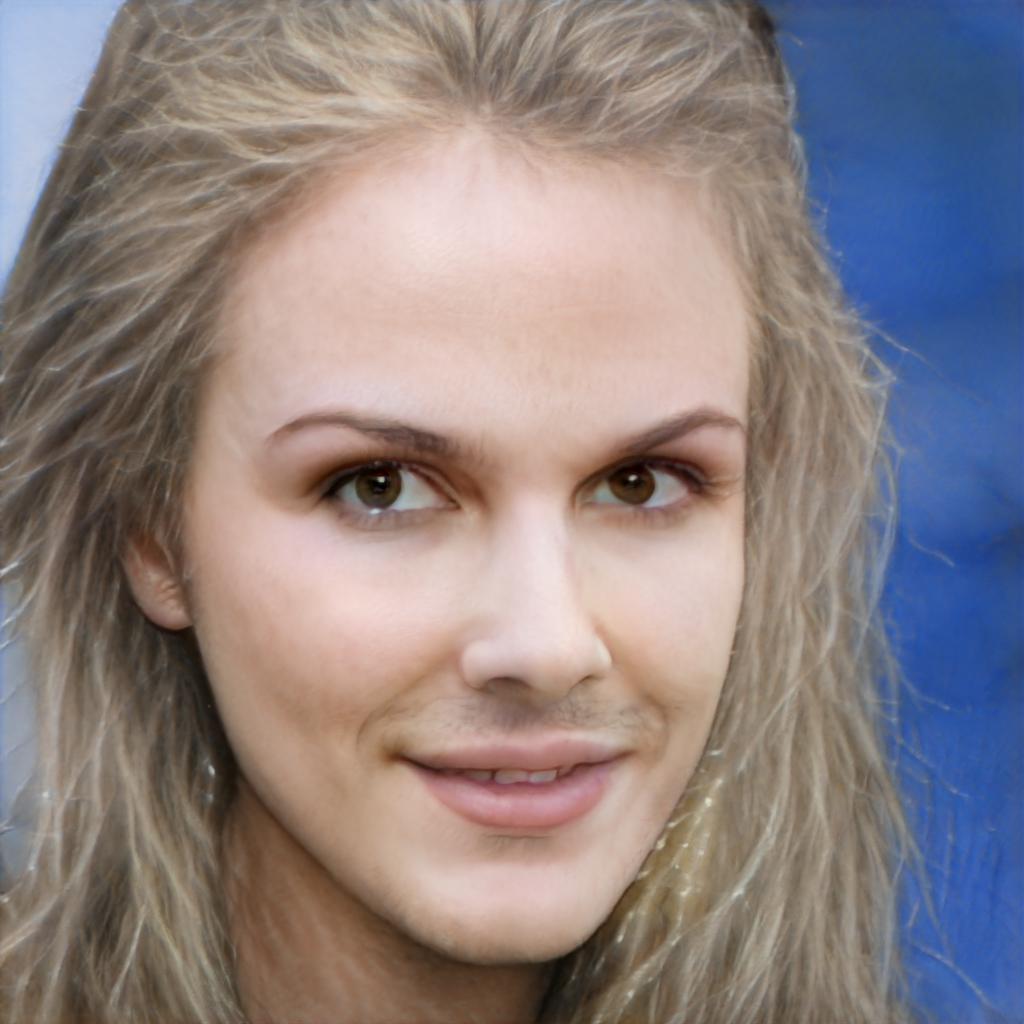} &
 \includegraphics[width=0.125\textwidth, ]{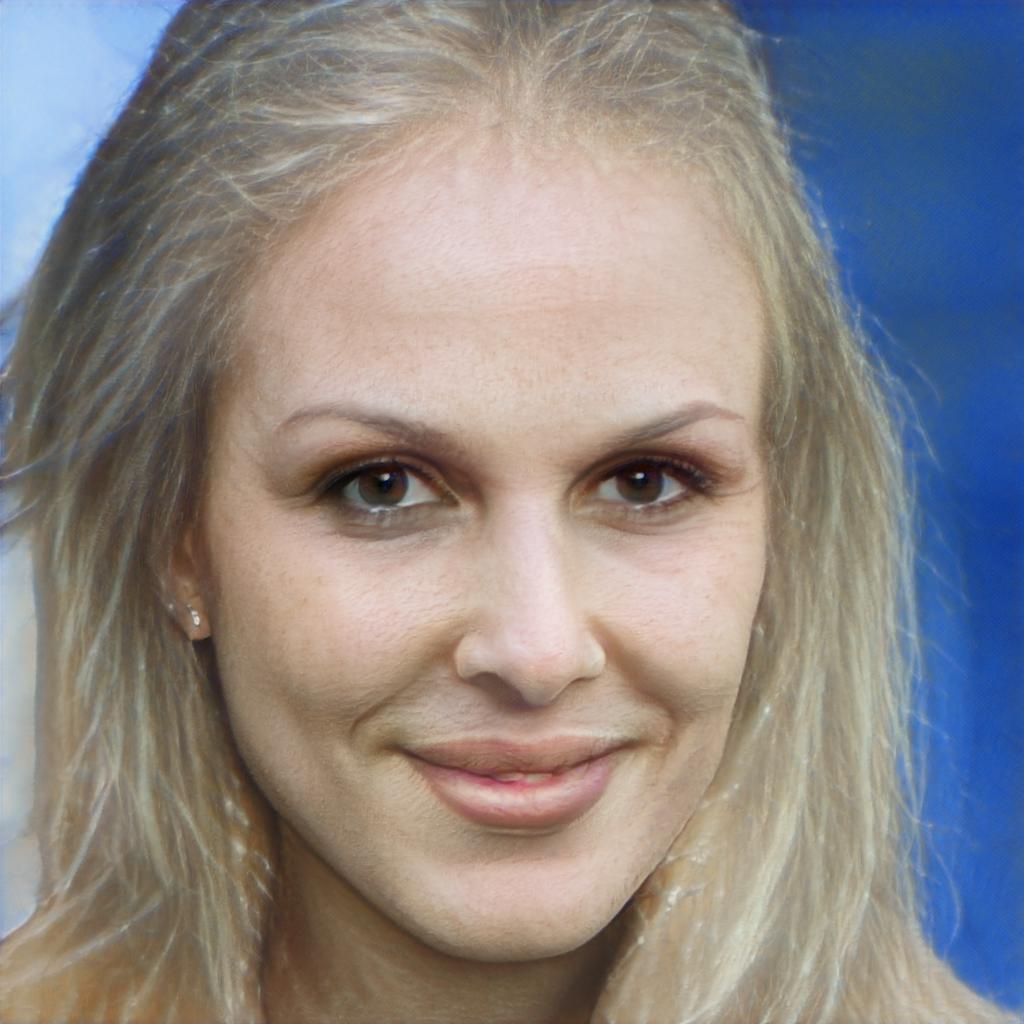} &
 \includegraphics[width=0.125\textwidth, ]{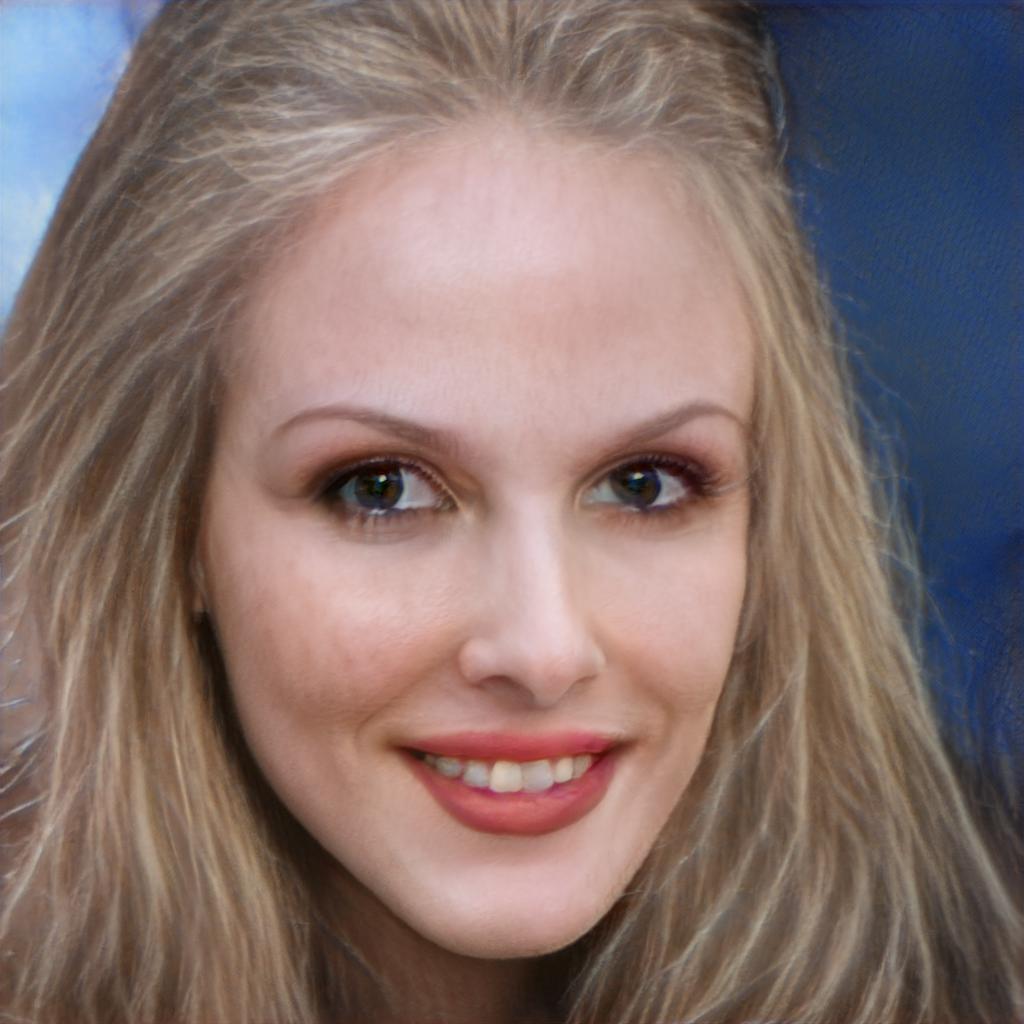} \\
 \\
\end{tabular}
\caption{Image editing at resolution $1024^2$ using InterFaceGAN in \wplus  and $\mathcal{W}^{\star}_{ID}$: The images are projected in the latent space of StyleGAN2, then the latent codes are moved in the direction that corresponds to changing one facial attribute. The editing is better in terms of attributes disentanglement in the new learned space ($\mathcal{W}^{\star}_{ID}$) compared to the editing done in the original latent space (\wplussp). Our method also leads to a better identity preservation.}
\label{fig:att_inter_results}
\end{figure}
\paragraph{Implementation details:} Real NVP consists of 3 coupling layers without batch normalization and the loss function was minimized with identity preserving loss \eqref{eq:att_id}. The same setup as in \ref{sec:implementation} was adopted. The editing directions were obtained after training an SVM on 15000 images of Celeba-HQ encoded using the pretrained encoder in \wplus and $\mathcal{W}^{\star}_{ID}$. The editing step is 6 for \wplus and it is between 10 to 15 for $\mathcal{W}^{\star}_{ID}$ (Note that with higher step the identity changes and the attributes become less disentangled). The model is trained on a Tesla V100 GPU (32GB) for 9 hours.

\paragraph{Results:} Figure \ref{fig:att_inter_results} displays the editing results on five attributes in $\mathcal{W}^{\star}_{ID}$ and \wplus and it shows that the editing results are visually better in $\mathcal{W}^{\star}_{ID}$ than in \wplussp. In particular, we see the following observation in \wplussp: gender is still entangled with adding Makeup and Lipstick (3rd row where the male gender is changed to female), changing the gender to Male is entangled with adding Beard and the Hair (column 4), and adding Mustache is entangled with gender (row 5).
While in $\mathcal{W}^{\star}_{ID}$ these attributes are better disentangled and the identity is better preserved. Finally, it is clear that we still obtain high quality images even if we did not retrain the generator and the editing is not done in \wplussp.

\setlength\tabcolsep{2 pt}
\begin{table}[h]
\begin{center}
\begin{tabular}{lcccccccccc}
\toprule
& 
\multicolumn{3}{c}{Linear separation} & &\multicolumn{3}{c}{Disentanglement} & 
&
\multicolumn{2}{c}{Dist. Unfolding} \\\cline{1-4} \cline{6-8} \cline{10-11}
Space & 
\begin{tabular}{@{}c@{}}Min Class.\\ Acc $\uparrow$\end{tabular} & 
\begin{tabular}{@{}c@{}}Max Class.\\ Acc $\uparrow$\end{tabular}& 
\begin{tabular}{@{}c@{}}Avg. Class.\\ Acc $\uparrow$\end{tabular}&
&
D $\uparrow$&
C $\uparrow$& 
I $\downarrow$&
& 
Mean $\downarrow$& 
STD $\downarrow$\\
\hline
\wstar (H) & \textbf{0.787} & 0.991 & \textbf{0.928} & & \textbf{0.82} & \textbf{0.56} & 0.27 & &  0.17 & 0.15\\
\wstar (L) & 0.777 & \textbf{0.992} & 0.913 & & 0.81  & 0.55  & \textbf{0.26} & & 0.13  & 0.14\\
\bottomrule
\end{tabular}
\end{center}
\caption{Ablation study of model architecture. \wstar (H) and \wstar (L) refer to Real NVP model with 13 and 3 coupling layers, respectively. Although the classification accuracy and the disentanglement metric show slightly better results in \wstar (H), the difference is not significant, advocating to retain the model with the smallest capacity.
}
\label{tab:ablation}
\end{table} 
\subsubsection{Ablation Study}\label{sec:ablation} 
In this section, we present the ablation study for attributes separation, distance unfolding and image editing. Implementation details and more results can be found in the suppl. material.
\paragraph{Attribute's separation and distance unfolding: }To assess the sensitivity of our proposed method with the number of coupling layers, we conducted experiments with higher and lower Real NVP model capacity and the results (Table \ref{tab:ablation}) shows that higher capacity model \wstar (H)  are slightly better than lower capacity one \wstar (L). 
\paragraph{Image Editing:} In Figure \ref{fig:edit_abl_1}, We can notice that the identity preservation loss helps indeed to preserve subject's identity and leads to a higher quality image editing. The latent distance unfolding loss benefit is less significant for editing.
\section{Discussion}
\label{sec:discussion}
 Our proposed method has enforced the linear separability and attribute disentanglement for image editing. Although other properties could be considered as well (\eg, pose preservation). Note that, these valuable properties could not be achieved in the original latent space without retraining the encoder and the generator. At the same time, we succeeded to unfold the distance in such a way that the latent Euclidean distance approximates the perceptual VGG16 distance. This method is generic and could be used for any perceptual or hand-crafted distance. We believe that this approach is useful when a specific distance is better than the latent Euclidean distance for a given application. In addition, our attribute separation approach could be extended in a straightforward way to other types of models (\eg VAEs, GANs) that manipulate the images in their latent space. For distance unfolding, the scope of models is larger as any model with a latent space could be adopted.
\begin{figure}[h]
\setlength\tabcolsep{2pt}%
\centering
\begin{tabular}{p{0.25cm}ccccccc}
\centering
&
 \textbf{Original} &
 \textbf{Inverted} &
 \textbf{Makeup} &
 \textbf{Male} &
 \textbf{Mustache} &
 \textbf{Chubby} &
 \textbf{Lipstick} \\
 \begin{turn}{90} \hspace{0.5cm} \wstara\end{turn} &
 \includegraphics[width=0.125\textwidth]{images/original/06002.jpg} & 
 \includegraphics[width=0.125\textwidth, ]{images/inversion/18_orig_img_1.jpg} &
 \includegraphics[width=0.125\textwidth, ]{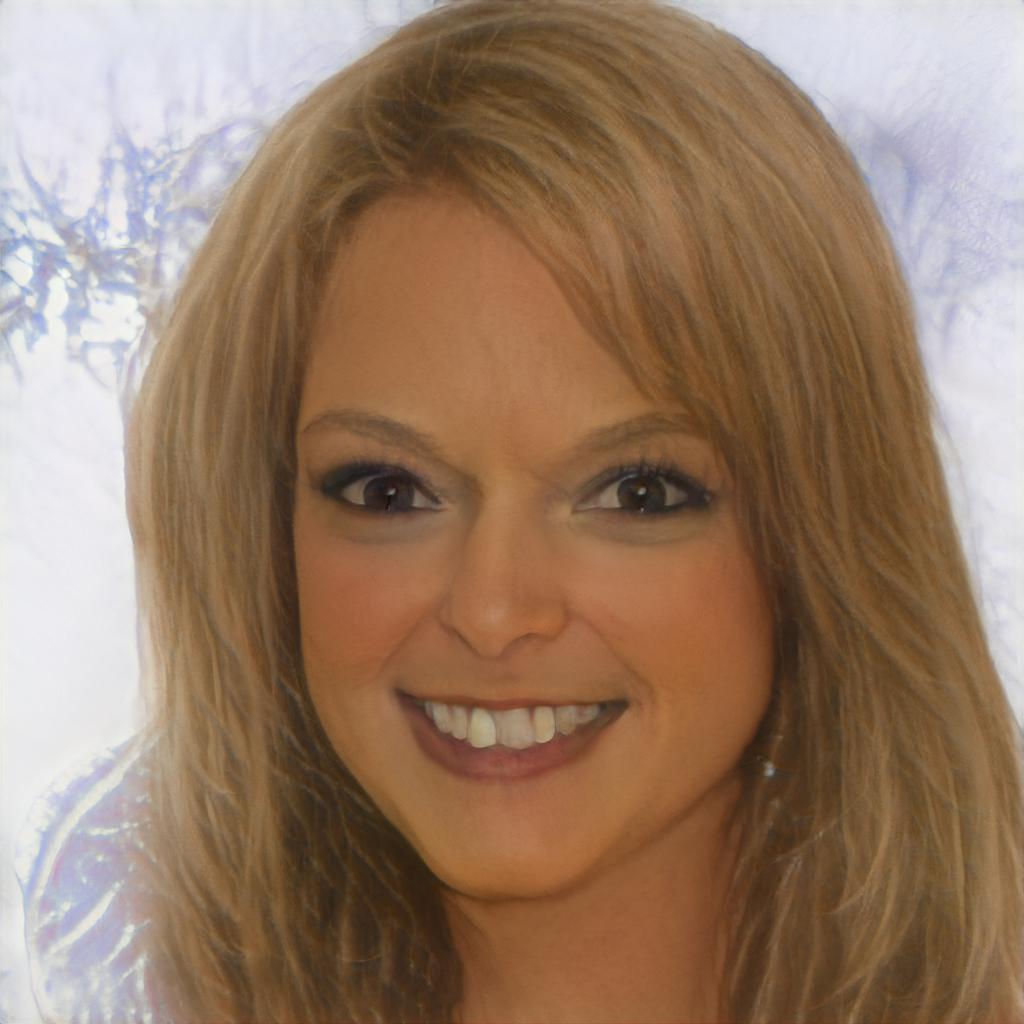} &
 \includegraphics[width=0.125\textwidth, ]{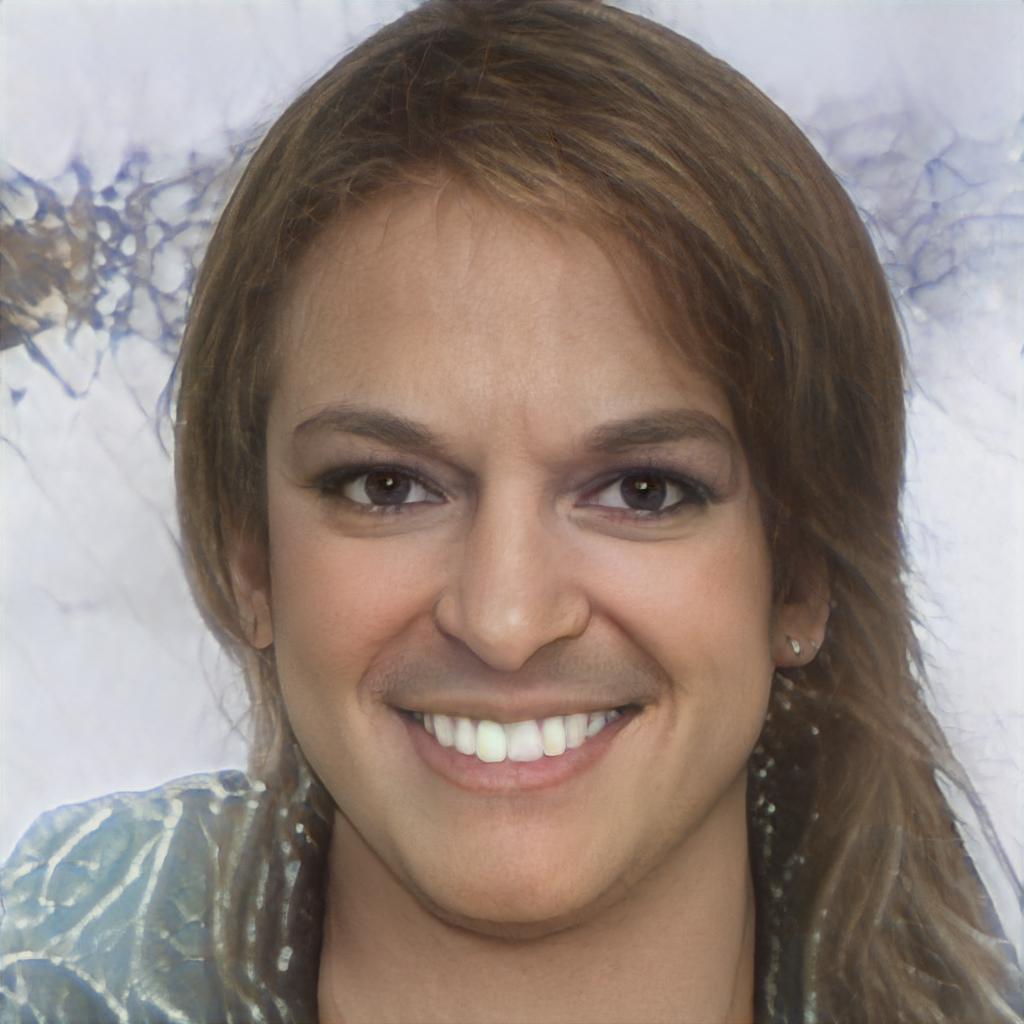} &
 \includegraphics[width=0.125\textwidth, ]{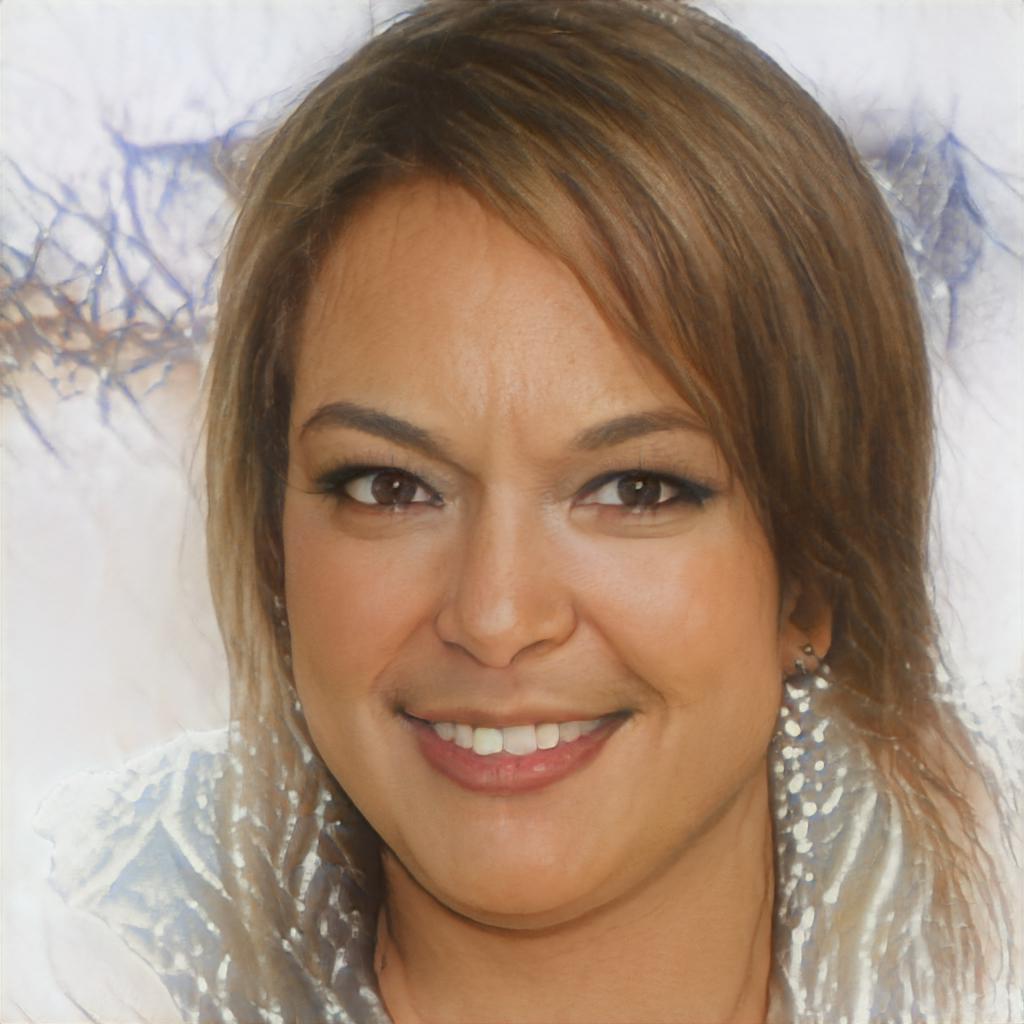} &
 \includegraphics[width=0.125\textwidth, ]{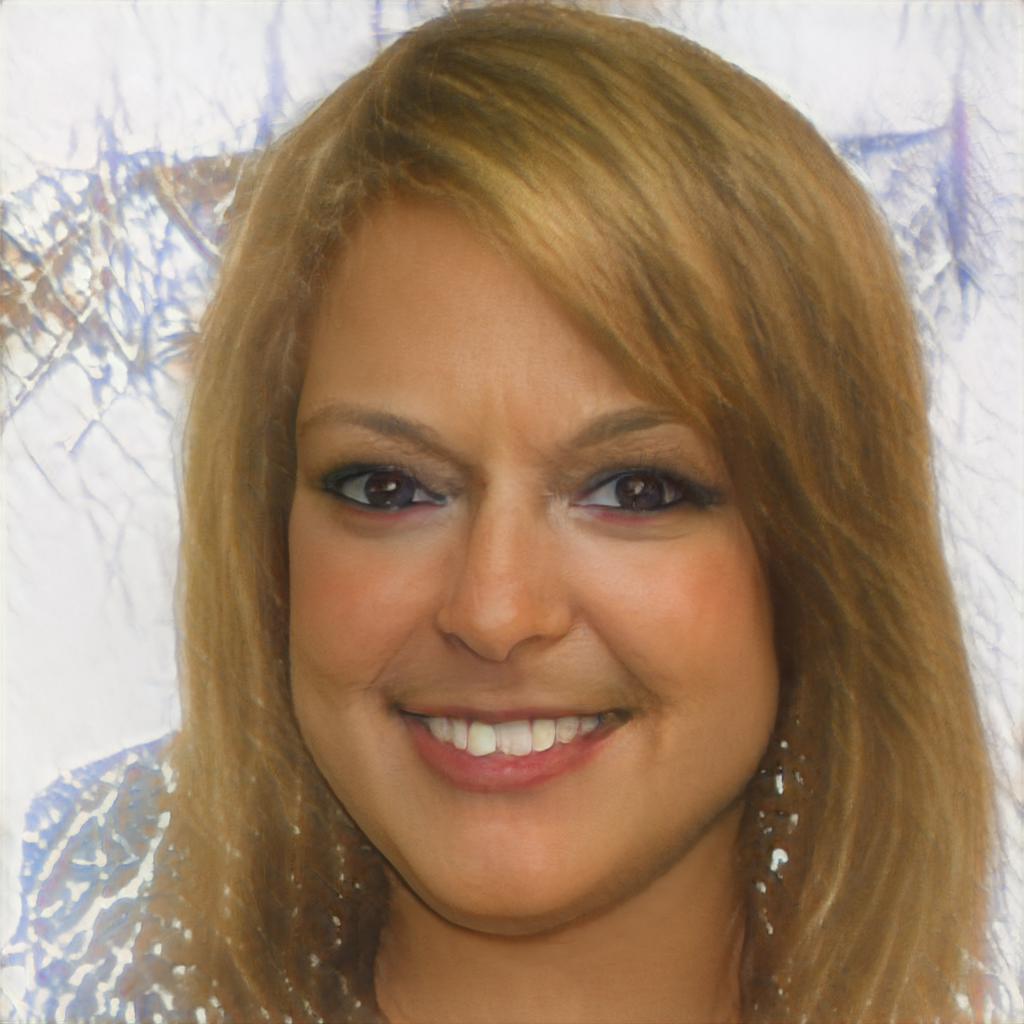} &
 \includegraphics[width=0.125\textwidth, ]{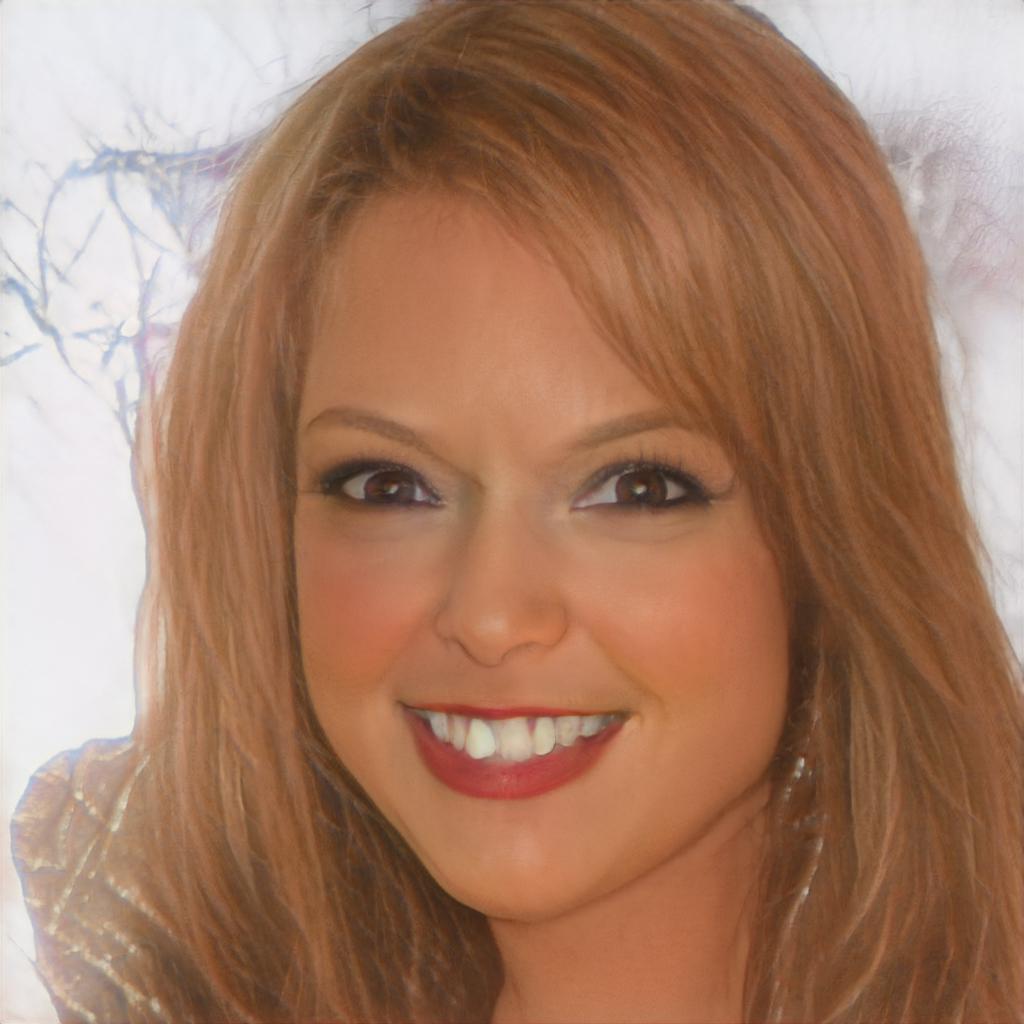} \\
 \begin{turn}{90} \hspace{0.5cm} $\mathcal{W}^{\star}_{ID}$ \end{turn} &
 \includegraphics[width=0.125\textwidth]{images/original/06002.jpg} & 
 \includegraphics[width=0.125\textwidth, ]{images/inversion/18_orig_img_1.jpg} &
 \includegraphics[width=0.125\textwidth, ]{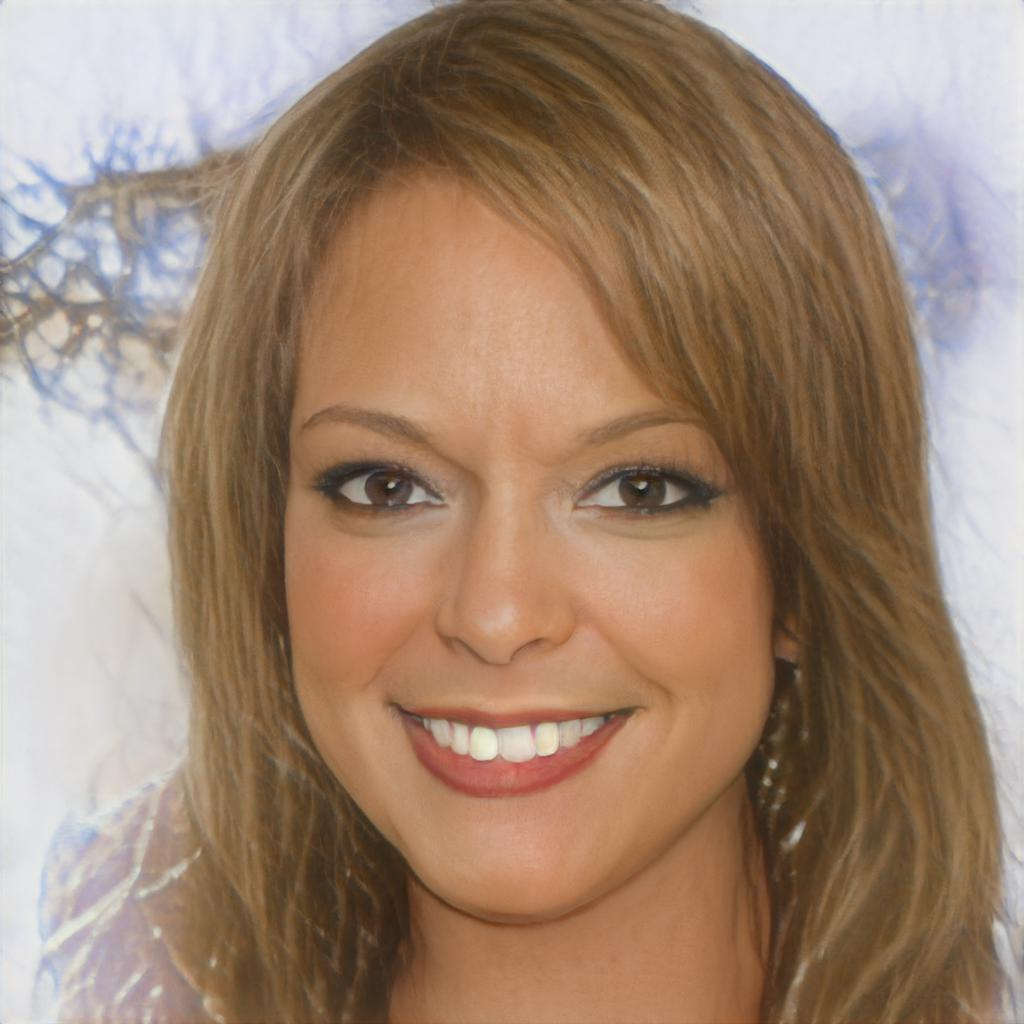} &
 \includegraphics[width=0.125\textwidth, ]{images/wstar_id_no_mag_3_coupl_dist_edit_train_ep5/20_img_1.jpg} &
 \includegraphics[width=0.125\textwidth, ]{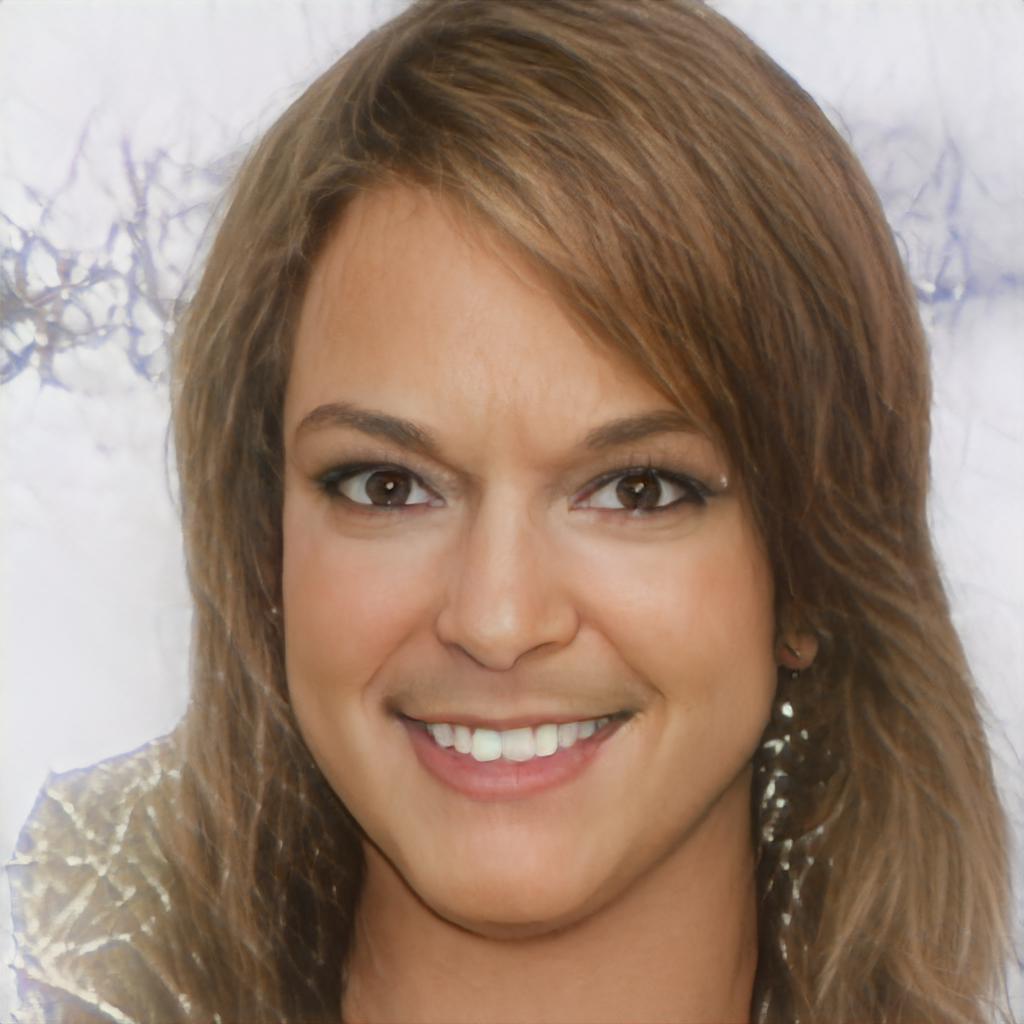} &
 \includegraphics[width=0.125\textwidth, ]{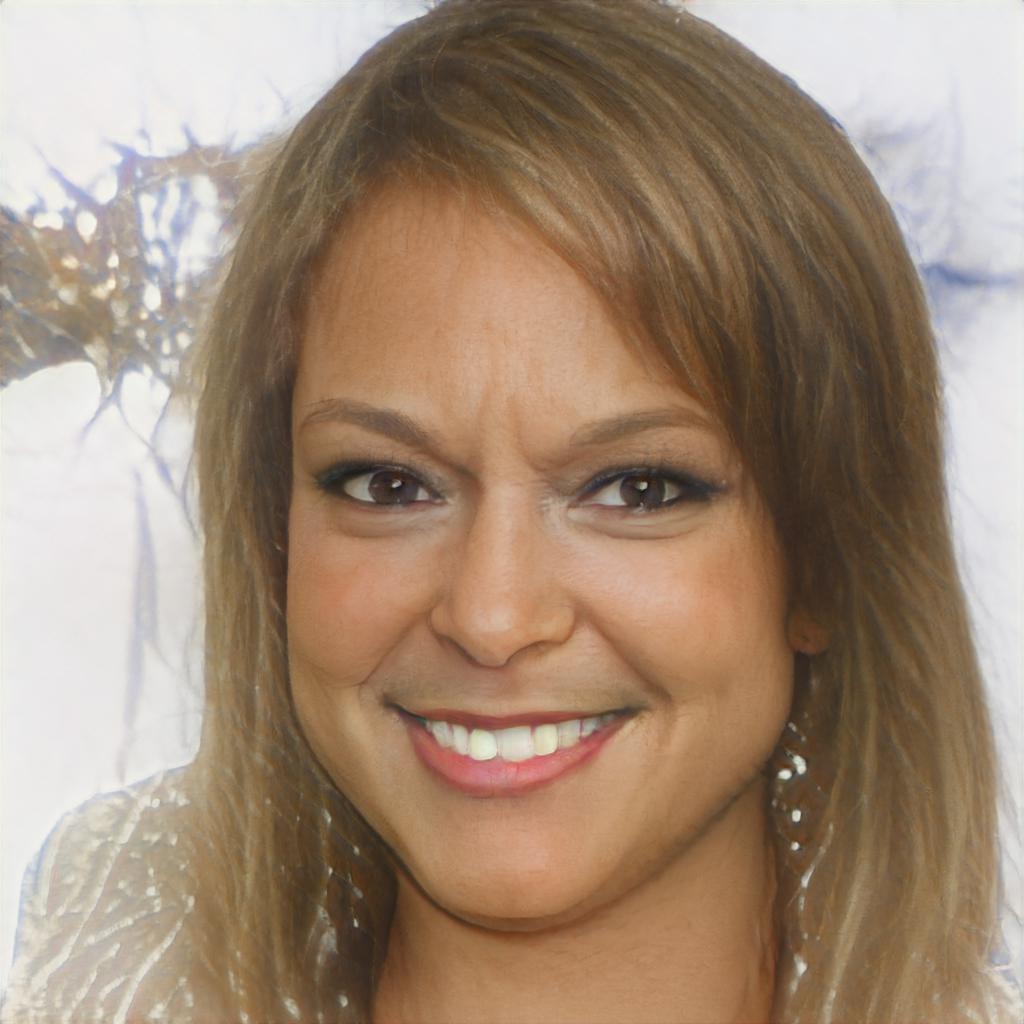} &
 \includegraphics[width=0.125\textwidth, ]{images/wstar_id_no_mag_3_coupl_dist_edit_train_ep5/36_img_1.jpg} \\
 \\
\end{tabular}
\caption{Ablation study of Image editing using InterFaceGAN.Top row: model trained with only the attribute separation loss \wstarasp. The attribute separation loss alone is clearly not sufficient to obtain a controllable and identity-preserving editing.}
\label{fig:edit_abl_1}
\end{figure}
\section{Conclusion}
\label{sec:conclusion}
We presented a general framework to enforce additional geometric and semantic properties to the latent space of generative models without the burden of retraining them. In particular, we trained a bijective transformation from the extended space of StyleGAN2 (\wplussp) to a proxy space (\wstarsp) where two properties are satisfied; the attributes are disentangled and the latent distance mimics the perceptual one in the image space. This generic method enables to bring additional supervision and regularization on top of any trained GAN. We validated our approach by surpassing the original latent space in terms of quantitative metrics and by showing better editing results in terms of attributes disentanglement. We plan to leverage this space for future applications that are affected by a well-conditioned latent distance.
{\small
\bibliographystyle{plainnat}
\bibliography{main}
}

\newpage
\section{Supplementary Material}

The supplementary material is organized as follows: Section \ref{appendix:latent_percept_issue} illustrates the inconsistency between the perceptual and the latent euclidean distance. In section \ref{appendix:linear_sep}, we visualize the linear separation of the attributes in the latent space.  Finally, section \ref{appendix:ablation} details the ablation study.
\subsection{Perceptual distance vs Latent distance}
\label{appendix:latent_percept_issue}

In this section, we illustrate the inconsistency between the latent Euclidean distance and the perceptual distance in the image space.

\noindent \textbf{Implementation details} \quad Real NVP consists of 13 coupling layers and is trained with the objective (3). $\lambda_d$=10. The latent distance is the MSE and the Perceptual one is VGG16 pretrained on Imagenet.
During training and test, the perceptual distance is rescaled (multiplied by $10$) to be in the same scale as the latent distance. We randomly sampled $600$ pairs of latent codes (images are from the Celeba-HQ dataset and were encoded using the pretrained encoder) and computed the latent distance in both spaces (\wplus, \wstarsp) and the perceptual distance of the corresponding image pairs.

\begin{figure}[h]
    \begin{center}
        \includegraphics[width=0.6\linewidth]{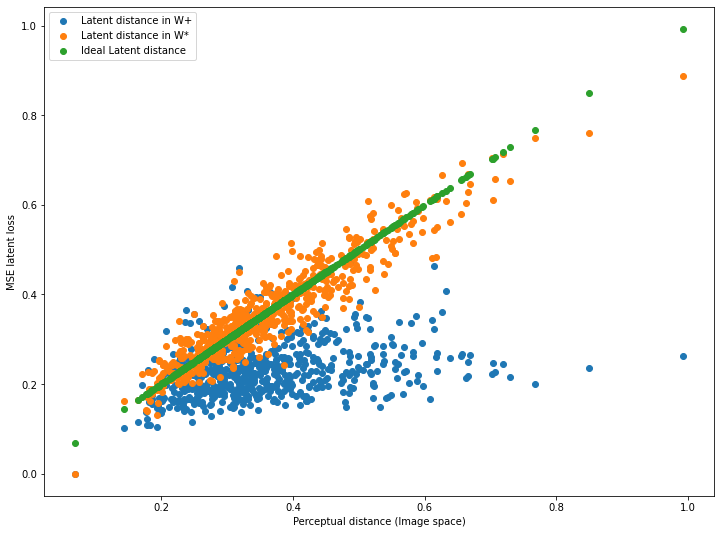}
    \end{center}
    \caption{An illustration of the inconsistency between the latent and the perceptual distance: each point represents pair of images. The latent distance is computed in \wplus (Blue) and \wstar (Orange). The green diagonal line represents the ideal relationship between the two distances.}
    \label{fig:lat_percept_issue}
\end{figure}

\begin{figure}[h]
    \begin{center}
    \includegraphics[width=0.6\linewidth]{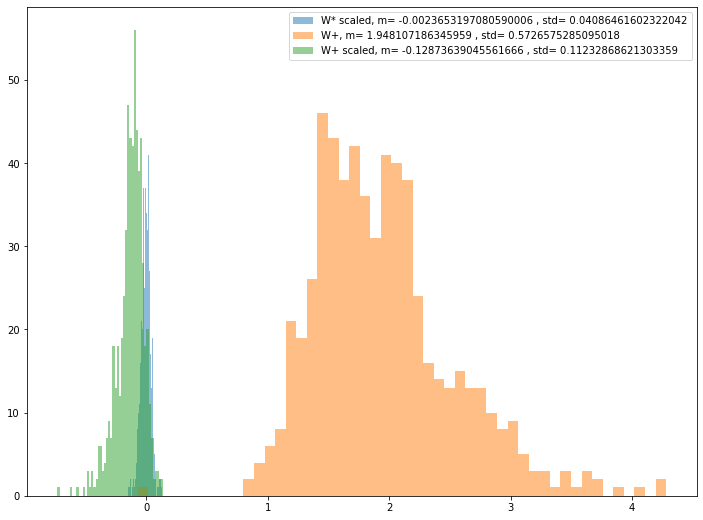}
    \end{center}
    \caption{An illustration of the inconsistency between the latent and the perceptual distance: the histograms show the difference between the perceptual and the latent distance of 600 pair of images. The latent distance is computed in \wplus (Green), \wplus without rescaling the perceptual loss (Orange) and \wstar (Blue).}
    \label{fig:lat_percept_issue2}
\end{figure}

\noindent \textbf{Results} \quad As we can see in Figure \ref{fig:lat_percept_issue} and \ref{fig:lat_percept_issue2}, the latent distance is not consistent with the perceptual one in \wplussp. In \wstar the latent distance is more similar to the perceptual one and closer to the ideal distance (equality between the two distances). \\

\subsection{Linear Separation of the attributes}
\label{appendix:linear_sep}
\begin{figure}[t]
\setlength\tabcolsep{2pt}%
\begin{center}
\begin{tabular}{p{0.25cm}ccc}

 &
 \textbf{Makeup} &
 \textbf{Male} &
 \textbf{Lipstick} \\
 &
 Acc:0.90 &
 Acc:0.97 &
 Acc:0.94 \\
 \begin{turn}{90} \hspace{1.5cm} \wplus\end{turn} &
 \includegraphics[width=0.3\textwidth]{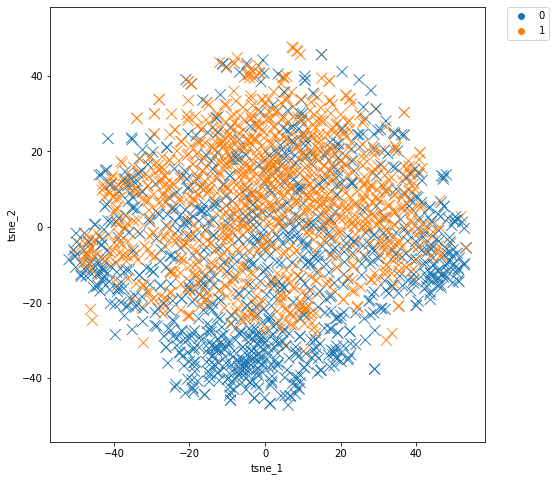} & 
 \includegraphics[width=0.3\textwidth]{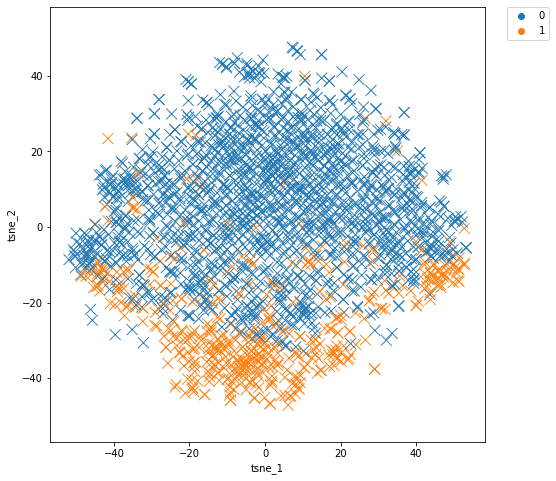} &
 \includegraphics[width=0.3\textwidth]{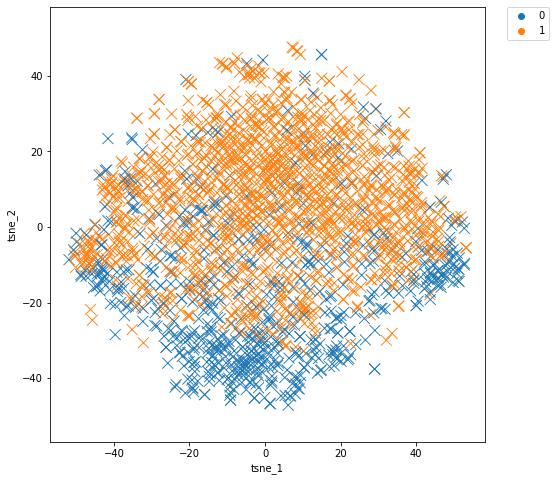} \\
 &
 Acc:0.94 &
 Acc:0.98 &
 Acc:0.96 \\
 \begin{turn}{90} \hspace{1.5cm} \wstar \end{turn} &
 \includegraphics[width=0.3\textwidth]{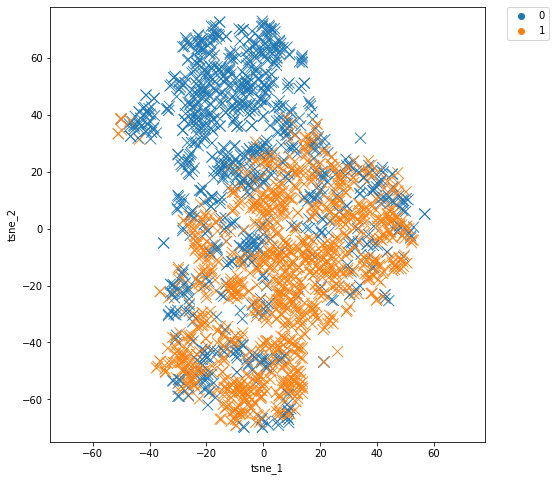} & 
 \includegraphics[width=0.3\textwidth]{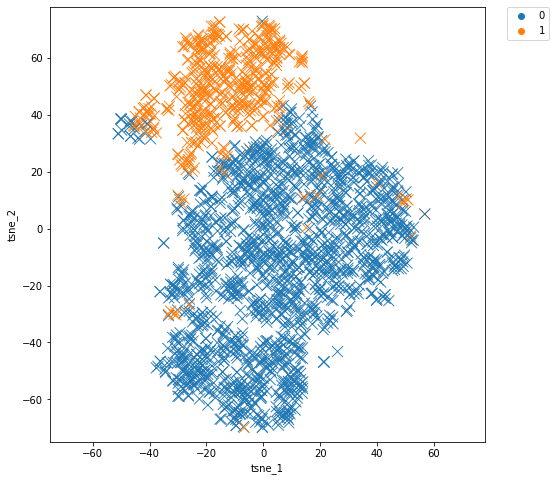} &
 \includegraphics[width=0.3\textwidth]{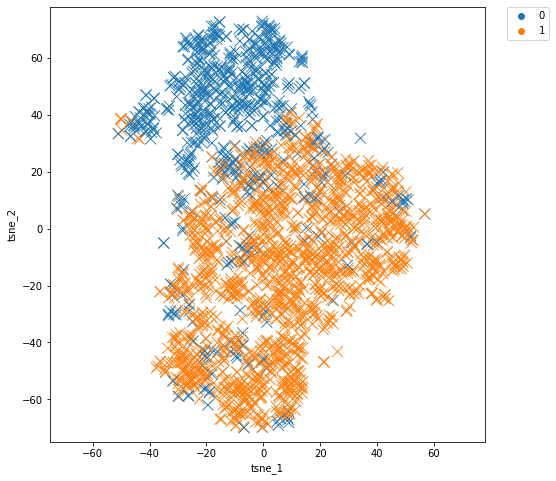} \\
 
\end{tabular}
\end{center}
\caption{t-SNE visualization in \wplus and \wstarsp: the classification accuracy is shown at the top (Acc). the 2D embedding enables to visualize that the attributes are better linearly disentangled in \wstarsp.}
\label{fig:tsne}

\end{figure}

In this section we use t-SNE \footnote{Used from scikit-learn library with default parameters: \href{https://scikit-learn.org/stable/modules/generated/sklearn.manifold.TSNE.html}{https://scikit-learn.org/stable/modules/generated/sklearn.manifold.TSNE.html}} \cite{tsne} to visualize the negative and positive examples for some attributes in both latent spaces (\ie \wstar and \wplussp), embedded onto the $2D$ plane. The training setup is the same as in Section 4.1 and 4.3.1. The dataset is composed of $2000$ latent codes averaged over the $18$ dimensions. \\
From Figure \ref{fig:tsne}, one can easily visualise that negative and positive attributes are more linearly separated in \wstar than in \wplus. 

\subsection{Ablation Study}
\label{appendix:ablation}
In this section, we detail the quantitative and qualitative ablation study. 
\subsubsection{Quantitative evaluation}

In this section, we investigate the effect of some design choices. The setup is the same as in Section 4.1 except when stated otherwise. We do not rescale the perceptual distance. We analyse 4 experiments which differ from the main setup with the following; H (High model capacity, 13 coupling layers), L (Low model capacity, 4 coupling layers), R (using random classifier, 13 coupling layers), 1R (using one random classifier with 1 linear layer for all the attributes at once. 13 coupling layers). From Table \ref{tab:ablation}:
\begin{itemize}
    \item Model Capacity (H/L): We can notice that higher model capacity gives slightly better improvement for attribute separation and latent distance unfolding.
    \item Random Classifier (R): Better initialization of the classifier leads to slightly better separation. 
    \item One Random Classifier (1R): Using one classifier for all the attributes gives worse separation, which can be explained by the capacity of the model (1 layer vs 40 layers) or the fact that optimizing for one class is easier than for 40 classes (even though all the classifiers are optimized jointly). In addition, we noticed that the classification loss is smaller compared to using 40 classifiers, which may have an equivalent effect as multiplying the unfolding loss by a higher $\lambda_d$, thus the distance unfolding results are better.
\end{itemize}

\setlength\tabcolsep{2 pt}
\begin{table}[h]
\begin{center}
\begin{tabular}{lcccccccccc}
\toprule
& 
\multicolumn{3}{c}{Linear separation} & &\multicolumn{3}{c}{Disentanglement} & 
&
\multicolumn{2}{c}{Dist. Unfolding} \\\cline{1-4} \cline{6-8} \cline{10-11}
Space & 
\begin{tabular}{@{}c@{}}Min Class.\\ Acc $\uparrow$\end{tabular} & 
\begin{tabular}{@{}c@{}}Max Class.\\ Acc $\uparrow$\end{tabular}& 
\begin{tabular}{@{}c@{}}Avg. Class.\\ Acc $\uparrow$\end{tabular}&
&
D $\uparrow$&
C $\uparrow$& 
I $\downarrow$&
& 
Mean $\downarrow$& 
STD $\downarrow$\\
\hline
\wplus & 0.635 & 0.979 & 0.834 & & 0.59 & 0.43 & 0.30 & & 1.95 & 0.970 \\
\wstar (H) & \textbf{0.787} & 0.991 & \textbf{0.928} & & \textbf{0.82} & \textbf{0.56} & 0.27 & &  0.17 & 0.15\\
\wstar (L) & 0.777 & \textbf{0.992} & 0.913 & & 0.81  & 0.55  & \textbf{0.26} & & 0.13  & 0.14\\
\wstar (R) & 0.654 & \textbf{0.992} & 0.927 & & 0.80 & 0.54 & 0.27 & & 0.25  & 0.21\\
\wstar (1R) & 0.686  & 0.988 & 0.873 & & 0.79 & 0.55  & 0.27 & & \textbf{0.0004}  & \textbf{0.04}\\
\bottomrule
\end{tabular}
\end{center}
\caption{Ablation study on Celeba-HQ dataset. \wstar (H) and \wstar (L) refer to Real NVP model with 13 and 4 coupling layers, respectively. For \wstar(R) and \wstar (1R) the classification model is not pretrained and one classification model for all attributes is used for the latter. Using one classification model per attribute gives significantly better results that using one model. Model capacity and weight initialization of the classification model are less significant.}
\label{tab:ablation}
\end{table}
Note that, scaling the perceptual distance is in favor of the distance unfolding, thus the linear attributes separation and disentanglement results are better than the one in the paper.

\subsubsection{Qualitative evaluation}

\begin{figure}[h]
\setlength\tabcolsep{2pt}%
\centering
\begin{tabular}{p{0.25cm}ccccccc}
\centering
&
 \textbf{Original} &
 \textbf{Inverted} &
 \textbf{Makeup} &
 \textbf{Male} &
 \textbf{Mustache} &
 \textbf{Chubby} &
 \textbf{Lipstick} \\
  \begin{turn}{90} \hspace{0.5cm} \wplus\end{turn} &
 \includegraphics[width=0.13\textwidth]{images/original/06002.jpg} & 
 \includegraphics[width=0.13\textwidth, ]{images/inversion/18_orig_img_1.jpg} &
 \includegraphics[width=0.13\textwidth, ]{images/wplus/18_img_3.jpg} &
 \includegraphics[width=0.13\textwidth, ]{images/wplus/20_img_3.jpg} &
 \includegraphics[width=0.13\textwidth, ]{images/wplus/22_img_3.jpg} &
 \includegraphics[width=0.13\textwidth, ]{images/wplus/13_img_3.jpg} &
 \includegraphics[width=0.13\textwidth]{images/wplus/36_img_3.jpg} \\
 \begin{turn}{90} \hspace{0.5cm} \wstara\end{turn} &
 \includegraphics[width=0.13\textwidth]{images/original/06002.jpg} & 
 \includegraphics[width=0.13\textwidth, ]{images/inversion/18_orig_img_1.jpg} &
 \includegraphics[width=0.13\textwidth, ]{images/wstar_no_mag_low_step/18_img_1.jpg} &
 \includegraphics[width=0.13\textwidth, ]{images/wstar_no_mag_low_step/20_img_1.jpg} &
 \includegraphics[width=0.13\textwidth, ]{images/wstar_no_mag_low_step/22_img_1.jpg} &
 \includegraphics[width=0.13\textwidth, ]{images/wstar_no_mag_low_step/13_img_1.jpg} &
 \includegraphics[width=0.13\textwidth, ]{images/wstar_no_mag_low_step/36_img_1.jpg} \\
\begin{turn}{90} \hspace{0.2cm} \wstara-ID\end{turn} &
 \includegraphics[width=0.13\textwidth]{images/original/06002.jpg} & 
 \includegraphics[width=0.13\textwidth, ]{images/inversion/18_orig_img_1.jpg} &
 \includegraphics[width=0.13\textwidth, ]{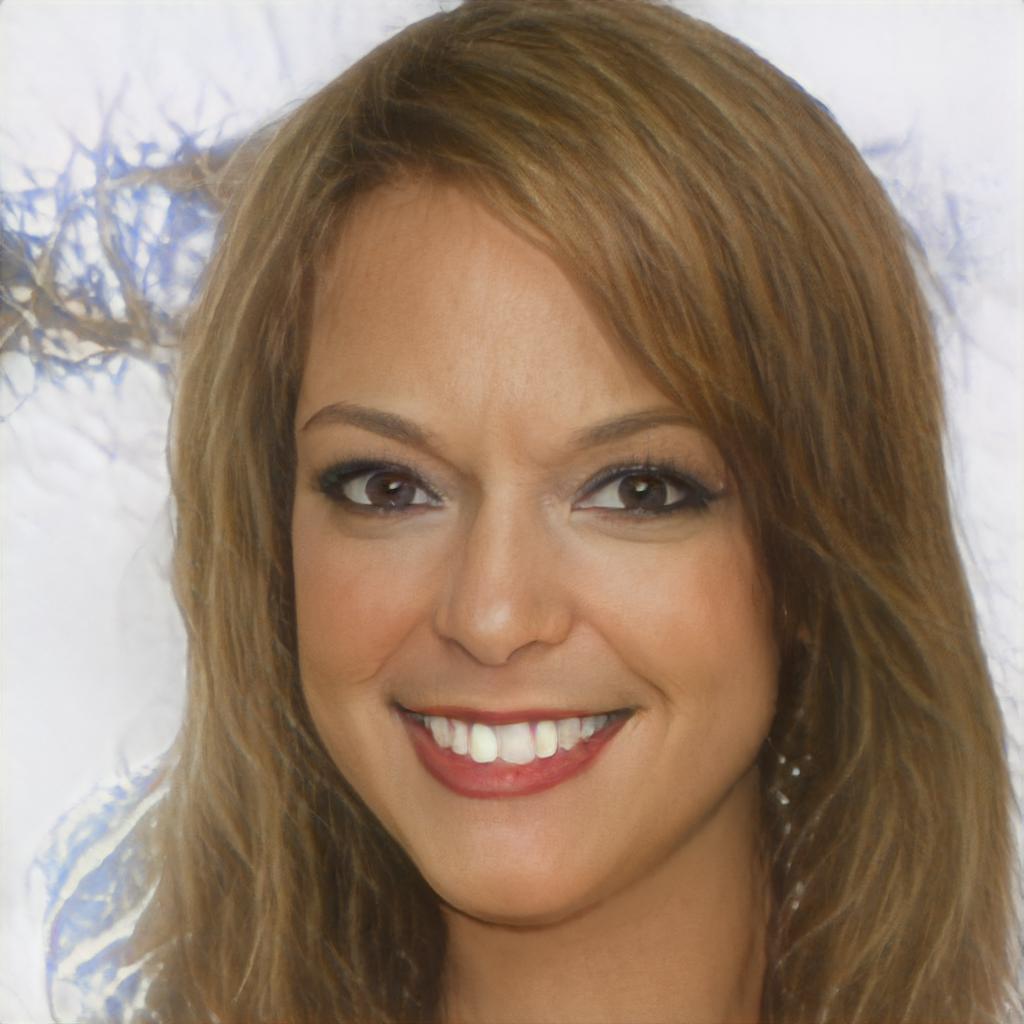} &
 \includegraphics[width=0.13\textwidth, ]{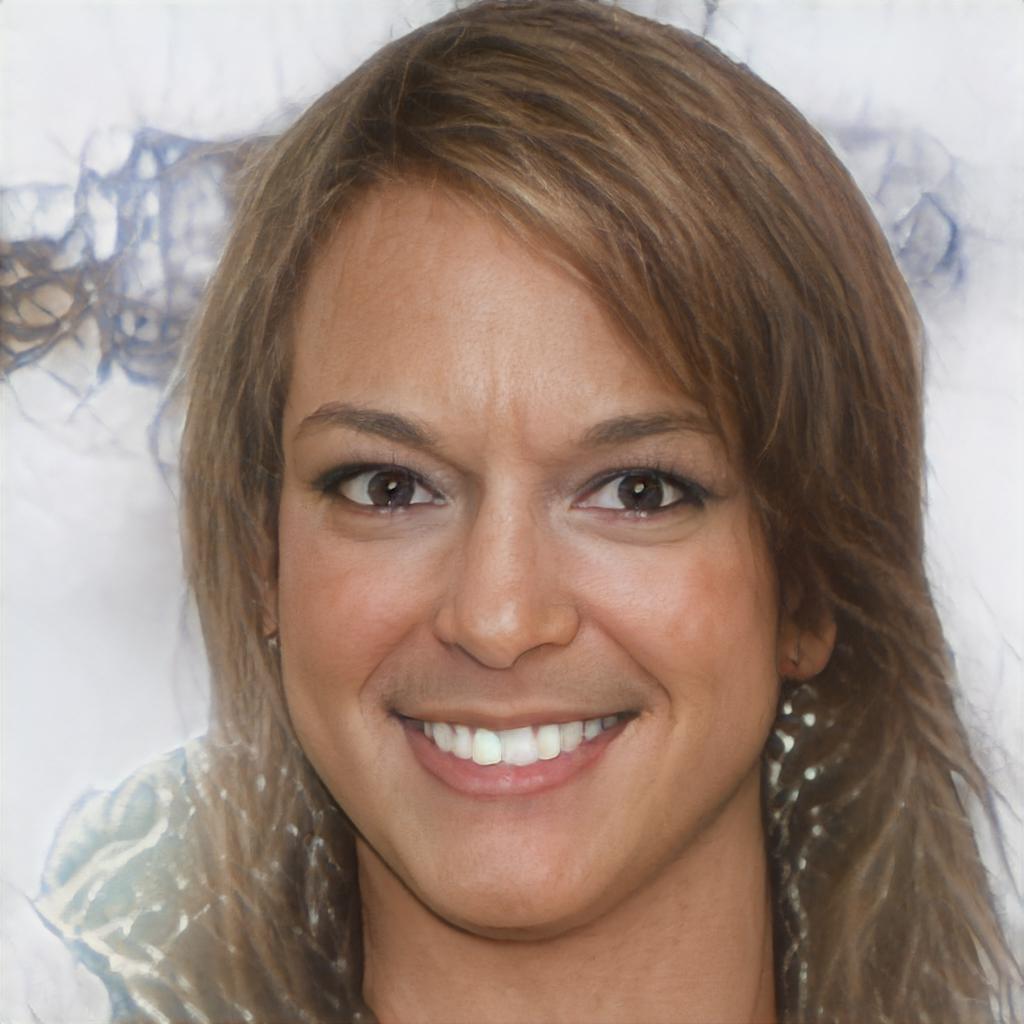} &
 \includegraphics[width=0.13\textwidth, ]{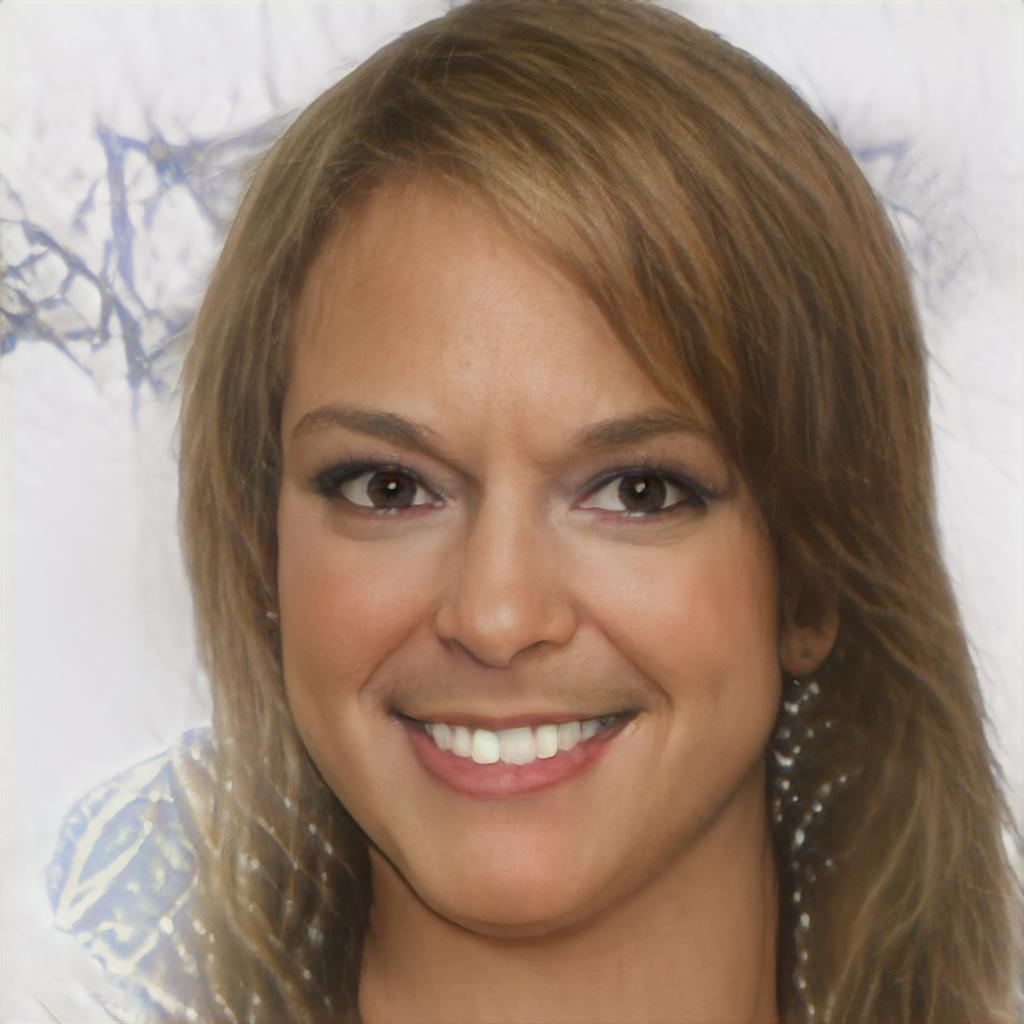} &
 \includegraphics[width=0.13\textwidth, ]{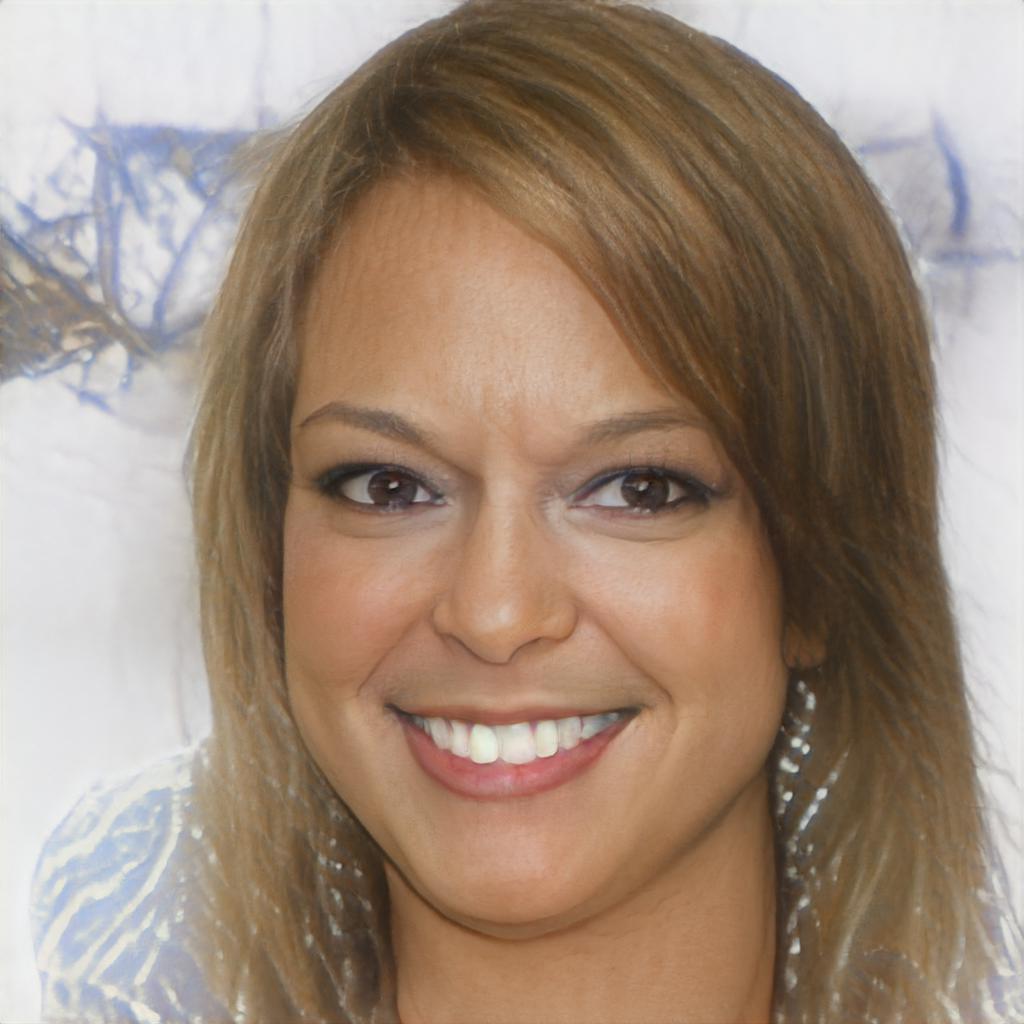} &
 \includegraphics[width=0.13\textwidth, ]{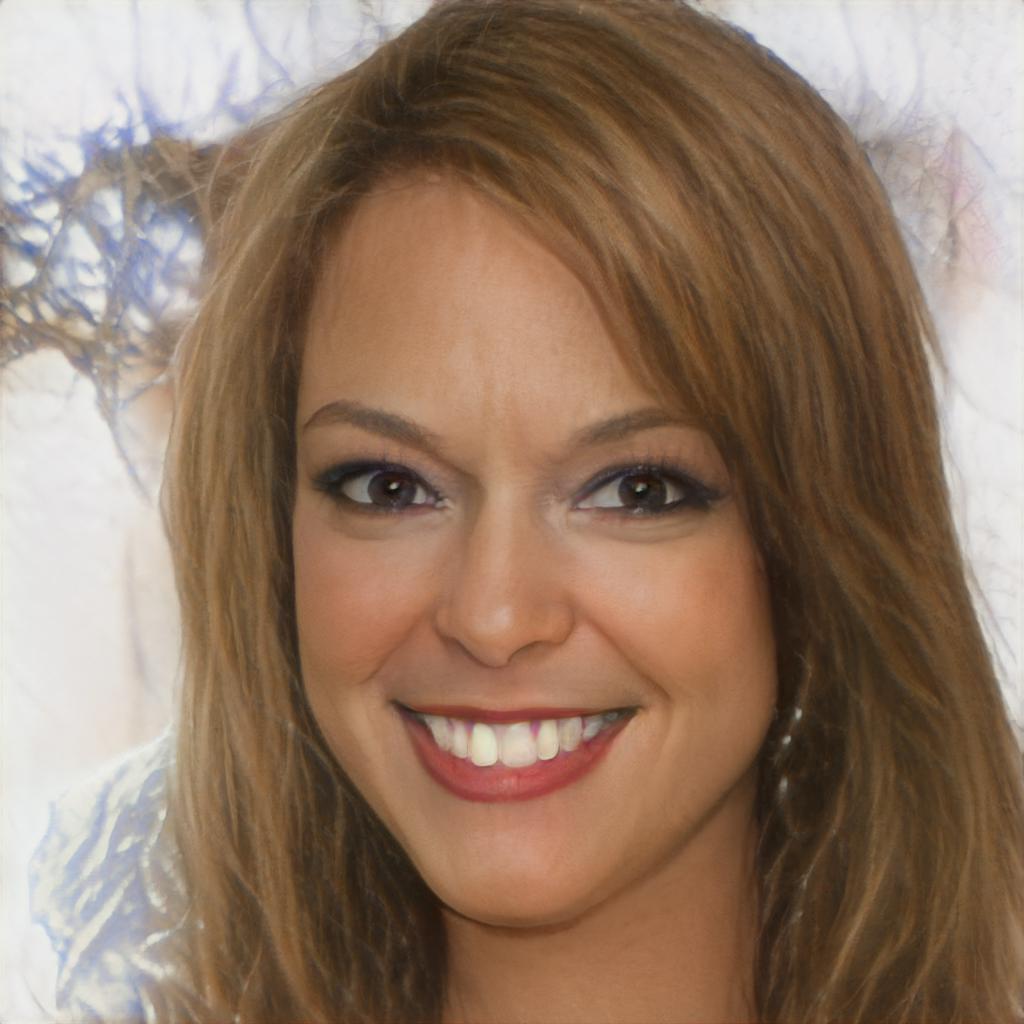} \\
 \begin{turn}{90}  \wstara$-ID^{\dagger}$\end{turn} &
 \includegraphics[width=0.13\textwidth]{images/original/06002.jpg} & 
 \includegraphics[width=0.13\textwidth, ]{images/inversion/18_orig_img_1.jpg} &
 \includegraphics[width=0.13\textwidth, ]{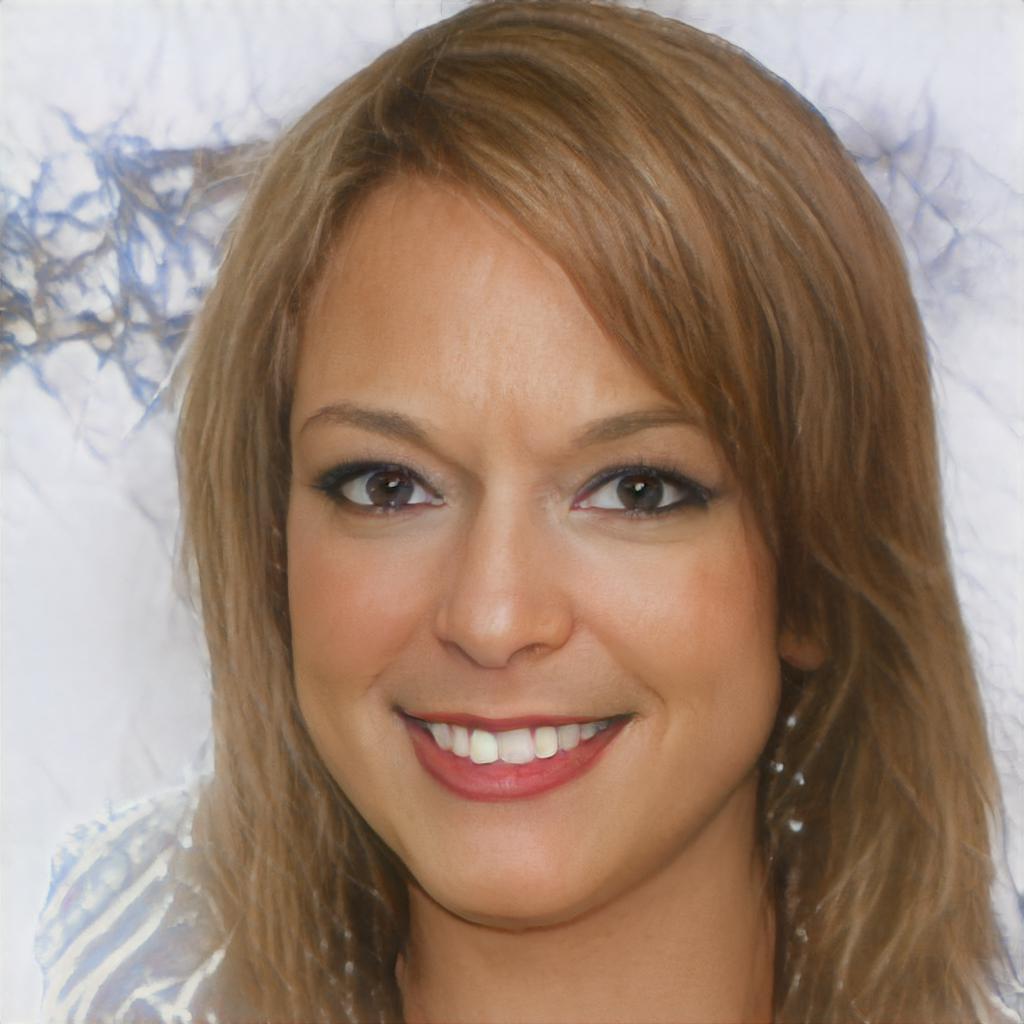} &
 \includegraphics[width=0.13\textwidth, ]{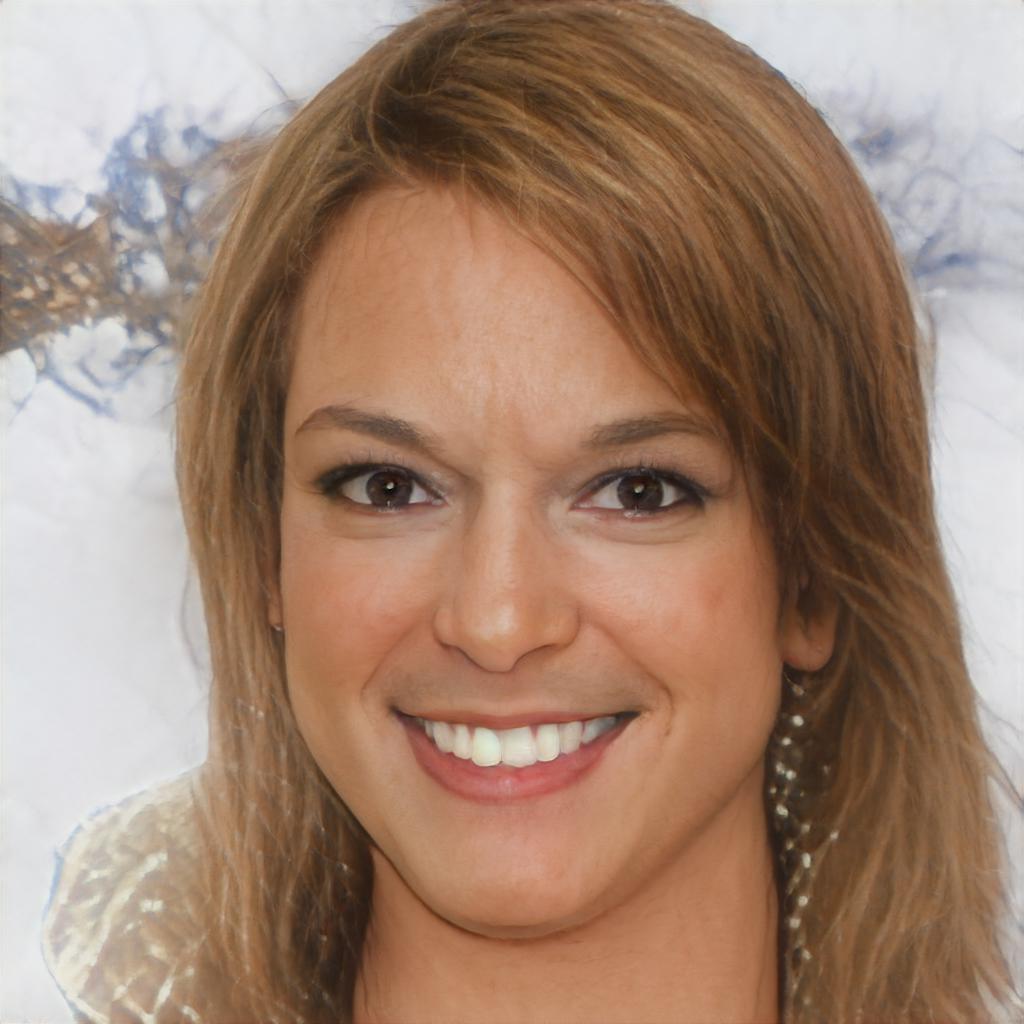} &
 \includegraphics[width=0.13\textwidth, ]{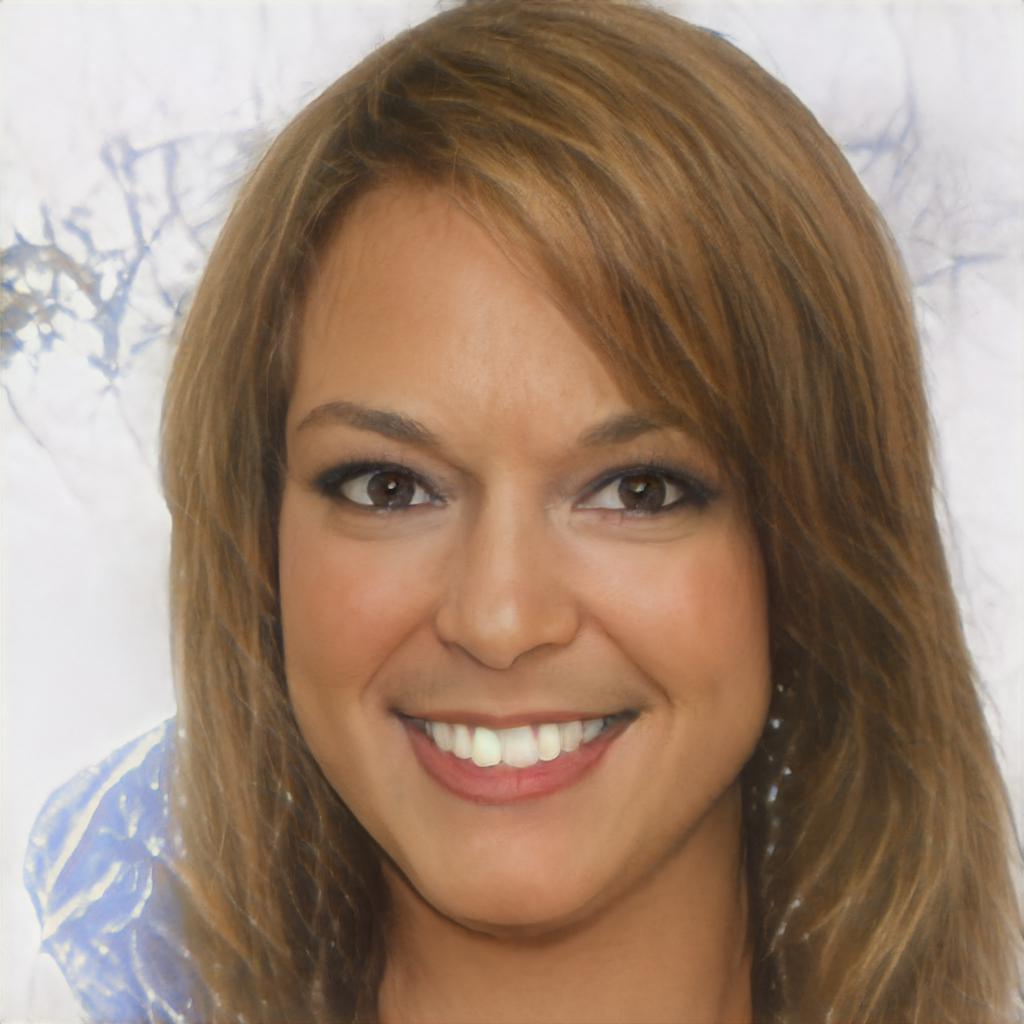} &
 \includegraphics[width=0.13\textwidth, ]{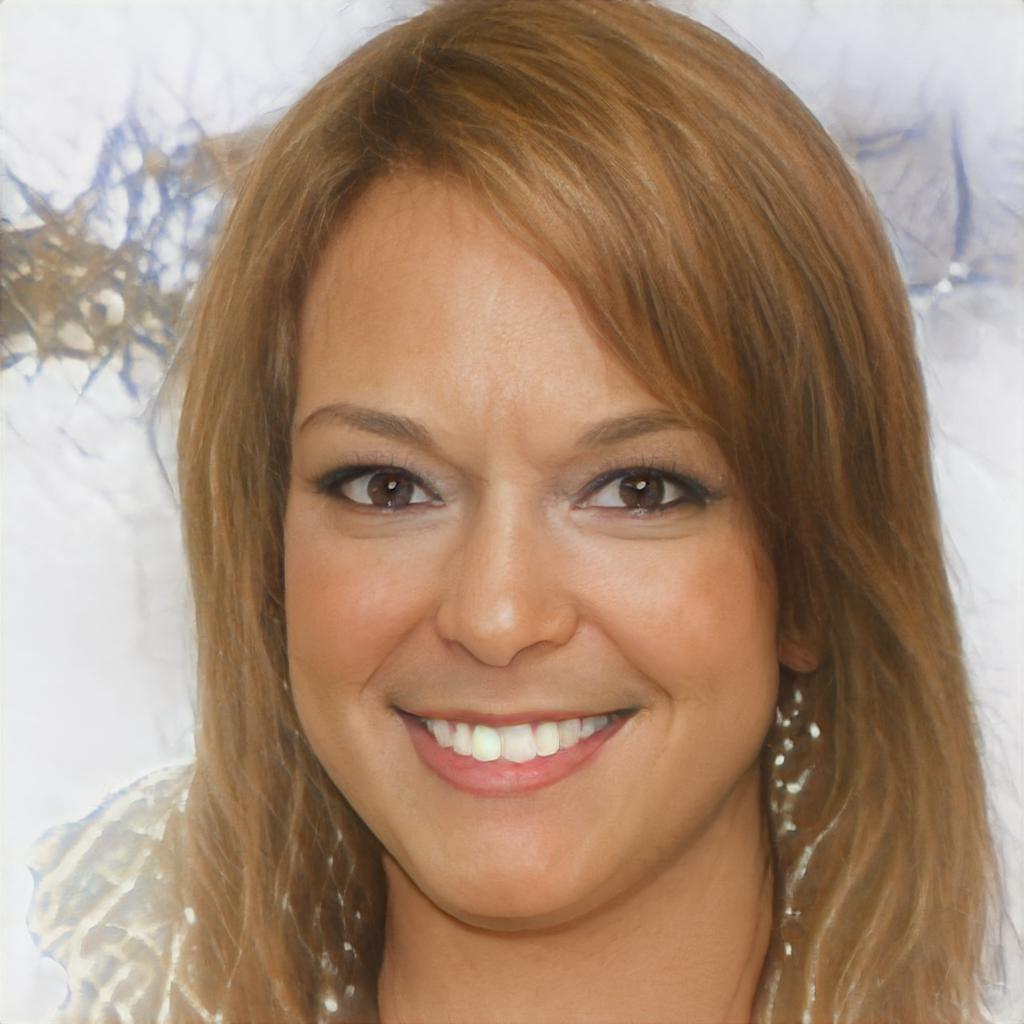} &
 \includegraphics[width=0.13\textwidth, ]{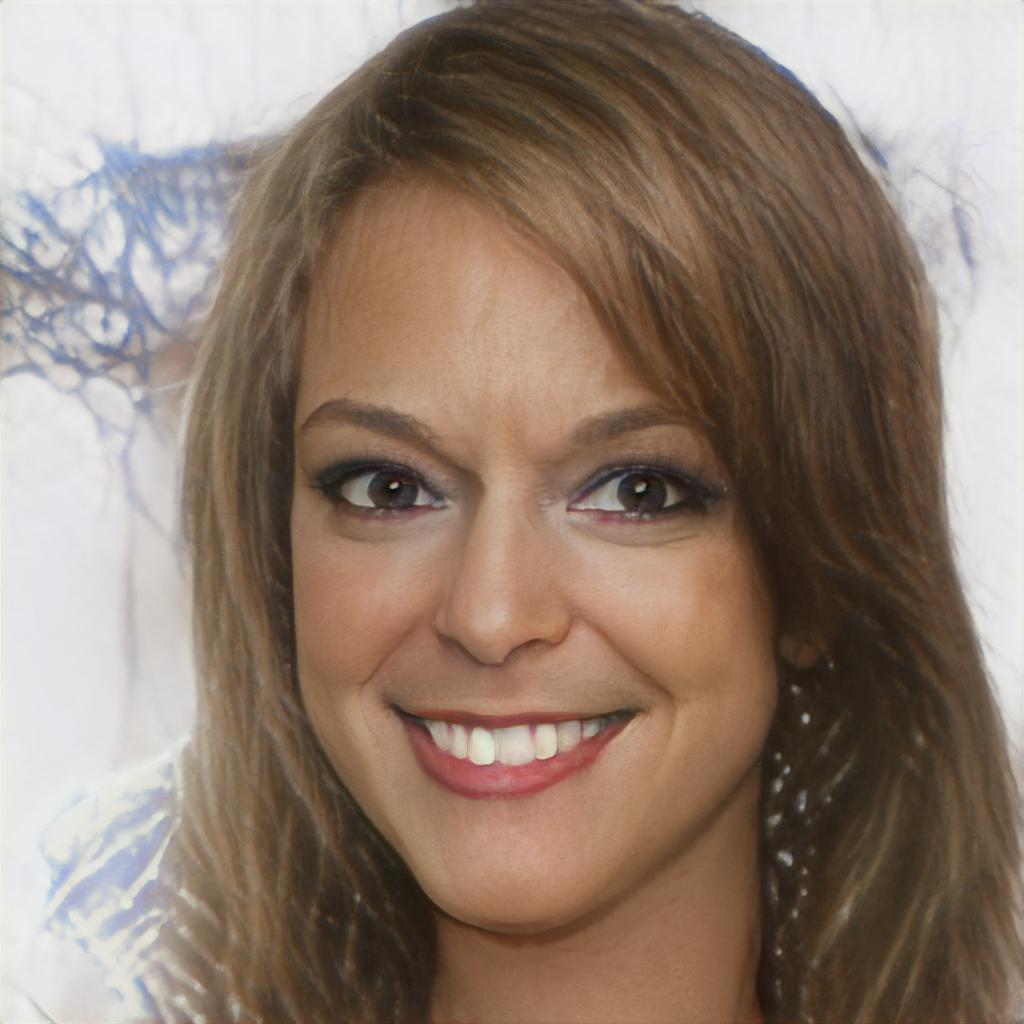} \\
 \begin{turn}{90}  \wstara$-ID^{\star}$\end{turn} &
 \includegraphics[width=0.13\textwidth]{images/original/06002.jpg} & 
 \includegraphics[width=0.13\textwidth, ]{images/inversion/18_orig_img_1.jpg} &
 \includegraphics[width=0.13\textwidth, ]{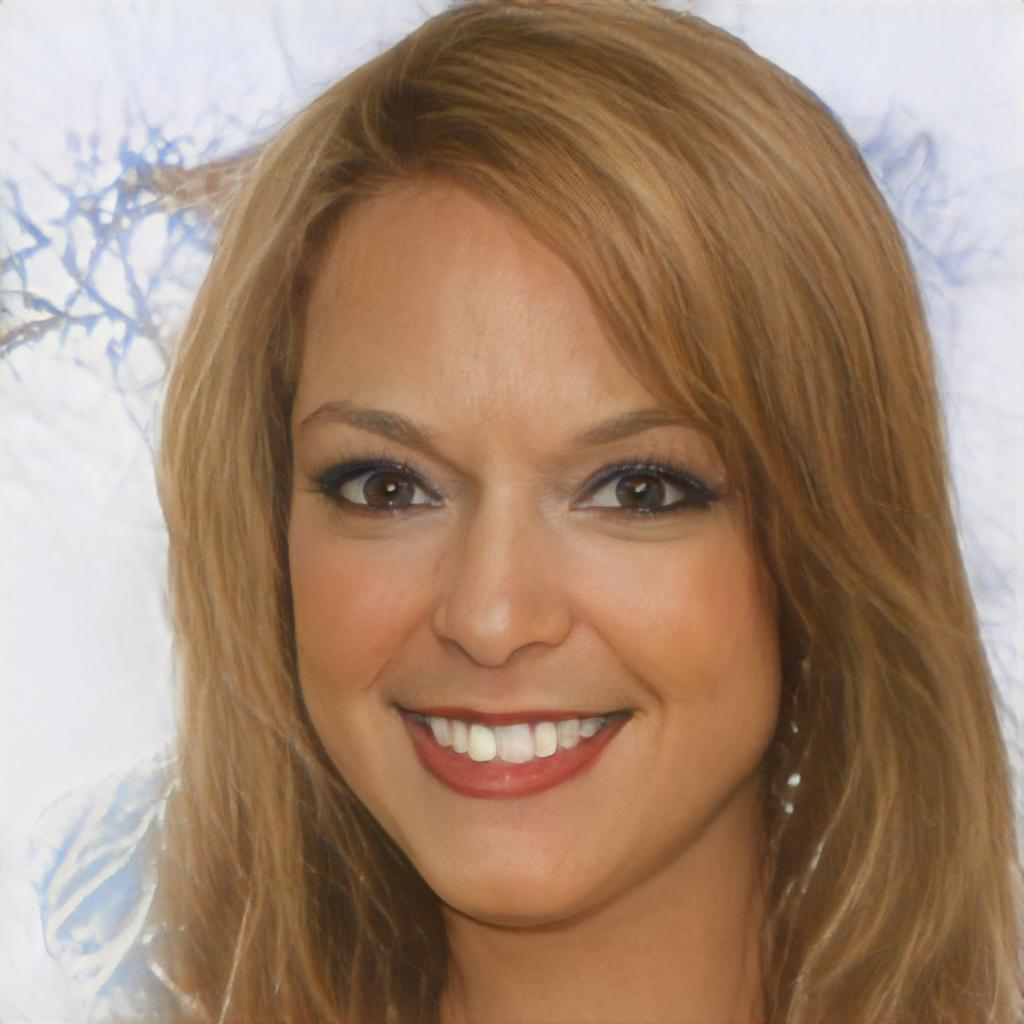} &
 \includegraphics[width=0.13\textwidth, ]{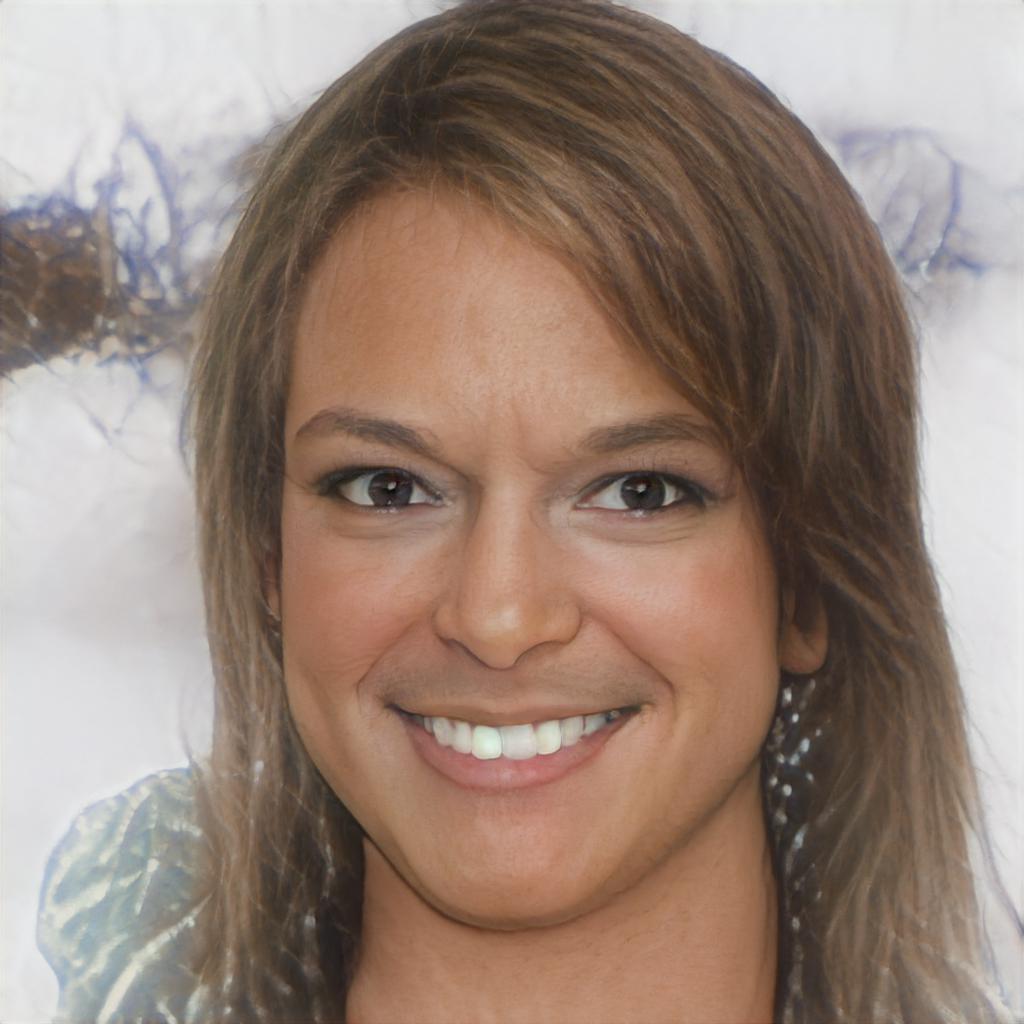} &
 \includegraphics[width=0.13\textwidth, ]{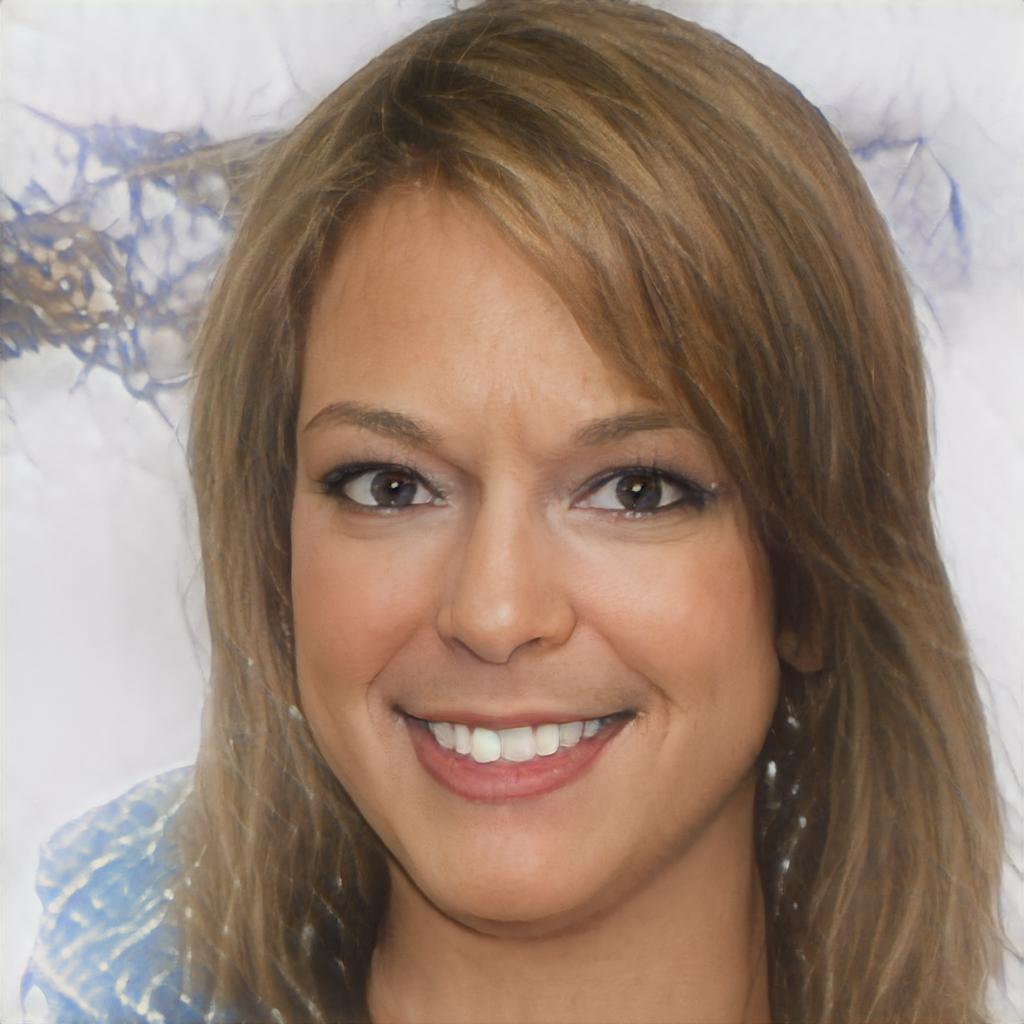} &
 \includegraphics[width=0.13\textwidth, ]{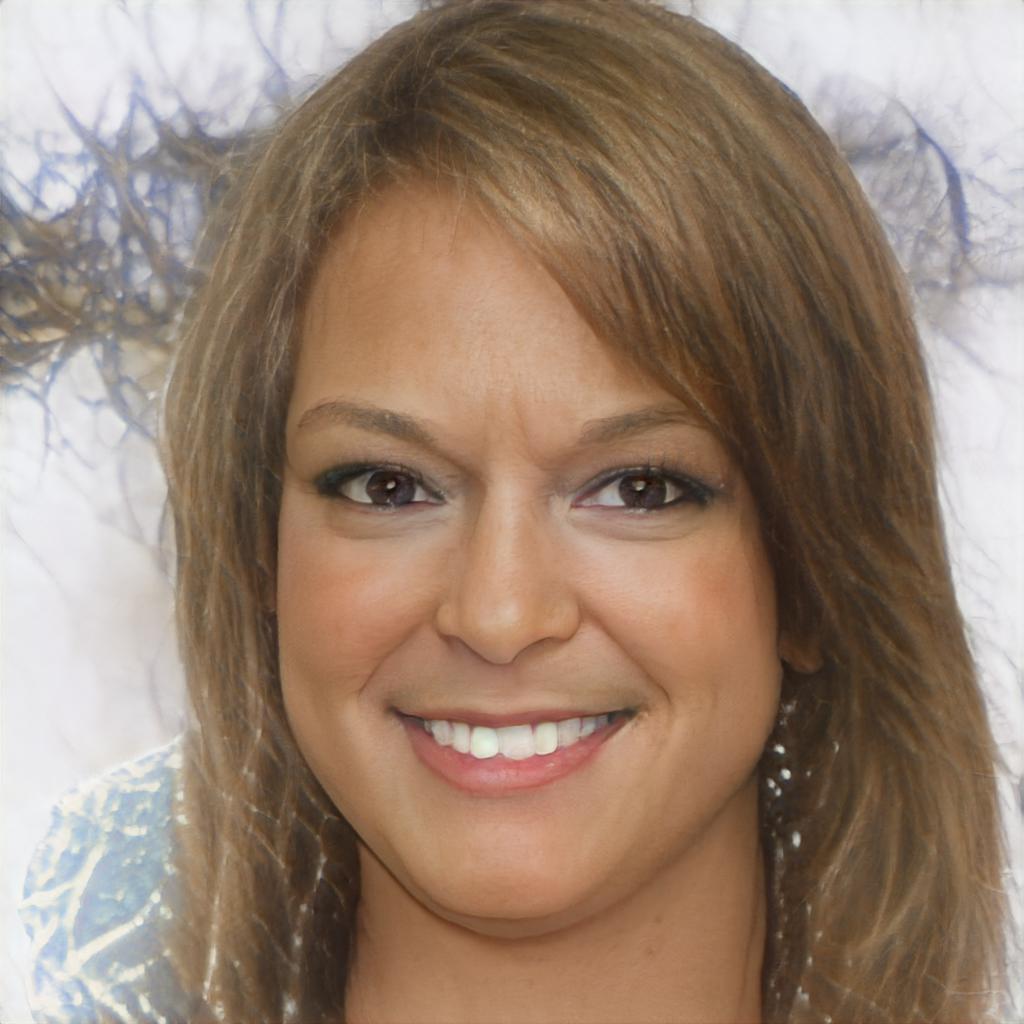} &
 \includegraphics[width=0.13\textwidth, ]{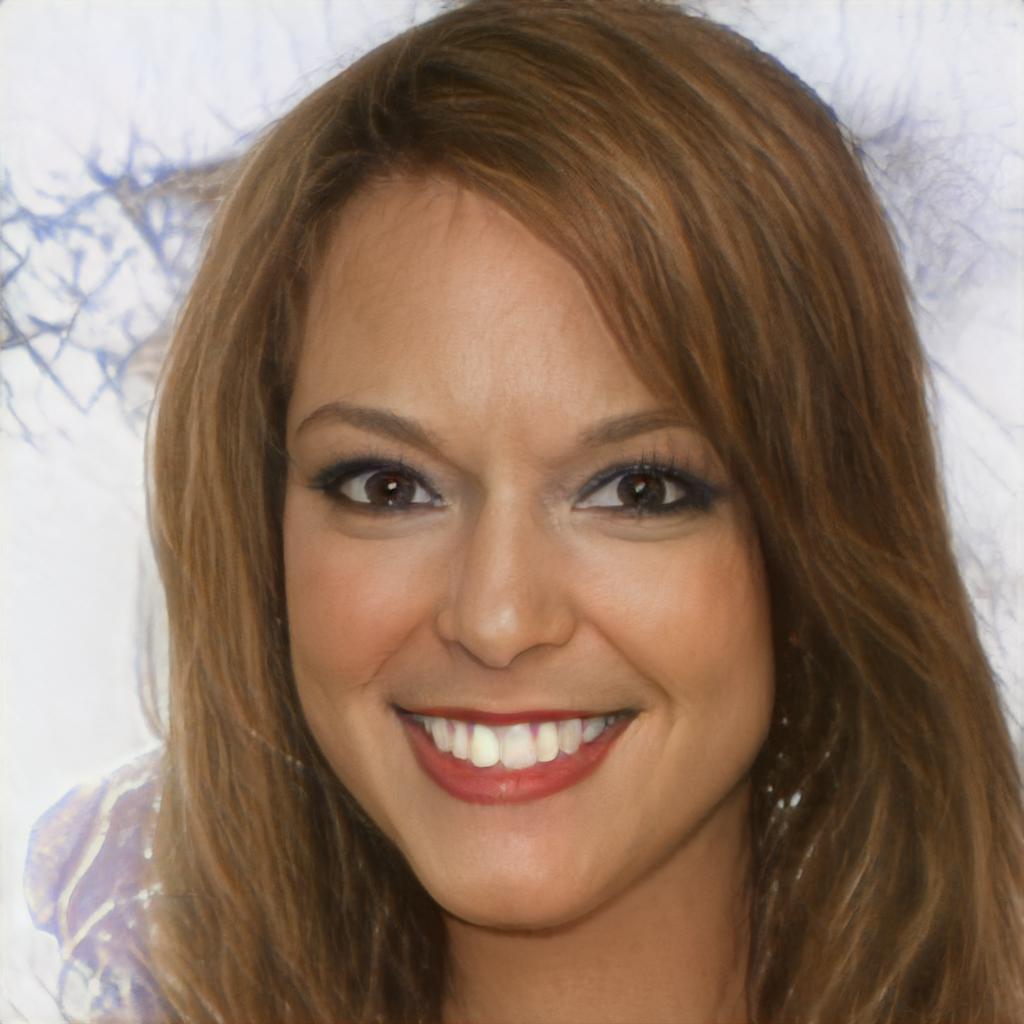} \\
 \begin{turn}{90} \hspace{0.4cm} $\mathcal{W}^{\star}_{ID}$  \end{turn} &
 \includegraphics[width=0.13\textwidth]{images/original/06002.jpg} & 
 \includegraphics[width=0.13\textwidth, ]{images/inversion/18_orig_img_1.jpg} &
 \includegraphics[width=0.13\textwidth, ]{images/wstar_id_no_mag_3_coupl_dist_edit_train_ep5/18_img_1.jpg} &
 \includegraphics[width=0.13\textwidth, ]{images/wstar_id_no_mag_3_coupl_dist_edit_train_ep5/20_img_1.jpg} &
 \includegraphics[width=0.13\textwidth, ]{images/wstar_id_no_mag_3_coupl_dist_edit_train_ep5/22_img_1.jpg} &
 \includegraphics[width=0.13\textwidth, ]{images/wstar_id_no_mag_3_coupl_dist_edit_train_ep5/13_img_1.jpg} &
 \includegraphics[width=0.13\textwidth, ]{images/wstar_id_no_mag_3_coupl_dist_edit_train_ep5/36_img_1.jpg} \\
 \\
\end{tabular}
\caption{Ablation study: Image editing using InterFaceGAN.}
\label{fig:edit_abl_2}
\end{figure}

\begin{figure}[h]
\setlength\tabcolsep{2pt}%
\centering
\begin{tabular}{p{0.25cm}ccccccc}
\centering
&
 \textbf{Original} &
 \textbf{Inverted} &
 \textbf{Makeup} &
 \textbf{Male} &
 \textbf{Mustache} &
 \textbf{Chubby} &
 \textbf{Lipstick} \\
  \begin{turn}{90} \hspace{0.5cm} \wplus\end{turn} &
 \includegraphics[width=0.13\textwidth]{images/original/06011.jpg} & 
 \includegraphics[width=0.13\textwidth, ]{images/inversion/18_orig_img_10.jpg} &
 \includegraphics[width=0.13\textwidth, ]{images/wplus/18_img_12.jpg} &
 \includegraphics[width=0.13\textwidth, ]{images/wplus/20_img_12.jpg} &
 \includegraphics[width=0.13\textwidth, ]{images/wplus/22_img_12.jpg} &
 \includegraphics[width=0.13\textwidth, ]{images/wplus/13_img_12.jpg} &
 \includegraphics[width=0.13\textwidth]{images/wplus/36_img_12.jpg} \\
 \begin{turn}{90} \hspace{0.5cm} \wstara\end{turn} &
 \includegraphics[width=0.13\textwidth]{images/original/06011.jpg} & 
 \includegraphics[width=0.13\textwidth, ]{images/inversion/18_orig_img_10.jpg} &
 \includegraphics[width=0.13\textwidth, ]{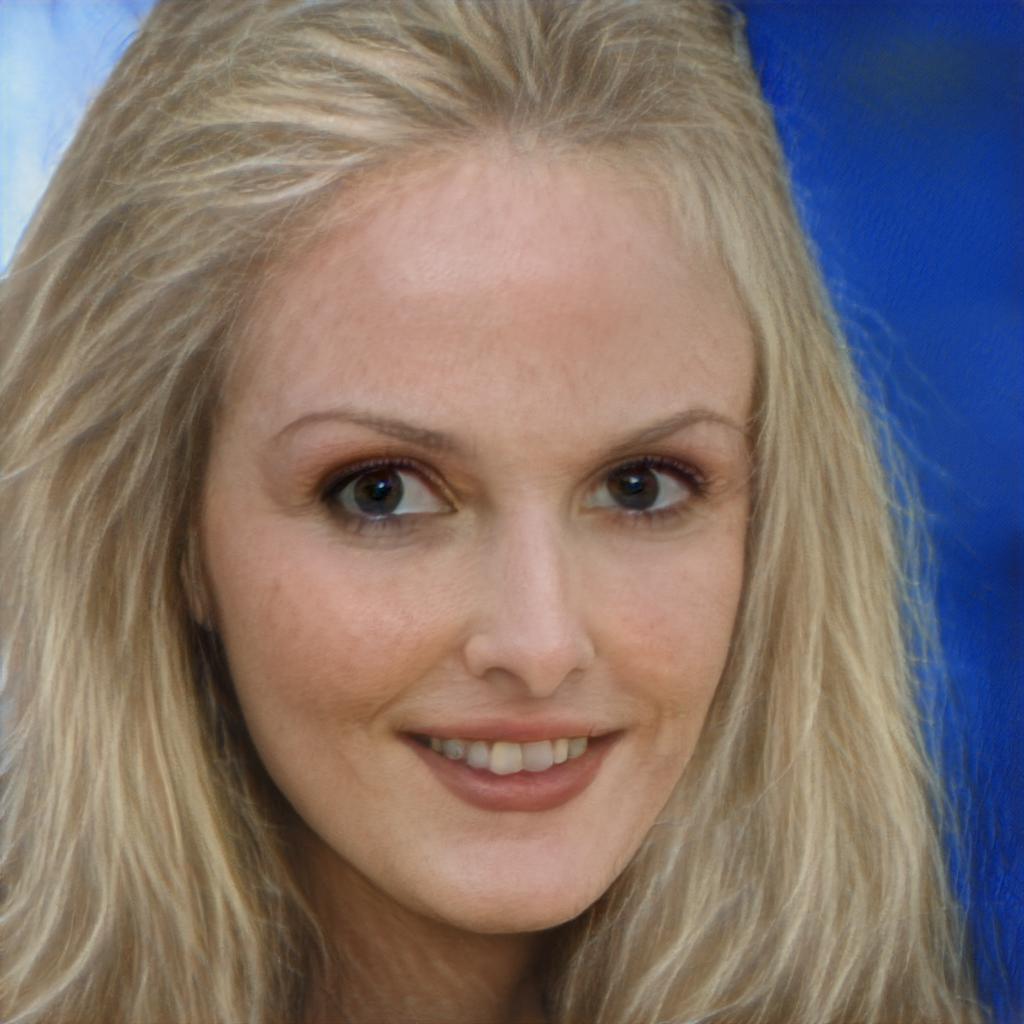} &
 \includegraphics[width=0.13\textwidth, ]{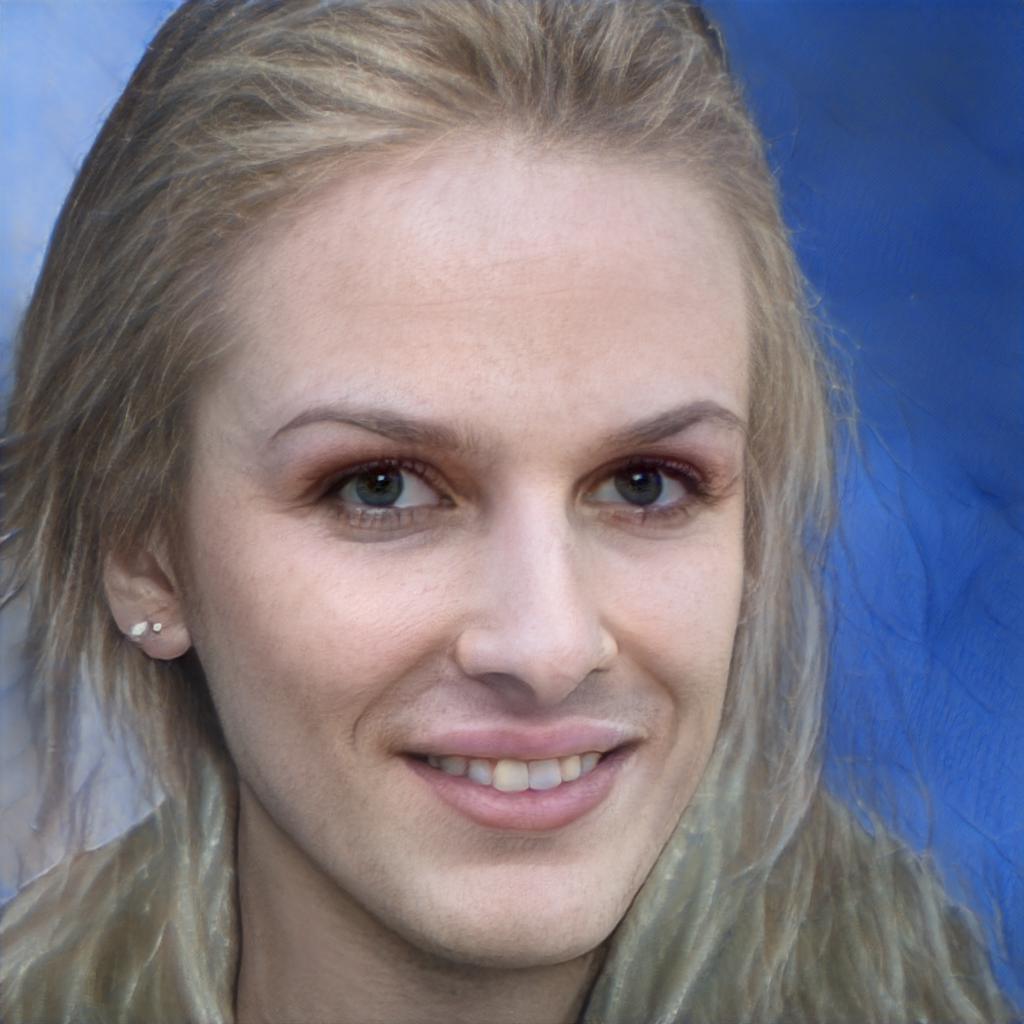} &
 \includegraphics[width=0.13\textwidth, ]{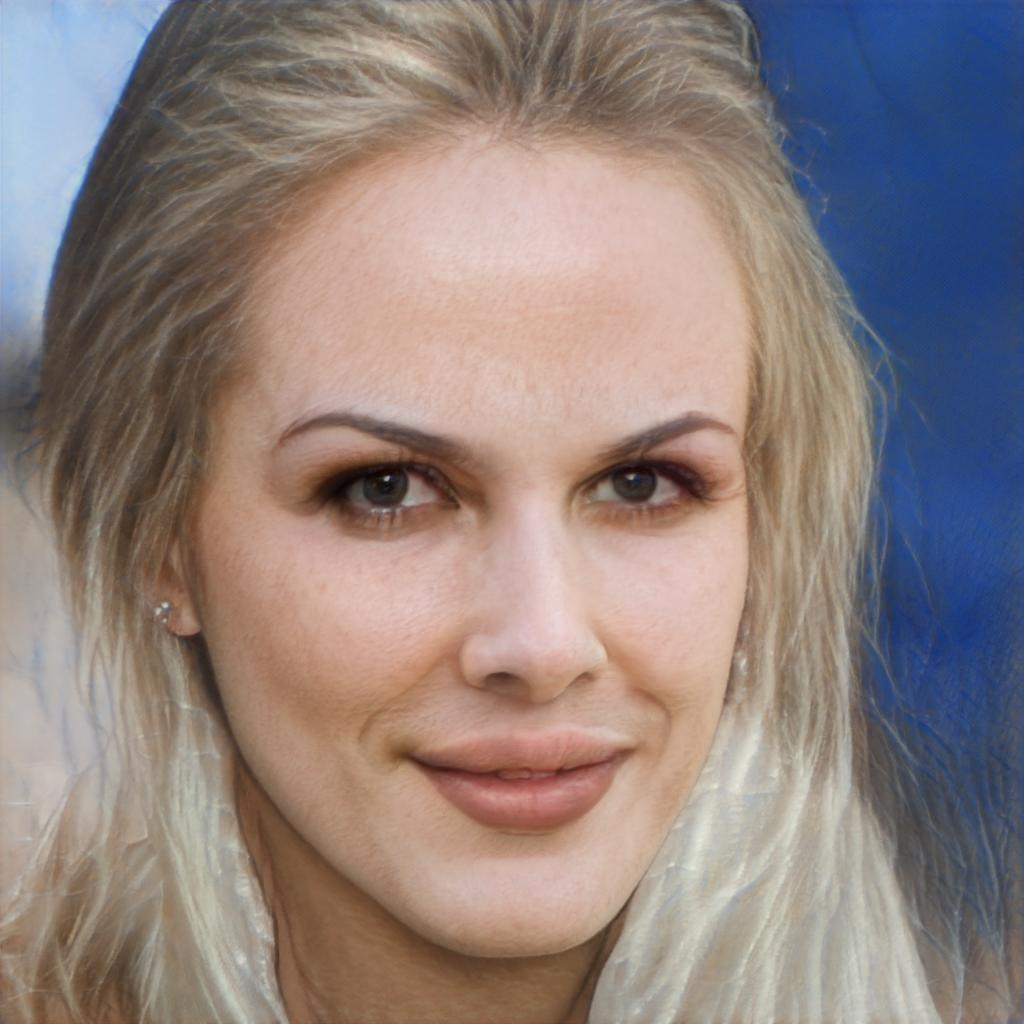} &
 \includegraphics[width=0.13\textwidth, ]{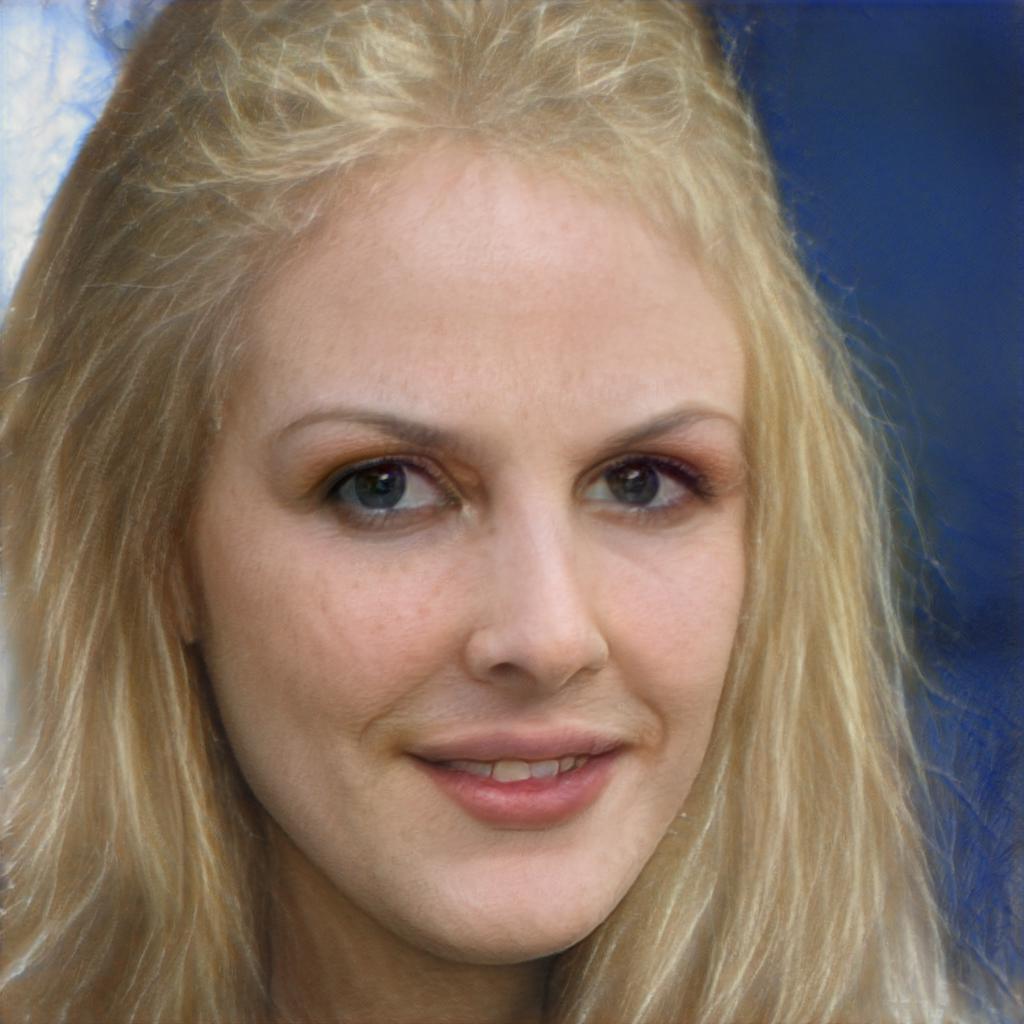} &
 \includegraphics[width=0.13\textwidth, ]{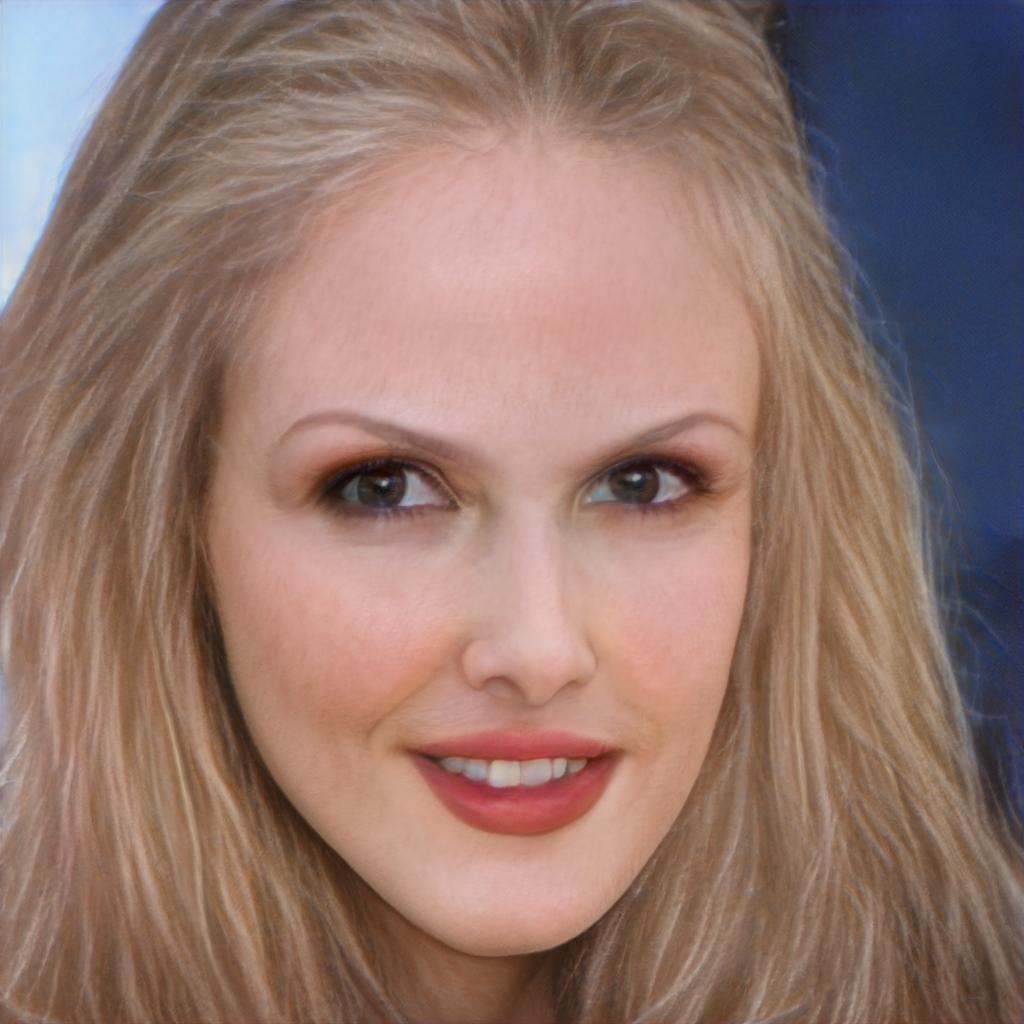} \\
 \begin{turn}{90} \hspace{0.2cm} \wstara-ID\end{turn} &
 \includegraphics[width=0.13\textwidth]{images/original/06011.jpg} & 
 \includegraphics[width=0.13\textwidth, ]{images/inversion/18_orig_img_10.jpg} &
 \includegraphics[width=0.13\textwidth, ]{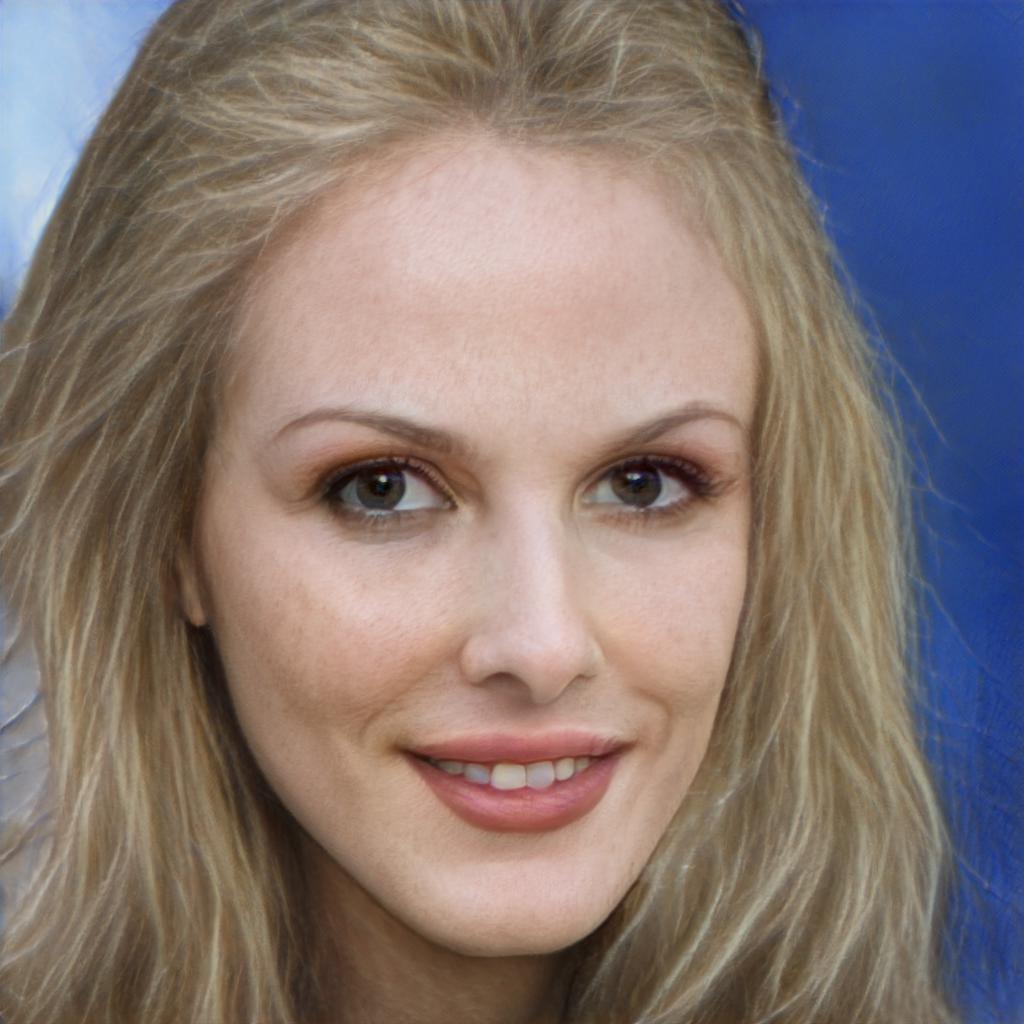} &
 \includegraphics[width=0.13\textwidth, ]{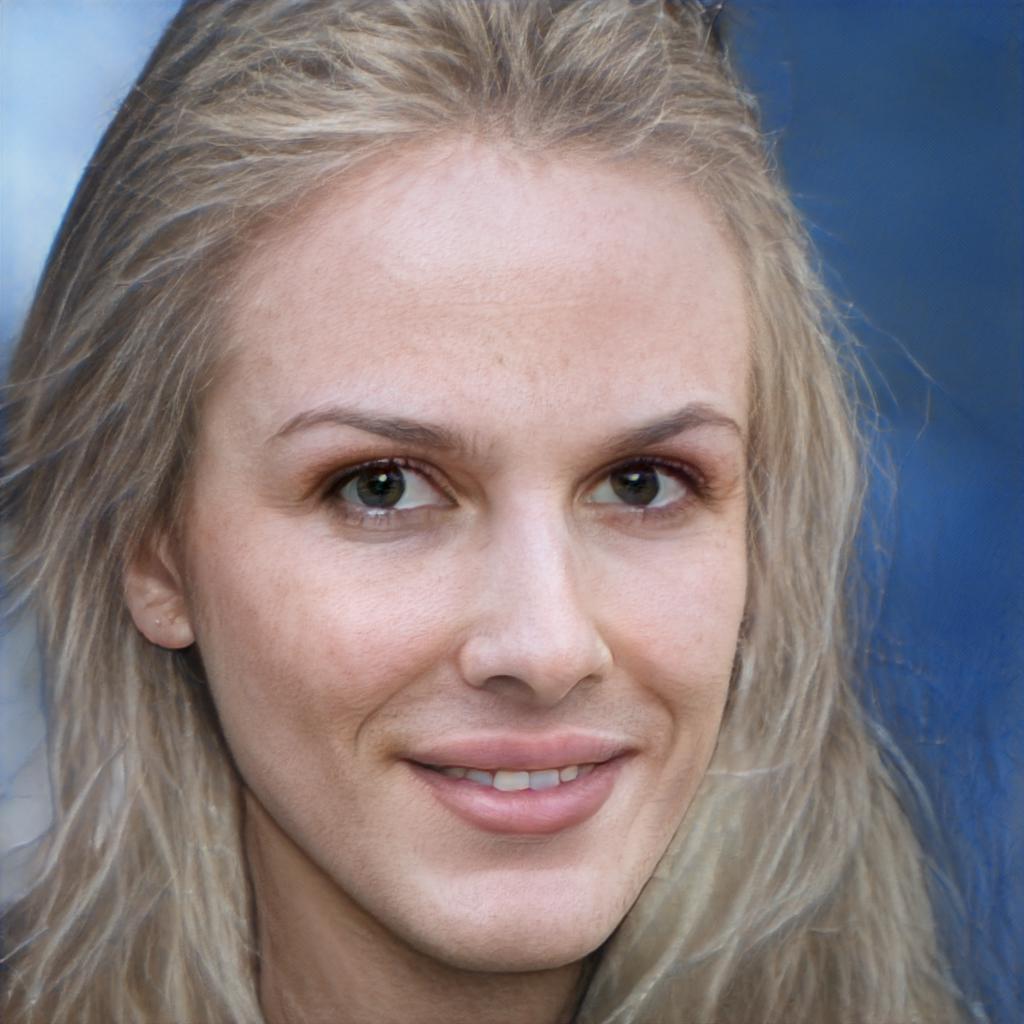} &
 \includegraphics[width=0.13\textwidth, ]{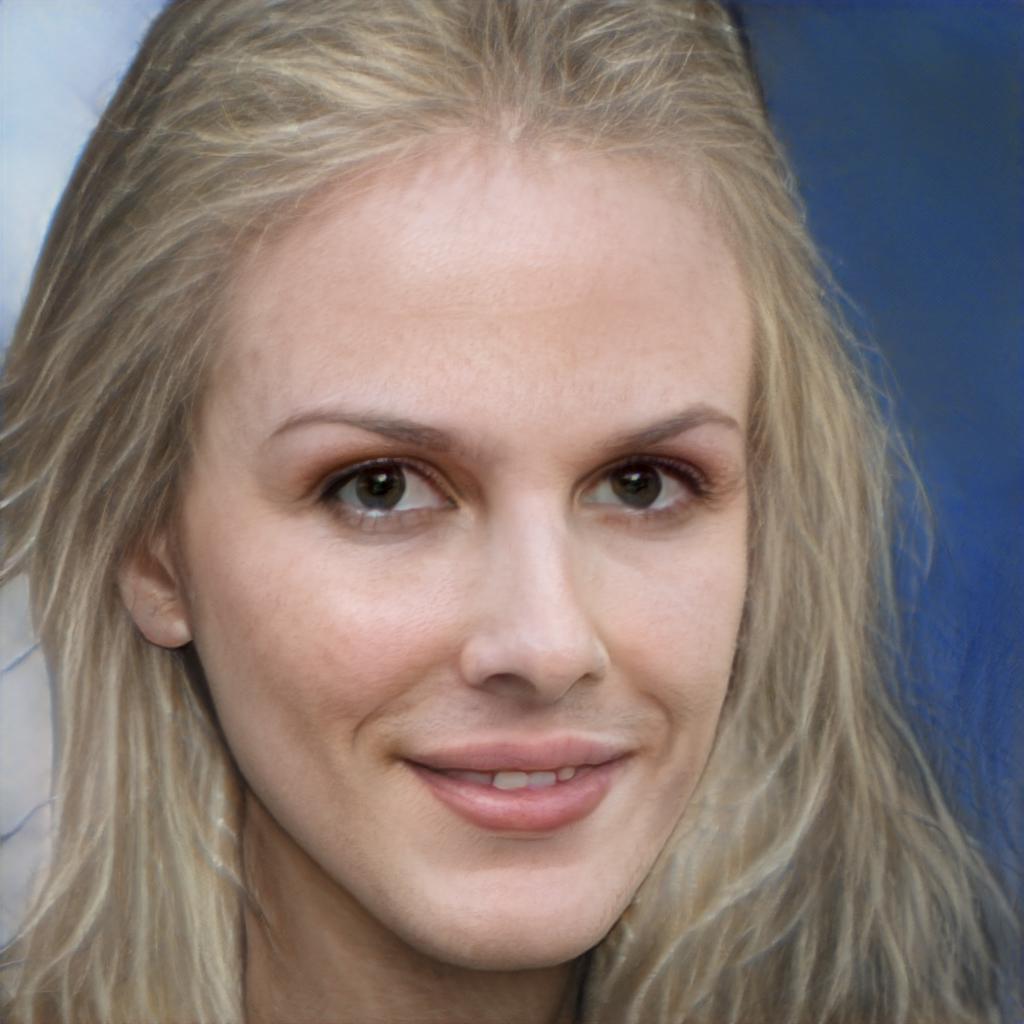} &
 \includegraphics[width=0.13\textwidth, ]{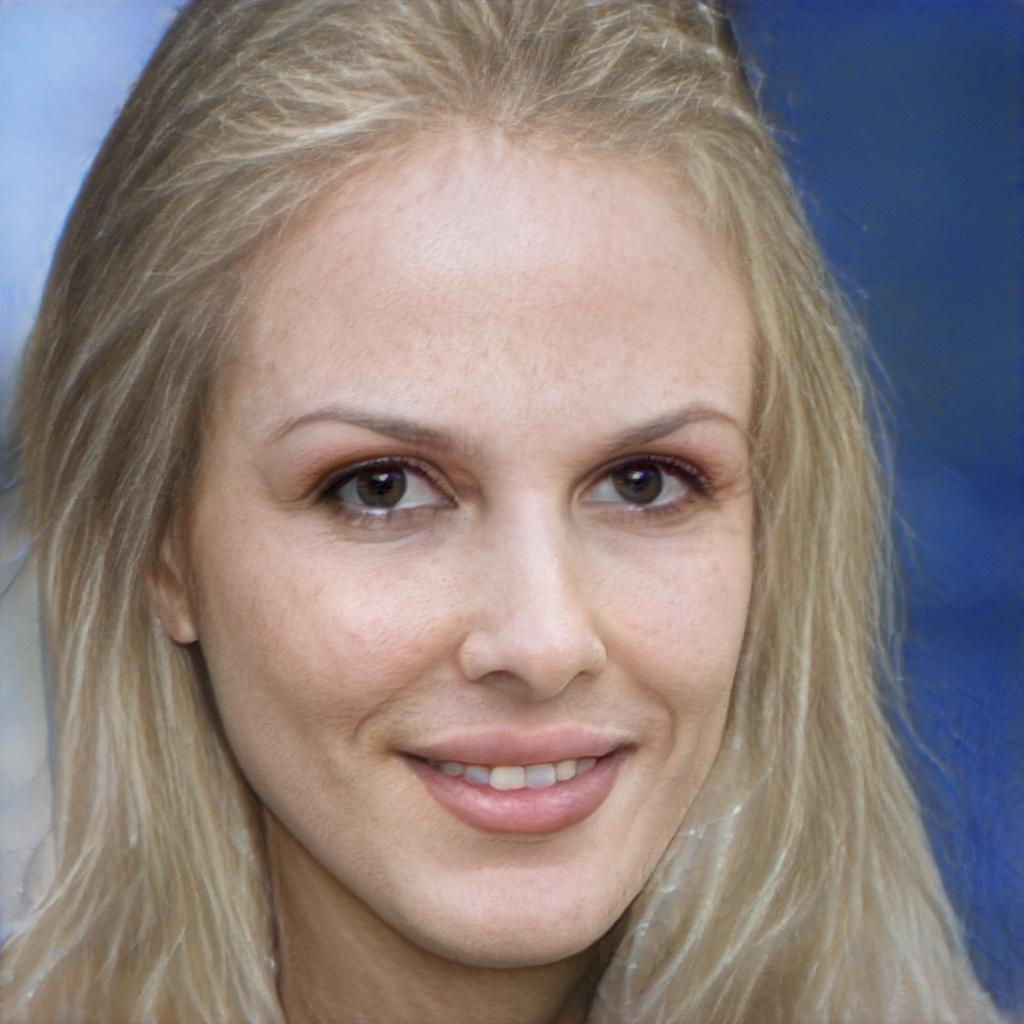} &
 \includegraphics[width=0.13\textwidth, ]{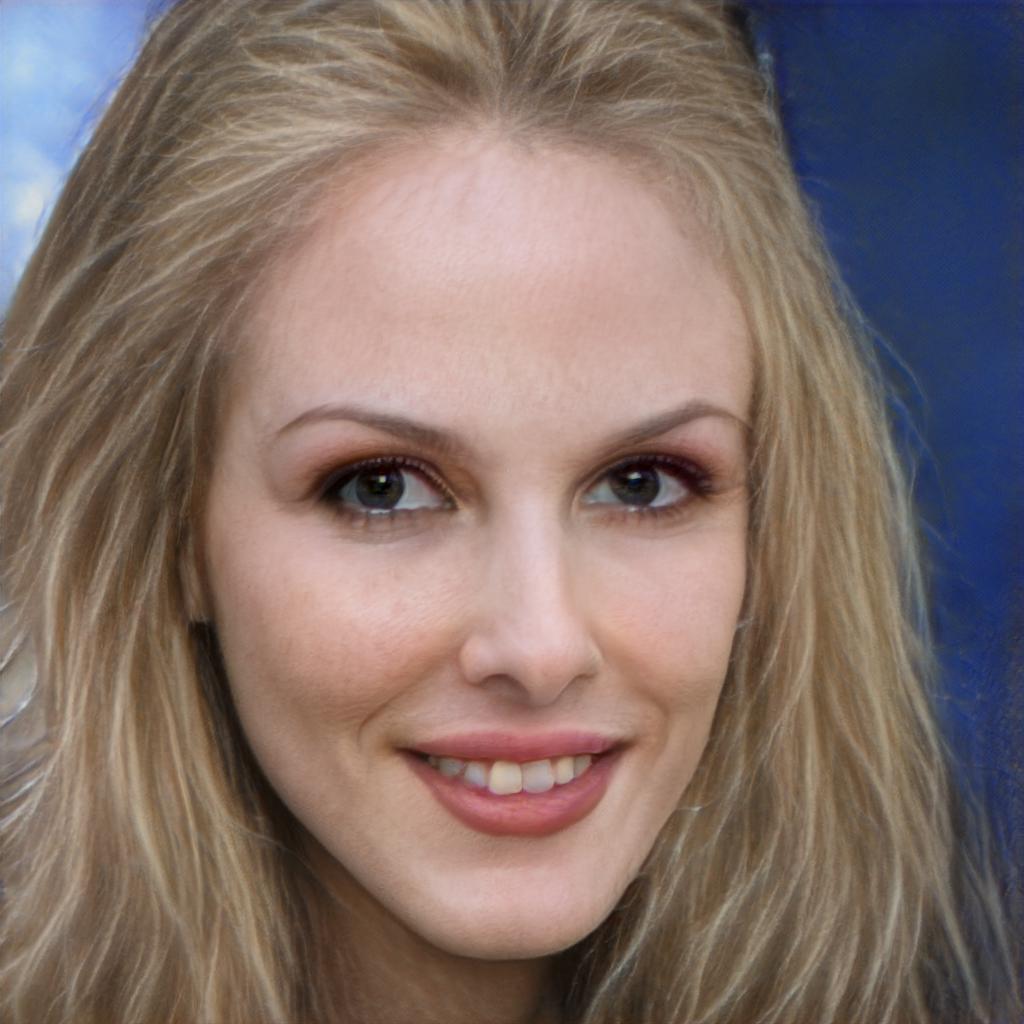} \\
 \begin{turn}{90}  \wstara$-ID^{\dagger}$\end{turn} &
 \includegraphics[width=0.13\textwidth]{images/original/06011.jpg} & 
 \includegraphics[width=0.13\textwidth, ]{images/inversion/18_orig_img_10.jpg} &
 \includegraphics[width=0.13\textwidth, ]{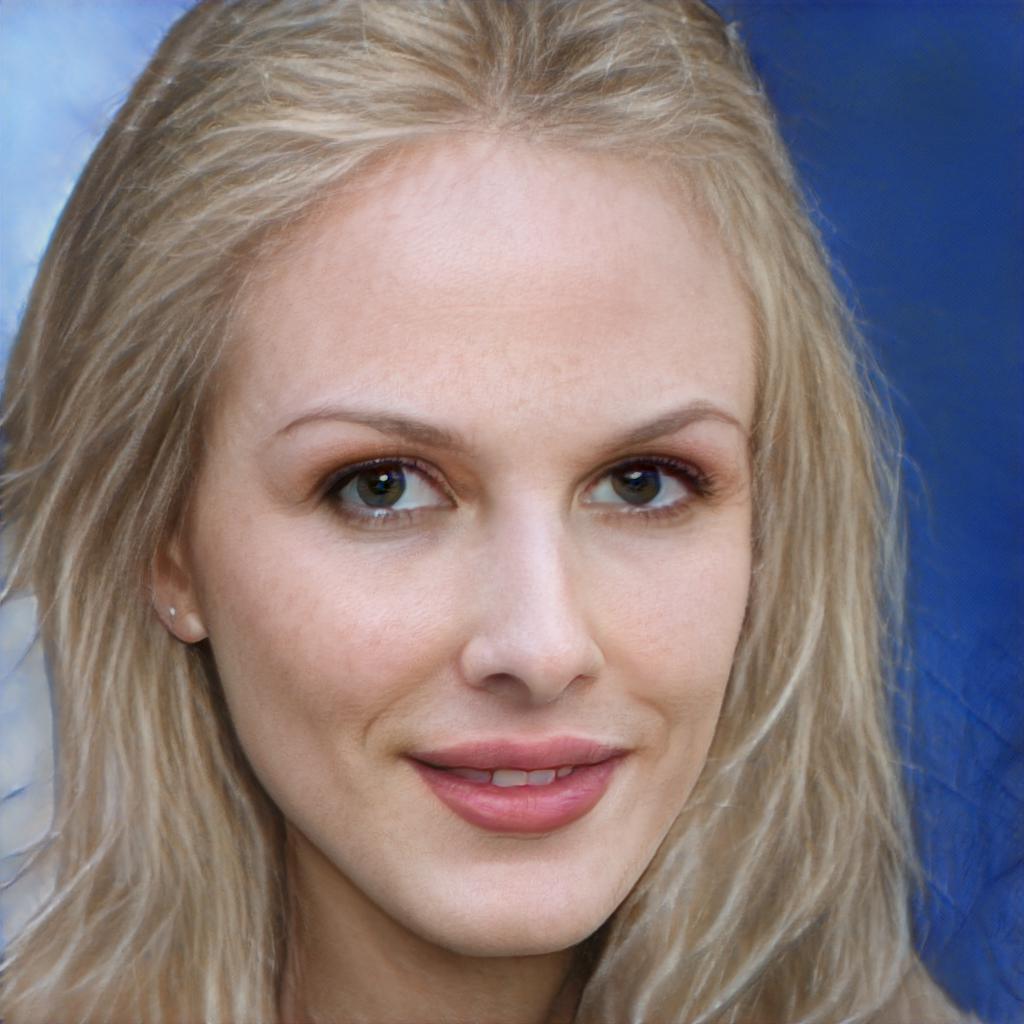} &
 \includegraphics[width=0.13\textwidth, ]{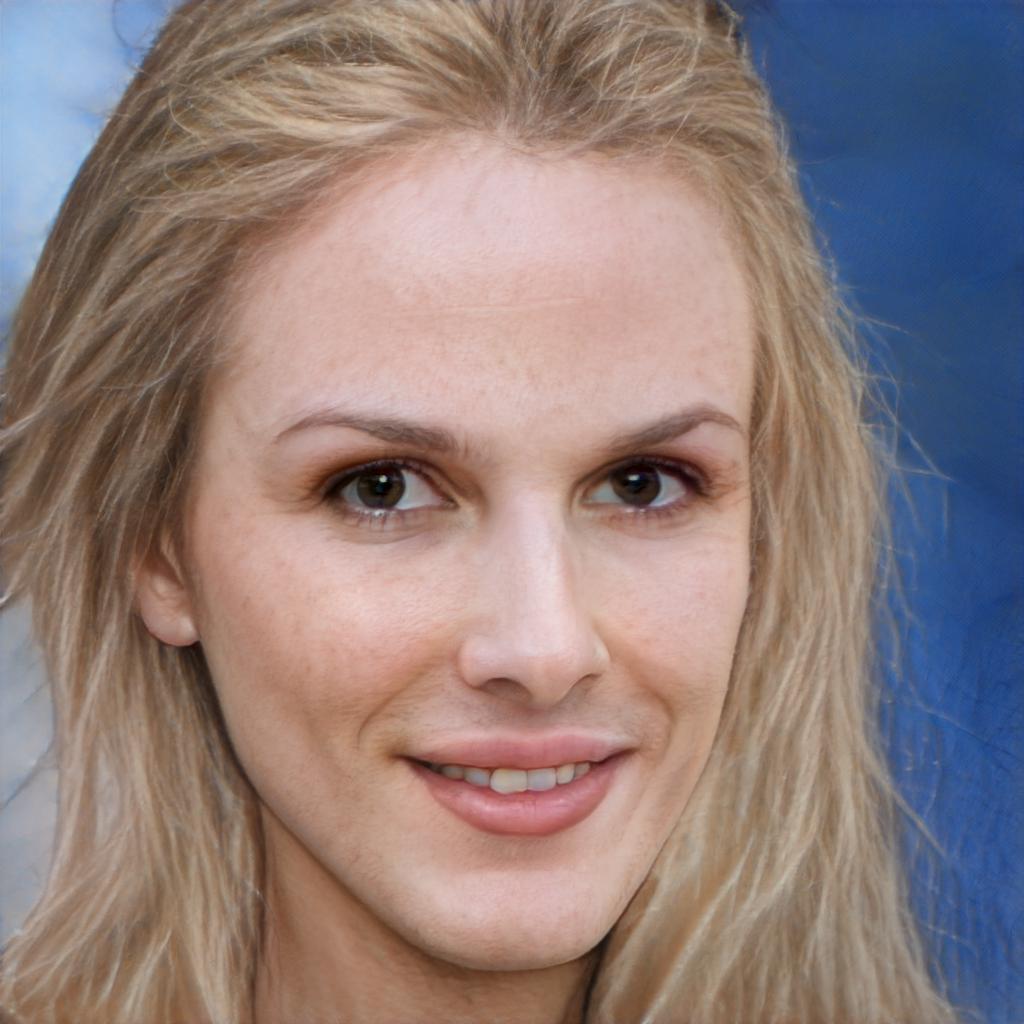} &
 \includegraphics[width=0.13\textwidth, ]{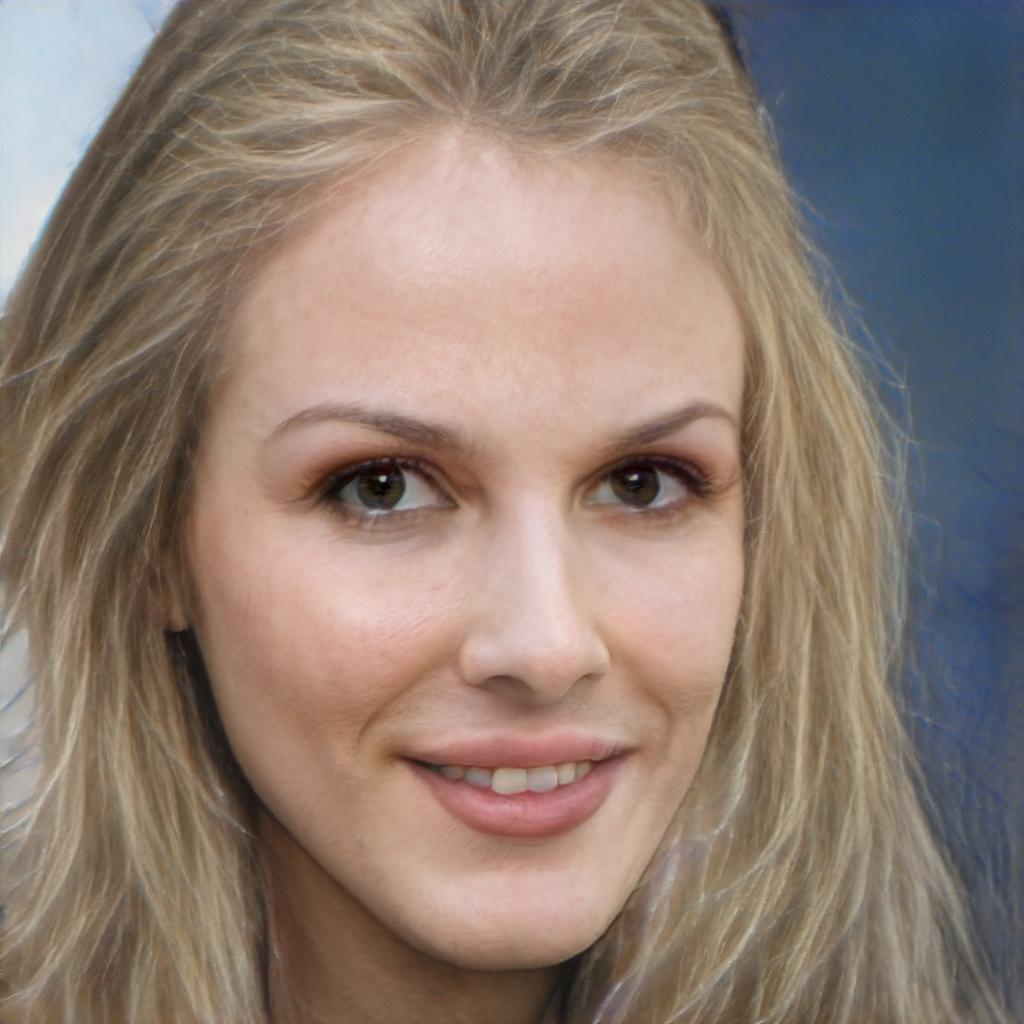} &
 \includegraphics[width=0.13\textwidth, ]{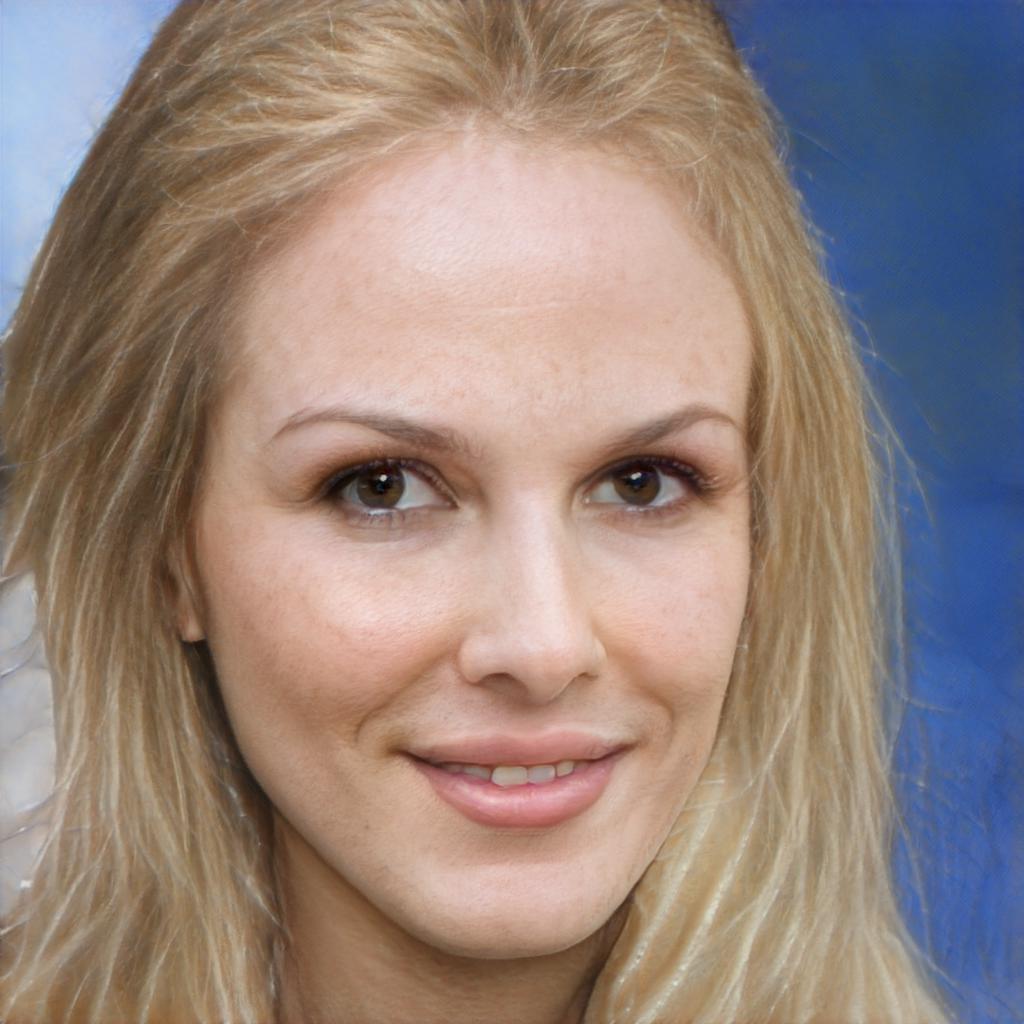} &
 \includegraphics[width=0.13\textwidth, ]{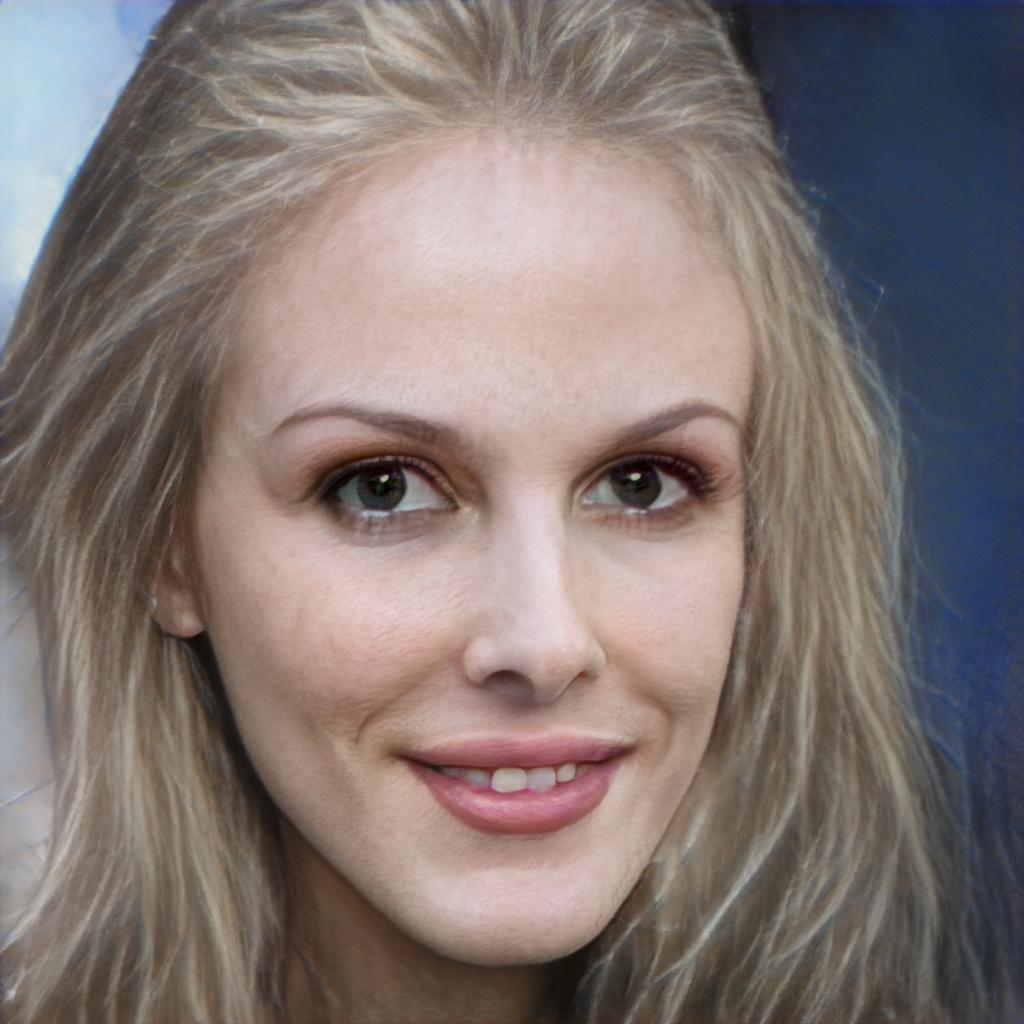} \\
 \begin{turn}{90}  \wstara$-ID^{\star}$\end{turn} &
 \includegraphics[width=0.13\textwidth]{images/original/06011.jpg} & 
 \includegraphics[width=0.13\textwidth, ]{images/inversion/18_orig_img_10.jpg} &
 \includegraphics[width=0.13\textwidth, ]{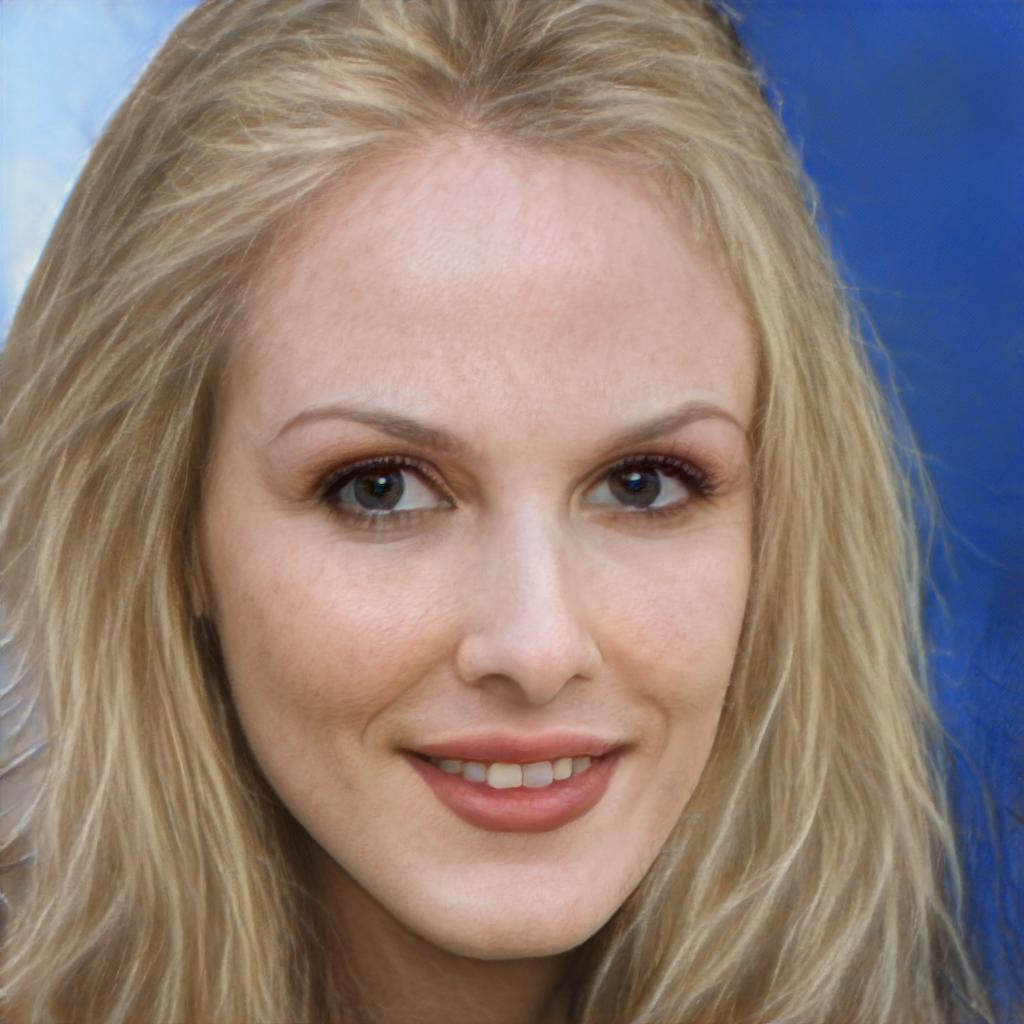} &
 \includegraphics[width=0.13\textwidth, ]{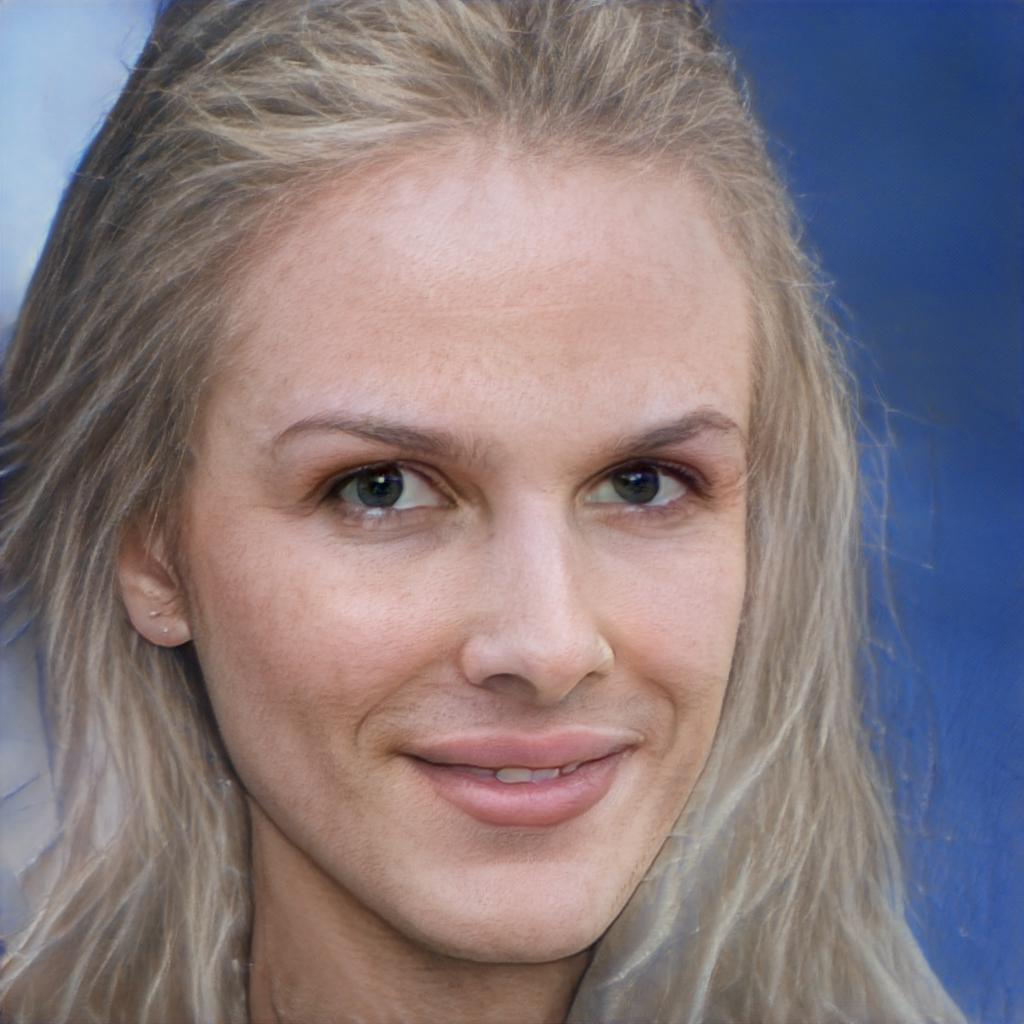} &
 \includegraphics[width=0.13\textwidth, ]{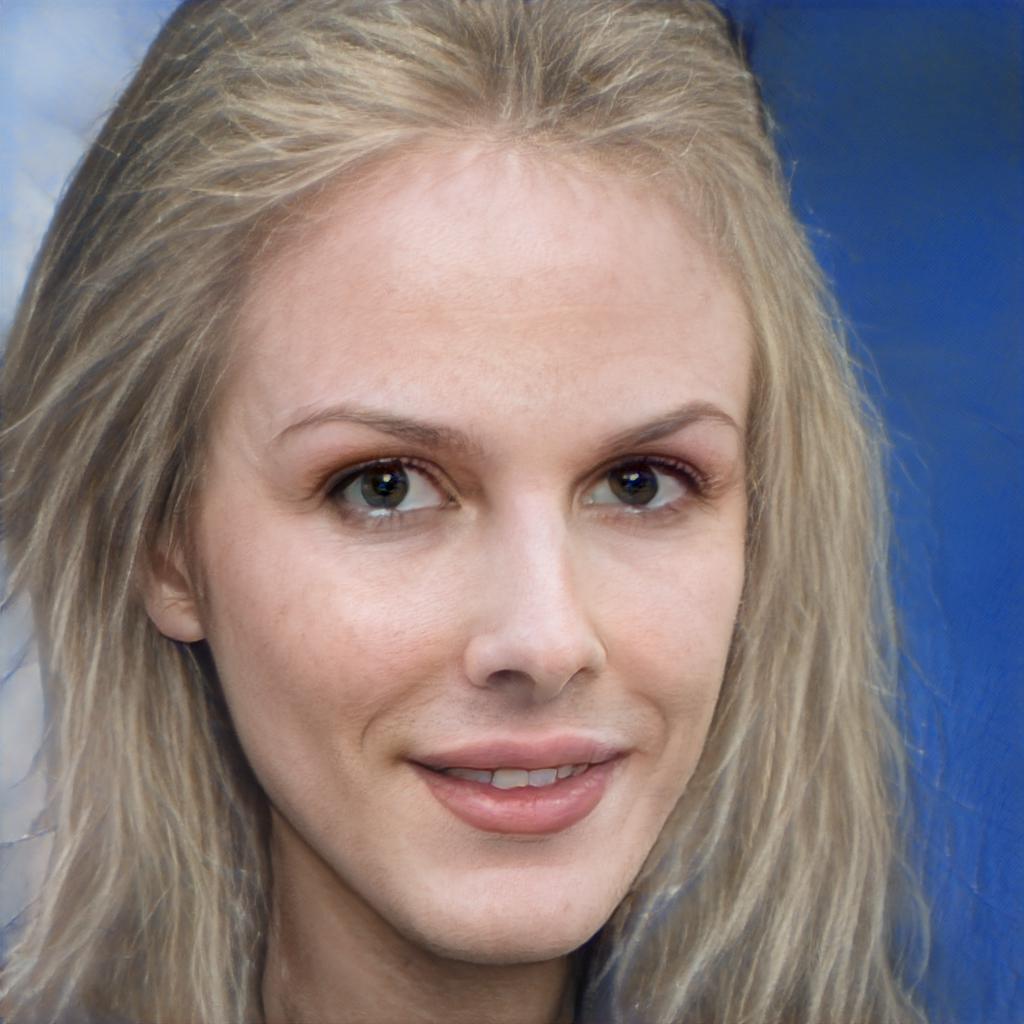} &
 \includegraphics[width=0.13\textwidth, ]{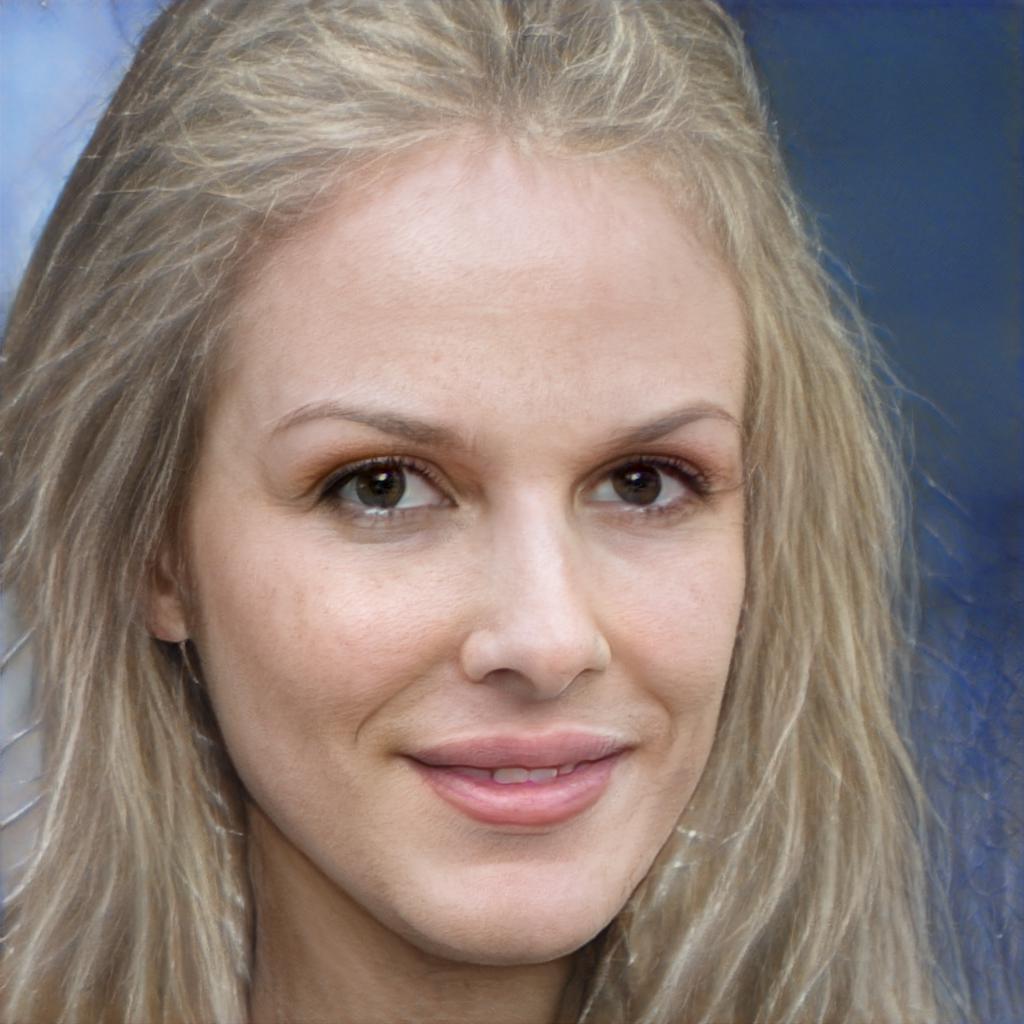} &
 \includegraphics[width=0.13\textwidth, ]{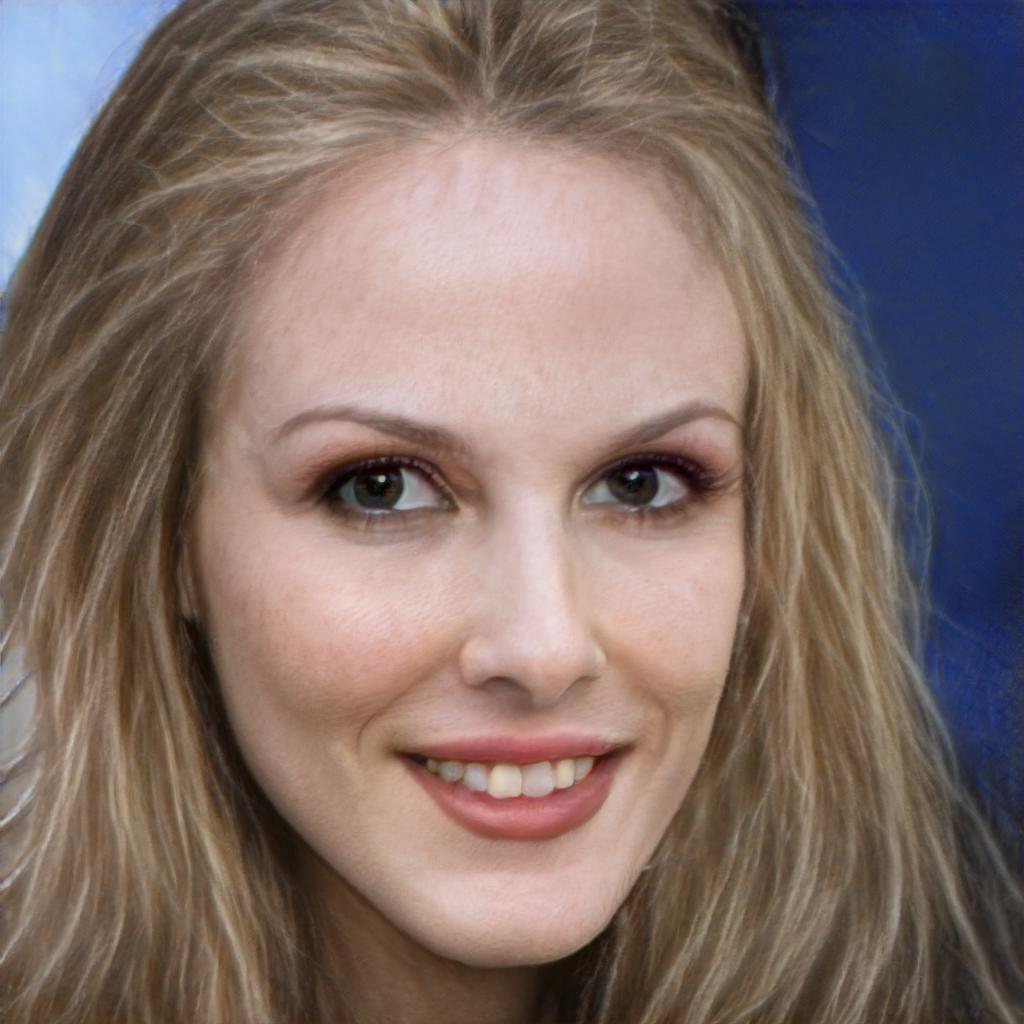} \\
 \begin{turn}{90} \hspace{0.4cm}  $\mathcal{W}^{\star}_{ID}$  \end{turn} &
 \includegraphics[width=0.13\textwidth]{images/original/06011.jpg} & 
 \includegraphics[width=0.13\textwidth, ]{images/inversion/18_orig_img_10.jpg} &
 \includegraphics[width=0.13\textwidth, ]{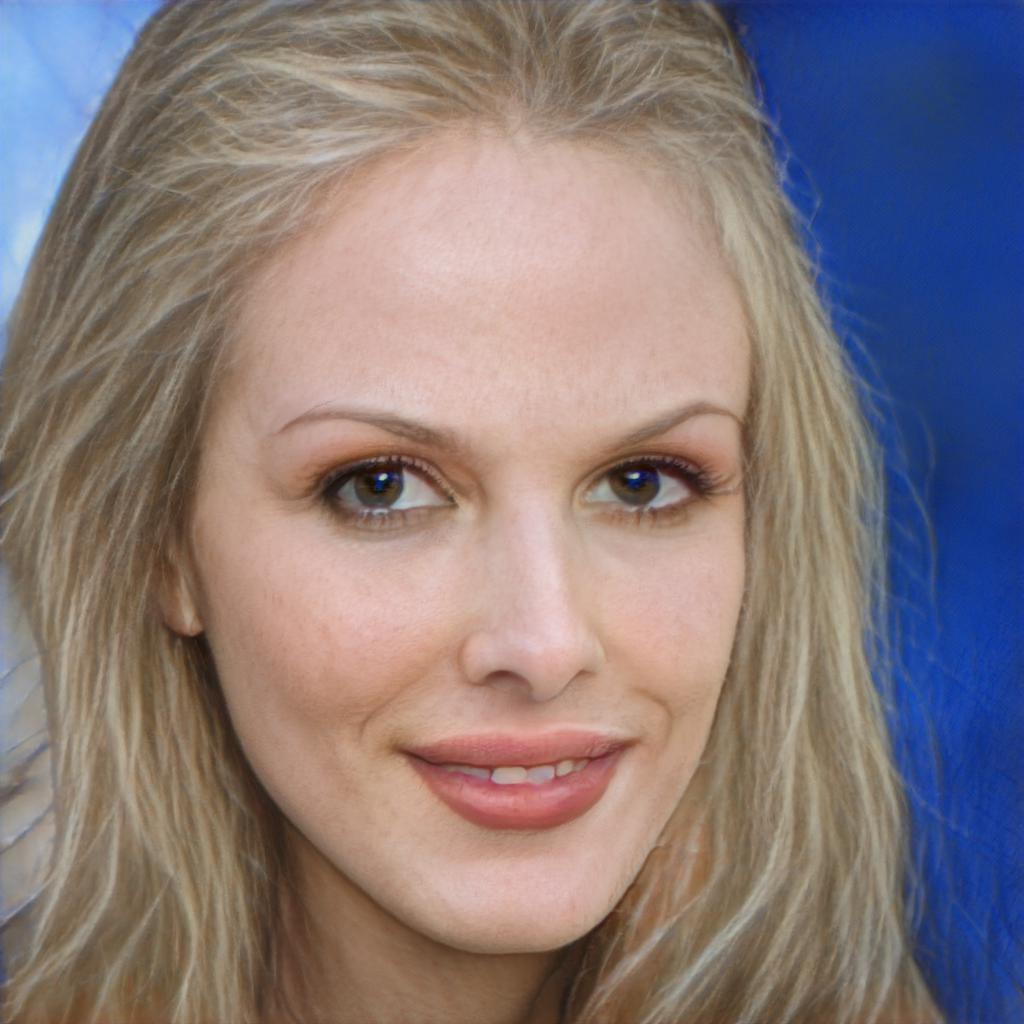} &
 \includegraphics[width=0.13\textwidth, ]{images/wstar_id_no_mag_3_coupl_dist_edit_train_ep5/20_img_10.jpg} &
 \includegraphics[width=0.13\textwidth, ]{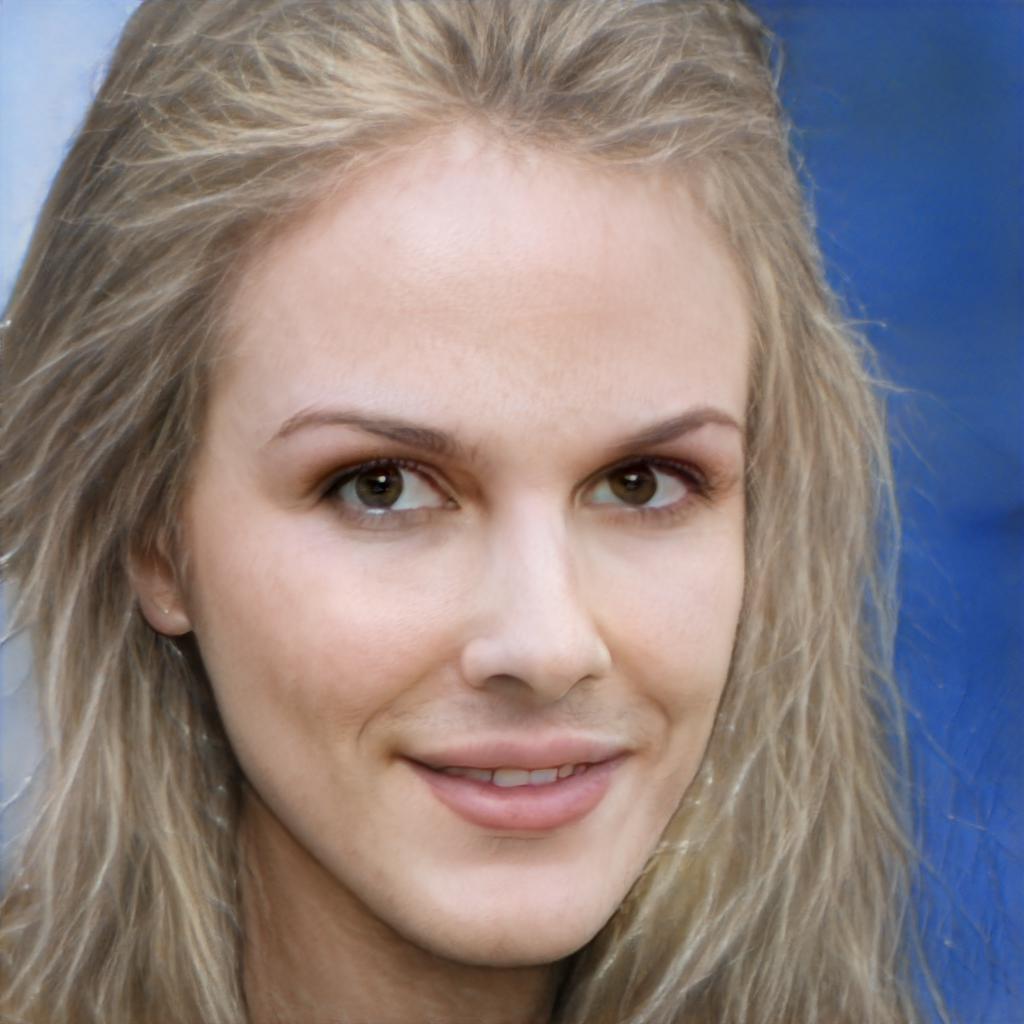} &
 \includegraphics[width=0.13\textwidth, ]{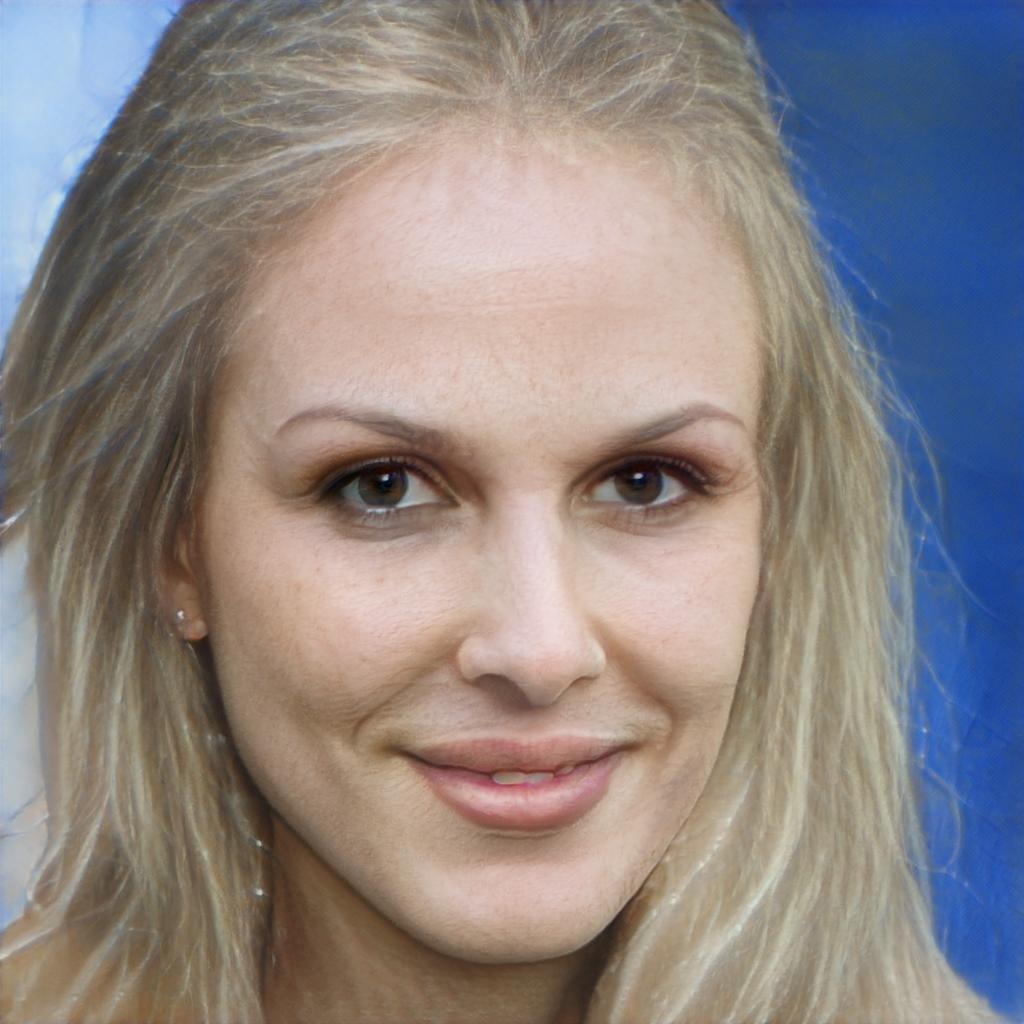} &
 \includegraphics[width=0.13\textwidth, ]{images/wstar_id_no_mag_3_coupl_dist_edit_train_ep5/36_img_10.jpg} \\
 \\
\end{tabular}
\caption{Ablation study: Image editing using InterFaceGAN. }
\label{fig:edit_abl_3}
\end{figure}

In this section, we will investigate the effect of different losses used for image editing. The same setup is used as in Section 4.3.2 except that we fix the editing step to $10$ in \wstar (except we put it $6$ for \wstarasp). We compare the following choices:
\begin{itemize}
    \item \wstara, where $T$ is trained only with the attribute separation loss eq. (2),
    \item \wstara-ID: \wstara with the identity preservation loss,
    \item \wstara-ID*: \wstara with lower weight for the identity loss $\lambda_{ID}=0.1$,
    \item \wstara-ID$^{\dagger}$: \wstara-ID with random classifiers. 
\end{itemize}  
From Figure \ref{fig:edit_abl_1} , \ref{fig:edit_abl_2}, \ref{fig:edit_abl_3}, we can conclude that:
\begin{enumerate}
    \item The identity loss helps to preserve the identity after editing (Row 2 vs 3, 4 and 5). In addition, there is a trade-off between preserving the identity and the amount of editing effect. Specifically, increasing $\lambda_{ID}$ lead to lower editing effect (Row 3 vs 5). 
    \item The latent distance unfolding loss has negligible effect on editing (Row 3 vs 6).
    \item Pretraining the classifiers does not seem to improve the editing results.
    
\end{enumerate}

\begin{figure}[h]
\setlength\tabcolsep{2pt}%
\centering
\begin{tabular}{p{0.25cm}ccccccc}
\centering
&
 \textbf{Original} &
 \textbf{Inverted} &
 \textbf{Makeup} &
 \textbf{Male} &
 \textbf{Mustache} &
 \textbf{Chubby} &
 \textbf{Lipstick} \\
 \begin{turn}{90} \hspace{0.5cm} \wplus\end{turn} &
 \includegraphics[width=0.13\textwidth]{images/original/06004.jpg} & 
 \includegraphics[width=0.13\textwidth, ]{images/inversion/18_orig_img_3.jpg} &
 \includegraphics[width=0.13\textwidth, ]{images/wplus/18_img_5.jpg} &
 \includegraphics[width=0.13\textwidth, ]{images/wplus/20_img_5.jpg} &
 \includegraphics[width=0.13\textwidth, ]{images/wplus/22_img_5.jpg} &
 \includegraphics[width=0.13\textwidth, ]{images/wplus/13_img_5.jpg} &
 \includegraphics[width=0.13\textwidth]{images/wplus/36_img_5.jpg} \\
 \begin{turn}{90} \hspace{0.5cm} \wstara\end{turn} &
 \includegraphics[width=0.13\textwidth]{images/original/06004.jpg} & 
 \includegraphics[width=0.13\textwidth, ]{images/inversion/18_orig_img_3.jpg} &
 \includegraphics[width=0.13\textwidth, ]{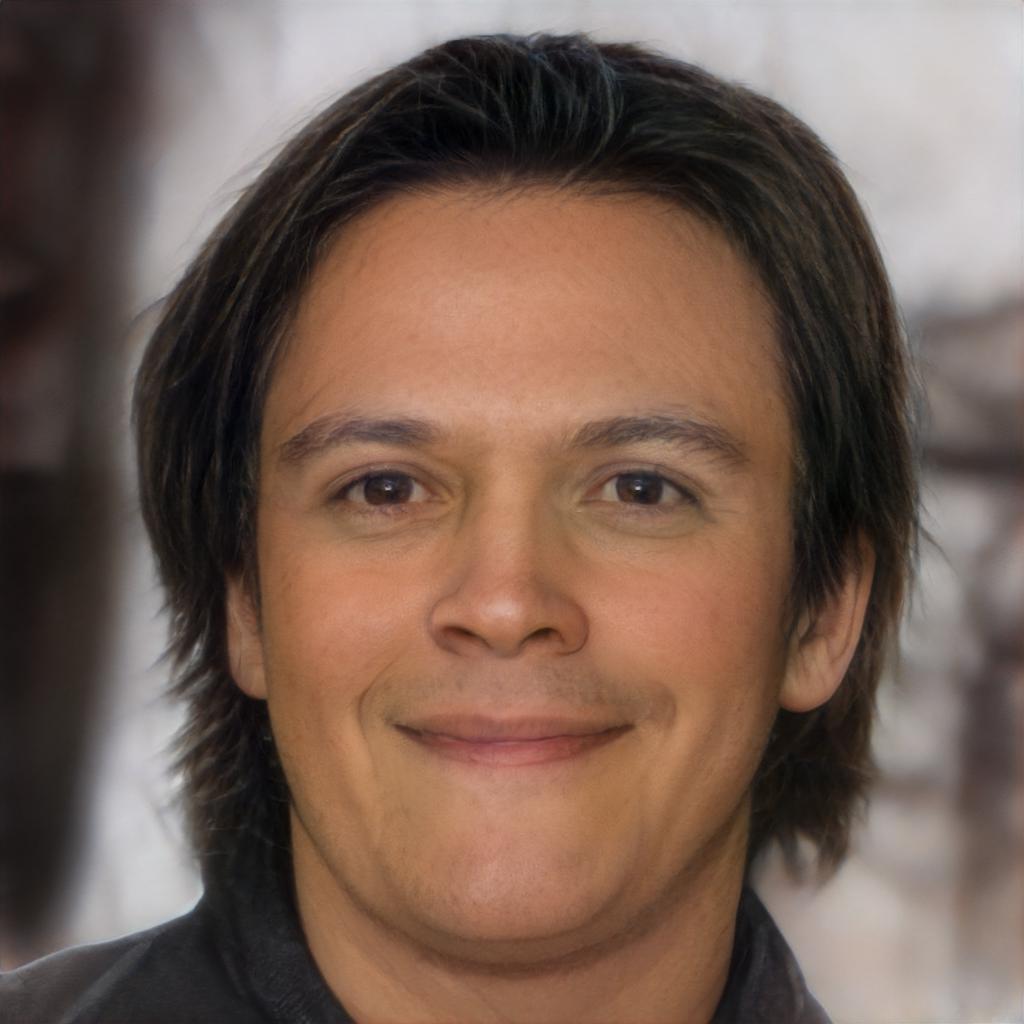} &
 \includegraphics[width=0.13\textwidth, ]{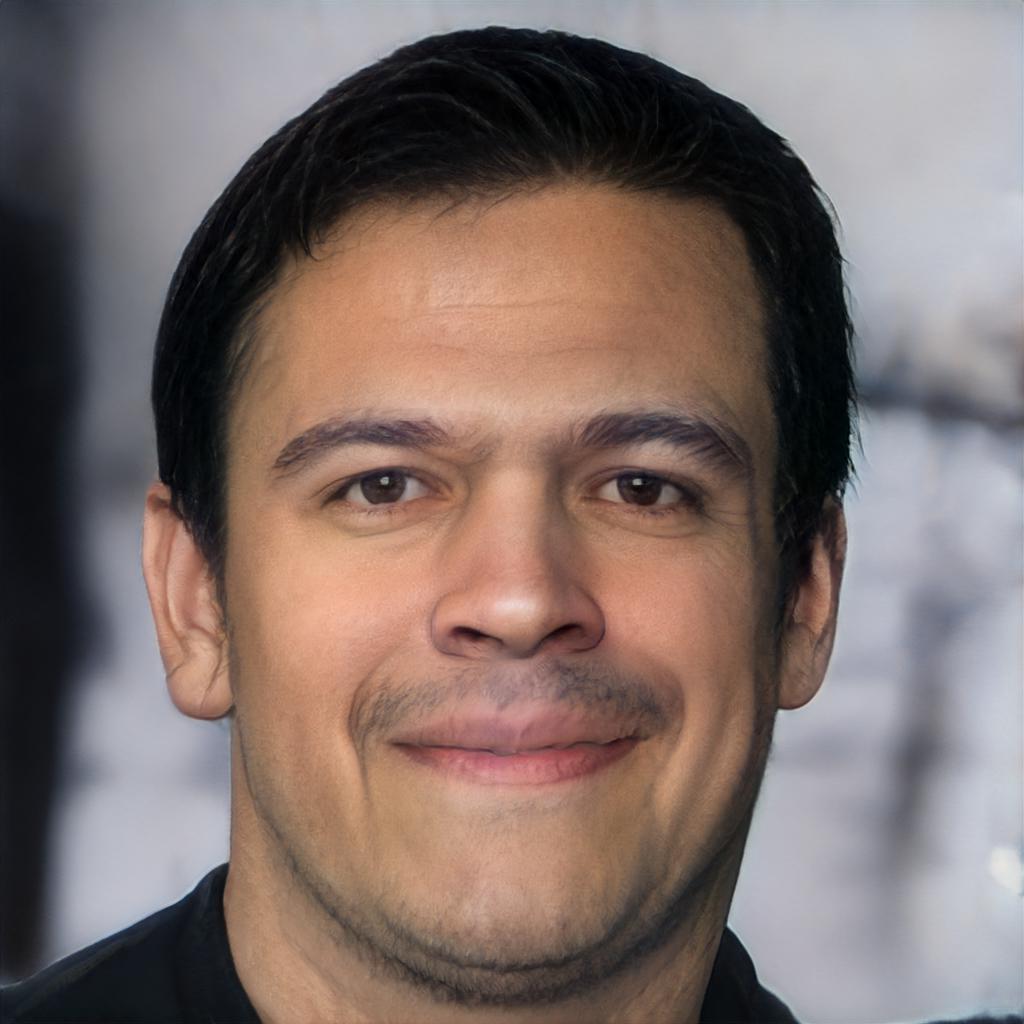} &
 \includegraphics[width=0.13\textwidth, ]{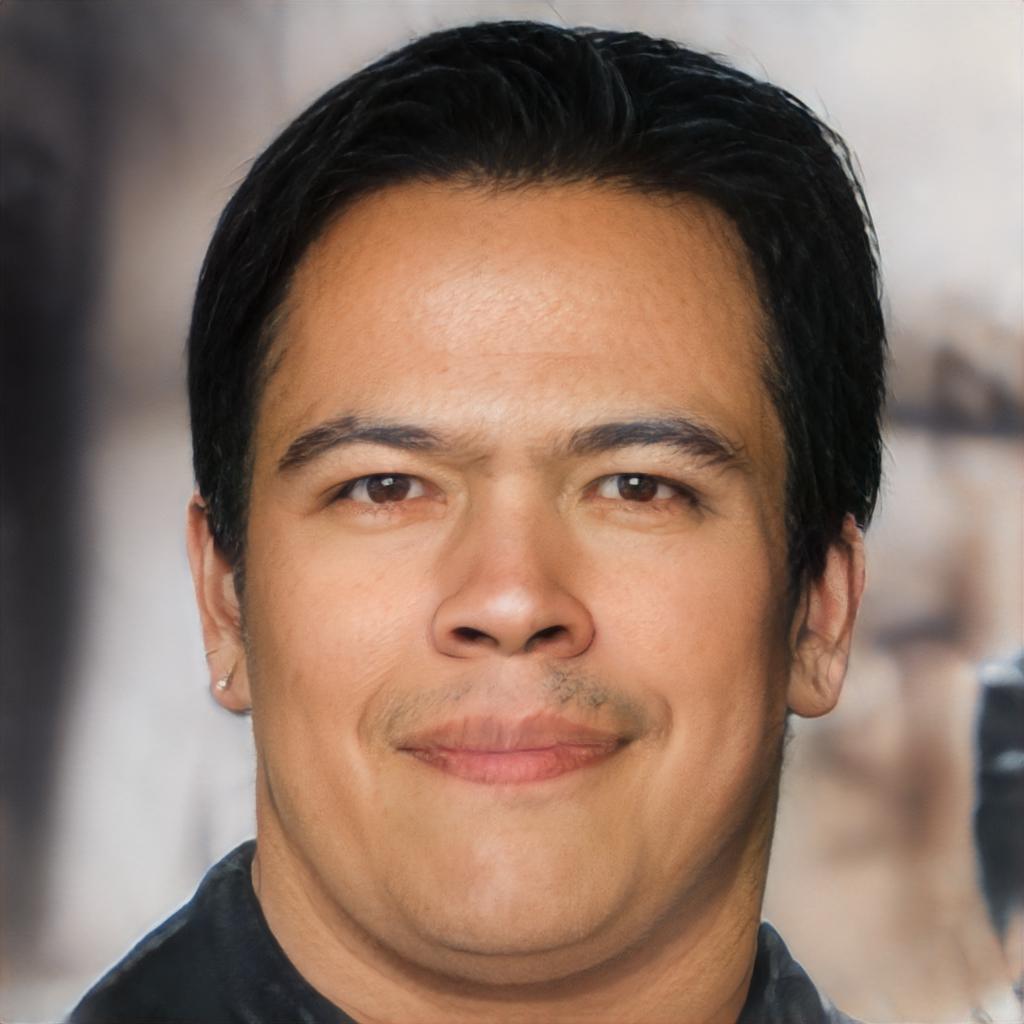} &
 \includegraphics[width=0.13\textwidth, ]{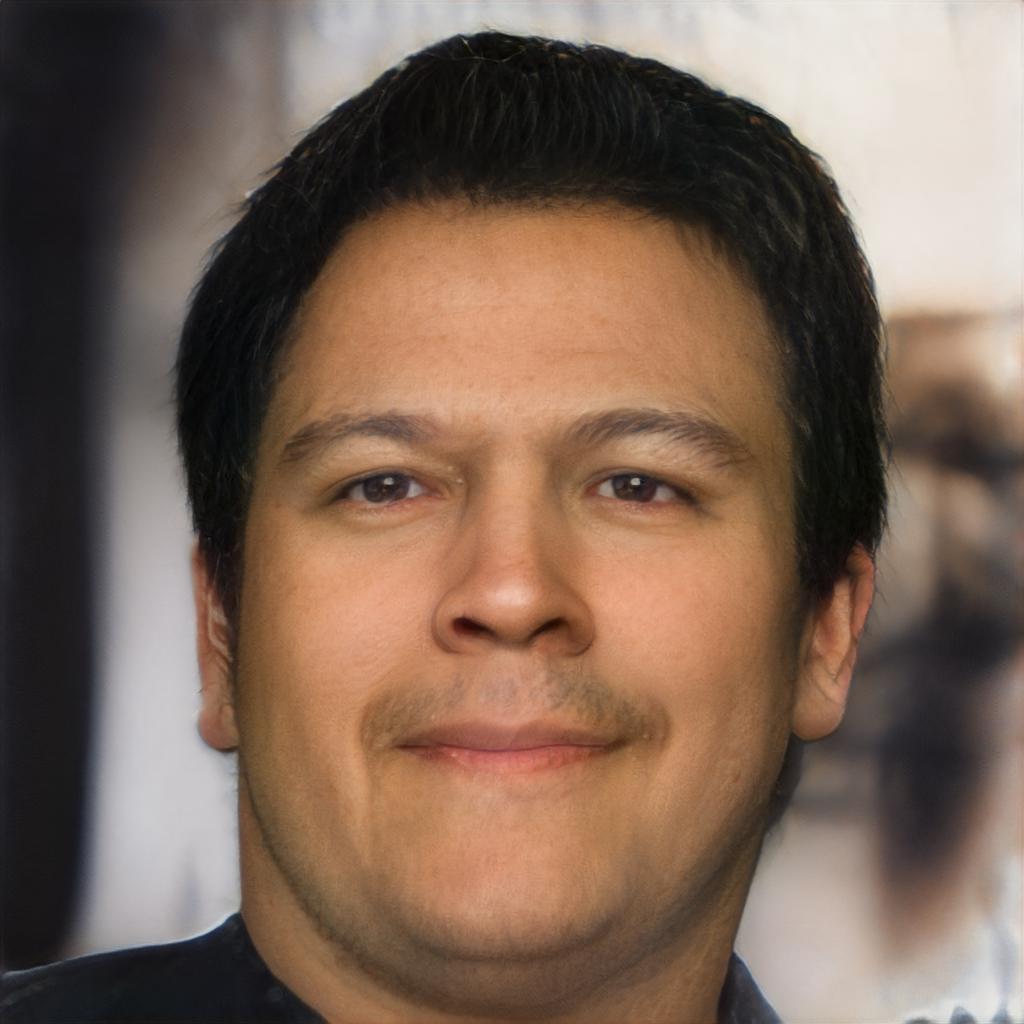} &
 \includegraphics[width=0.13\textwidth, ]{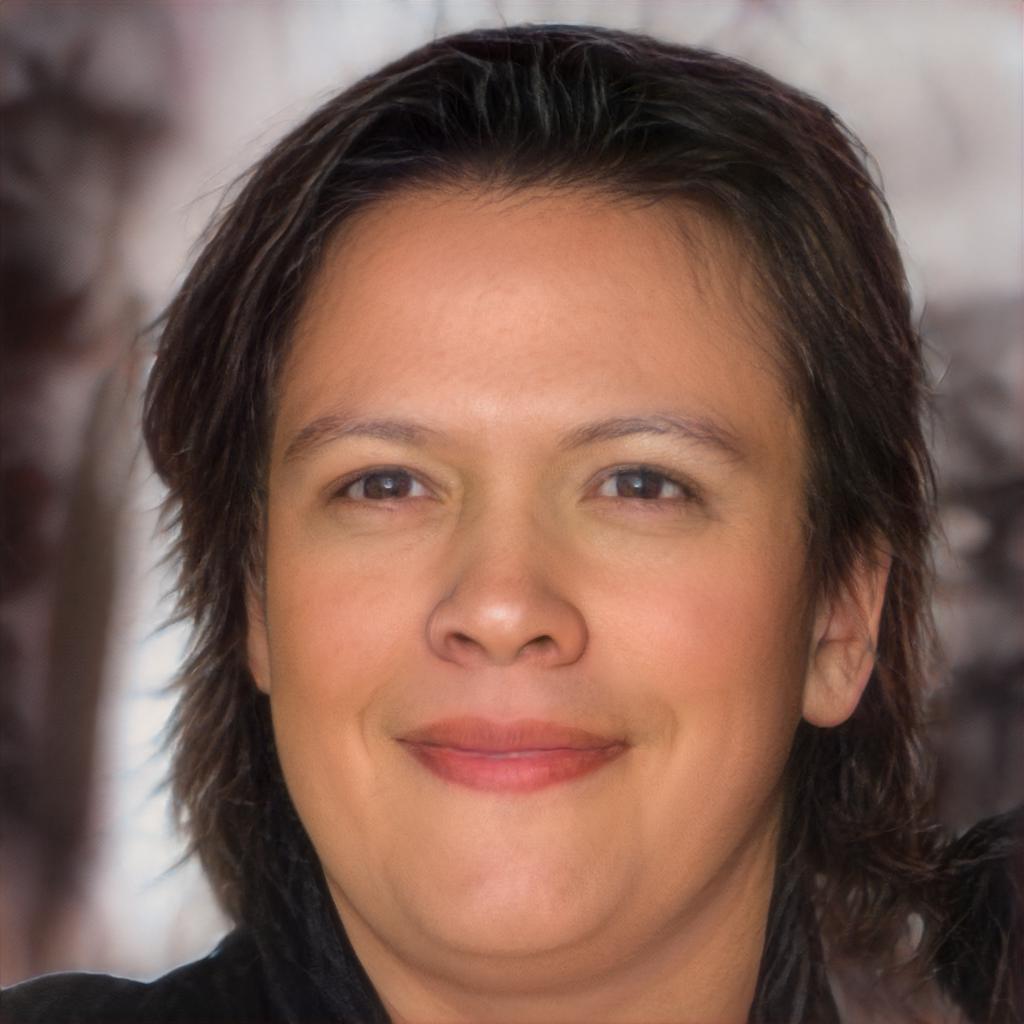} \\
 \begin{turn}{90} \hspace{0.2cm} \wstara-ID\end{turn} &
 \includegraphics[width=0.13\textwidth]{images/original/06004.jpg} & 
 \includegraphics[width=0.13\textwidth, ]{images/inversion/18_orig_img_3.jpg} &
 \includegraphics[width=0.13\textwidth, ]{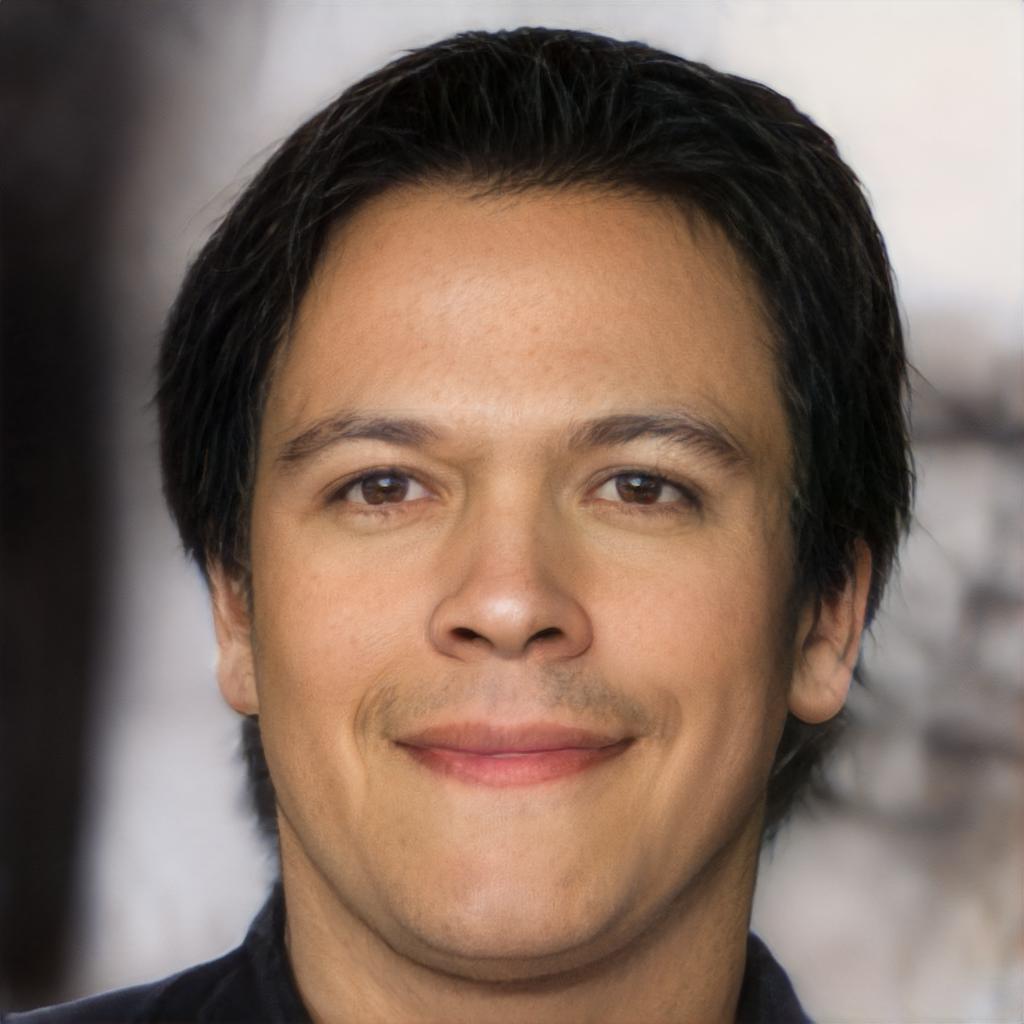} &
 \includegraphics[width=0.13\textwidth, ]{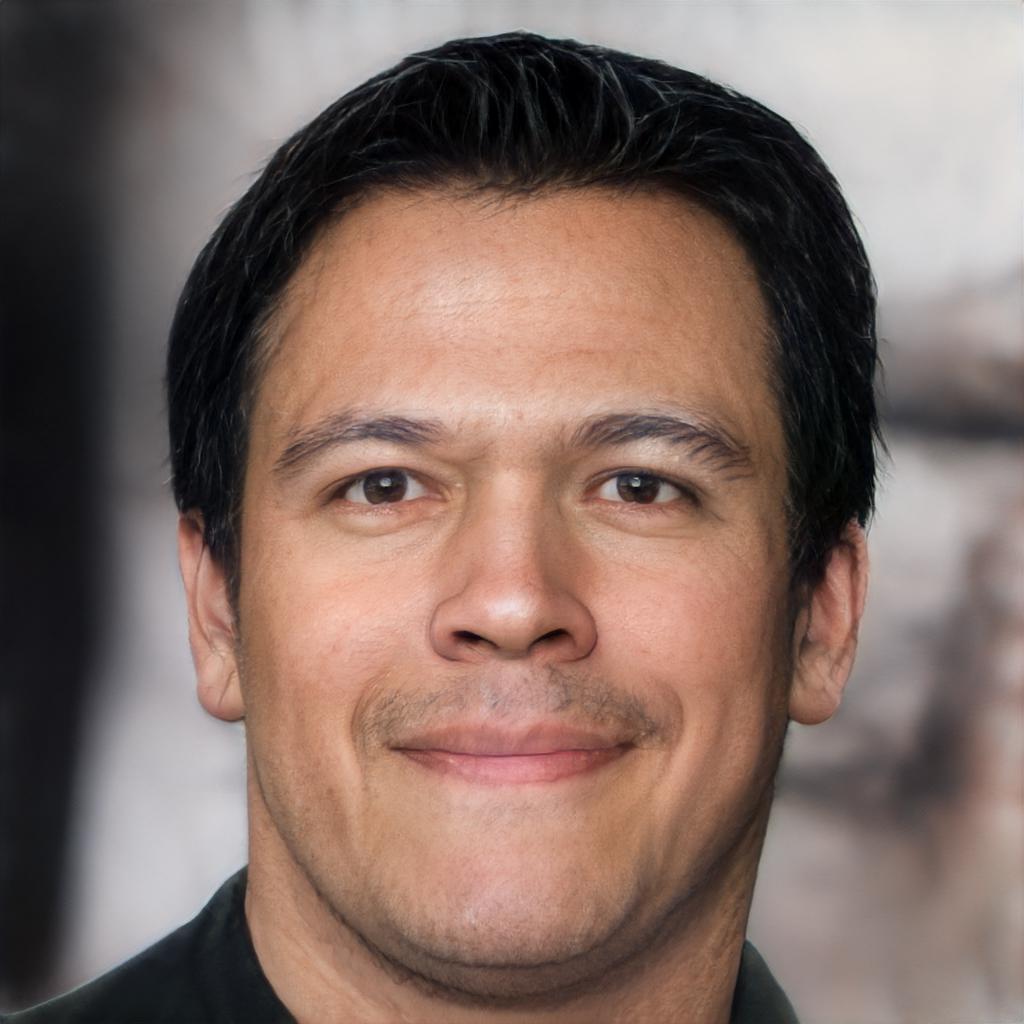} &
 \includegraphics[width=0.13\textwidth, ]{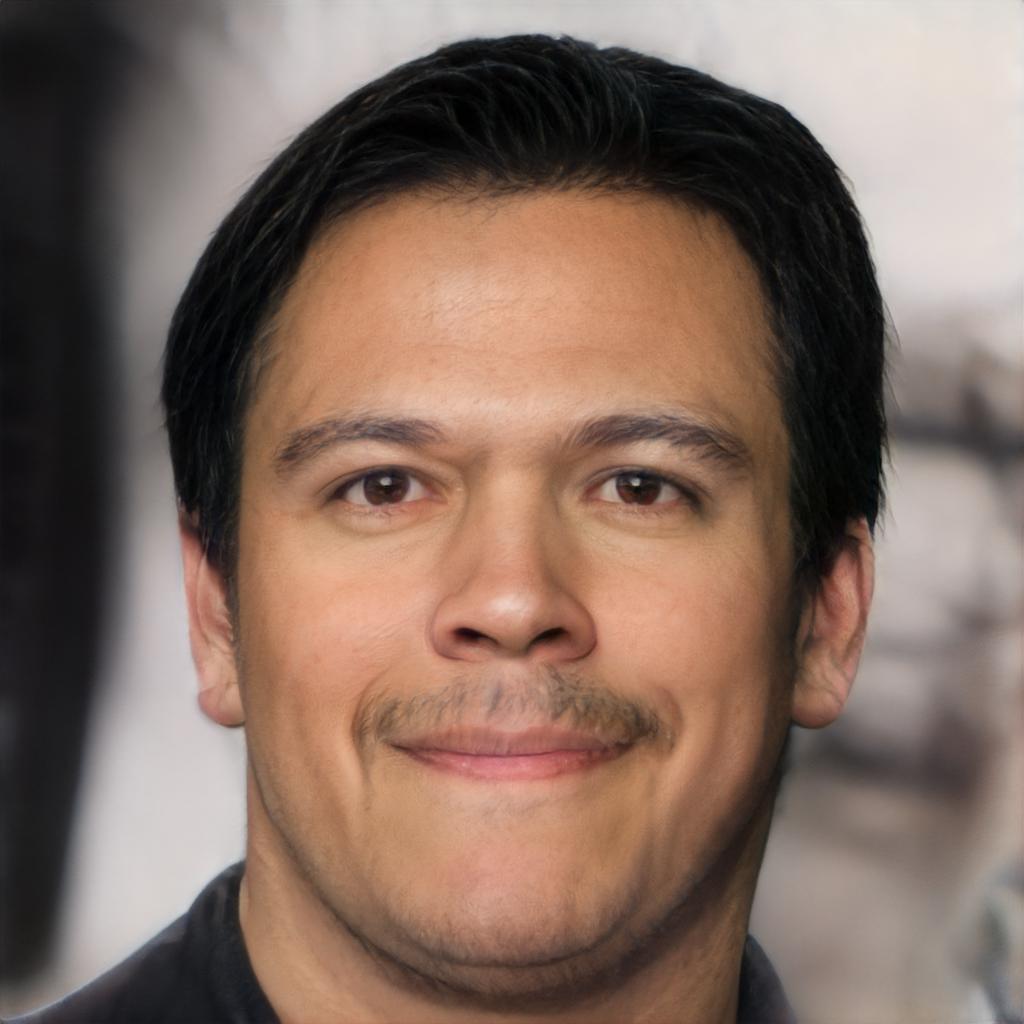} &
 \includegraphics[width=0.13\textwidth, ]{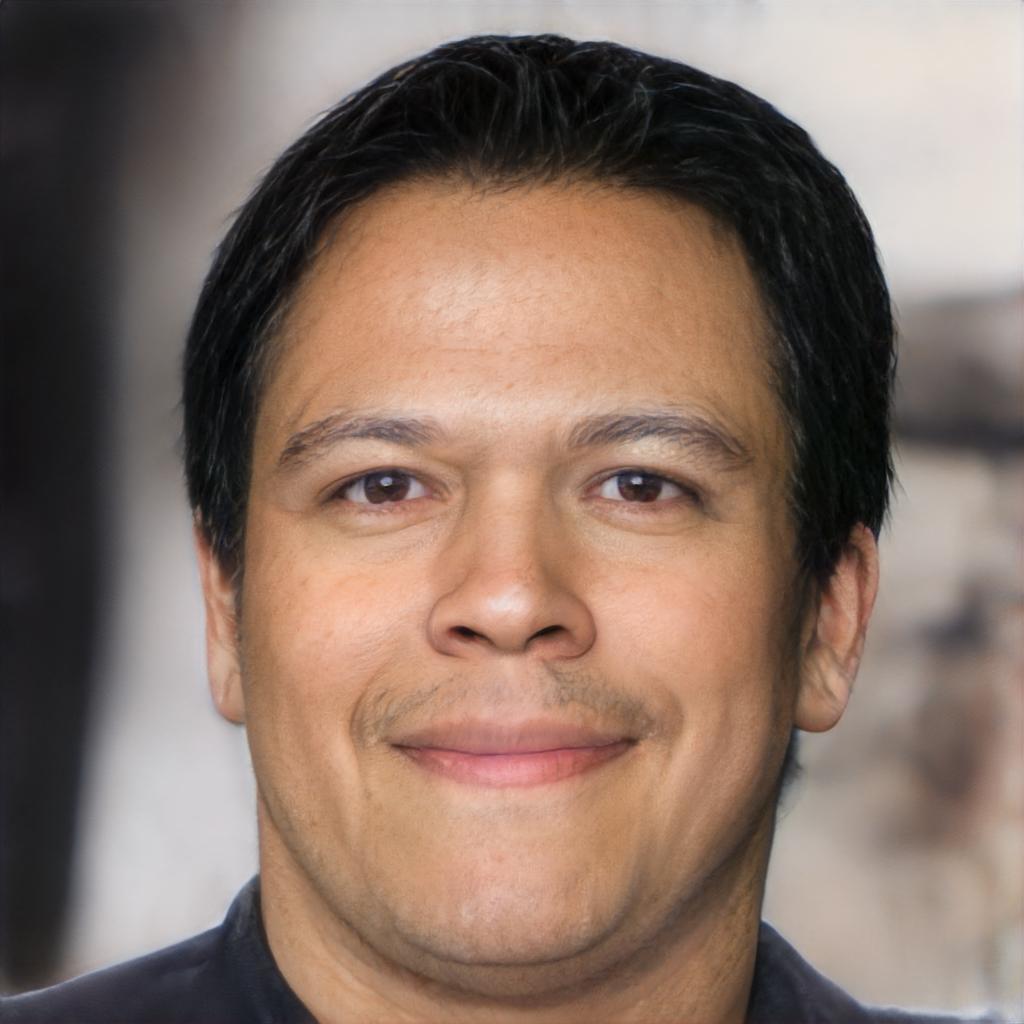} &
 \includegraphics[width=0.13\textwidth, ]{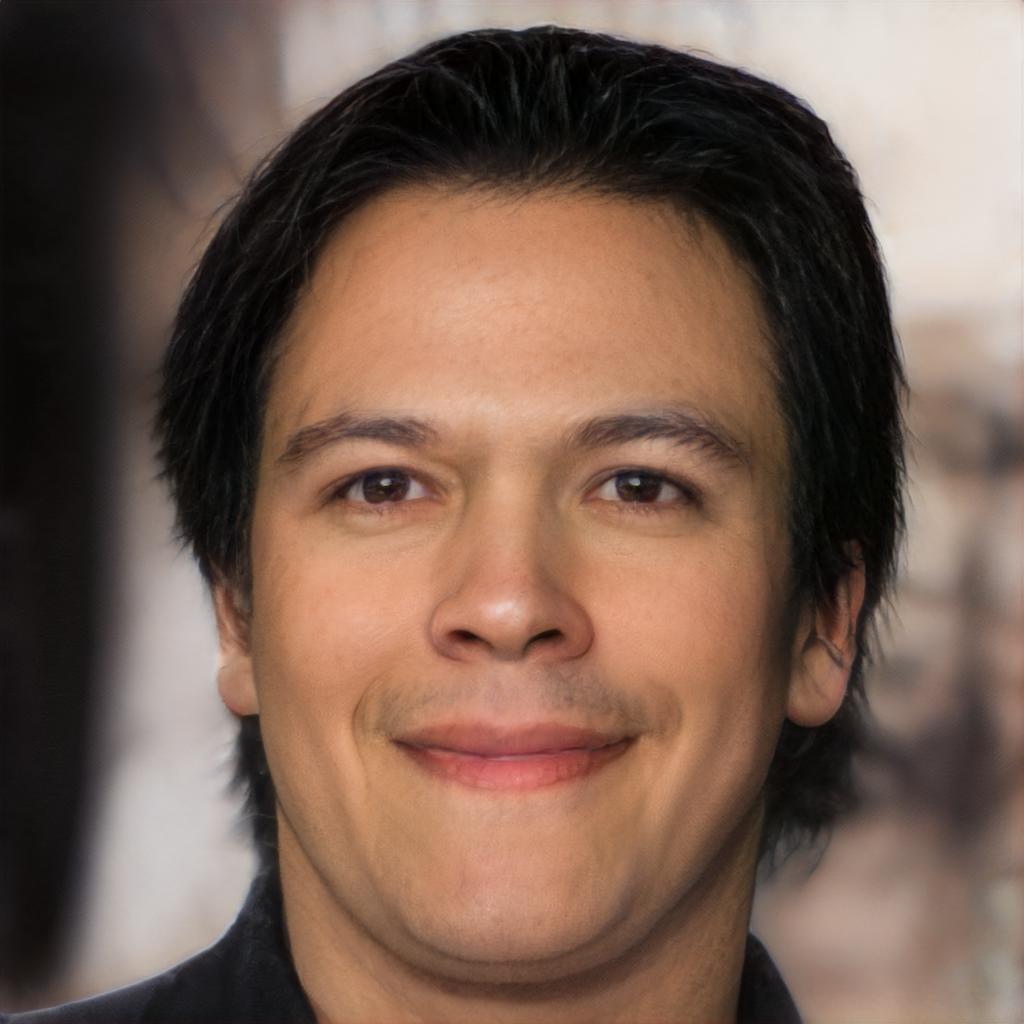} \\
 \begin{turn}{90} \wstara$-ID^{\dagger}$\end{turn} &
 \includegraphics[width=0.13\textwidth]{images/original/06004.jpg} & 
 \includegraphics[width=0.13\textwidth, ]{images/inversion/18_orig_img_3.jpg} &
 \includegraphics[width=0.13\textwidth, ]{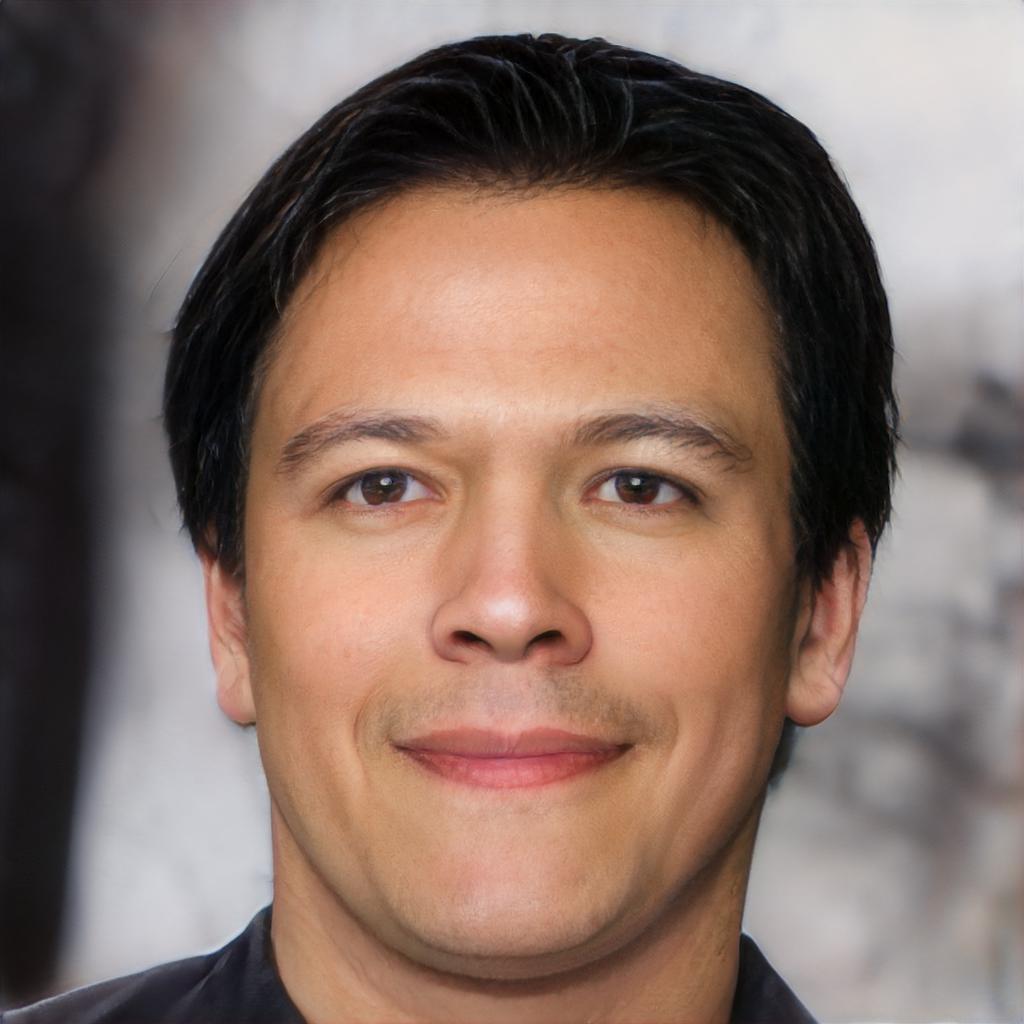} &
 \includegraphics[width=0.13\textwidth, ]{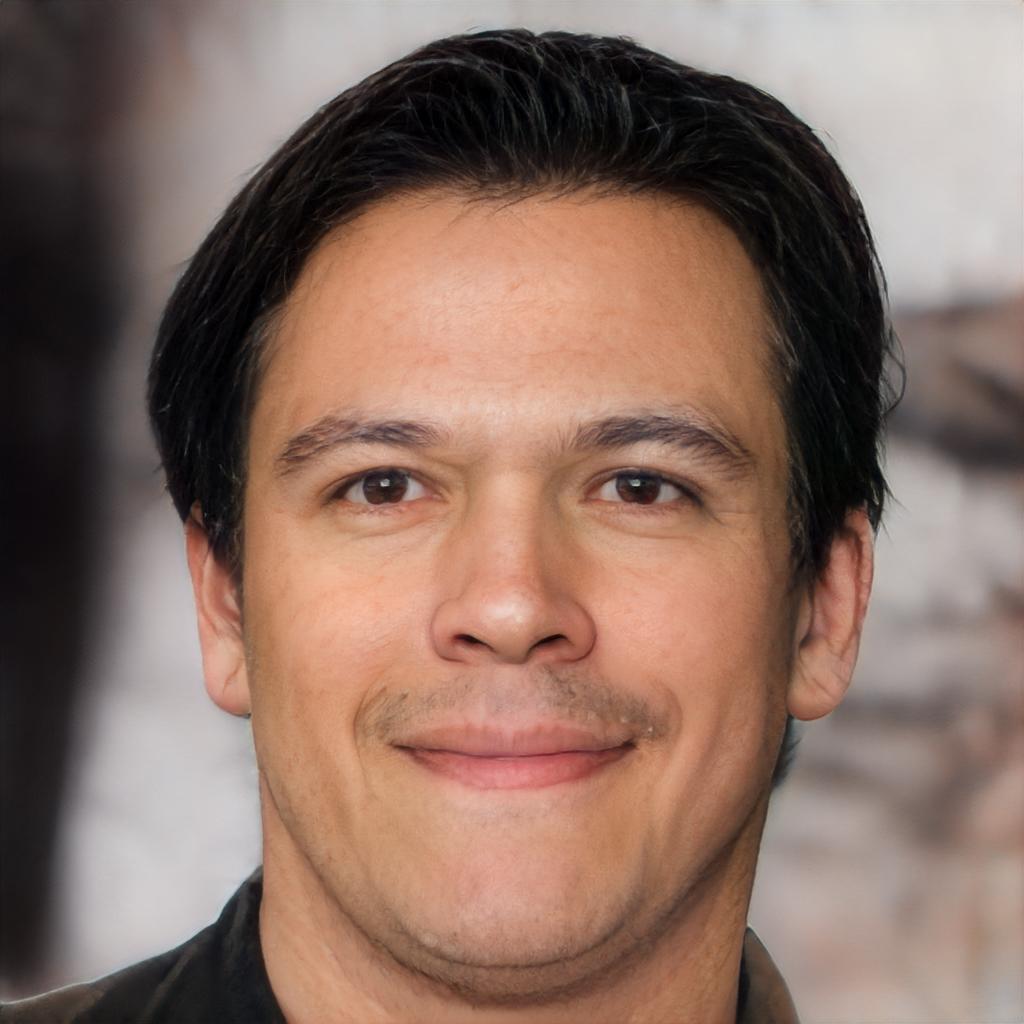} &
 \includegraphics[width=0.13\textwidth, ]{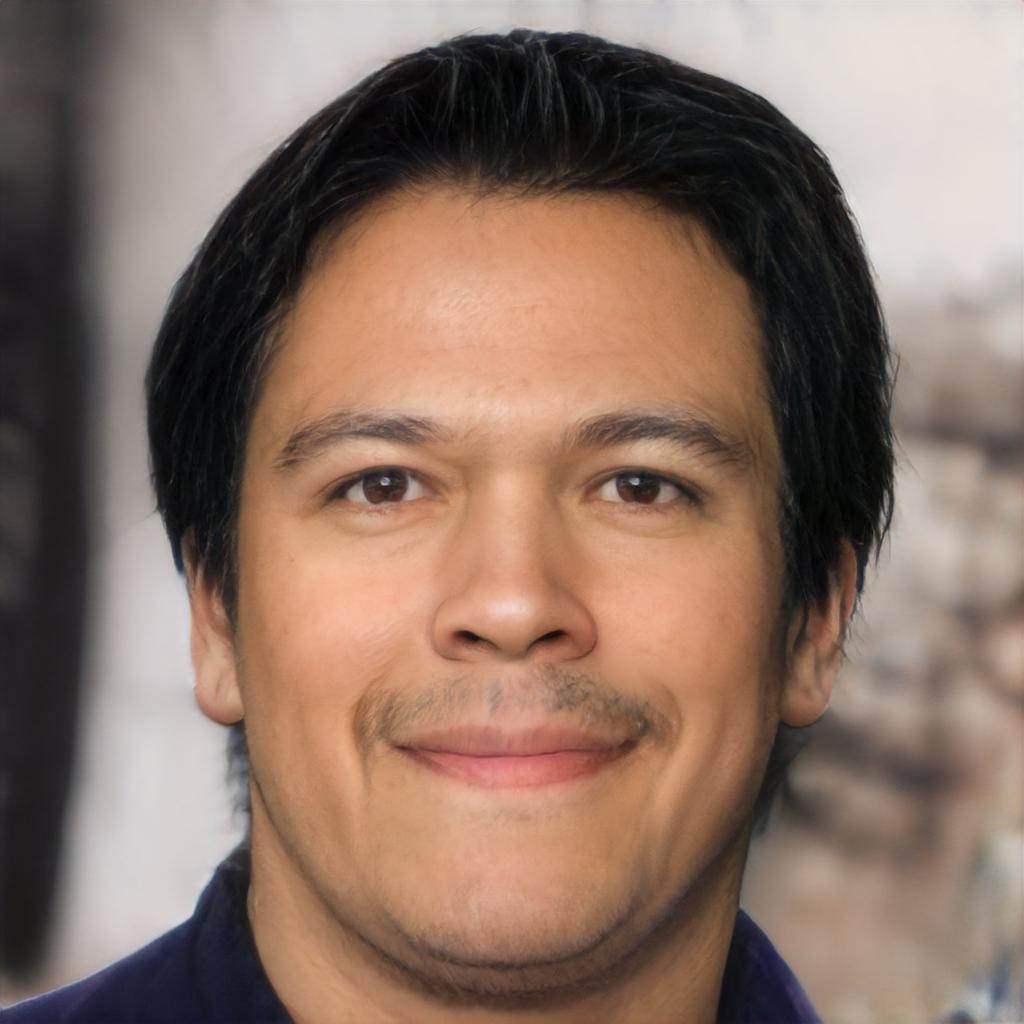} &
 \includegraphics[width=0.13\textwidth, ]{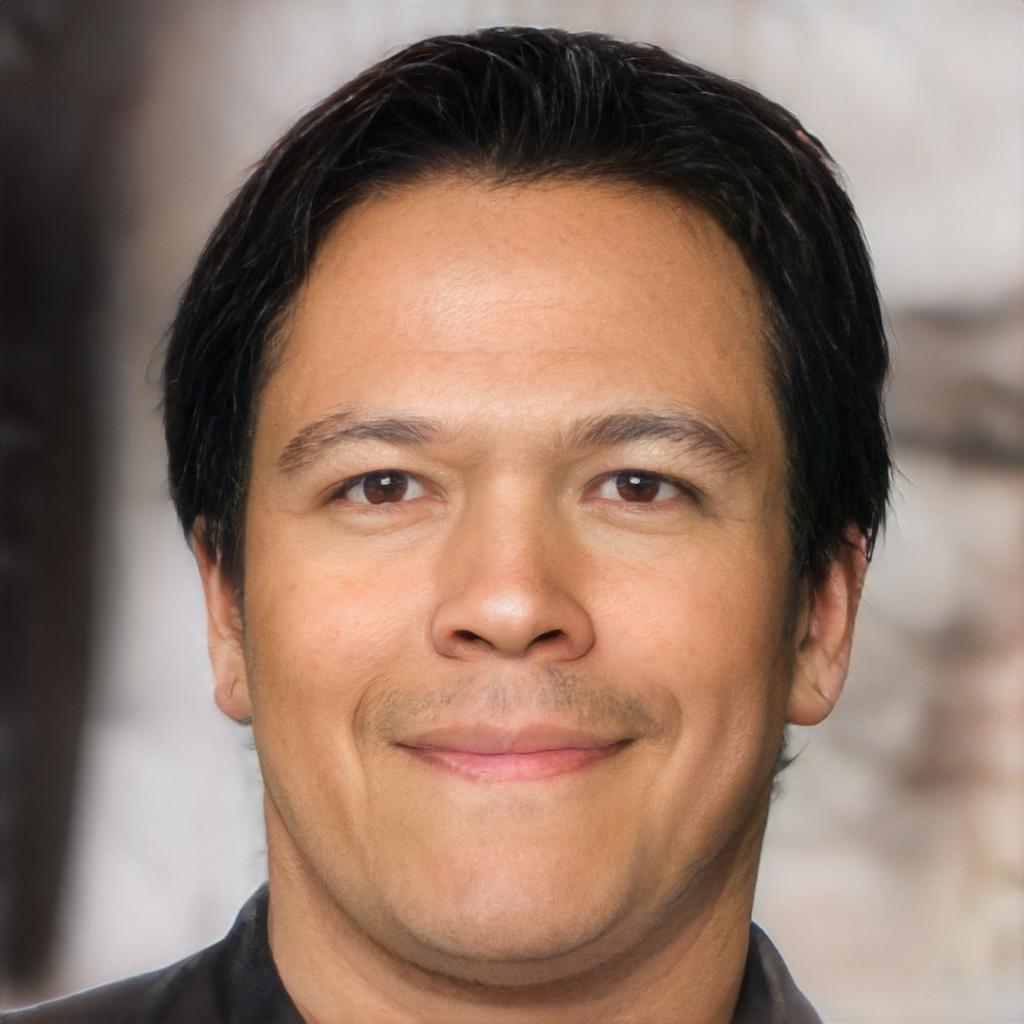} &
 \includegraphics[width=0.13\textwidth, ]{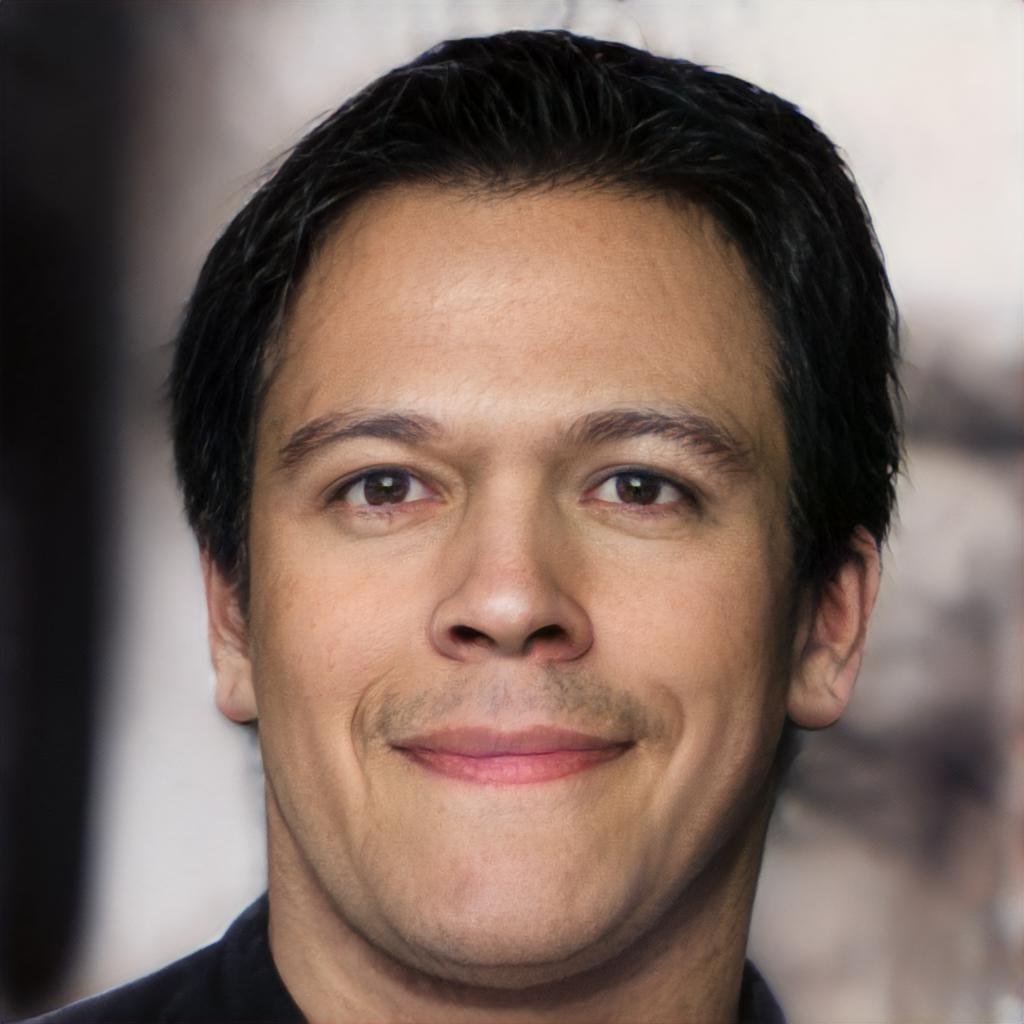} \\
 \begin{turn}{90} \wstara$-ID^{\star}$\end{turn} &
 \includegraphics[width=0.13\textwidth]{images/original/06004.jpg} & 
 \includegraphics[width=0.13\textwidth, ]{images/inversion/18_orig_img_3.jpg} &
 \includegraphics[width=0.13\textwidth, ]{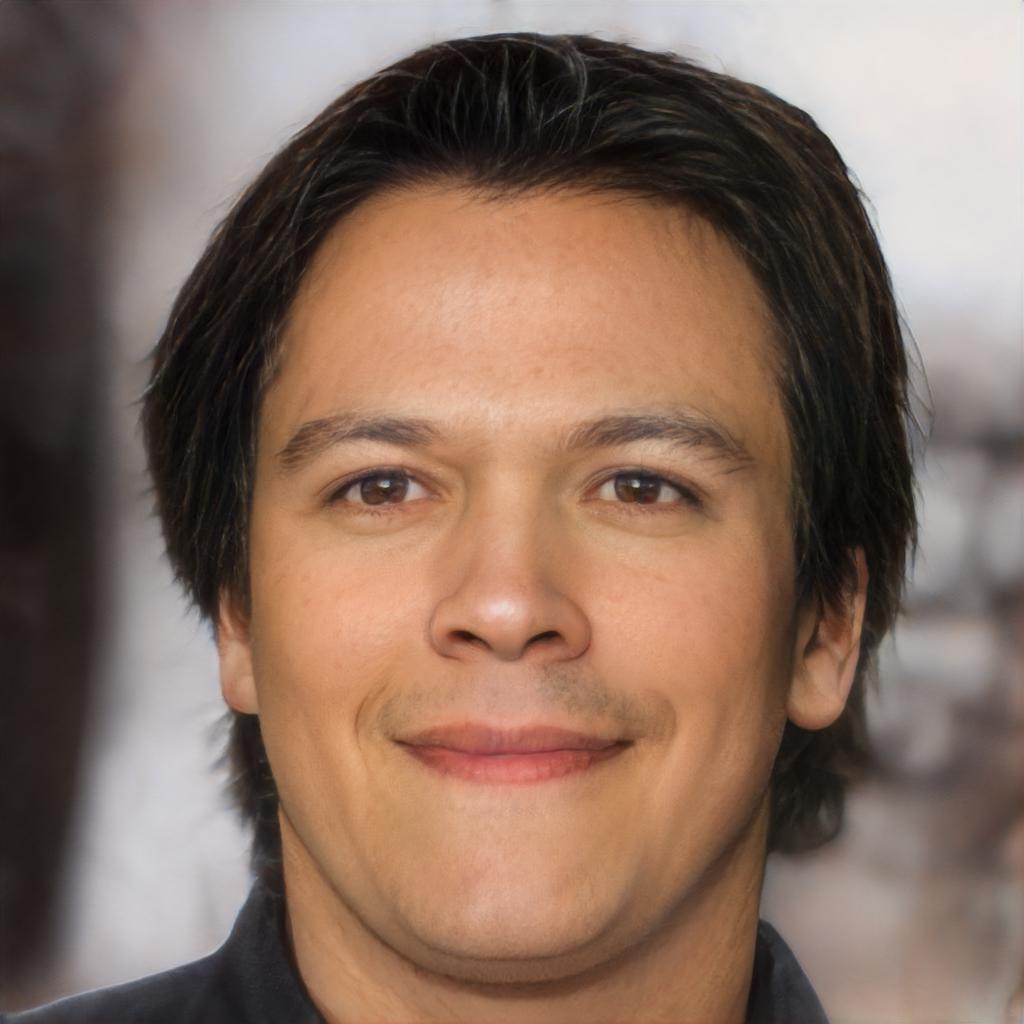} &
 \includegraphics[width=0.13\textwidth, ]{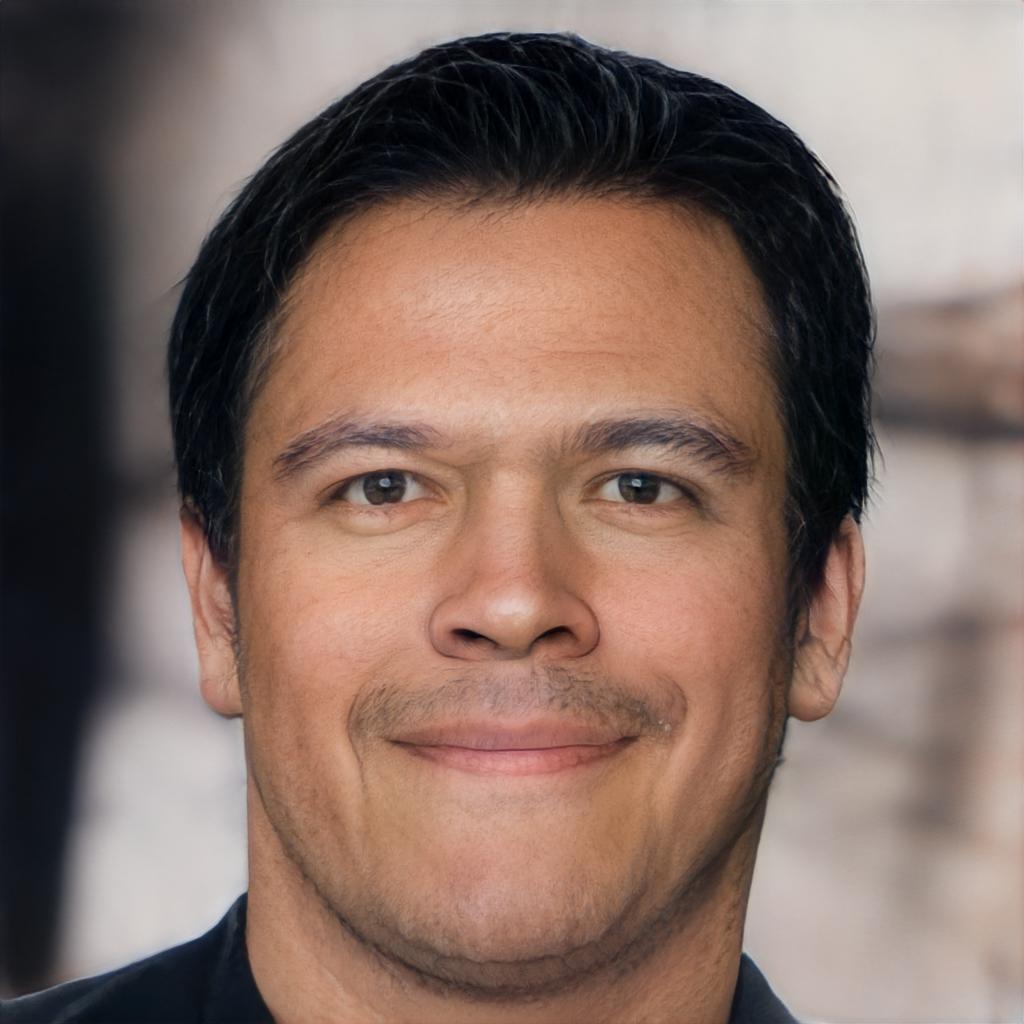} &
 \includegraphics[width=0.13\textwidth, ]{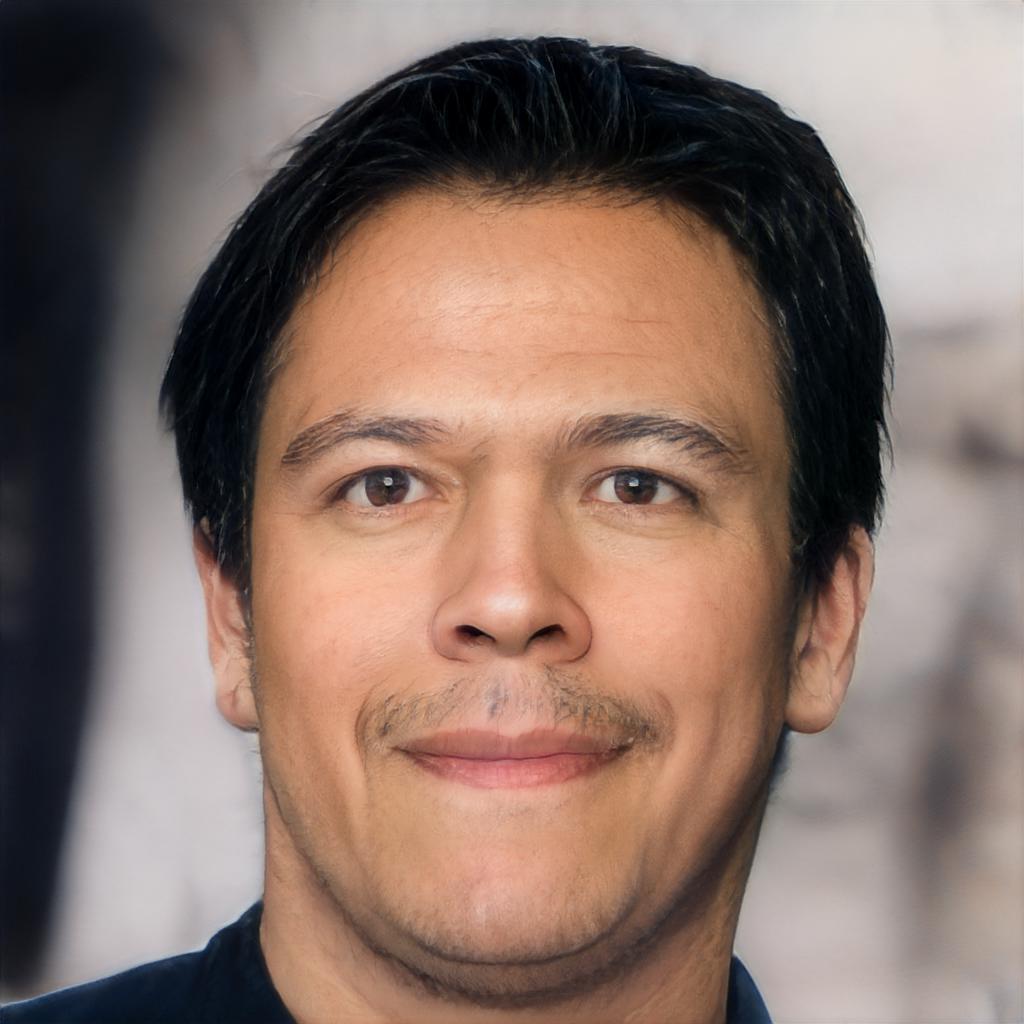} &
 \includegraphics[width=0.13\textwidth, ]{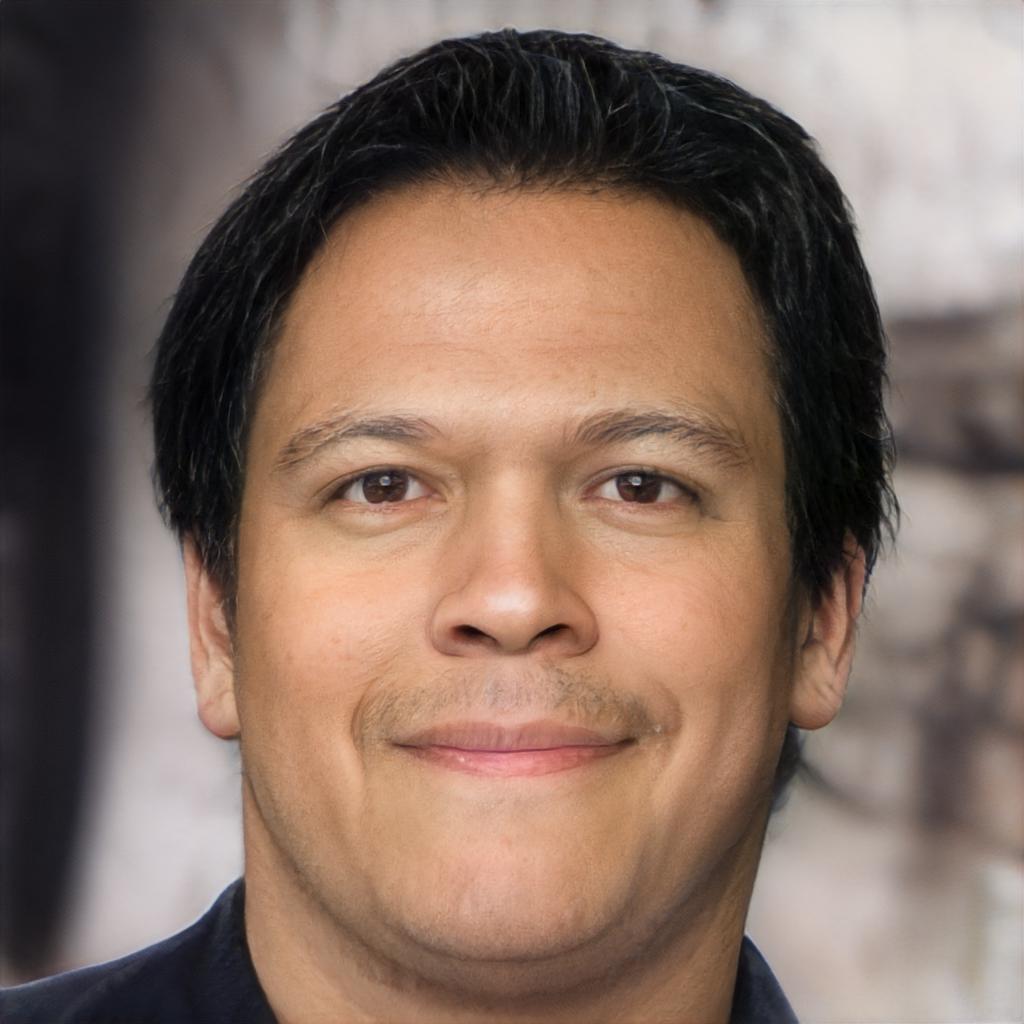} &
 \includegraphics[width=0.13\textwidth, ]{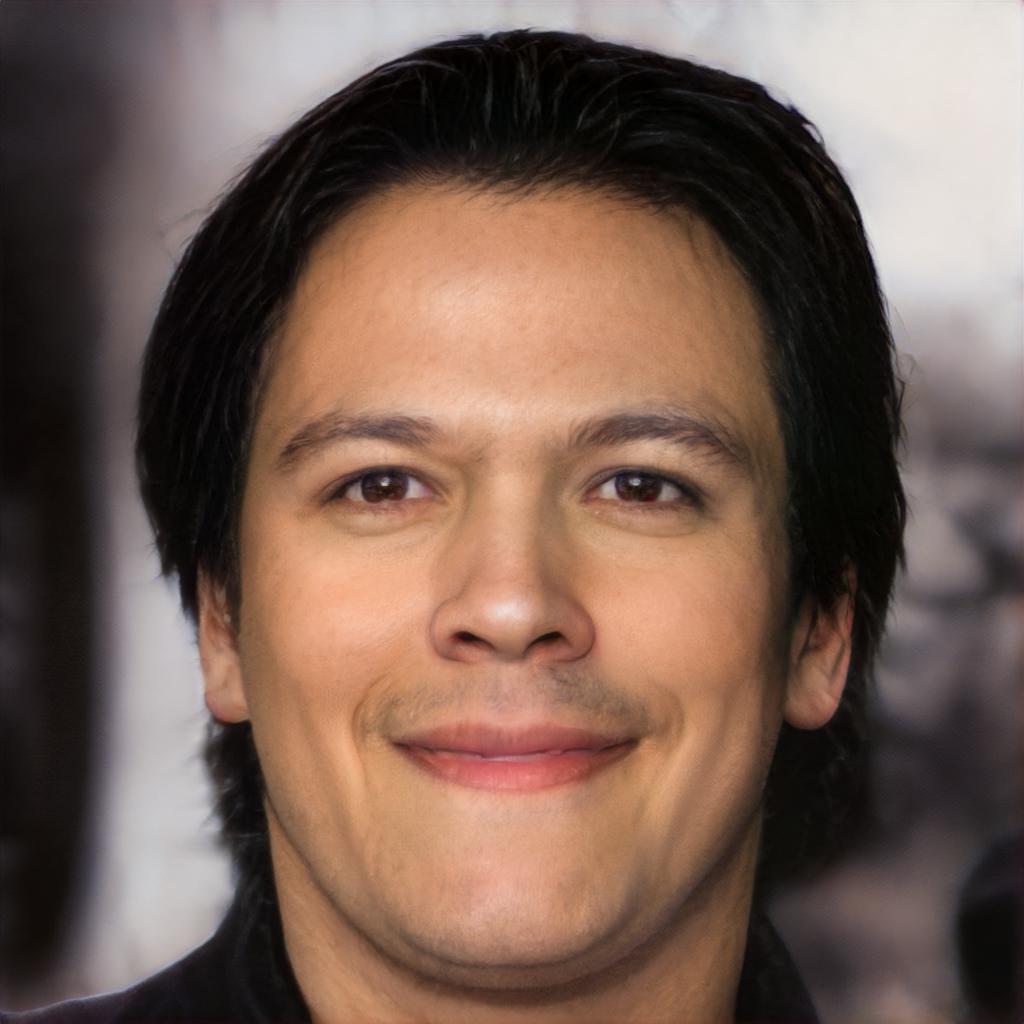} \\
 \begin{turn}{90} \hspace{0.4cm} $\mathcal{W}^{\star}_{ID}$ \end{turn} &
 \includegraphics[width=0.13\textwidth]{images/original/06004.jpg} & 
 \includegraphics[width=0.13\textwidth, ]{images/inversion/18_orig_img_3.jpg} &
 \includegraphics[width=0.13\textwidth, ]{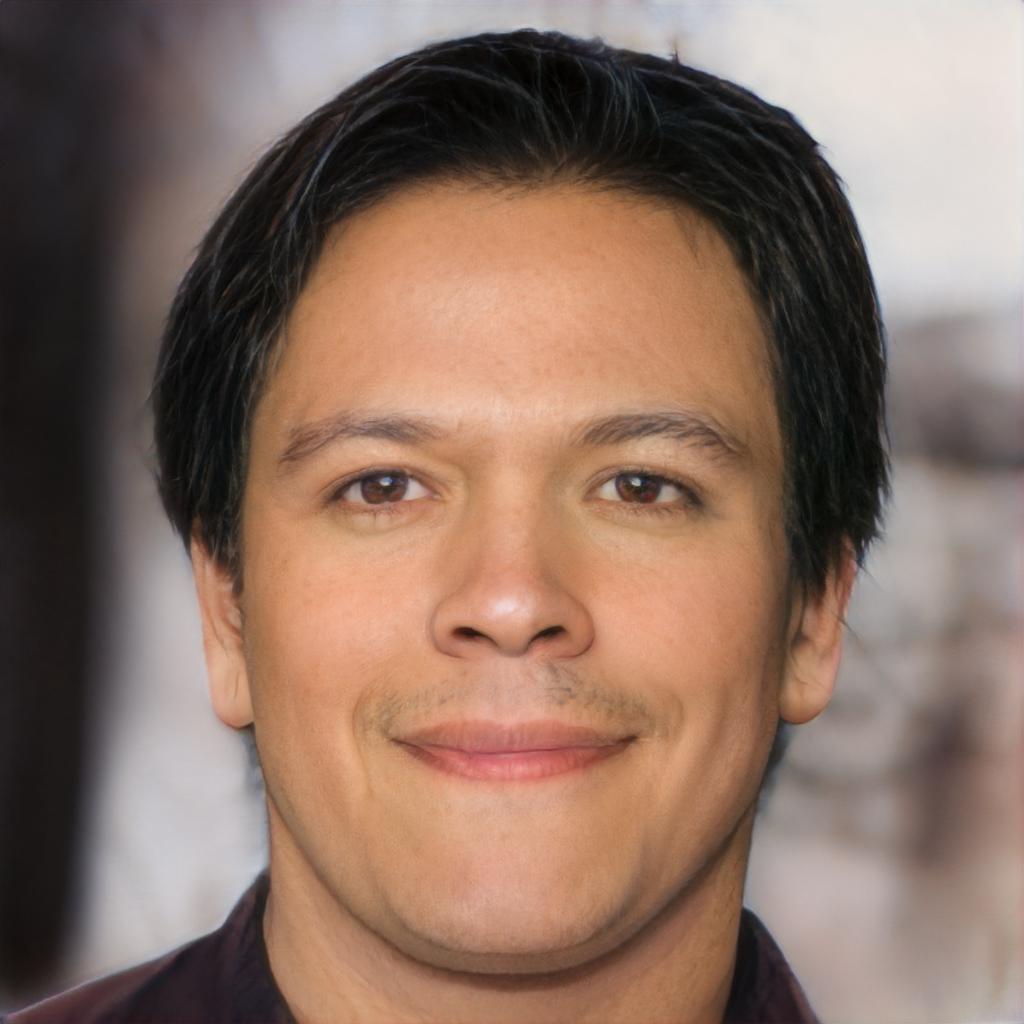} &
 \includegraphics[width=0.13\textwidth, ]{images/wstar_id_no_mag_3_coupl_dist_edit_train_ep5/20_img_3.jpg} &
 \includegraphics[width=0.13\textwidth, ]{images/wstar_id_no_mag_3_coupl_dist_edit_train_ep5/22_img_3.jpg} &
 \includegraphics[width=0.13\textwidth, ]{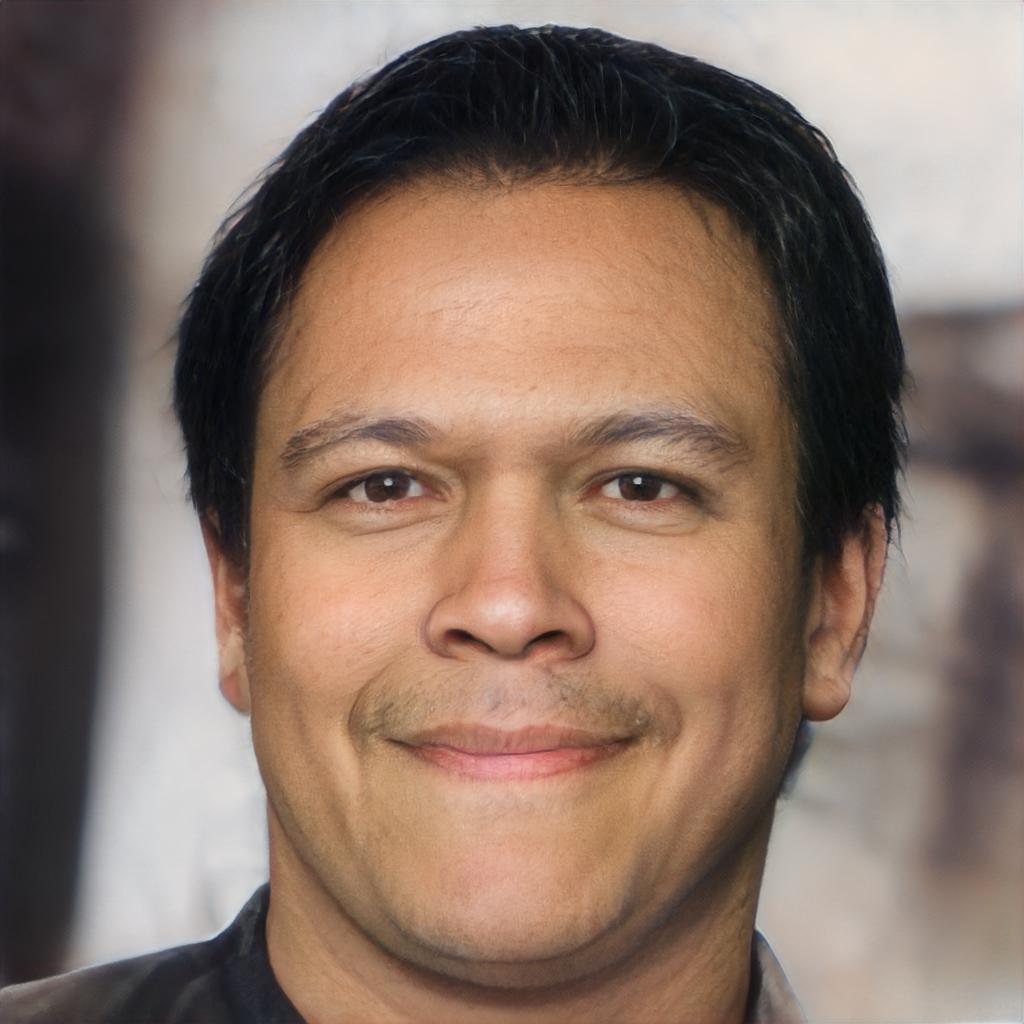} &
 \includegraphics[width=0.13\textwidth, ]{images/wstar_id_no_mag_3_coupl_dist_edit_train_ep5/36_img_3.jpg} \\
 
 \\
\end{tabular}
\caption{Ablation study: Image editing using InterFaceGAN.}
\label{fig:edit_abl_1}

\end{figure}

\begin{figure}[h]
\setlength\tabcolsep{2pt}%
\centering
\begin{tabular}{p{0.25cm}ccccccc}
\centering
&
 \textbf{Original} &
 \textbf{Inverted} &
 \textbf{Makeup} &
 \textbf{Male} &
 \textbf{Mustache} &
 \textbf{Chubby} &
 \textbf{Lipstick} \\
 \begin{turn}{90} \hspace{0.5cm} \wplus\end{turn} &
 \includegraphics[width=0.13\textwidth]{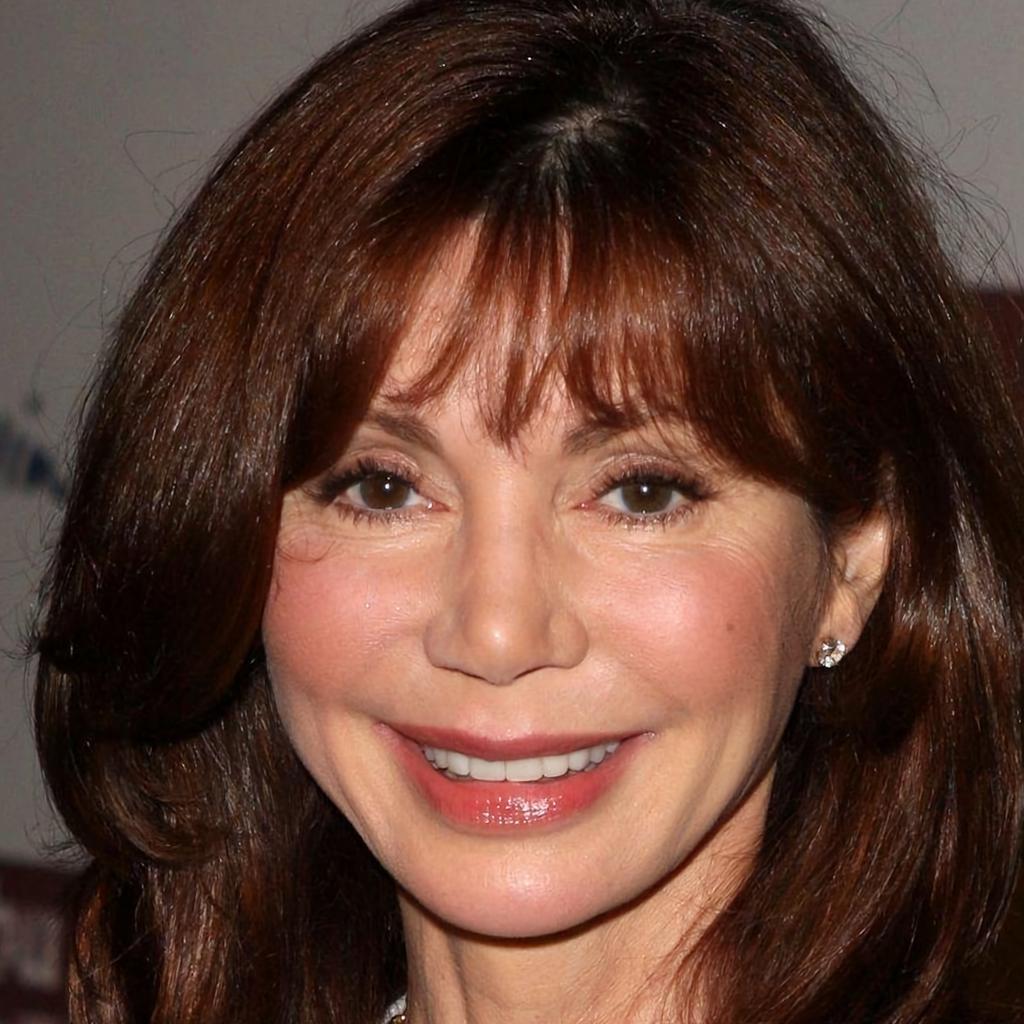} & 
 \includegraphics[width=0.13\textwidth]{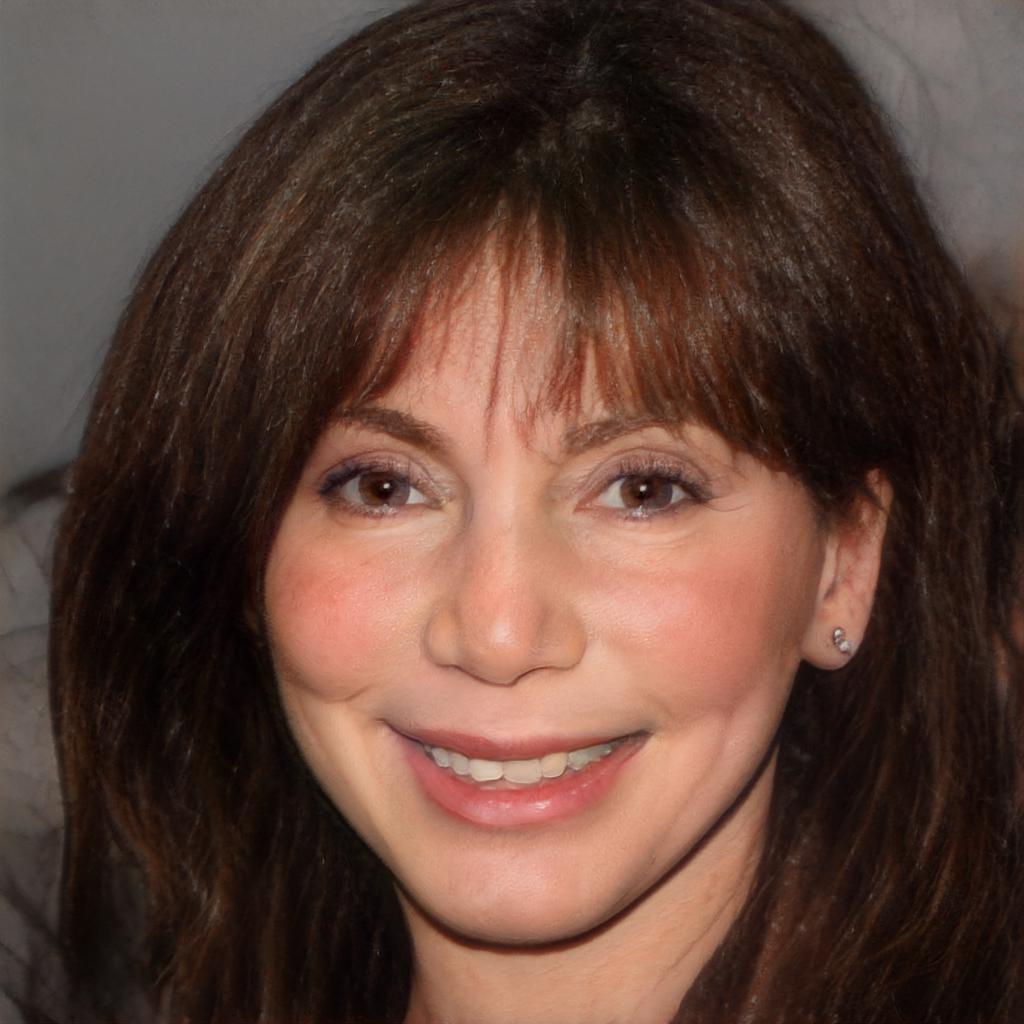} &
 \includegraphics[width=0.13\textwidth]{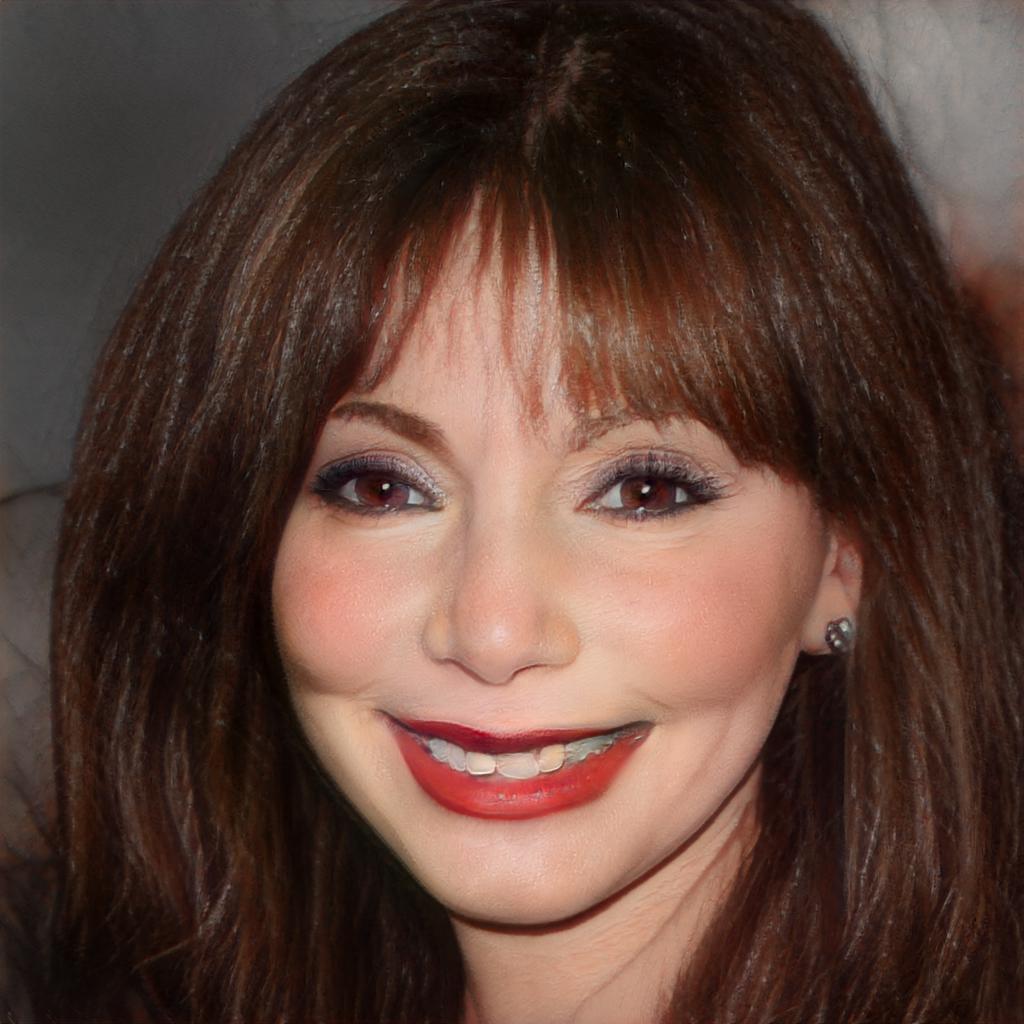} &
 \includegraphics[width=0.13\textwidth]{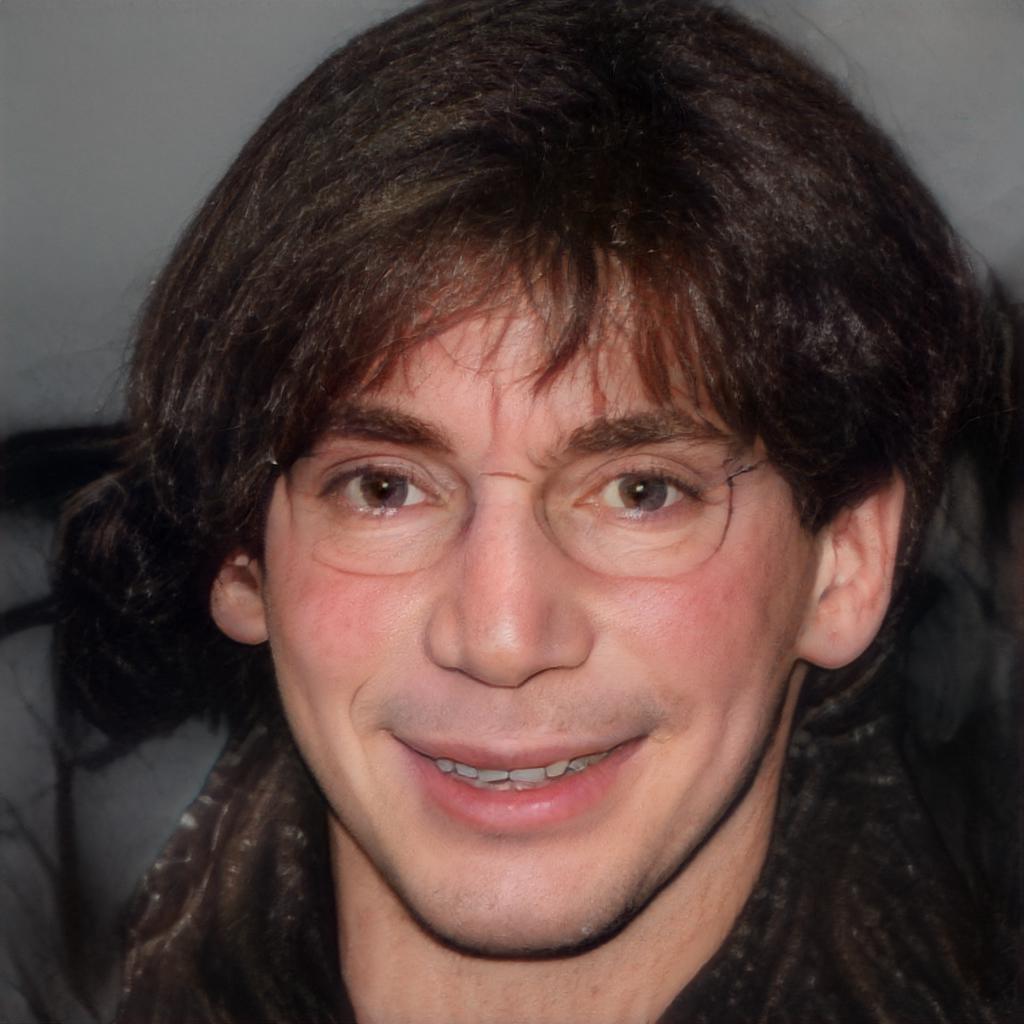} &
 \includegraphics[width=0.13\textwidth]{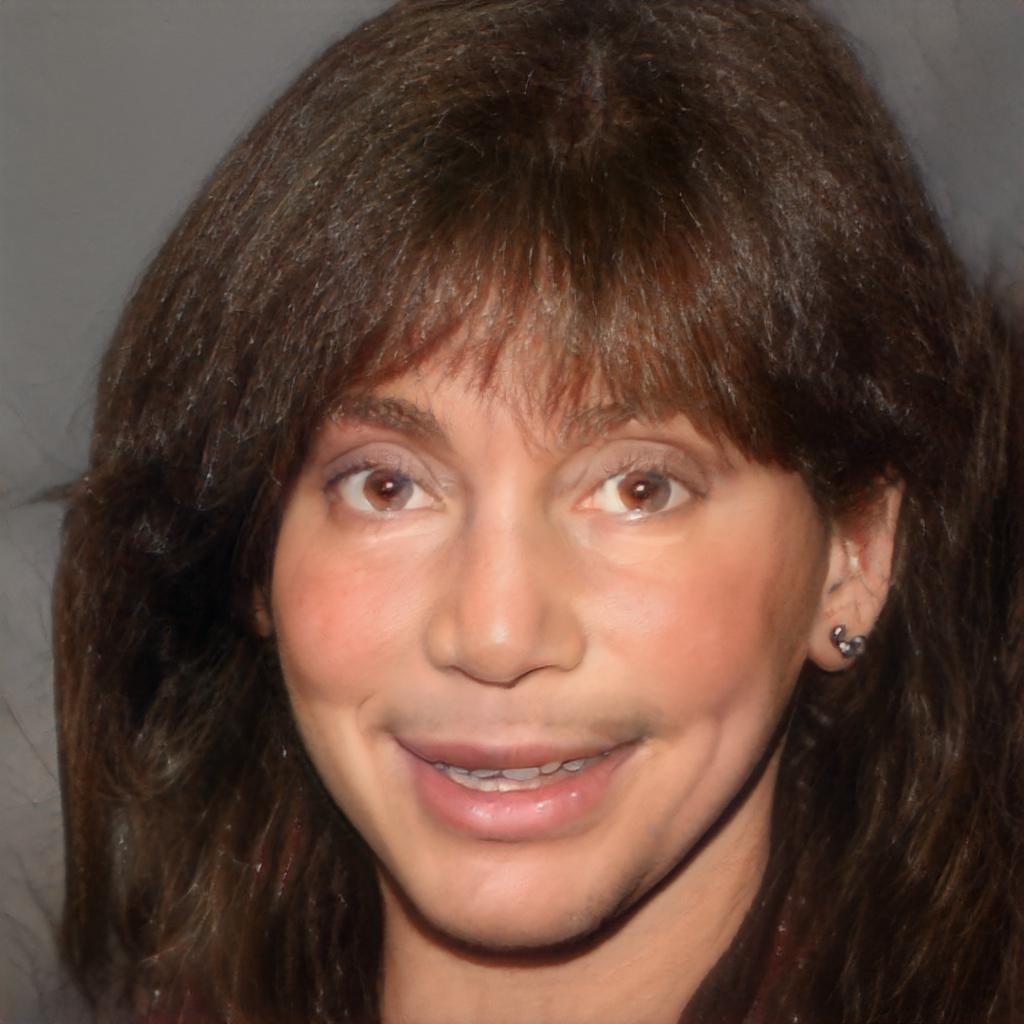} &
 \includegraphics[width=0.13\textwidth]{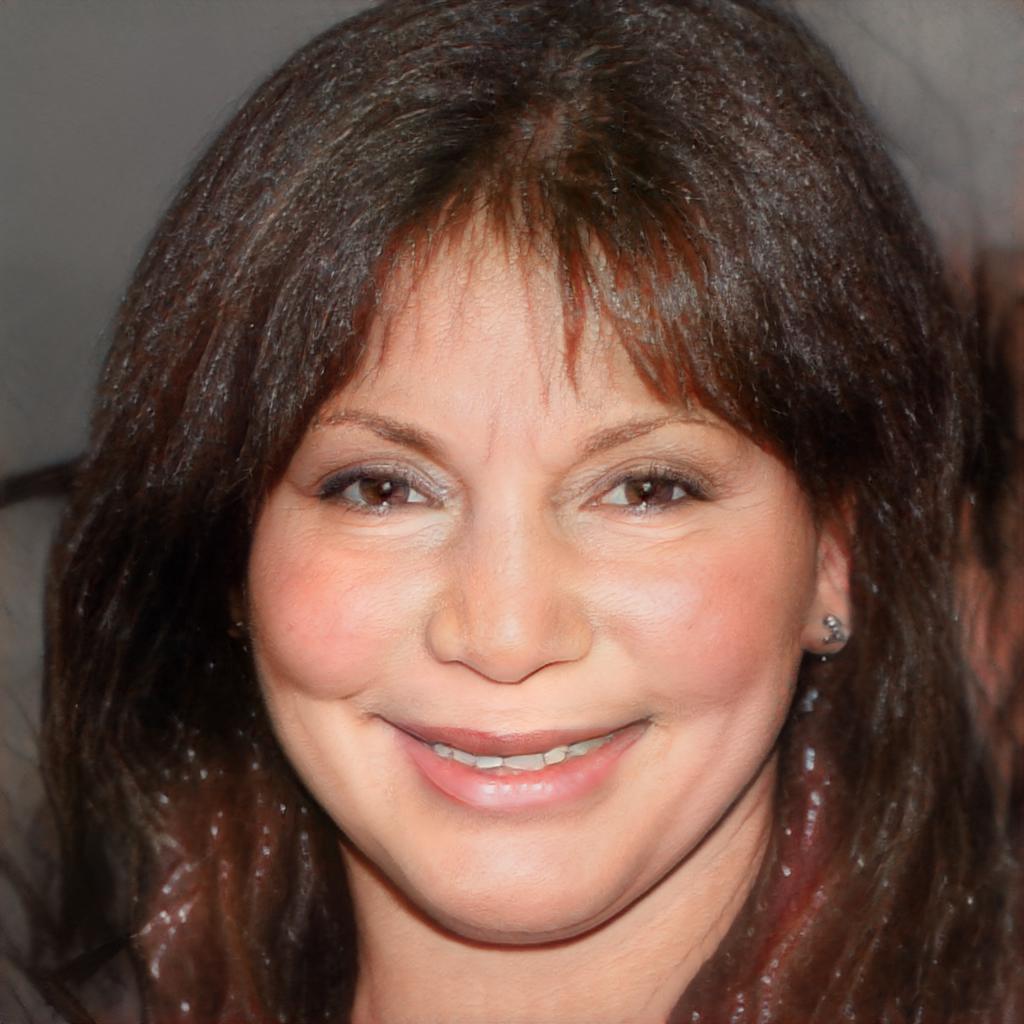} &
 \includegraphics[width=0.13\textwidth]{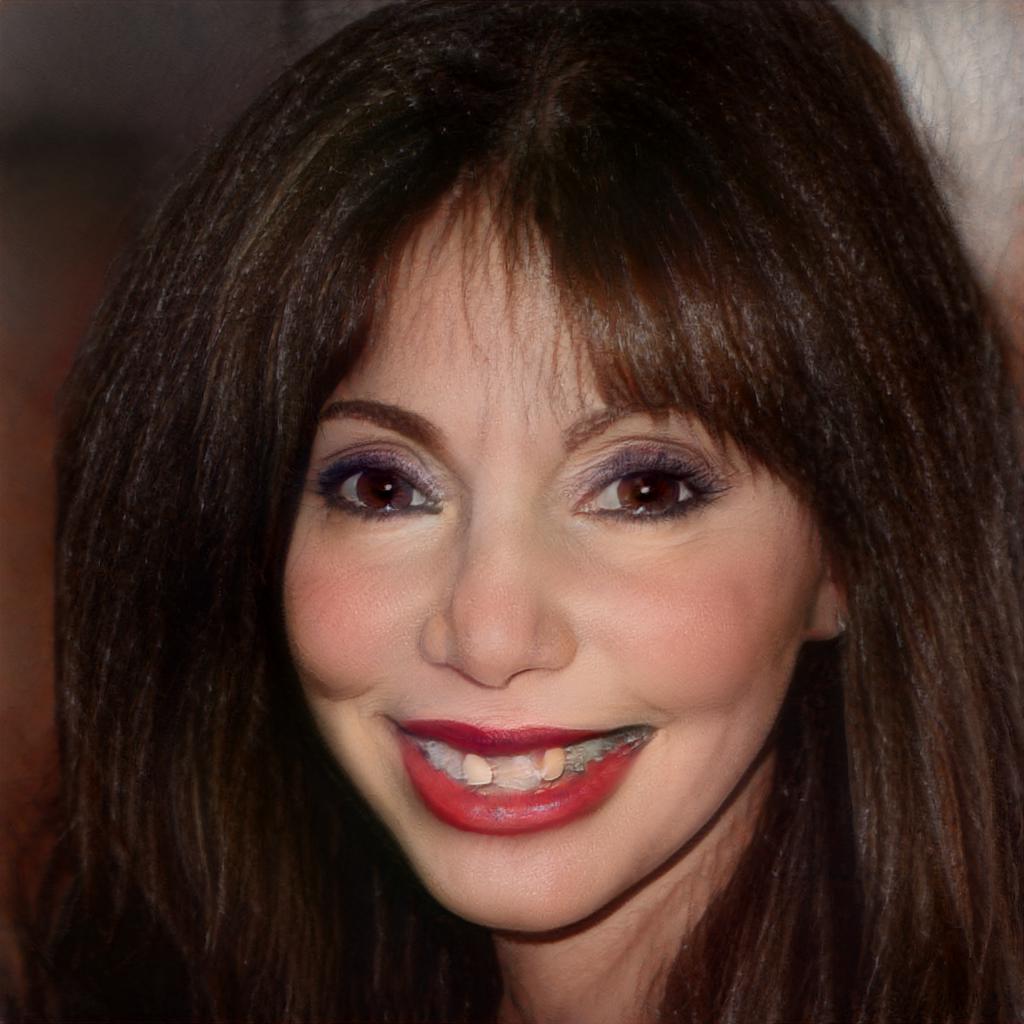} \\
\begin{turn}{90} \hspace{0.5cm} $\mathcal{W}^{\star}_{ID}$ \end{turn} &
 \includegraphics[width=0.13\textwidth]{images/original/06001.jpg} & 
 \includegraphics[width=0.13\textwidth]{images/inversion/13_orig_img_2.jpg} &
 \includegraphics[width=0.13\textwidth, ]{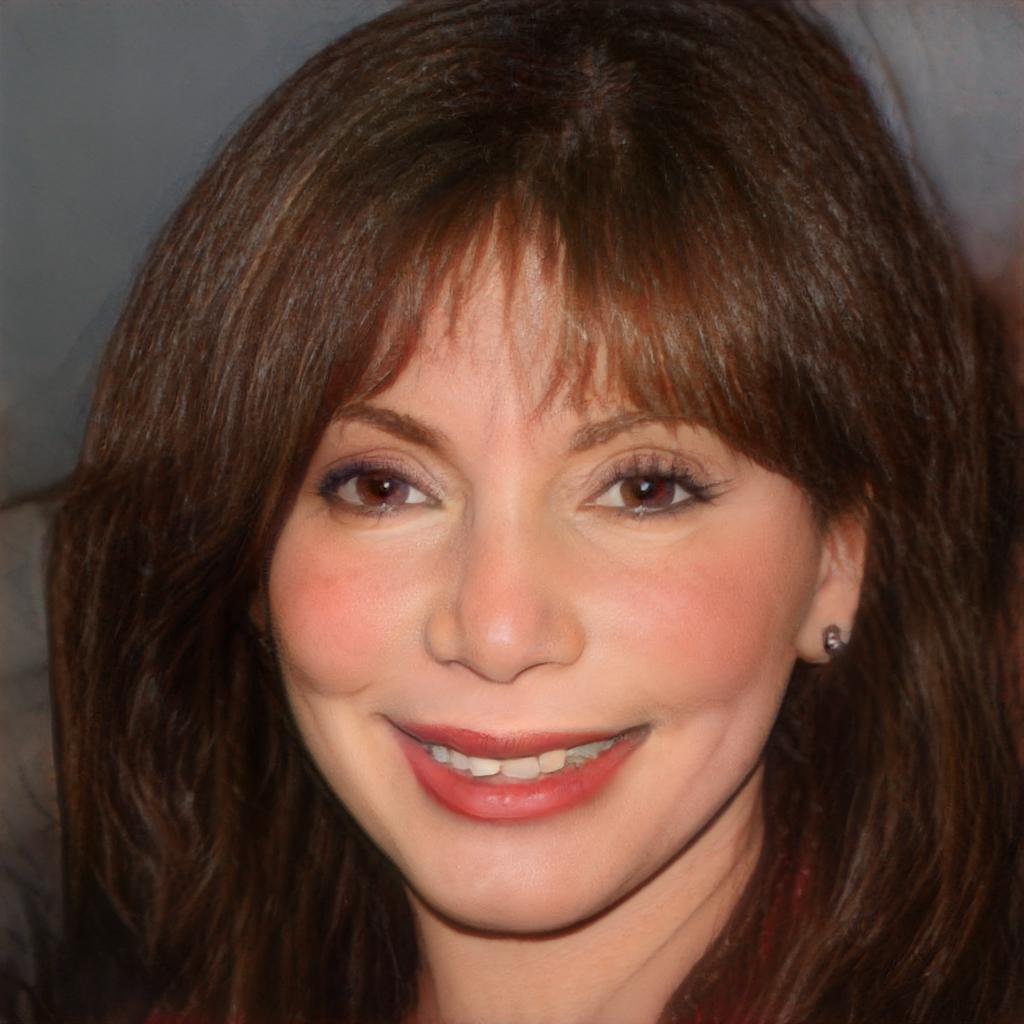} &
 \includegraphics[width=0.13\textwidth, ]{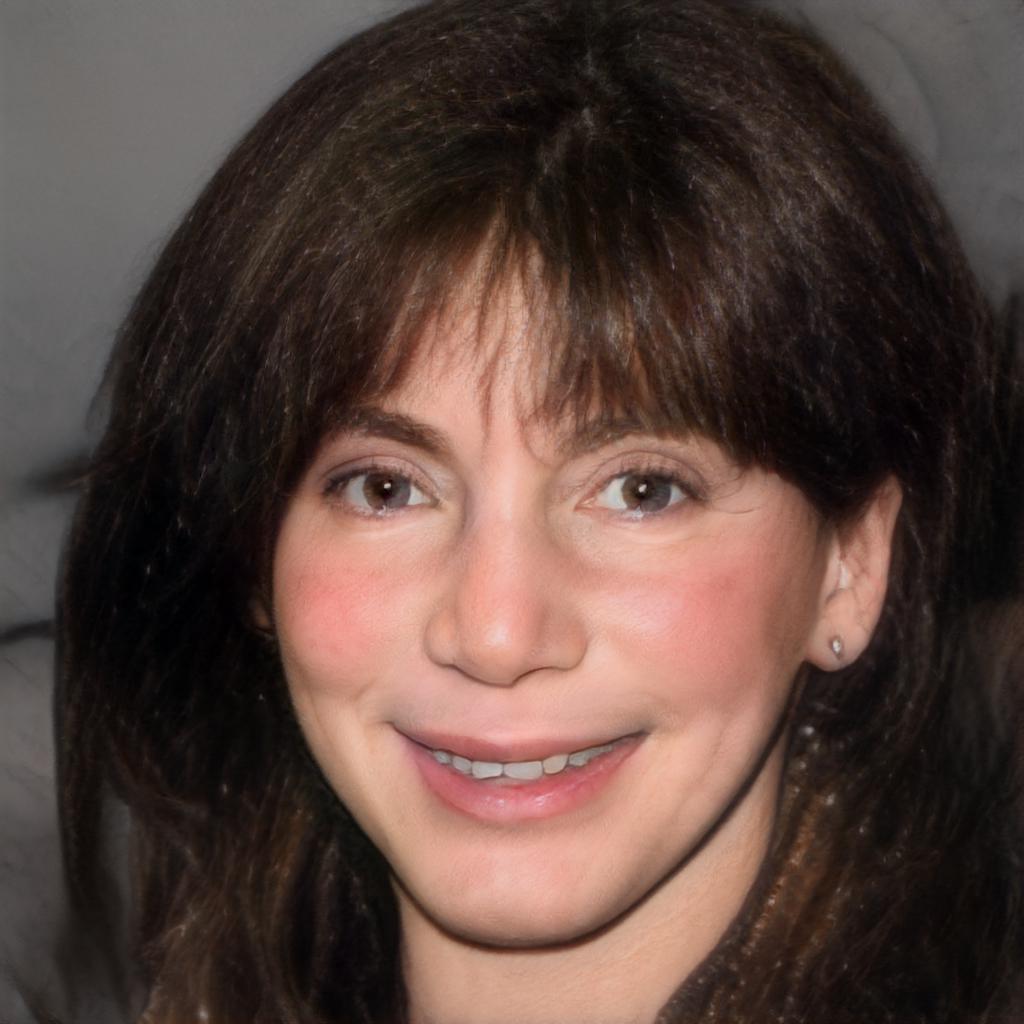} &
 \includegraphics[width=0.13\textwidth, ]{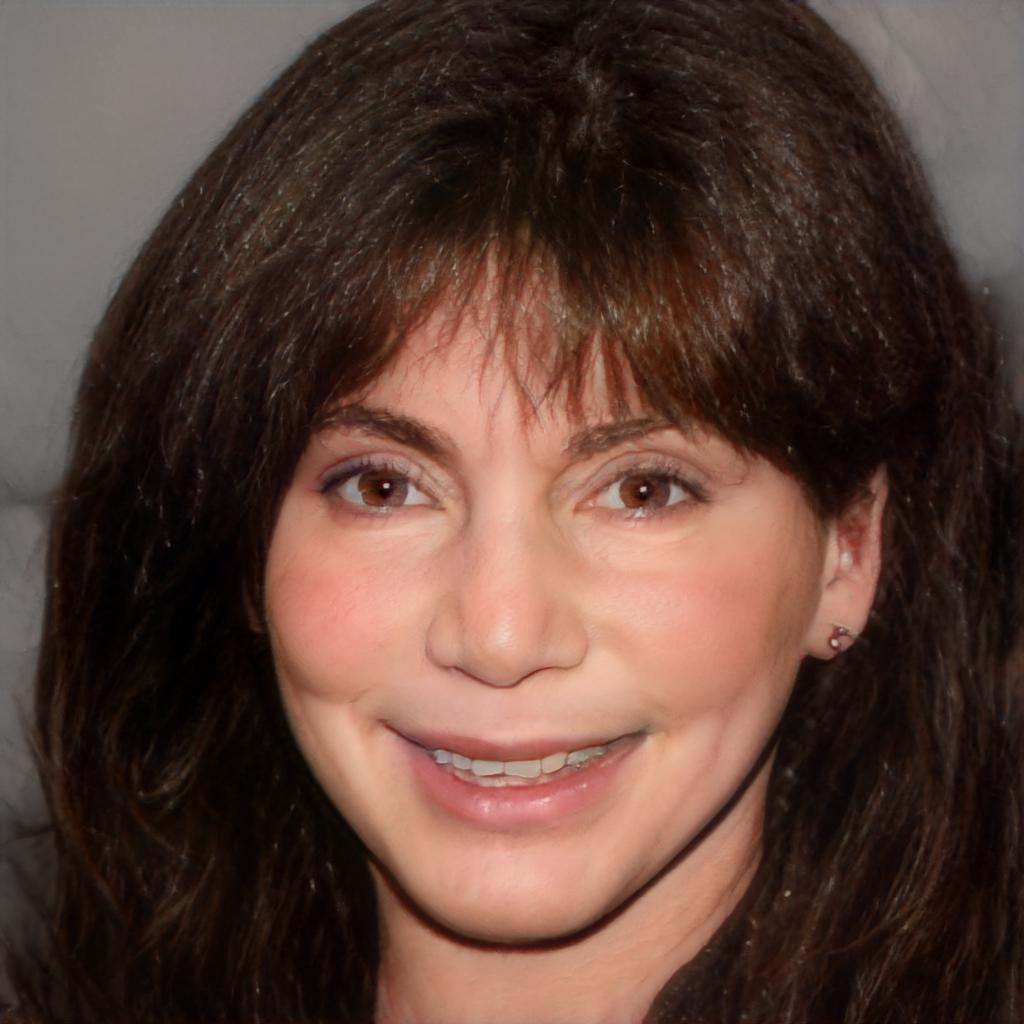} &
 \includegraphics[width=0.13\textwidth, ]{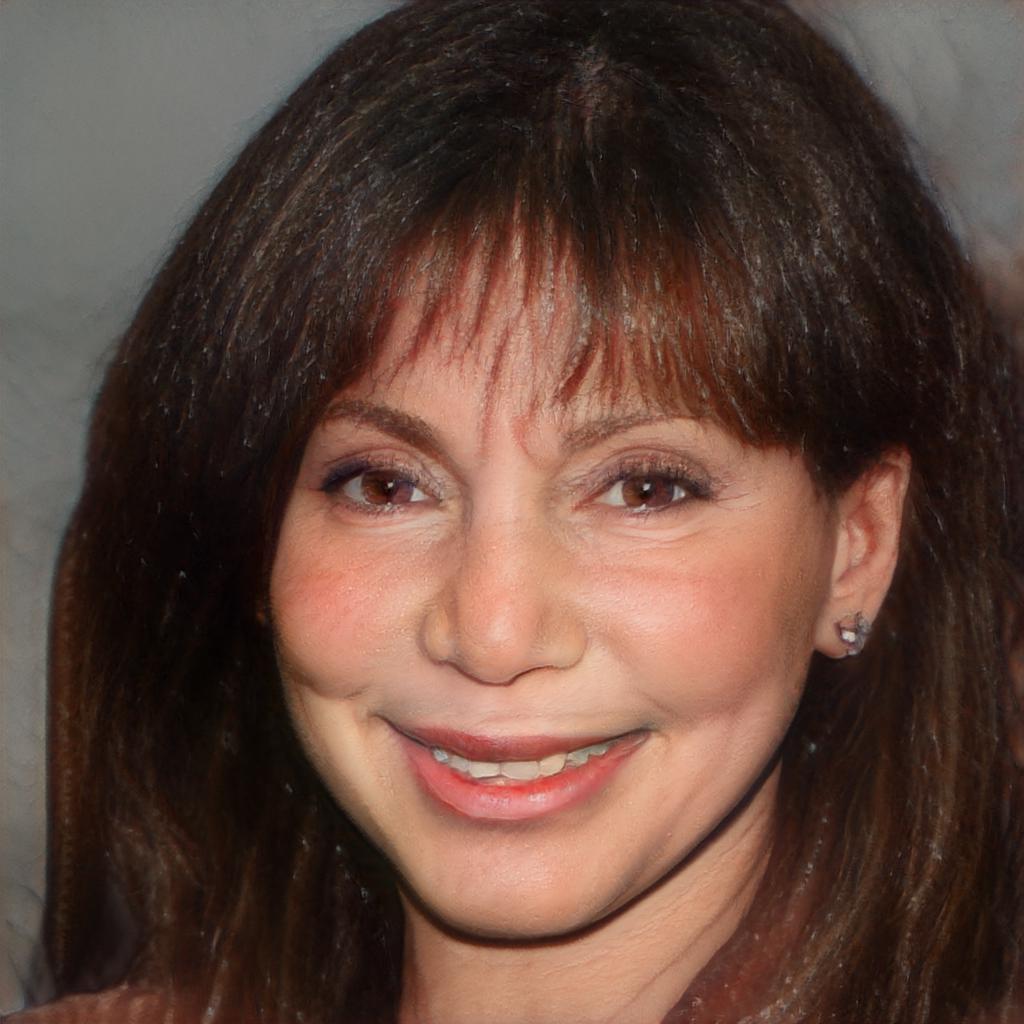} &
 \includegraphics[width=0.13\textwidth, ]{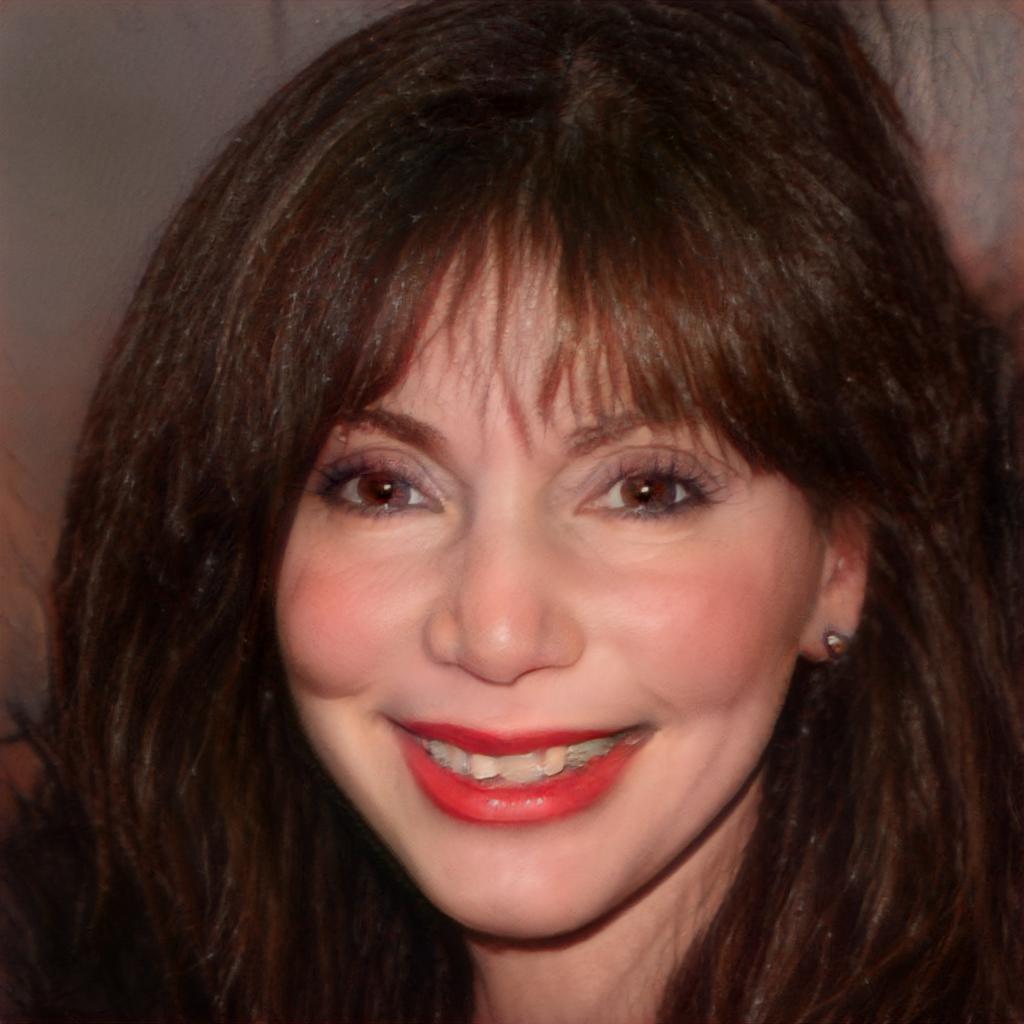} \\
 \midrule
 \begin{turn}{90} \hspace{0.5cm} \wplus\end{turn} &
 \includegraphics[width=0.13\textwidth]{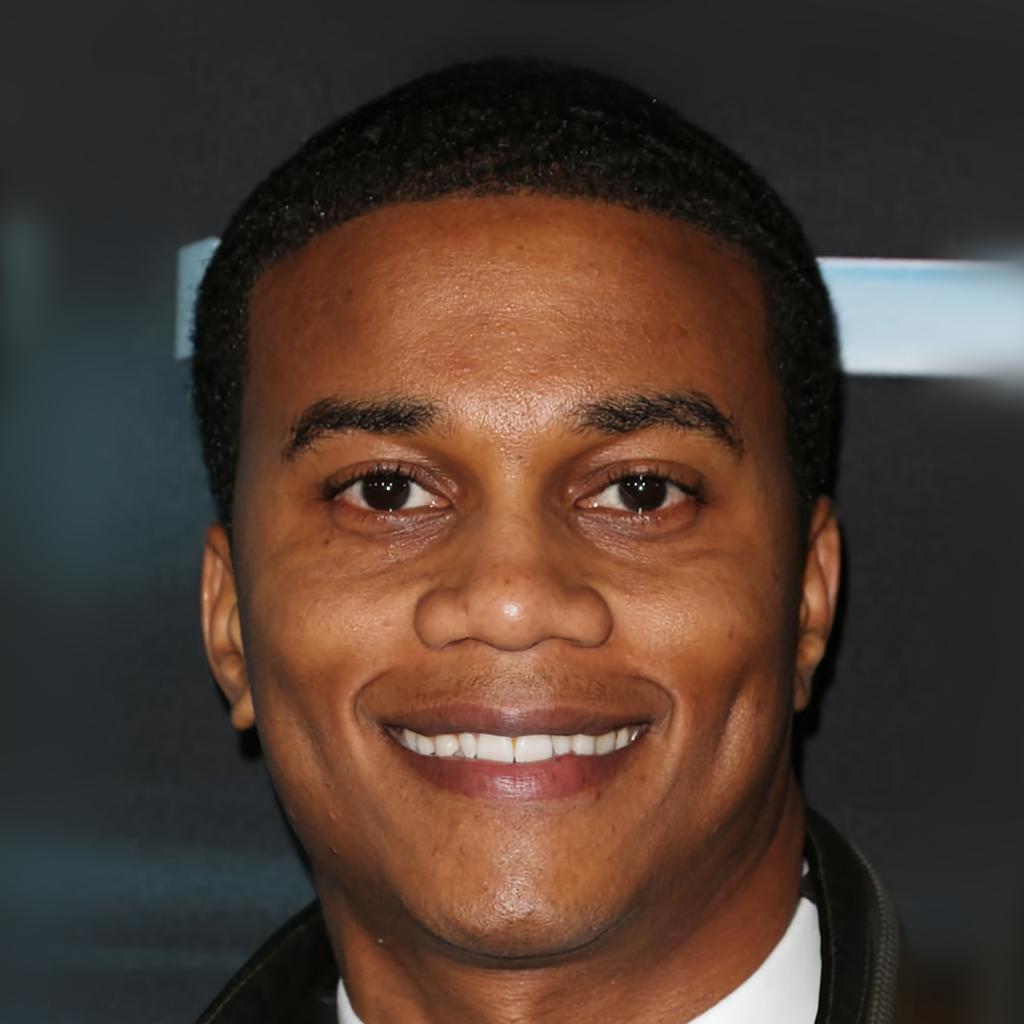} & 
 \includegraphics[width=0.13\textwidth]{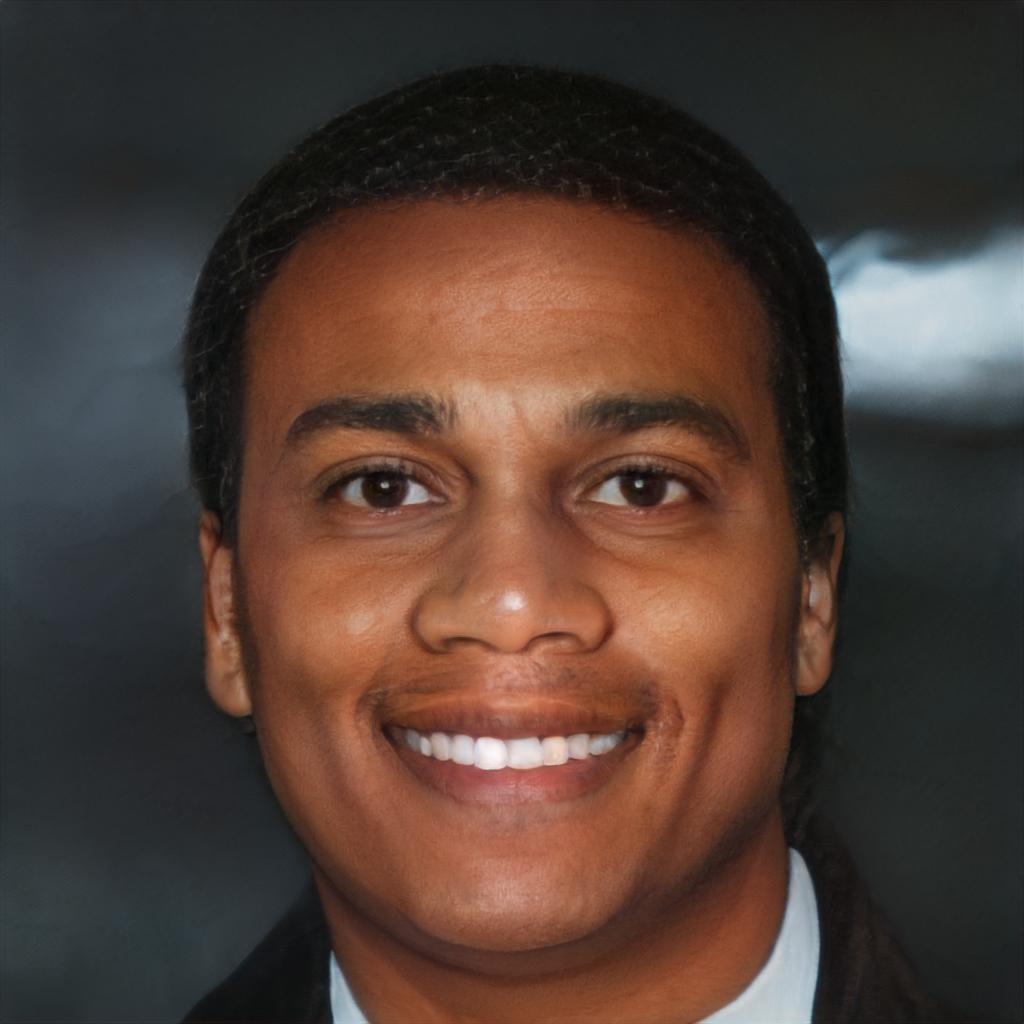} &
 \includegraphics[width=0.13\textwidth, ]{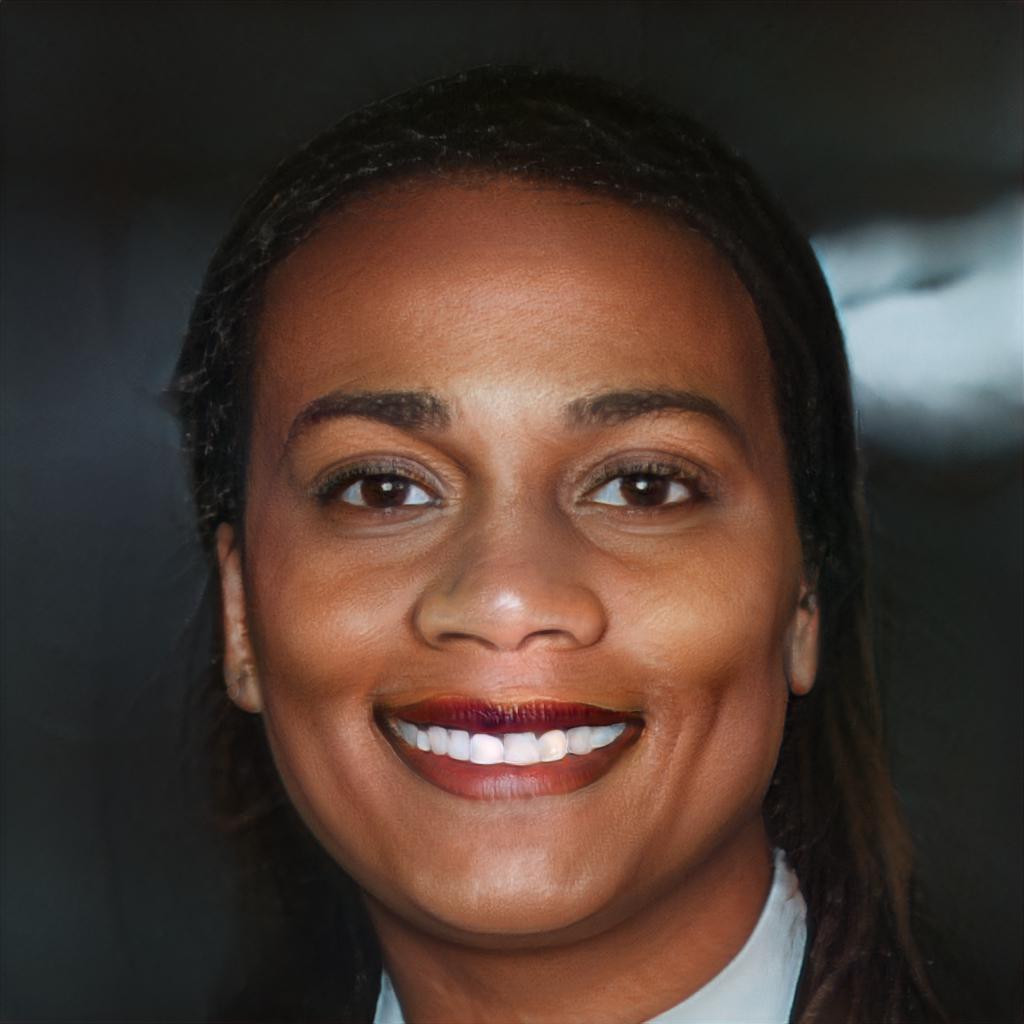} &
 \includegraphics[width=0.13\textwidth, ]{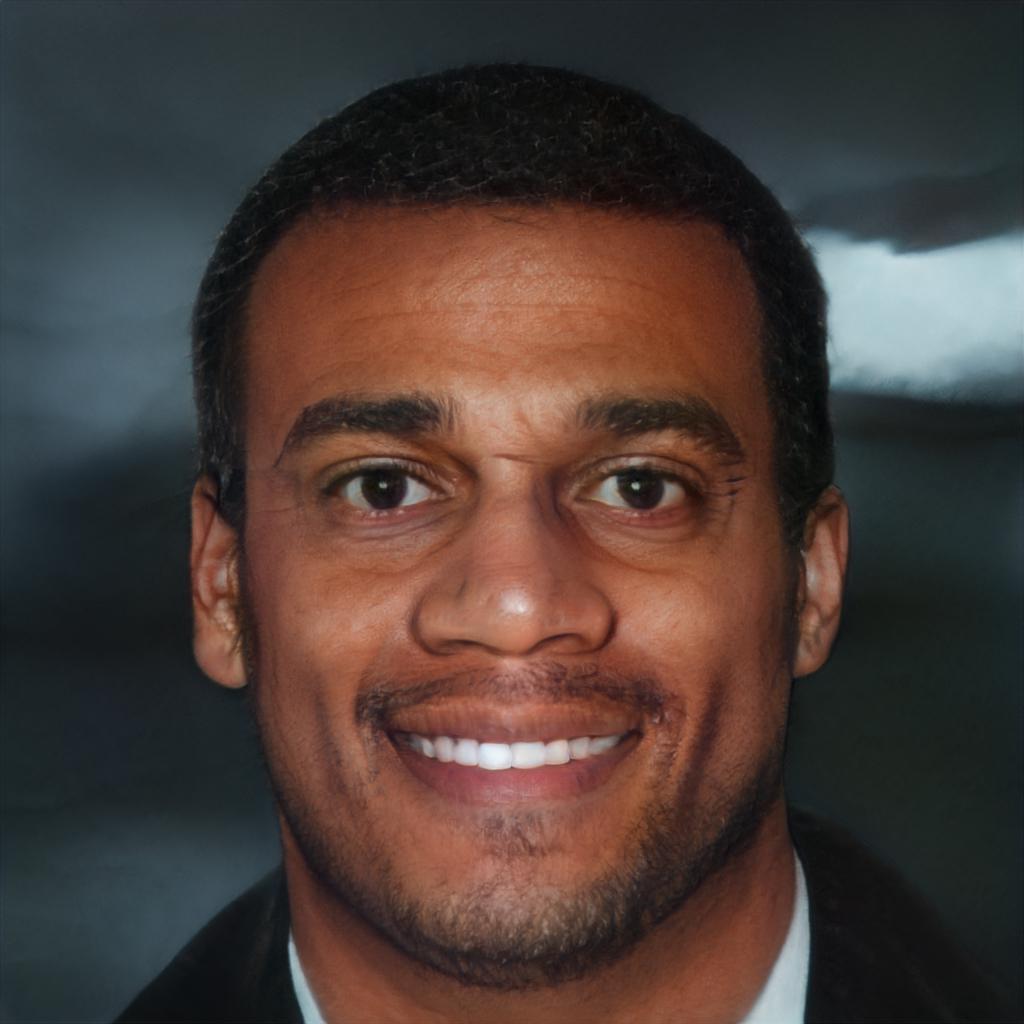} &
 \includegraphics[width=0.13\textwidth, ]{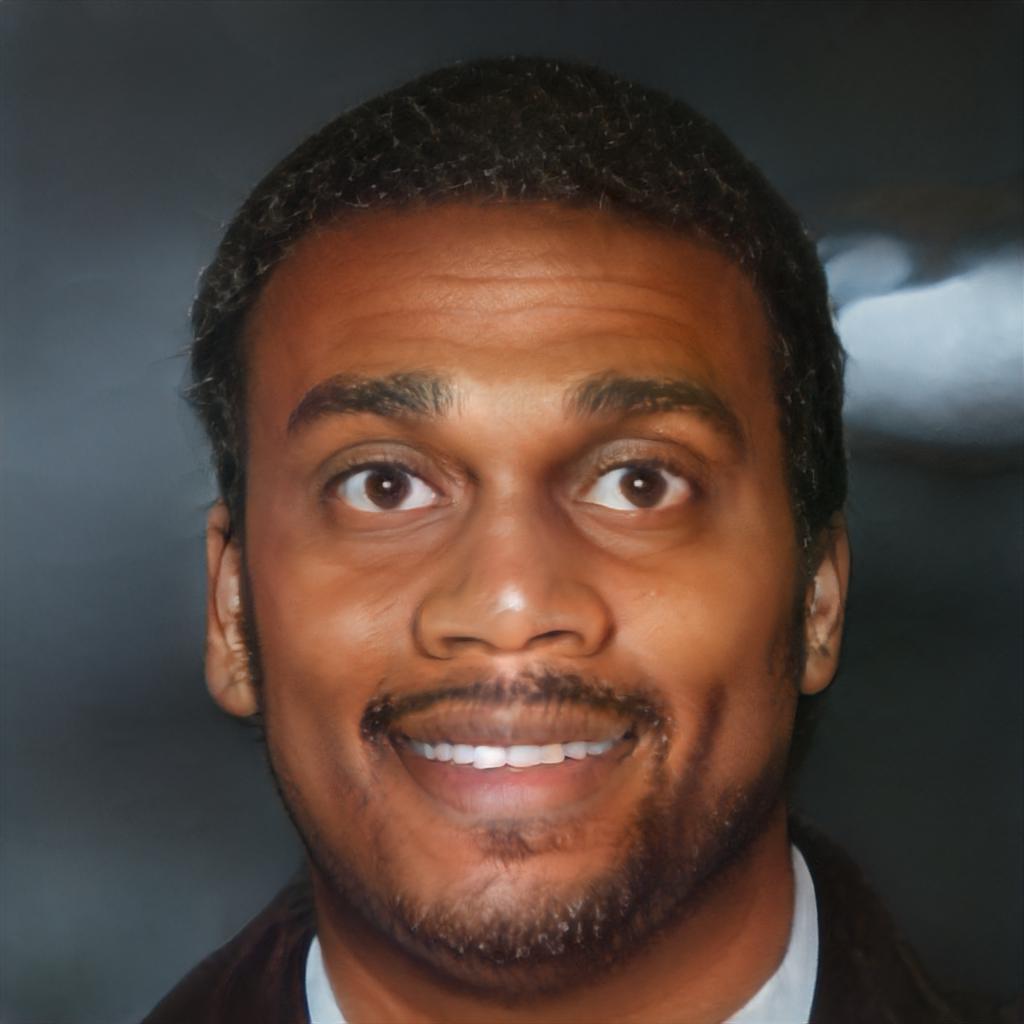} &
 \includegraphics[width=0.13\textwidth, ]{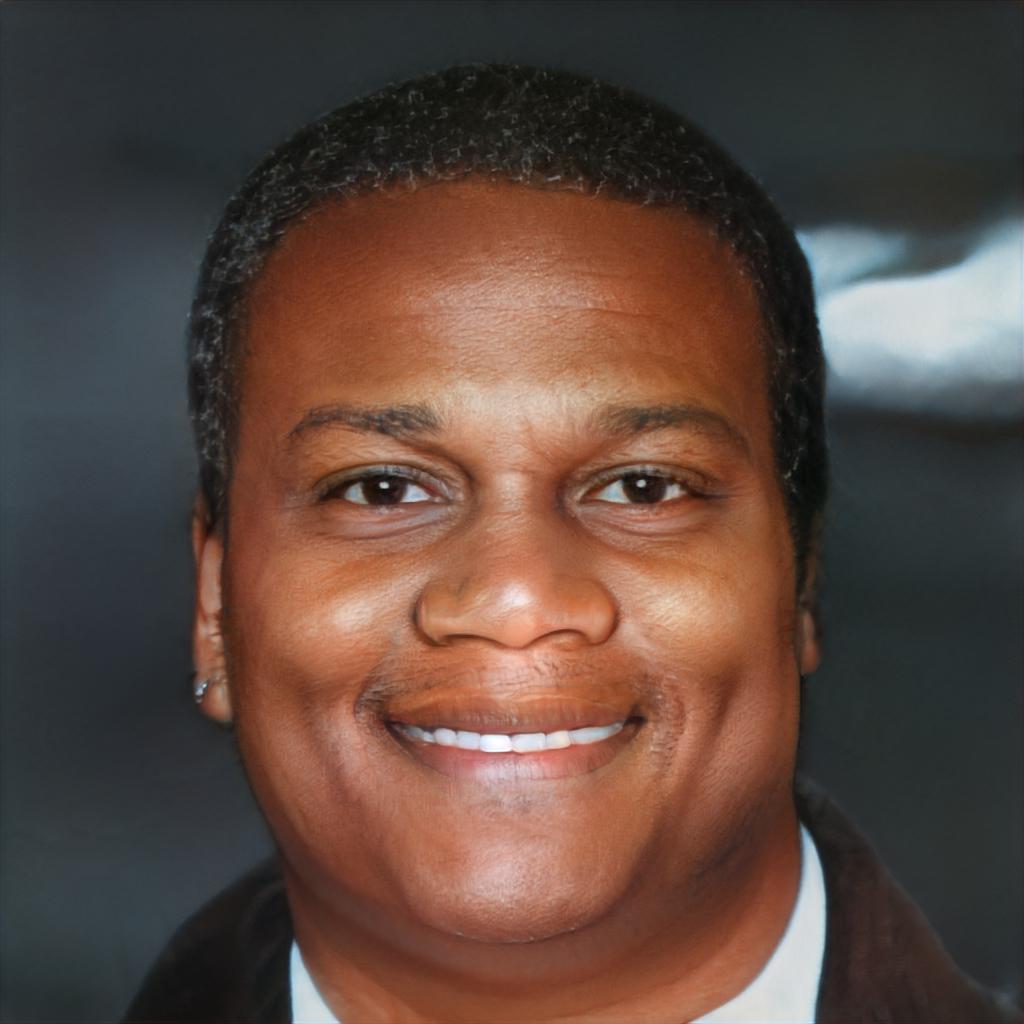} &
 \includegraphics[width=0.13\textwidth]{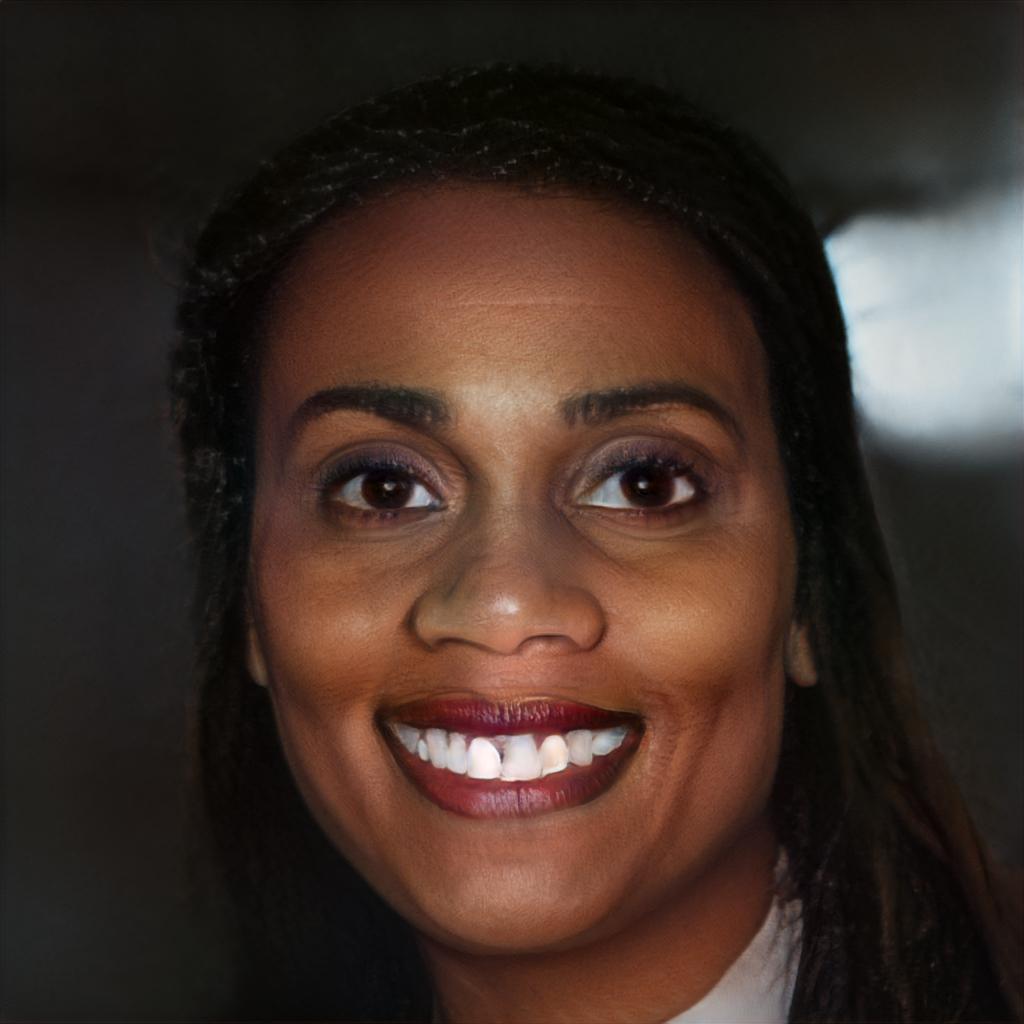} \\
\begin{turn}{90} \hspace{0.5cm} $\mathcal{W}^{\star}_{ID}$ \end{turn} &
 \includegraphics[width=0.13\textwidth]{images/original/06019.jpg} & 
 \includegraphics[width=0.13\textwidth]{images/inversion/13_orig_img_20.jpg} &
 \includegraphics[width=0.13\textwidth, ]{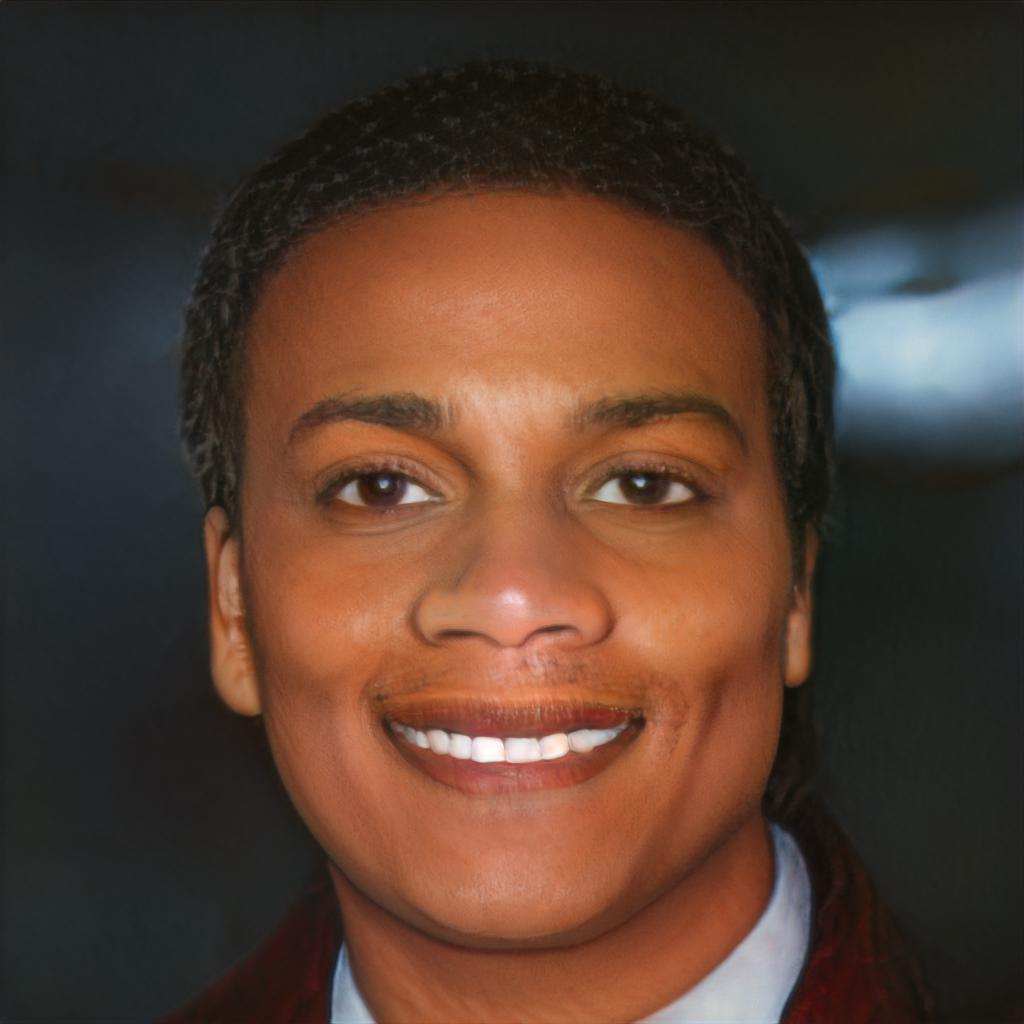} &
 \includegraphics[width=0.13\textwidth, ]{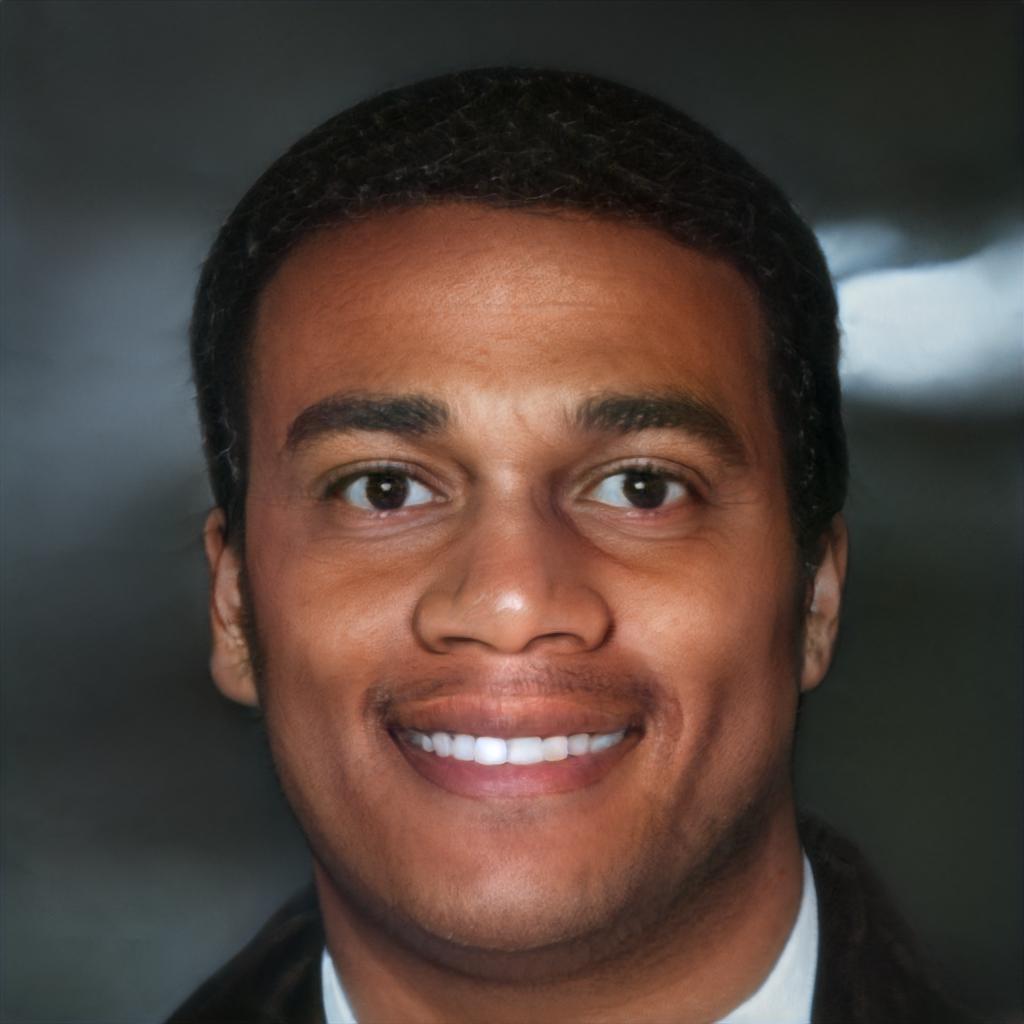} &
 \includegraphics[width=0.13\textwidth, ]{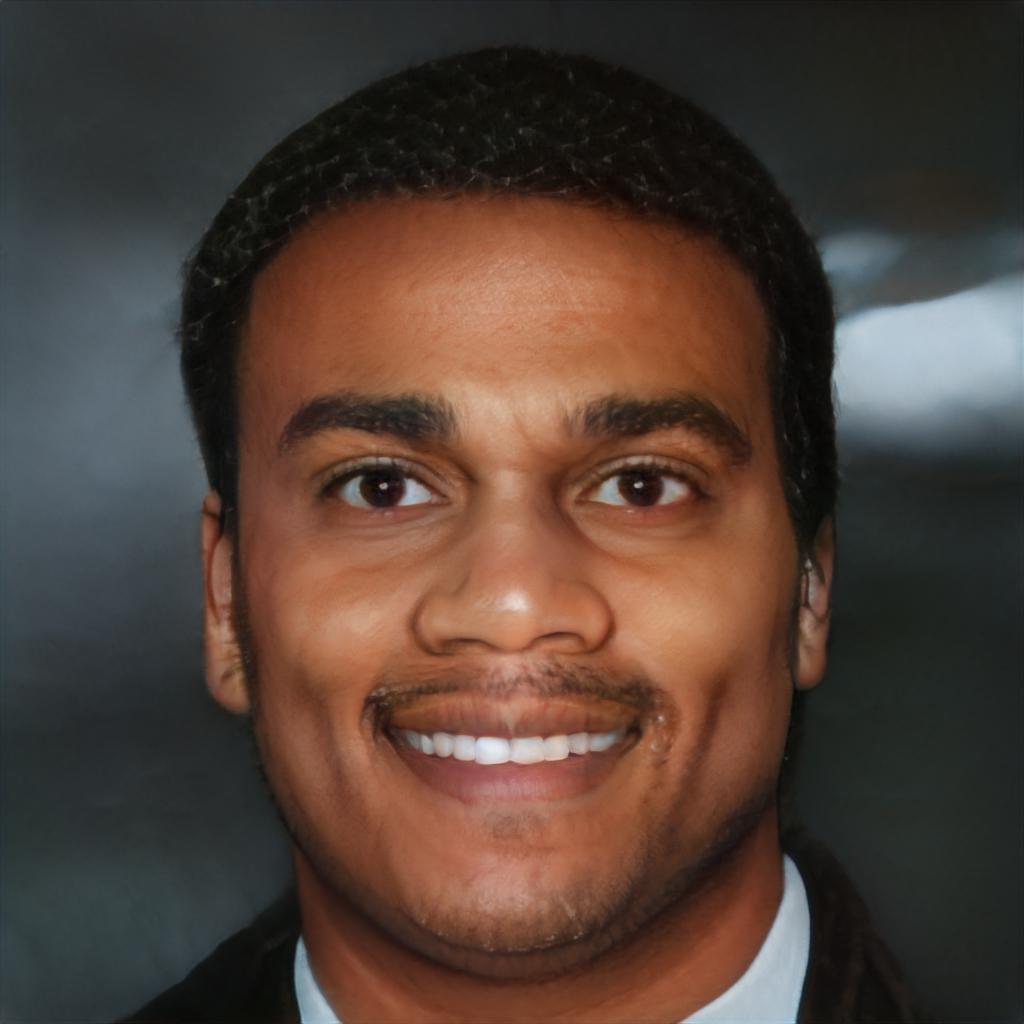} &
 \includegraphics[width=0.13\textwidth, ]{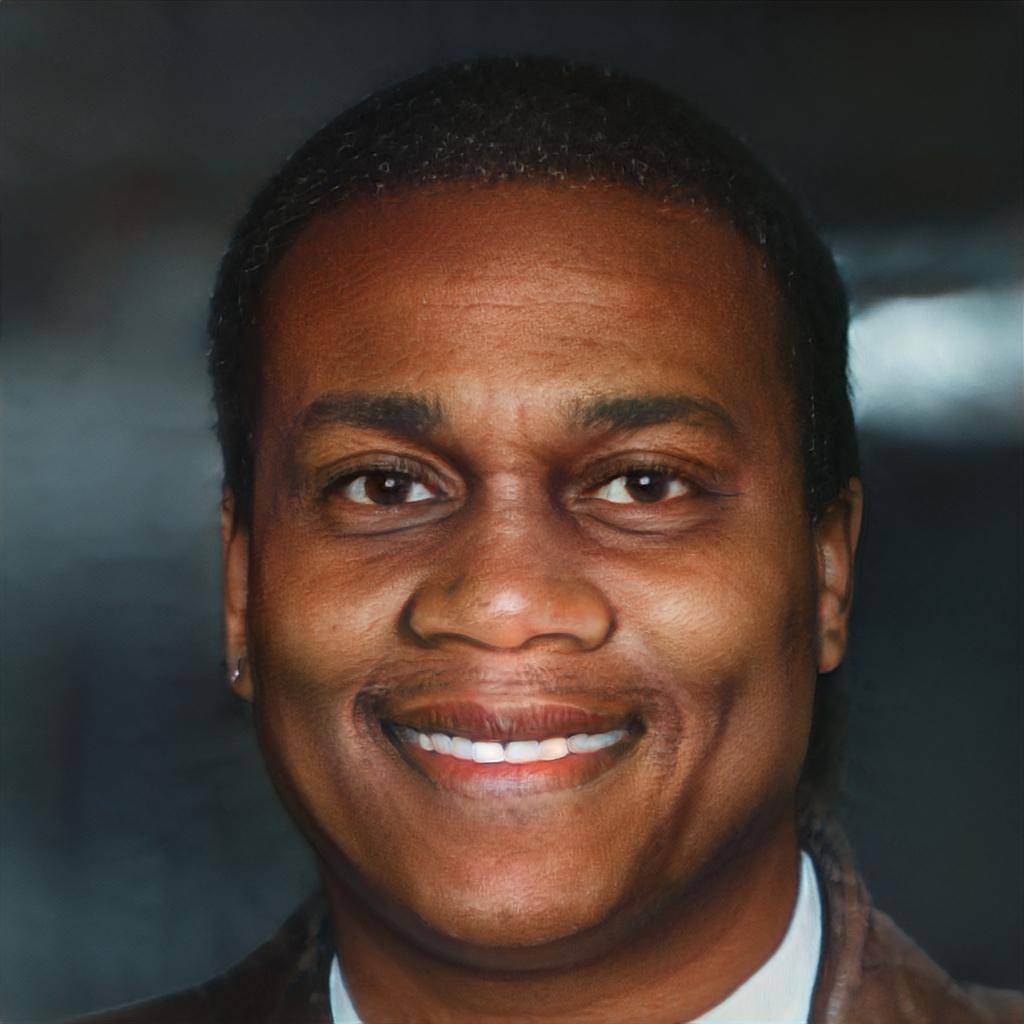} &
 \includegraphics[width=0.13\textwidth, ]{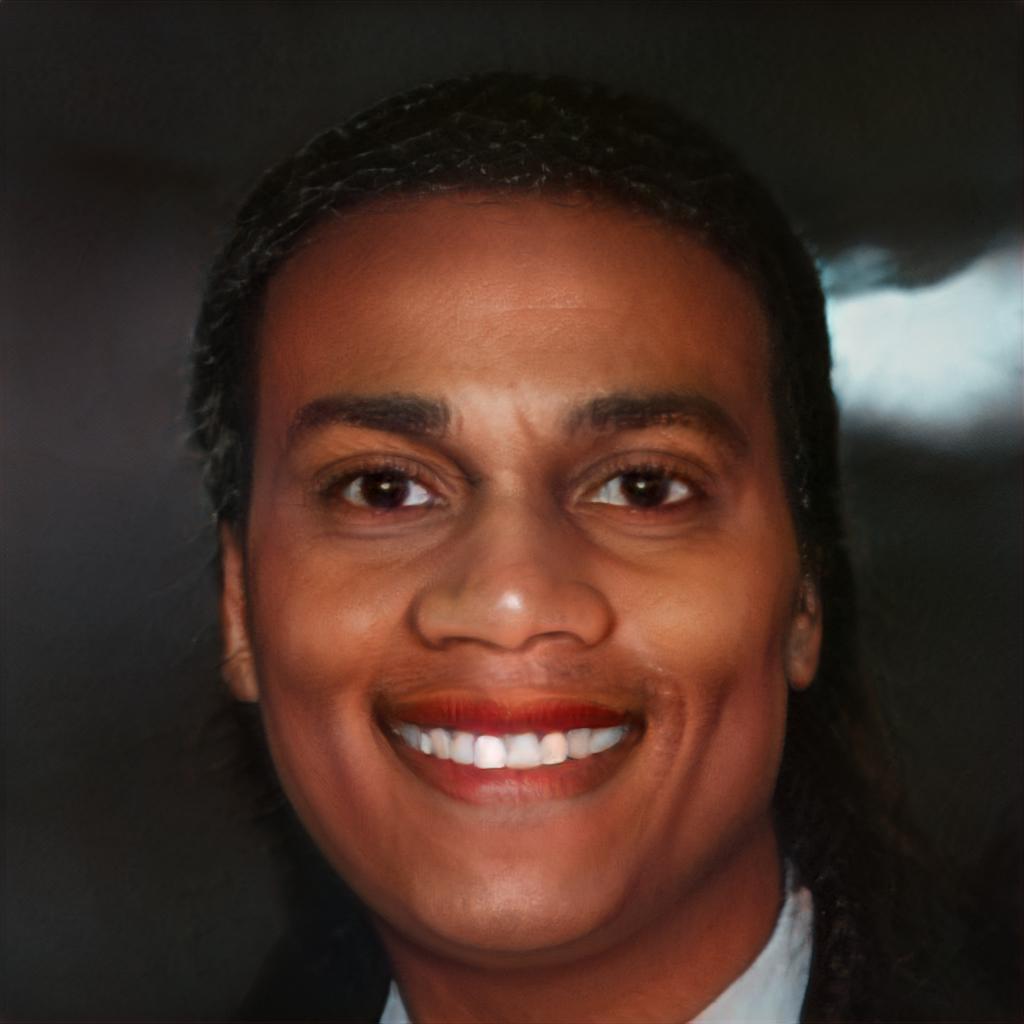} \\
 \midrule
 \begin{turn}{90} \hspace{0.5cm} \wplus\end{turn} &
 \includegraphics[width=0.13\textwidth, ]{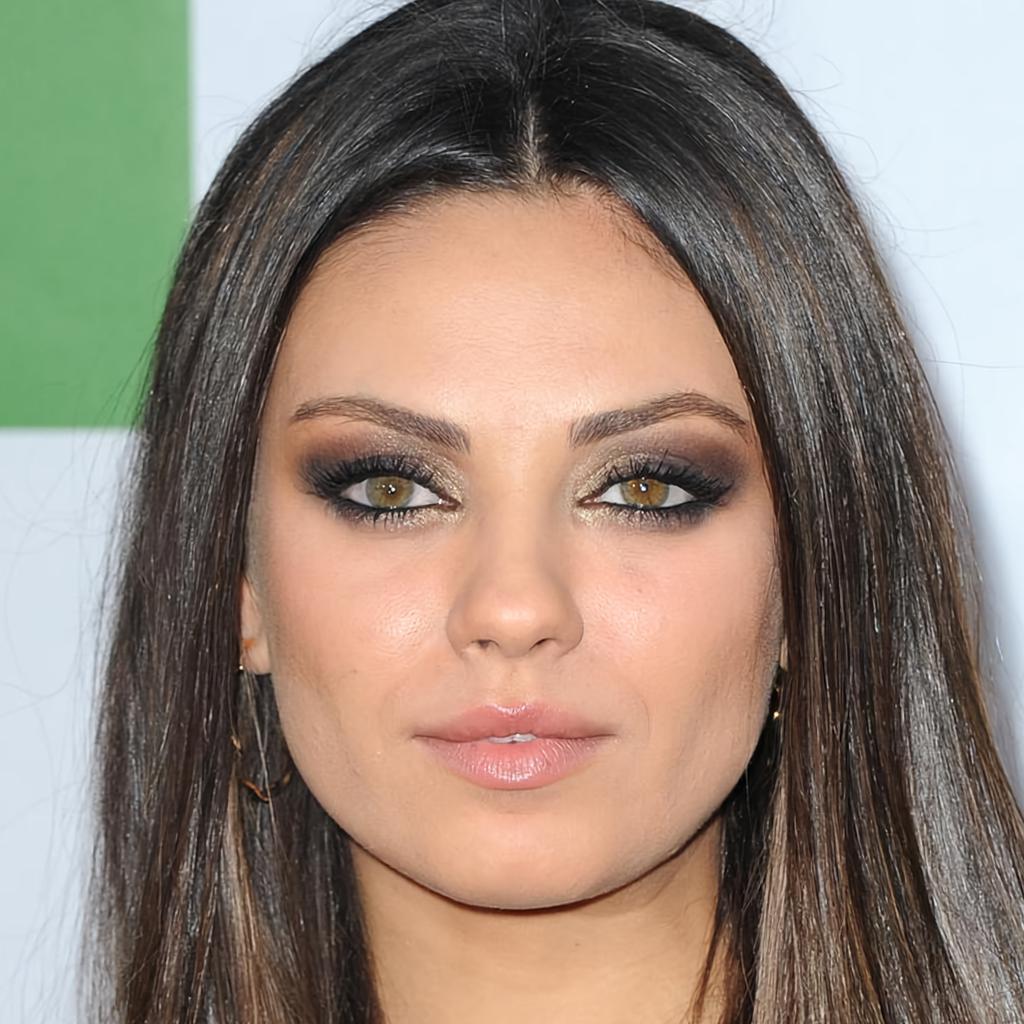} & 
 \includegraphics[width=0.13\textwidth, ]{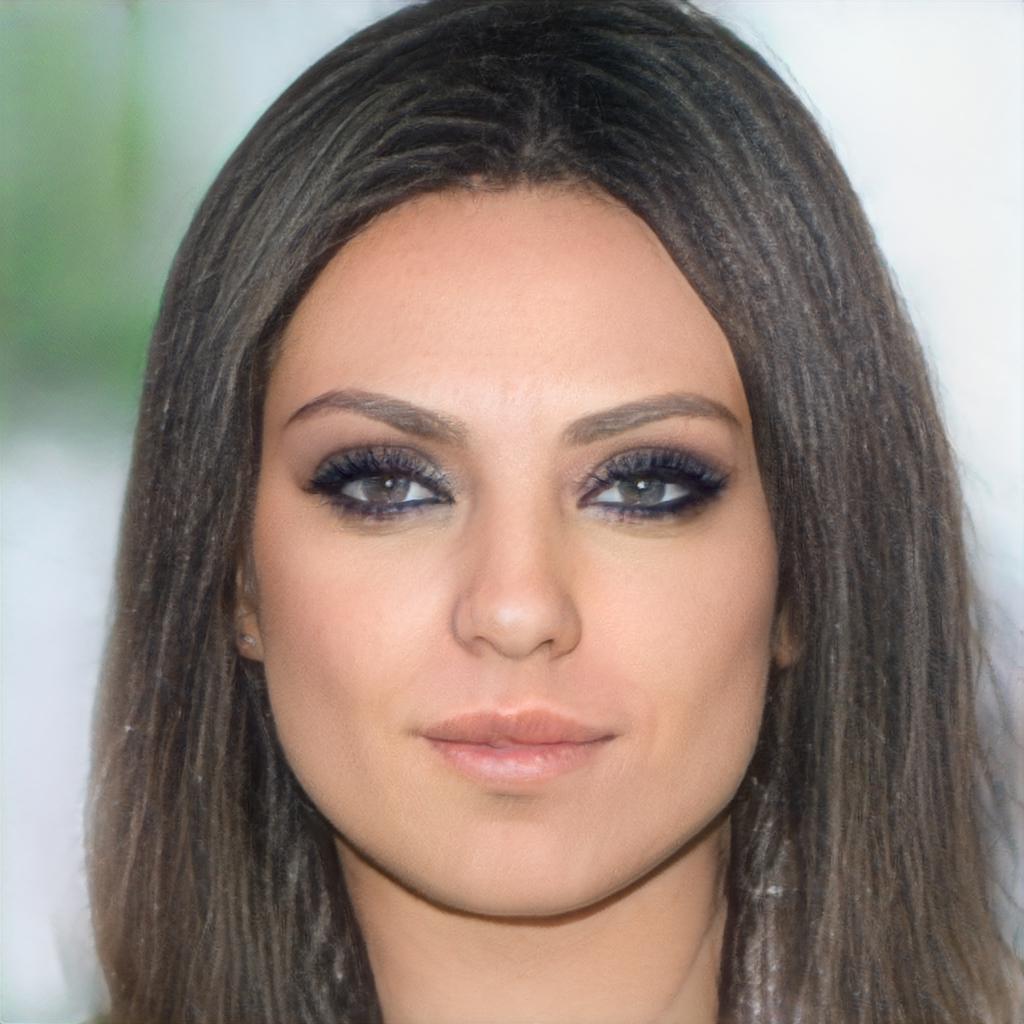} &
 \includegraphics[width=0.13\textwidth, ]{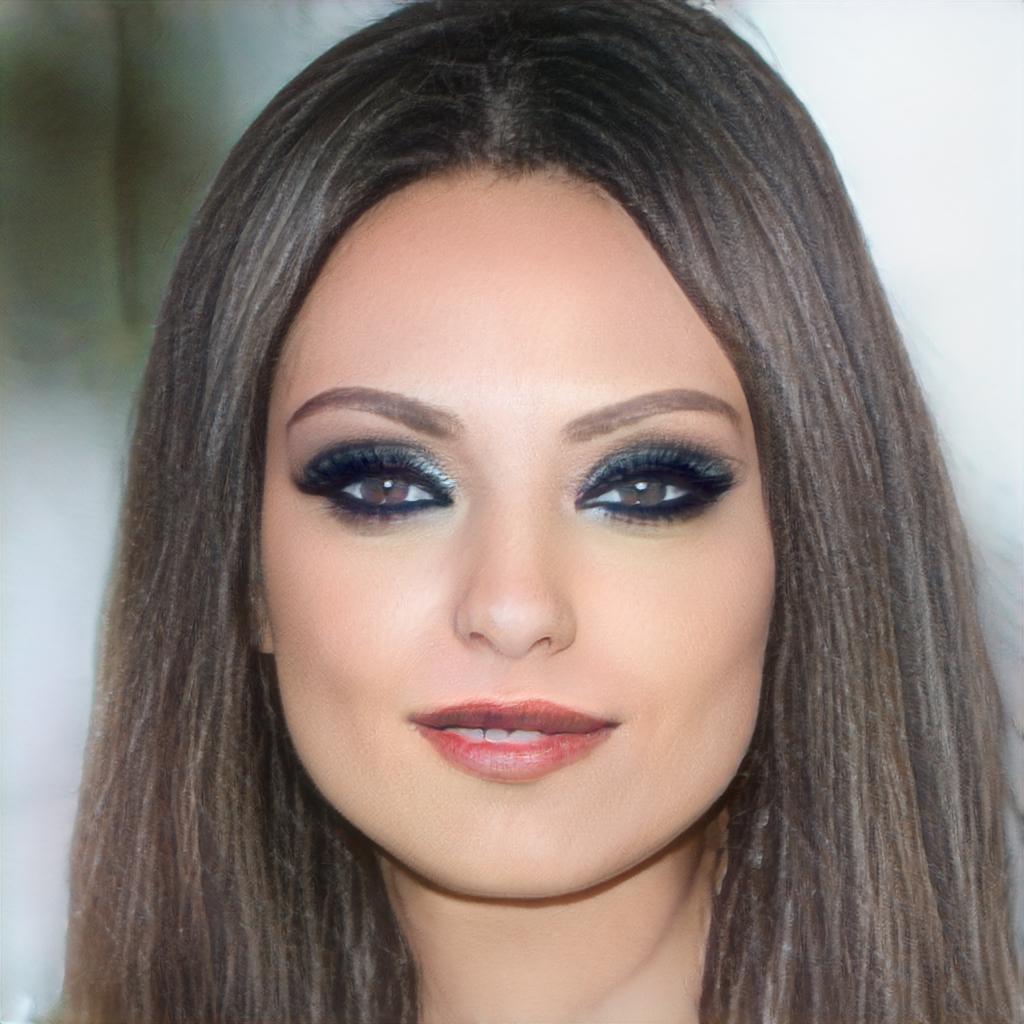} &
 \includegraphics[width=0.13\textwidth, ]{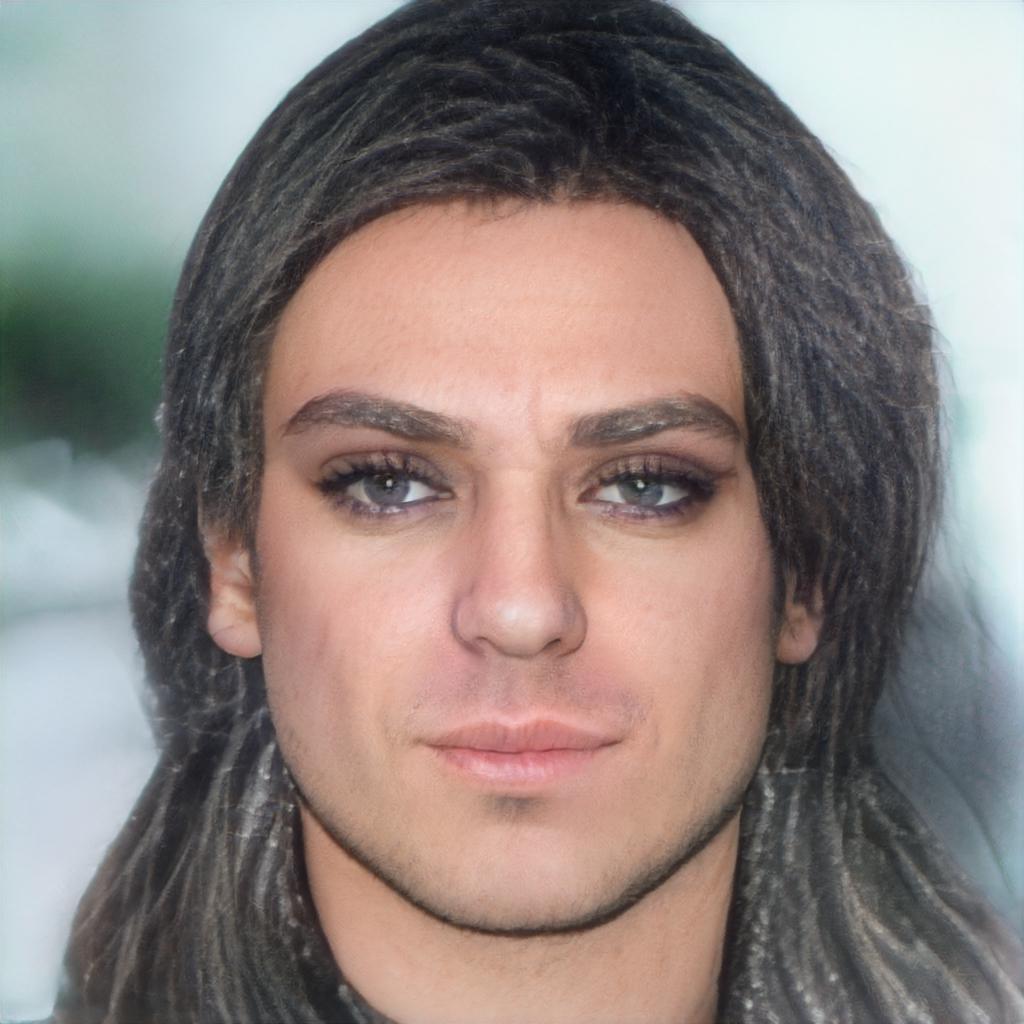} &
 \includegraphics[width=0.13\textwidth, ]{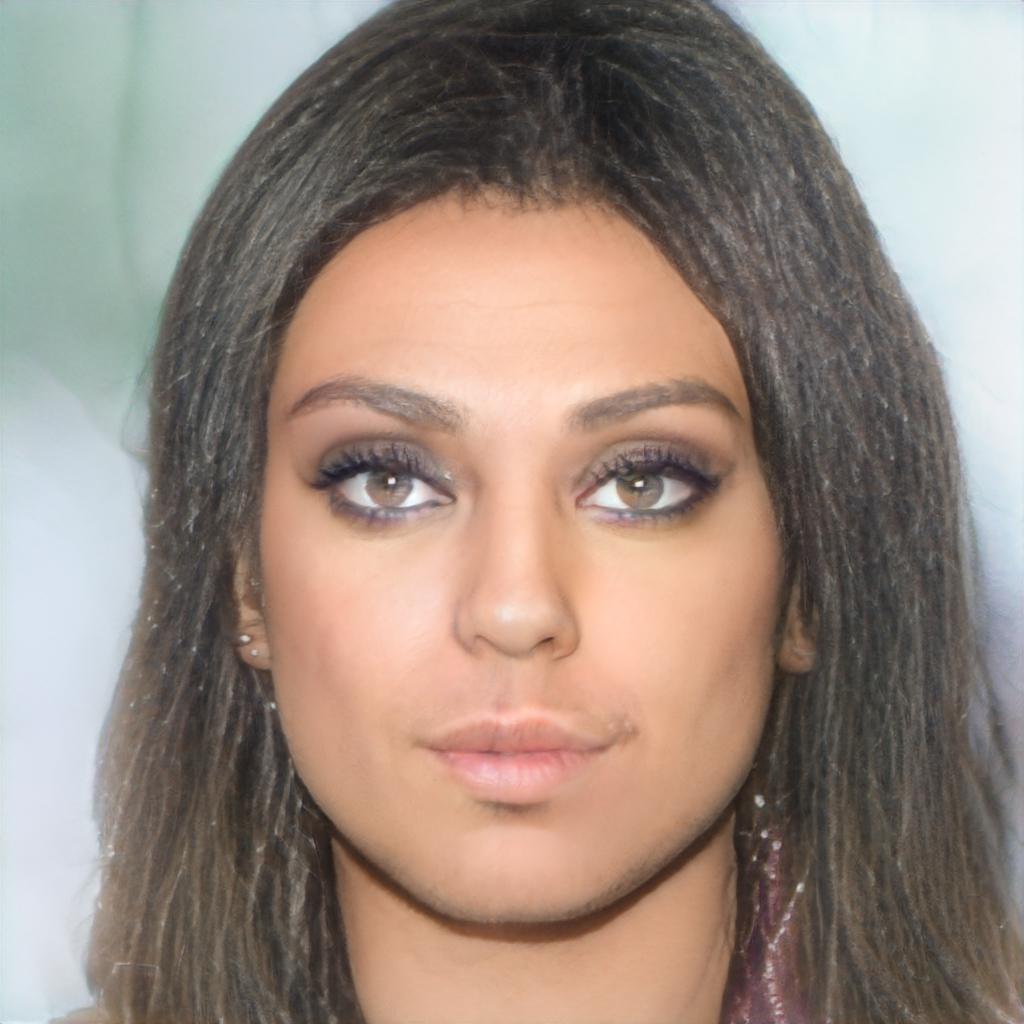} &
 \includegraphics[width=0.13\textwidth, ]{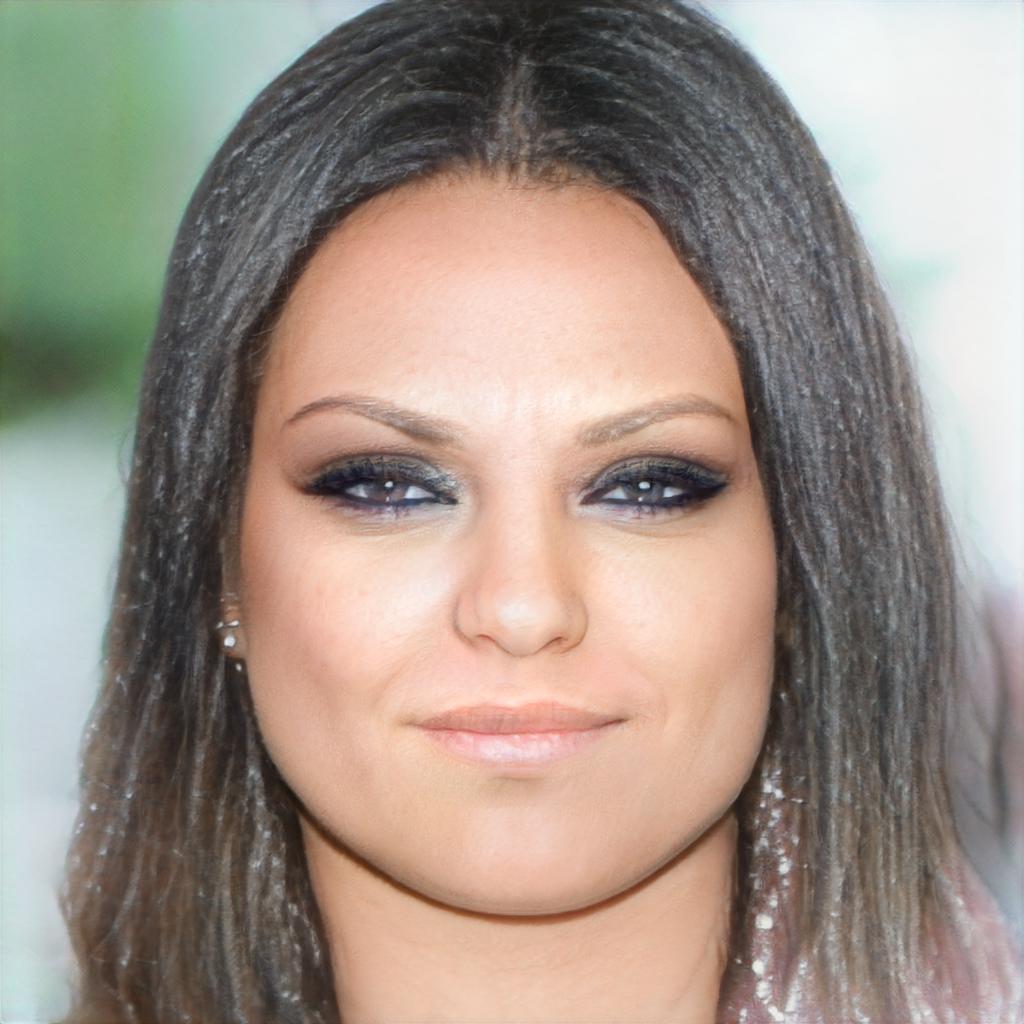} &
 \includegraphics[width=0.13\textwidth]{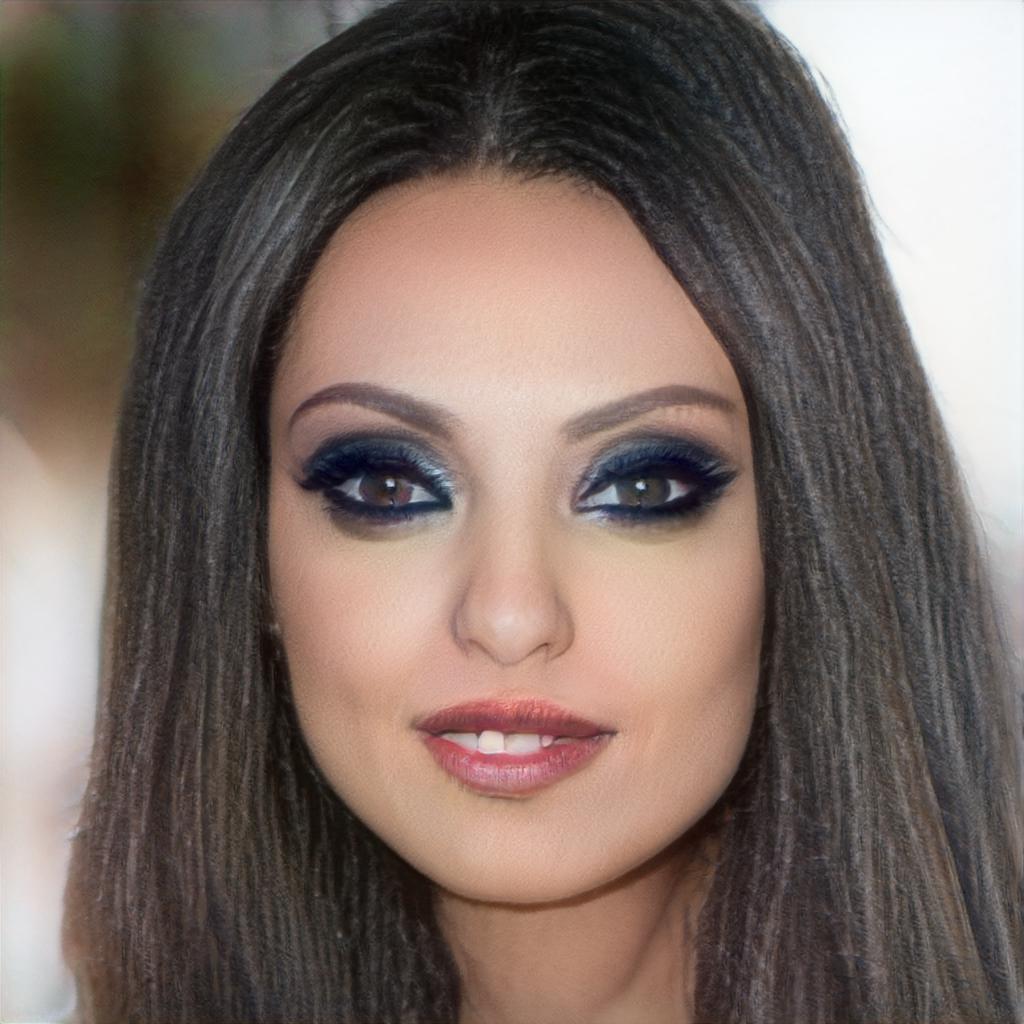} \\
 \begin{turn}{90} \hspace{0.5cm} $\mathcal{W}^{\star}_{ID}$ \end{turn} &
 \includegraphics[width=0.13\textwidth, ]{images/original/06026.jpg} & 
 \includegraphics[width=0.13\textwidth, ]{images/inversion/13_orig_img_27.jpg} &
 \includegraphics[width=0.13\textwidth,  ]{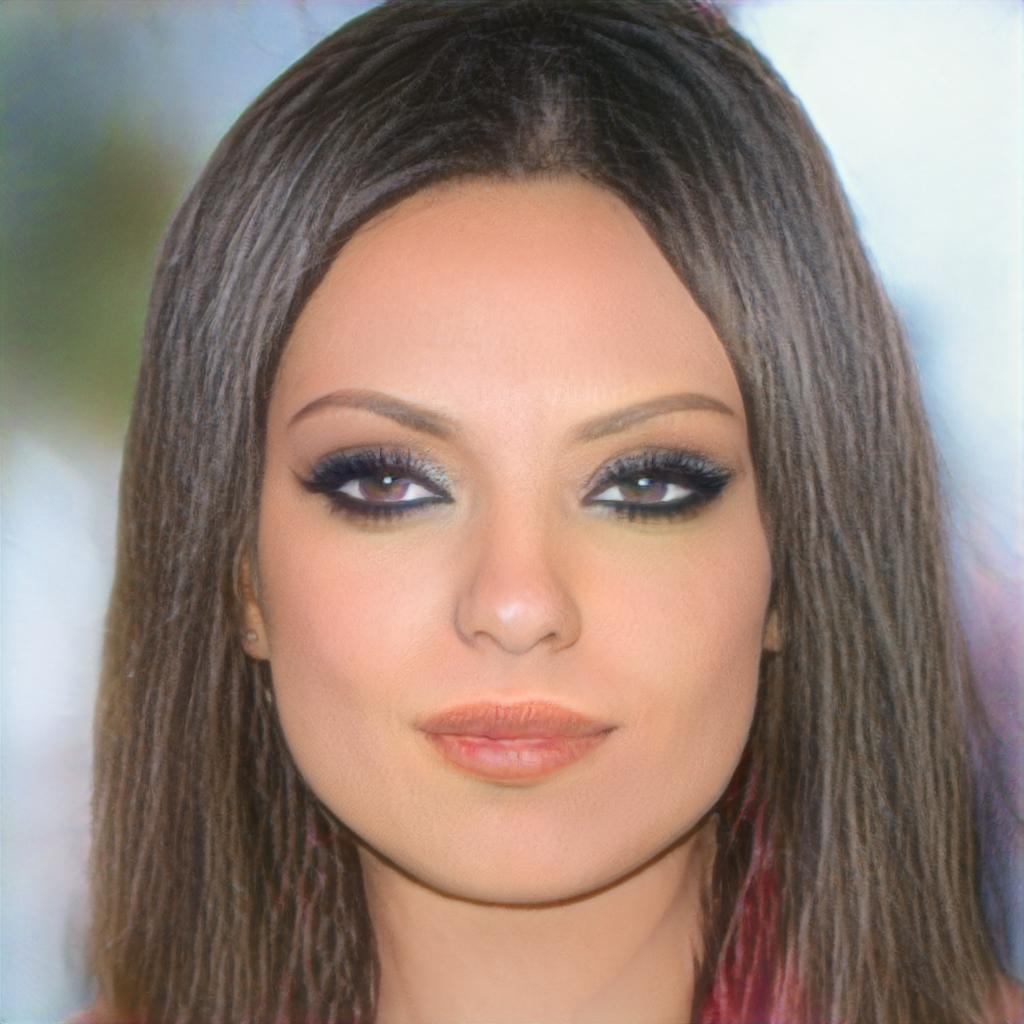} &
 \includegraphics[width=0.13\textwidth, ]{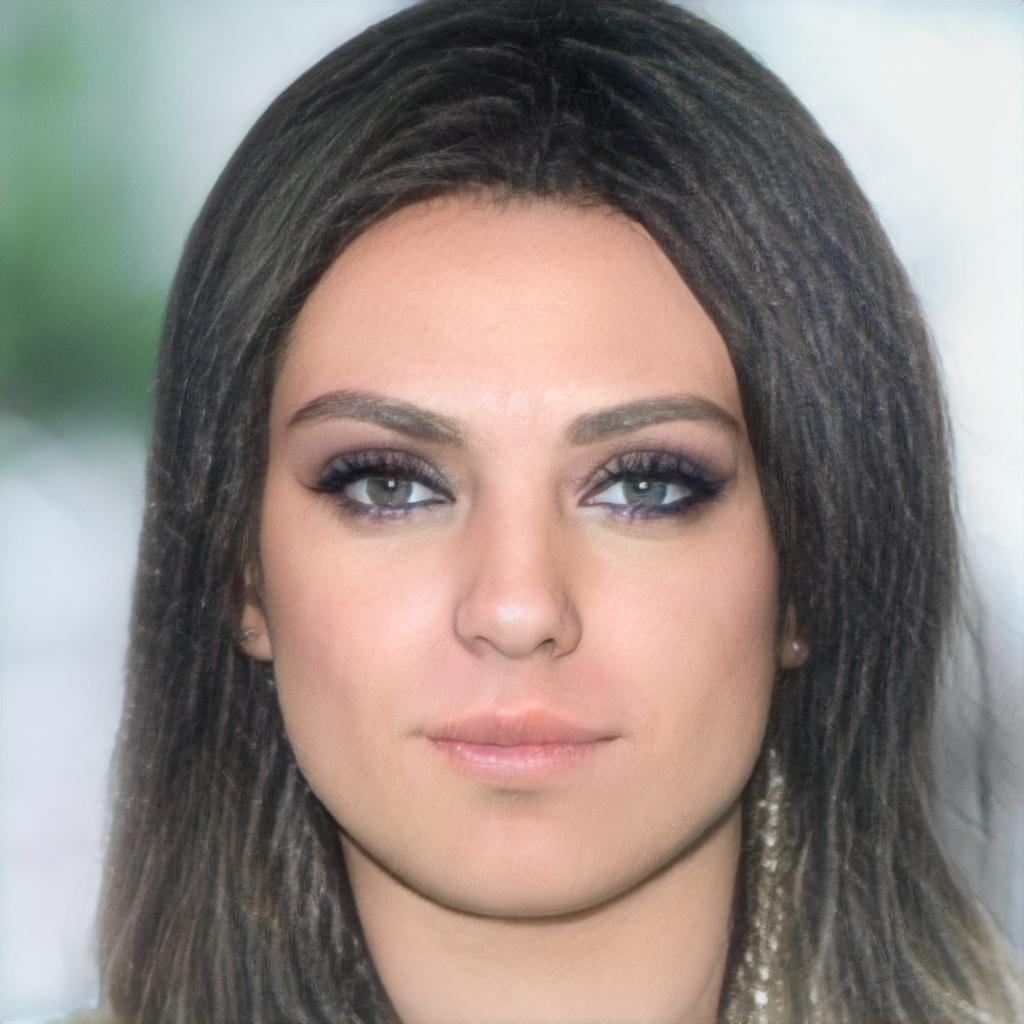} &
 \includegraphics[width=0.13\textwidth, ]{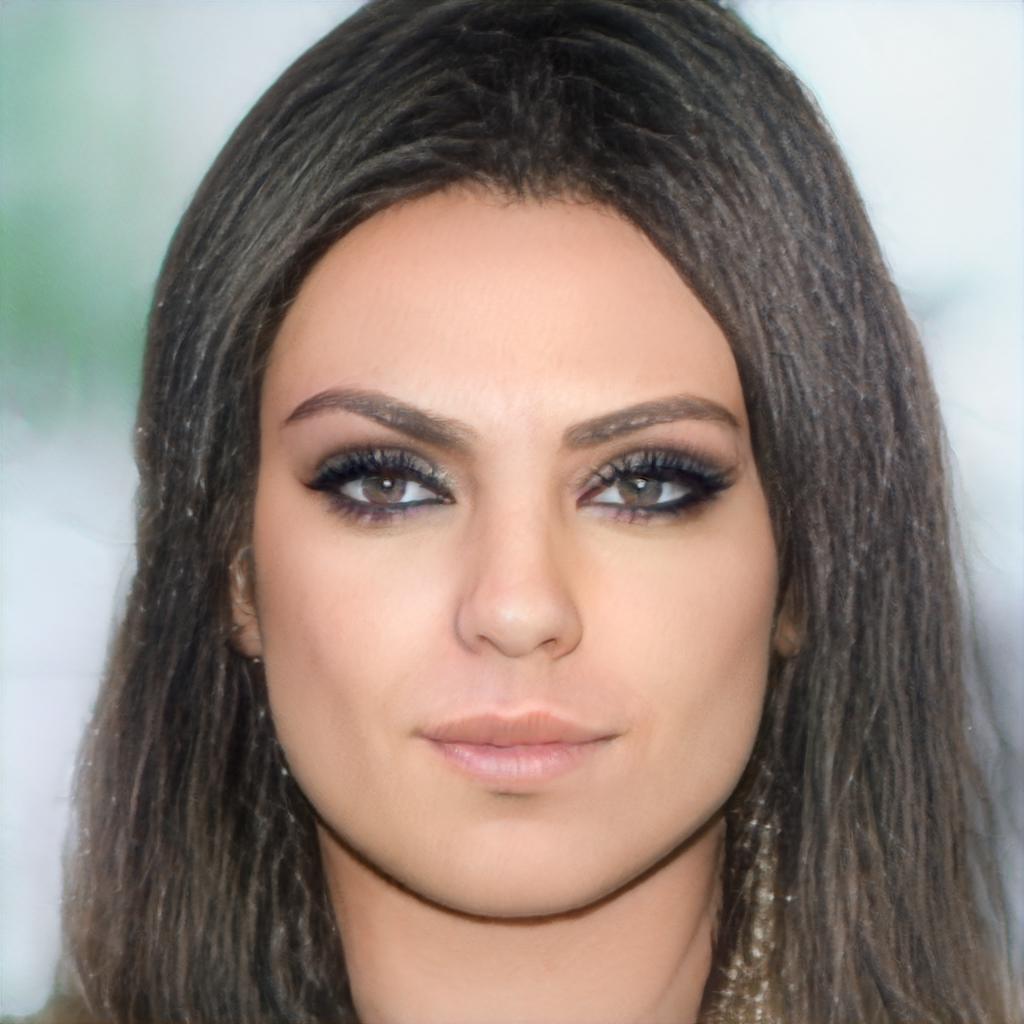} &
 \includegraphics[width=0.13\textwidth, ]{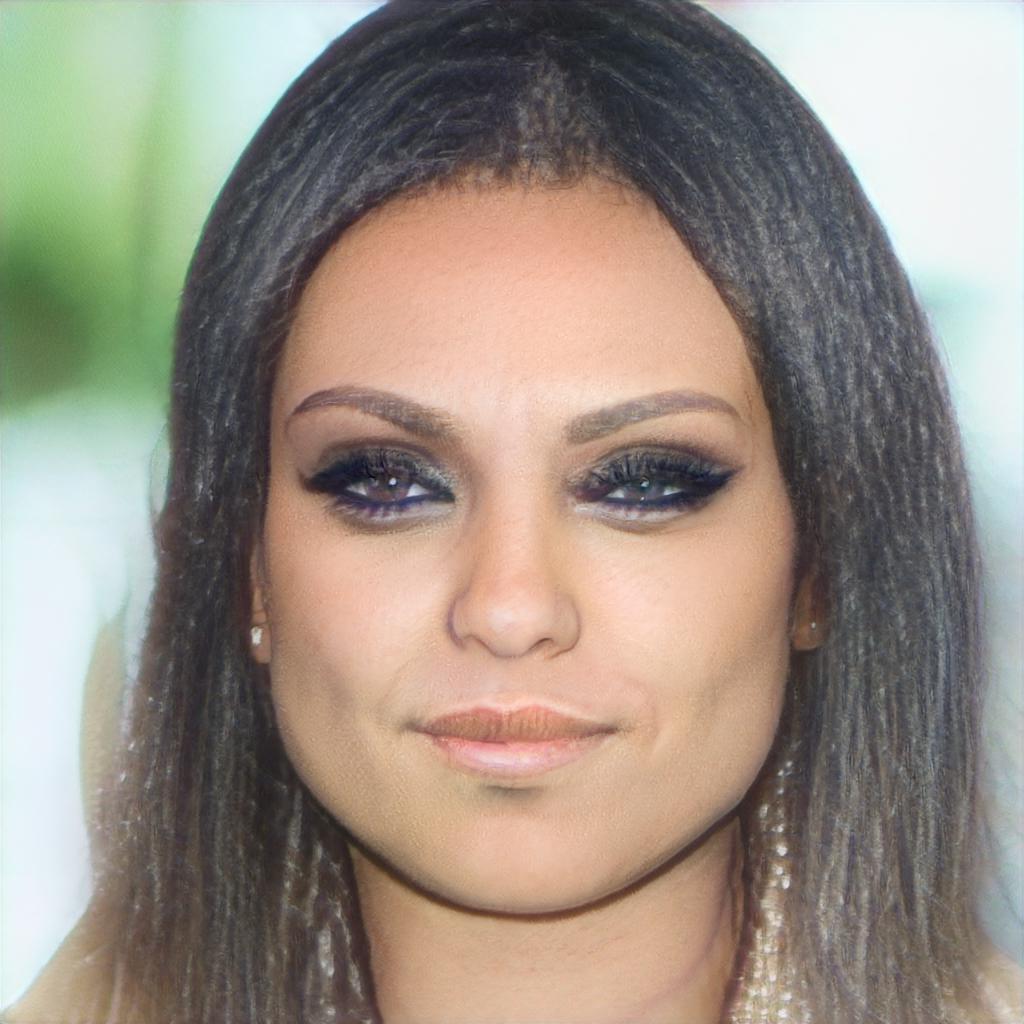} &
 \includegraphics[width=0.13\textwidth, ]{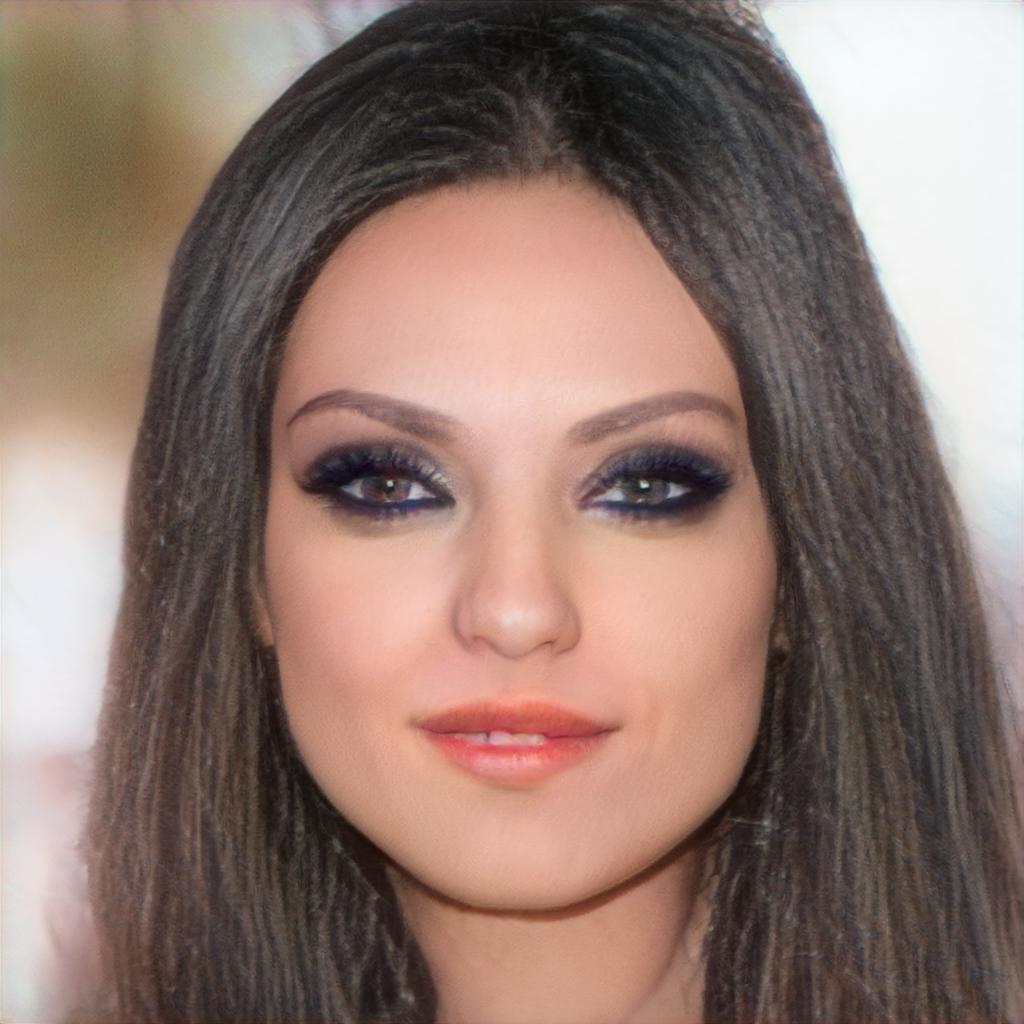} \\
\\
\end{tabular}
\caption{Image editing at resolution $1024^2$ using InterFaceGAN in \wplus  and $\mathcal{W}^{\star}_{ID}$: The images are projected in the latent space of StyleGAN2, then the latent codes are moved in the direction that corresponds to changing one facial attribute. The editing is better in terms of attributes disentanglement in the new learned space ($\mathcal{W}^{\star}_{ID}$) compared to the editing done in the original latent space (\wplussp). Our method also leads to a better identity preservation.}
\label{fig:att_inter_results2}

\end{figure}

\end{document}